\documentclass{article}

\usepackage[utf8]{inputenc}
\usepackage[T1]{fontenc}
\usepackage{lmodern}
\usepackage{textcomp}
\usepackage{slantsc}
\usepackage{PRIMEarxiv}
\usepackage{fancyhdr}
\usepackage{framed}
\usepackage{amsmath}
\usepackage{amsfonts}
\usepackage{amssymb}
\usepackage{mathtools}
\usepackage{bbm}
\usepackage{bm}
\usepackage{dsfont}
\usepackage{booktabs}
\usepackage{multirow}
\usepackage{enumitem} 
\usepackage{graphicx}
\usepackage{tikz}
\usepackage{mwe}
\usepackage{varwidth}
\usepackage{ragged2e}
\usepackage[most]{tcolorbox}
\usepackage{color}
\usepackage[misc]{ifsym}
\usepackage{caption}
\usepackage{subcaption}
\usepackage{cite}
\usepackage{url}
\usepackage{hyperref}   
\usepackage{cleveref}

\makeatletter
\newcommand{\pfnsymbol}[1]{%
  \textsuperscript{\@fnsymbol{#1}}%
}
\makeatother

\title{Missing value imputation with adversarial random forests---MissARF}

\author{
Pegah Golchian$^{1,2}$, Jan Kapar$^{1,2}$, David S. Watson$^{3}$, Marvin N. Wright$^{1,2,4}$
\and
$^1$Leibniz Institute for Prevention Research \& Epidemiology – BIPS, Germany
\and
$^2$Faculty of Mathematics and Computer Science, University of Bremen, Germany 
\and
$^3$Department of Informatics, King's College London, United Kingdom
\and
$^4$Department of Public Health, University of Copenhagen, Denmark \\
\texttt{wright@leibniz-bips.de} \\
}

\begin{document}

\maketitle

\begingroup
\renewcommand\thefootnote{}
\footnotetext{%
  \makebox[0pt][l]{%
    \hspace*{-1.8em}%
    \parbox[t]{\textwidth}{%
      Published as an open access article in:
      \textit{Statistics in Medicine}, 45(3--5):e70379, 2026.
      DOI: \url{https://doi.org/10.1002/sim.70379}.
    }%
  }%
}
\endgroup

\begin{abstract}
Handling missing values is a common challenge in biostatistical analyses, typically addressed by imputation methods. 
We propose a novel, fast, and easy-to-use imputation method called \emph{missing value imputation with adversarial random forests (MissARF)}, based on generative machine learning, that provides both single and multiple imputation. MissARF employs adversarial random forest (ARF) for density estimation and data synthesis. To impute a missing value of an observation, we condition on the non-missing values and sample from the estimated conditional distribution generated by ARF.
Our experiments demonstrate that MissARF performs comparably to state-of-the-art single and multiple imputation methods in terms of imputation quality and fast runtime with no additional costs for multiple imputation.

\keywords{Missing Data, Single Imputation, Multiple Imputation, Generative Modeling, Adversarial Learning, Tree-based Machine Learning Methods}
\end{abstract}

\textbf{Code --- } {\url{https://github.com/bips-hb/arf}} (Software)
\\
\phantom{\textbf{Code --- }} {\url{https://github.com/bips-hb/MissARF-imputation-benchmark}} (Experiments)

\section{Introduction}\label{sec1}
Working with real data can be challenging because we often have to face the problem of missing values. If we want to analyze data, e.g. regarding inference or prediction tasks, it is crucial to properly address this issue. Missing values can arise from various causes, such as human error, equipment failure, or participants withdrawing from studies. Behind the missing values in a dataset are different missing data patterns that Rubin \cite{rubin1976inference} defines as: missing completely at random (MCAR), missing at random (MAR) and missing not at random (MNAR). It is assumed that each data point has a probability of being missing. While with MCAR this probability is the same for all data points, with MAR it depends on observed data. MNAR is the most challenging pattern, since its missingness can depend on observed but also unobserved data, including the missing variable itself. Improper handling of these missingness types can introduce bias, weaken the statistical power as well as prediction performance of machine learning models \cite{van2018flexible,molenberghs2014handbook}.

While some biostatistical methods can handle missing values internally \cite{breimann_Cart_1984, xgboost_chen2016, ke2017lightgbm, tang2017random, josse2024consistency}, most require dealing with them beforehand. Many methods to approach missing data are available, yet there is no single best-suited method overall \cite{slade2020fair,cao2022review,afkanpour2024identify,woznica2020does}. The easiest way to handle missing values is \emph{complete-case analysis (listwise deletion)}, where all observations containing missing values are omitted. In the case of a few missings and MCAR, complete-case analysis can give unbiased estimates of, e.g., means and regression coefficients, but standard errors are generally overestimated \cite{van2018flexible}. In cases other than MCAR it can lead to biased estimates \cite{Rubin1987a,van2018flexible}. Furthermore, with growing missingness and other factors \cite{Little_Rubin}, more information gets lost, up to the point where not a single non-missing observation remains.

Methods that fill missing values are called \emph{imputation methods}. The literature distinguishes between \emph{single imputation}, which outputs a single complete dataset, and \emph{multiple imputation}, which outputs multiple ones. A simple example would be to impute by sampling a random value from the marginal empirical distributions of each variable once for a single imputation and multiple times for multiple imputation (\emph{random imputation}). While with single imputation the dataset is just analyzed once, multiple imputation introduces imputation uncertainty by analyzing each of the imputed datasets and pooling the results into a final point estimate with standard error estimated by ``Rubin's rules'' \cite{Rubin1987a,white2011multiple} and therefore reduces bias. Depending on the task and data structure, different imputation methods are preferred. For prediction tasks, single imputation is often favored for its simplicity and computational efficiency whereas for statistical inference, multiple imputation is typically recommended \cite{van2018flexible,white2011multiple}. 

Other simple examples of single imputation methods are \emph{mean} or \emph{median imputation}, where each missing value is filled with the mean or median of the observed values of that variable. These can lead to biased estimates as well, except for the mean, which is unbiased for MCAR in some cases\cite{van2018flexible}. Furthermore, the variance can be underestimated and the relations between the variables distorted\cite{van2018flexible}. In contrast to complete-case analysis, single imputations tend to underestimate standard errors\cite{van2018flexible}. More sophisticated machine learning methods, such as MissForest \cite{stekhoven2012missforest}---an iterative method based on the random forest (RF) algorithm \cite{Breiman2001} that treats the missing data problem as a prediction problem---are becoming increasingly popular. Machine learning methods have the benefit of tackling complex interactions and nonlinearities of variables and yield superior and fast results in terms of prediction performance\cite{tang2017random}. However, these approaches still do not account for imputation uncertainty, resulting in smaller $p$-values, narrower confidence intervals, and stronger relationships between variables than in the real dataset \cite{van2018flexible}.

Multiple imputation \cite{Rubin1987a} \cite{rubin1996multiple} accounts for imputation uncertainty and can, under suitable conditions, lead to unbiased estimates with appropriately estimated standard errors and valid statistical inference \cite{van2018flexible}. A popular implementation of multiple imputation is given by \emph{multivariate imputation by chained equations (MICE)} \cite{van2007multiple}. The \texttt{R} package \texttt{mice} \cite{mice} offers different ways to create multiple datasets, e.g., with \emph{predictive mean matching (PMM)}---a nearest neighbor approach, which is the current default for numerical data---or with tree-based methods such as RFs. Depending on the data structure, different built-in univariate imputation methods are recommended \cite{van2018flexible}. However, with real data, the data distribution is typically unknown and for a user it is not always clear which option to choose. Some single imputation methods can be extended to multiple imputation, as long as there is sufficient variability across the different datasets. For example, as mentioned above, random imputation can simply be repeated several times and an approach has been made to extend MissForest to multiple imputation by adding PMM to the out-of-bag (OOB) predictions\cite{missranger}. %https://cran.r-project.org/web/packages/missRanger/vignettes/missRanger.html

Generating synthetic data and imputing missing data address a similar question -- finding plausible values for unobserved data. In this paper, we establish this link and consider imputation with generative models. Alongside the current hype about generative machine learning for text generation (e.g., ChatGPT \cite{OpenAI2023GPT4}) and image creation (e.g., DALL-E \cite{Ramesh2022DALLE}), a few approaches have been proposed to use generative models for imputation \cite{yoon2018gain,li2018learning,wibisono2024natural,zhang2023systematic,shahbazian2023generative}. However, those methods are almost exclusively based on deep learning approaches, making them data-hungry, computationally intensive and difficult to train for non-experts. Further, deep learning often struggles with tabular data \cite{grinsztajn2022,shwartz2022tabular} and most of the aforementioned deep learning approaches are not available in \texttt{R}. We argue that an imputation method for everyday biostatistical practice should perform well on tabular data, be computationally fast, easy to use without much hyperparameter tuning and be available in \texttt{R}. With these goals in mind, we present a novel imputation method based on the adversarial random forest (ARF)\cite{watsonAdversarialRandomForests2023} algorithm, a recently proposed tree-based generative model that has been shown to perform well in data synthesis on tabular data \cite{watsonAdversarialRandomForests2023,qian2024synthcity}. Our method \emph{missing value imputation with adversarial random forests (MissARF)} offers both single and multiple imputation. 

We introduce MissARF in Section~\ref{sec: 2 Methods}, which outlines how ARFs can be used for single and multiple imputation. In Section~\ref{sec: 3 Experiments}, we compare MissARF with other imputation methods such as MICE and MissForest for single and multiple imputation on simulated data and in Section \ref{sec: 4 Real data example} on a real data example. Lastly, in Section~\ref{sec: 5 Discussion} we summarize and discuss our results and provide an outlook.

\section{Methods}\label{sec: 2 Methods}
\subsection{Adversarial random forests}
\emph{Adversarial random forests (ARFs)} \cite{watsonAdversarialRandomForests2023} are a tree-based machine learning algorithm for generative modeling. They implement an iterative variant of unsupervised RFs \cite{shi2006unsupervised} that gradually learns the structural properties of the data, building a basis for density estimation and generative modeling. ARFs bear some resemblance to \emph{generative adversarial networks (GANs)} \cite{goodfellow2014generative}.
The difference is that in ARF the discriminator is a RF and the generator leverages the parameters of the discriminator instead of being a separate model. Intuitively, ARFs work as follows: (1) Train a RF to distinguish between naïve synthetic data (created by sampling from the product of the marginals) and real data. (2) Sample again from the product of marginals inside the leaves of this RF and train another RF to distinguish between this new, less naïve synthetic data and the original data. (3) Repeat step (2) until the synthetic data cannot be distinguished from the original data. (4) Apply a univariate density estimation procedure to each feature in each leaf. (5) For data synthesis, draw a leaf according to leaf weights (see below) and sample from the local density estimated in step (4).

In more detail, let $\mathbf{X}_{\text{real}} \in \mathbb{R}^{n \times p}$ be a dataset with $n \in \mathbb{N}$ instances of a set of $p \in \mathbb{N}$ features, where each row represents a sample from the feature space $\mathcal{X} \subseteq \mathbb{R}^p$. To train an ARF, we first create a naïve synthetic dataset $\mathbf{X}_{\text{synth}}^{(0)} \in \mathbb{R}^{n \times p}$ by drawing from the original values with replacement separately within each column, effectively treating all features as independent. The real data points are labeled with 1 and the synthetic data points with 0. In the first discriminator step, a RF is trained on the binary classification setting with $\mathbf{X}^{(0)}=\{(\mathbf{X}_{\text{real}},\mathbf{X}_{\text{synth}}^{(0)})\}$ and $Y=\{\bm{1}_n,\bm{0}_n\}$, i.e., the RF is trained to distinguish the naïve synthetic data from the real data. 
%and a target $y_i \in Y = \{0,1\} \subseteq \mathbb{R}$. 
In the next generating step, we exploit the splits of the forest to create a more realistic synthetic dataset $\mathbf{X}_{\text{synth}}^{(1)}$. 
For this, we again sample the original values with replacement separately within each column -- but this time we do so within each leaf of the RF (keeping the same size) resulting in a dataset that more closely resembles the original.
Now, at the next discriminator step, we train a new RF on $\mathbf{X}^{(1)}=\{(\mathbf{X}_{\text{real}},\mathbf{X}_{\text{synth}}^{(1)})\}$. We continue the generating and discriminating steps until the RF cannot distinguish between real and synthetic data. We measure this convergence with the out-of-bag (OOB) accuracy \cite{Breiman2001} 
and stop if the OOB accuracy is less than $0.5 + \delta$ with a small $\delta \geq 0$. Our final model is the last RF before the OOB accuracy falls below this threshold. 
Let $T$ be the number of trees in the final forest of ARF. Let $L$ be the number of leaves over all $T$ trees in the RF and $L_t$ the number of leaves in a tree $t \in [T]$. For each tree, each leaf $l\in [L_t]$ represents a unique hyperrectangular subspace $\mathcal{X}_l \subset \mathcal{X}$ where all subspaces together span the whole data manifold $\mathcal{X} = \bigcup_{l}\mathcal{X}_l$. Leaves are characterized by a conjunction of tree splits that define the complete path starting from the root.
Let $n_{t}$ be the number of ``real'' training samples for tree $t$ and $n_{tl}$ the number of those real samples that fall into leaf $l$. Then the probability that an observation $\bm{x}$ falls within $\mathcal{X}_l$ in tree $t$---the leaf's \textit{coverage}---can be empirically estimated by $n_{tl} / n_{t}$. We now assign equal weights to all trees in the RF and define the \emph{leaf weights} over the whole RF as $\omega_l = \frac{n_{tl}}{Tn_{t}}$. 

In case of convergence, we can assume that all features are mutually independent within the leaves. For all $\bm{x} \in \mathcal{X}$, the \emph{local independence criterion} can be formulated as 
\begin{align}
    \hat{p}_l(\bm{x}) =\prod_{j = 1}^p \hat{p}_{lj}(x_j),
\end{align}
where $p_l$ denotes the multivariate density within a leaf and $p_{lj}$ the univariate density for a feature within a leaf. The proof can be found in Watson et al\cite{watsonAdversarialRandomForests2023}. Intuitively, the ARF procedure can only learn dependency structures of the data by ``splitting out'' the dependencies in the trees and as soon as it is unable to learn any further dependencies, we can assume the data to be independent (within the leaves). 

The local independence criterion simplifies the task of multivariate density estimation, which can now be achieved by learning $p$ separate univariate density estimation functions within each leaf for each feature based on the original values. Watson et al. \cite{watsonAdversarialRandomForests2023} implemented a maximum likelihood-based truncated Gaussian approach for continuous variables and multinomial distributions for categorical ones. The estimated density of ARF at a given point $\bm{x} \in \mathcal{X}$ is defined as:
\begin{equation}\label{eq:arf_dens}
    \hat{p}_{\mathrm{ARF}}(\bm{x}) = \sum_{l=1}^L \omega_l \hat{p}_l(\bm{x})= \sum_{l=1}^L \omega_l \prod_{j = 1}^p \hat{p}_{lj}(x_j).
\end{equation}
It takes the weighted average of the densities of all the leaves into which $\bm{x}$ falls, i.e., where the split criteria are met. 
Since ARF uses intra-leaf distributions truncated to the leaf bounds in practice, leaves where $\bm{x} \not \in \mathcal{X}_l$ naturally receive zero weights $\omega_l = 0$. Watson et al.\cite{watsonAdversarialRandomForests2023} prove under mild conditions that $\hat{p}_{\mathrm{ARF}}$ converges to the real data distribution of $\mathbf{X}_{\text{real}}$ for infinite data.
Generating new synthetic data points using ARF works as follows: We first sample leaves with probabilities $\omega_l$. Then we create synthetic data points by sampling independently from the estimated univariate densities within the leaves for every variable. 

We can also perform conditional density estimation and sampling with ARF under a set of conditions by filtering out the leaves that match the conditions and updating the leaf weights $\omega_l$. This procedure was first used to generate counterfactual explanations with ARFs \cite{dandl2024countarfactuals}. Conditional density estimation and sampling are necessary for our missing value imputation method and will be described in detail below. 

\subsection{MissARF: Missing value imputation with adversarial random forests} \label{sec: MissARF}
\emph{Missing value imputation with adversarial random forests (MissARF)} is an imputation method based on generative modeling that offers single and multiple imputation. The general idea is to condition on non-missing values and impute missing values by sampling from the conditional distribution, estimated by an ARF. To estimate the conditional distribution, we filter the leaves where the condition on non-missing values matches the splits in the trees and apply the local density estimation on those leaves for each feature on the real observations. Next, we adjust the leaf weights with the filtered leaves. Finally, we sample a leaf according to the new leaf weights and impute by sampling a value from the estimated conditional distribution.

Let $\bm{x}$ be a data point that contains missing values at the index positions $\overline{C} = \{j \mid x_j = \mathrm{NA}\}$, where $\mathrm{NA}$ symbolizes a missing value. Further, let $X_C$ denote the random variable of the non-missing values $\bm{x}_C$ with $C = \{j \mid x_j \neq \mathrm{NA}\}$. 
Then for all $\bm{x}_{\overline{C}} \in \mathcal{X_{\overline{C}}}$, the conditional density given $\bm{x}_{C} $ can be estimated exploiting Bayes' theorem and the local independence assumption within the leaves as 
\begin{align}\label{eq:arf_dens_cond}
    \hat{p}_{\mathrm{ARF}}\left(\bm{x}_{\overline{C}} \,\middle|\, X_C = \bm{x}_C\right)
    &= \frac{\hat{p}_{\mathrm{ARF}}(\bm{x}_{\overline{C}}, \bm{x}_C)}{\hat{p}_{\mathrm{ARF}}(\bm{x}_C)} = \frac{\hat{p}_{\mathrm{ARF}}(\bm{x})}{\hat{p}_{\mathrm{ARF}}(\bm{x}_C)} = \frac{\sum_{l=1}^L \omega_l \hat{p}_l( \bm{x}_C) \hat{p}_l( \bm{x}_{\overline{C}})}{\hat{p}_{\mathrm{ARF}}(\bm{x}_C)} = \sum_{l=1}^L 
    \omega_l \frac{\hat{p}_l(\bm{x}_C)}{\hat{p}_{\mathrm{ARF}}(\bm{x}_C)} \prod_{j \in \overline{C}} \hat{p}_{lj}(x_j) \notag \\
    &= \sum_{l=1}^L 
    \tilde{\omega}_l \prod_{j \in \overline{C}} \hat{p}_{lj}(x_j)
\end{align}
with adjusted leaf weights $\tilde{\omega}_l \coloneqq \tilde{\omega}_l(\bm{x}_C) \coloneqq \omega_l\frac{\hat{p}_l( \bm{x}_C)}{\hat{p}_\mathrm{ARF}(\bm{x}_C)}$. Intuitively, the weights are rescaled by the intra-leaf densities $\hat{p}_l( \bm{x}_C)$ and normalized by the marginal global density $\hat{p}_\mathrm{ARF}(\bm{x}_C)$ to ensure that leaves where the condition is more likely fulfilled receive higher weights and that the weights sum to 1.  As above, because ARF uses intra-leaf distributions truncated to the leaf bounds in practice, leaves with ranges not satisfying the condition naturally receive zero weights $\tilde{\omega}_l = 0$.

We can get an imputed data point $\dot{\bm{x}}$ analogous to the unconditional case by first sampling a leaf that matches the condition based on the adjusted leaf weights $\tilde{\omega}_l$ and then sampling a value from the distribution estimated in that leaf for the missing feature.\footnote{If no leaves can be found, e.g., because no leaf satisfies the condition, then the default in the \texttt{arf} package would sample from the entire range of the feature.} To impute a whole dataset with MissARF, we consider all the rows that contain missing values and condition on the non-missing values for each of them.

For single imputation, we can directly use the imputed data point $\dot{\bm{x}}$, drawn from the conditional distribution (Eq.~\ref{eq:arf_dens_cond}), for each row and by that create a complete dataset. However, since we estimated the full conditional density, we can also calculate summary statistics such as the mean, median or mode in a leaf, to achieve a more stable imputation result. 
Alternatively, for continuous features, we could impute the expected value of the conditional distribution estimated by ARF, i.e., the expected value over all leaves that match the condition, a method we adopt by default in our experiments. 
Let $\mu_{lj}$ be the mean of the real and non-missing feature values of $x_j$ within a leaf $l$ conditioned on the non-missing values $\bm{x}_C$. With the adjusted leaf weights $\tilde{\omega}_l$, we can then impute the value $\dot{x}_j$ by:
\begin{align}
    \dot{x}_j = \mathbb{E}\left[X_j \,\middle|\, X_C = \bm{x}_C \right] 
    = \sum_{l=1}^L \tilde{\omega}_l ~\mu_{lj}.
\end{align}
To address categorical features, we impute by the most frequent category in the selected leaves, weighted by $\tilde{\omega}_l$. For multiple imputation, we can sample several times from the conditional distribution (Eq.~\ref{eq:arf_dens_cond}) to get a range of imputed datasets. Note that we do not sample several times from the same distribution estimated in a single leaf, but sample a new leaf based on the (adjusted) leaf weights for each new imputation. 

To use MissARF, the RFs in the ARF procedure have to be trained on the original dataset, containing missing values. However, by construction, RFs cannot handle missing values without further modifications. Several such modifications have been proposed \cite{tang2017random, josse2024consistency}, but a systematic comparison of these methods is lacking. Here, we implemented a slightly modified version of the \emph{missingness incorporated in attributes}\cite{twala2008good,twala2010ensemble} approach in \texttt{ranger}\cite{ranger}\footnote{Implemented in ranger version $\geq0.17.0$, available on \url{https://github.com/imbs-hl/ranger} and on \texttt{CRAN}.}, which learns an optimal child node assignment of missing values during the calculation of the node split criterion. For categorical features, this is done by simply treating the missing values as a separate category. For numerical features, the following two splits are compared based on the split criterion:
\begin{description}
\item[Split A:] $\{ x_j \leq s \text{ or } x_j = \mathrm{NA} \}$ versus $\{ x_j > s \}$
\item[Split B:] $\{ x_j \leq s\}$ versus $\{ x_j > s \text{ or } x_j = \mathrm{NA} \}$,
\end{description}
where $s$ is the split value selected for feature $x_j$. The finally selected split point is the one that maximizes the split criterion, i.e., minimizes the loss function. 

\subsubsection*{Example and intuitive explanation}
\begin{figure}
\centering
\includegraphics[trim={2cm 2.3cm 3.5cm 2.3cm},clip,scale=.5]{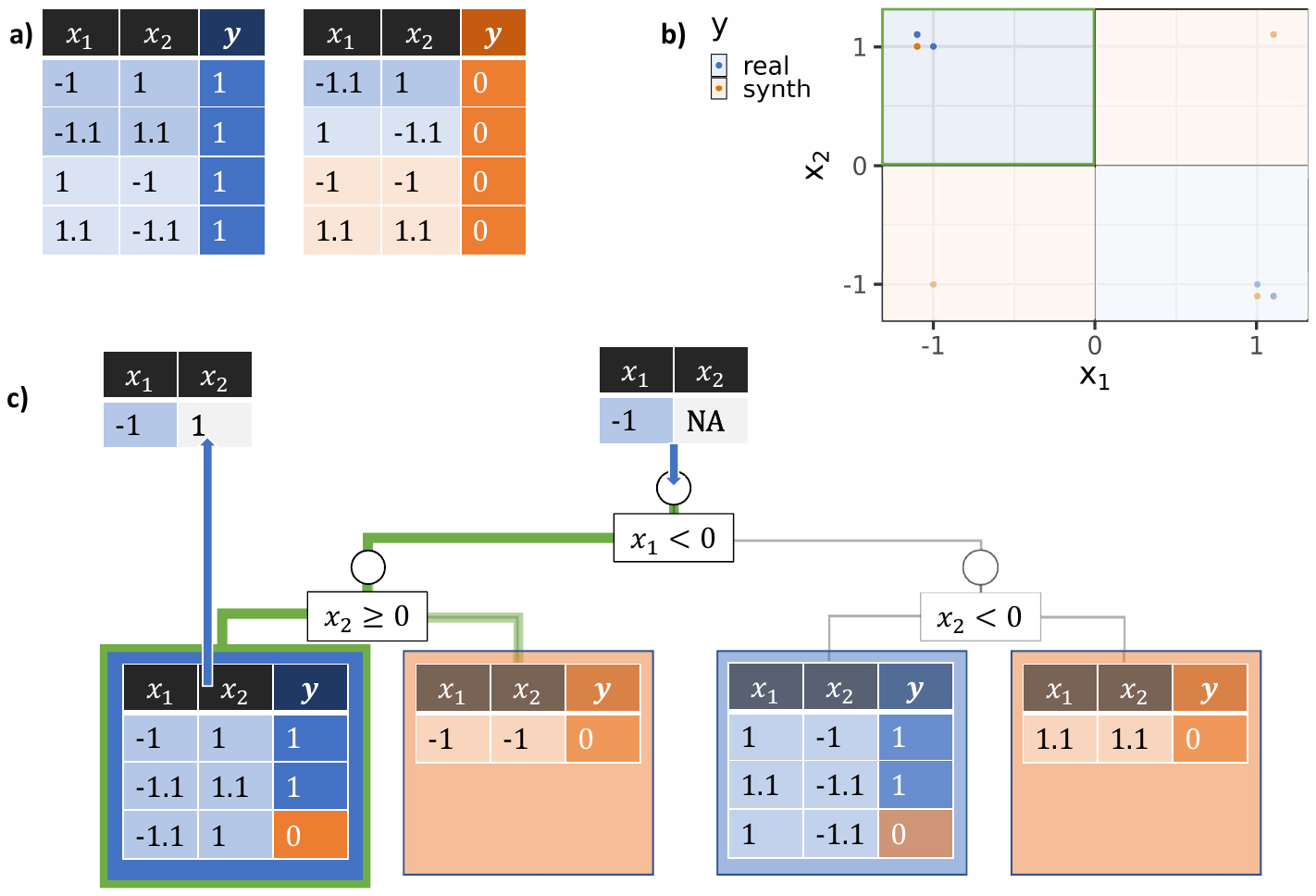}
\caption{Single imputation example with MissARF of the point $\bm{x}=(-1,\mathrm{NA})$ on one exemplary tree of a converged ARF of a two-dimensional dataset consisting of clusters around $(-1,1)$ and $(1,-1)$. Real data points are labeled with $y=1$ (blue) and synthetic data points are labeled with $y=0$ (orange). If the tree predicts real observations, it is colored in blue and orange otherwise. a) Shows the reduced example dataset in a tabular form. b) The partition plot of the exemplary tree of ARF in c), where the imputation example is visualized.}
\label{fig: example MissARF}
\end{figure}

To illustrate MissARF, we use a simple two-dimensional dataset with two features with points in clusters around $(-1,1)$ and $(1,-1)$. In Figure~\ref{fig: example MissARF}a) we show such an example with only four observations labeled as $y=1$ (blue) and the resulting synthetic dataset labeled as $y=0$ (orange), which are plotted in Figure~\ref{fig: example MissARF}b). Assume that we have a data point $\bm{x}=(-1, \mathrm{NA})$ that we would like to impute. For that, we filter the leaves, where the condition on non-missing values matches the splits in the trees. In this exemplary tree of the RF (Figure~\ref{fig: example MissARF}c)), we would end up in the two left leaves since $x_1 <0$ and $x_2$ is undefined. In the next step, we sample one of the filtered leaves weighted by the share of real data points (leaf weights $\tilde{\omega}_l$) and then impute by sampling from the estimated distribution for that feature in this leaf for the missing feature. In this example, we sample the left leaf and then sample marginally from $x_2$ and could then end up with, e.g., $\dot{x}_2=1$.

\subsection*{Application in \texttt{R}}
The implementation of MissARF is integrated into the \texttt{arf}\cite{arf_r} package\footnote{Implemented in \texttt{arf} version $\geq0.2.3$, available on \url{https://github.com/bips-hb/arf} and on \texttt{CRAN}.} itself. Assume the R object \texttt{data} is a dataset with missing values. It can be imputed with MissARF as follows:

\begin{verbatim}
   library(arf)

   # Single imputation
   data_single_imp <- arf::impute(data)

   # Multiple imputation
   data_multiple_imp <- arf::impute(data, m = 20)
\end{verbatim}

\section{Simulation studies} \label{sec: 3 Experiments}
\subsection{Setup} \label{sec: setup}
We compare our novel imputation method MissARF with state-of-the-art single and multiple imputation methods in terms of imputation quality and runtime. In the following, we refer to setting I for single imputation, and to setting II for multiple imputation. We replicate each experiment 1000 times in both settings.

\subsubsection*{Simulated data}
To cover different settings, we generate different correlated multivariate data for $\mathbf{X}\in \mathbb{R}^{n\times p}$ using a Gaussian copula-based approach \cite{goldfeld2020simstudy}, where each dataset is generated from one of the following univariate distributions: $\mathcal{N}(0,1)$, $\mathrm{Binom}(0.5)$, $ \mathrm{Pois}(2)$, $\mathrm{Gamma}(2,0.5)$, and $\mathcal{U}(-1,1)$. The correlation structure is defined by a Toeplitz matrix $\mathrm{Corr}(\mathbf{X})_{ij}=0.5^{|j-i|}$, where neighboring features are correlated the most. To simulate data, we utilize the \texttt{R} package \texttt{simstudy}\cite{simstudy_R}. We choose different numbers of features $p\in \{4,10,20\}$ and different sample sizes $n\in \{500, 1000, 10\,000\}$. Further, we simulate a linear and squared effect on a binary outcome $Y \in \mathbb{R}^n$, with an effect size of the features chosen as $\bm{\beta} = (-0.5, \dots, 0.5) \in \mathbb{R}^p$ with an equidistant step size. More precisely, let $\bm{x}_i$ denote a row of $\mathbf{X}$ with an outcome $y_i$. For each $n$ simulated observations $\bm{x}_i$, we sample $y_i$ from a binomial distribution $y_i \sim \mathrm{Binom}(\pi_i)$ with $\pi_i=\sigma(\bm{x}_i\bm{\beta})$ or $\pi_i=\sigma(\bm{x}_i^2\bm{\beta})$ where $\sigma$ is the pointwise standard logistic function $\sigma(x)=\frac{1}{1+e^{-x}}$. To measure downstream prediction performance (see below), we simulate an additional test dataset of the same size $n$.

\subsubsection*{Missing data}\label{sec:missing_data}
For all datasets, we simulate the missing data patterns MCAR, MAR and MNAR for missingness proportions of 0.1, 0.2 and 0.4. To simulate missing values in a selected feature by MAR or MNAR, the feature values are divided into two groups defined by a cut-off value (here, the median). For MAR, the cut-off value is calculated on another fully observed variable and for MNAR on the variable itself. One of the two groups, either above or below the median, is then chosen at random and missing values are introduced within this group. For our implementation, we use the \texttt{R} package \texttt{missMethods} \cite{missMethods_R}. To maintain consistency across all cases and comply with the MAR definition, where missingness depends on an observed variable, we introduce missing values in only half of the features. For the prediction task, missingness is introduced separately in both the training and test data.

\subsubsection*{Metrics}\label{sec:metrics}
In setting I (single imputation), we consider the \emph{normalized root mean squared error (NRMSE)} as a measure of data dissimilarity and the Brier score \cite{brier1950verification} as a measure of downstream prediction performance. The NRMSE evaluates the point-wise similarity of an imputed $\dot{\mathbf{X}}$ and the original dataset $\mathbf{X}$. To achieve comparability for features measured on different scales, we first standardize the features and then calculate the RMSE over all points. To get a consistent standardization over all imputed datasets, we always choose the same $\mu_X$ and $\sigma_X$ from the original dataset $\mathbf{X}$ yielding a standardized imputed dataset as $\tilde{\dot{\mathbf{X}}} = \frac{\dot{\mathbf{X}}-\mu_X}{\sigma_X}$.
We take the intuitive extension of the one-dimensional NRMSE, which is a Euclidean distance, to a high-dimensional NRMSE by the Frobenius norm. Let $\tilde{\dot{x}}_{ij}\in \tilde{\dot{\mathbf{X}}} \subseteq \mathbb{R}^{n \times p}$ be an entry in the normalized imputed dataset and $\tilde{x}_{ij} \in \tilde{\mathbf{X}} \subseteq \mathbb{R}^{n \times p}$ in the normalized ground truth dataset, then the NRMSE can be calculated as
\begin{align}
    \mathrm{NRMSE}(\dot{\mathbf{X}},\mathbf{X}) = \sqrt{\frac{1}{n \cdot p}\sum_{j=1}^{p} \sum_{i=1}^{n} \left( \tilde{\dot{x}}_{ij} - \tilde{x}_{ij}\right)^2}.
\end{align}
The Brier score assesses calibration as well as discrimination of predictions. First, we fit a generalized linear model to the imputed training data. We then use this model to predict the probability for $Y=1$ with the imputed test data. For each of the $n$ observations in the test data, we then compare the true $y$ value with the predicted probability $\hat{\pi}$ using the Brier score:
\begin{equation}\label{eq:brier}
    \mathrm{Brier\,Score}(\hat{\pi}, y) = \frac{1}{n} \sum_{i=1}^n \left(\hat{\pi}_i - y_i\right)^2.
\end{equation}

In setting II (multiple imputation), we are interested in whether we can derive reliable statistical inferences with our imputed datasets. Here, the parameters of interest are the regression coefficients $\bm{\beta}$ of a logistic regression. For simulated data, these values are given by the data-generating process (see above) itself. With ``Rubin's rules'' we obtain a parameter estimate $\hat{\bm{\beta}}$ as the mean of the $m$ estimates of the imputed datasets and a standard error for this estimate. 
From this, we can calculate $(1-\alpha)$-confidence intervals for each regression coefficient of a feature $\beta_j$, such that for the random variables $A_j$ and $B_j$, with $A_j < B_j$, the probability $P(A_j < \beta_j < B_j) = 1-\alpha$ holds. With the realizations $A_j =a_j$ and $B_j = b_j$ we can define the confidence interval as $CI_j = [a_j, b_j]$. We assess the quality of statistical inference by examining the coverage rate, average width of the confidence intervals and the root mean squared error (RMSE) of the regression coefficients.

The \emph{coverage rate} indicates how often, over $K$ replications, confidence intervals encompass the actual value of the parameter of a feature $j$: %$\frac{1}{K} \sum_{k=1}^K \mathds{1}(\beta \in \mathrm{CI}^{(k)})$. 
\begin{align}
    \hat{P}_\mathrm{Cov}(A_j < \beta_j < B_j)=\frac{1}{K} \sum_{k=1}^K \mathds{1}\left(a_j^{(k)} < \beta_j < b_j^{(k)}\right),
\end{align}
with the indicator function $\mathds{1}(\cdot)$ that returns 1 if the $\beta_j$ is in the estimated confidence interval and 0 otherwise. The coverage should be close to the nominal rate, which in our experiments is 95\%. Empirical coverage below the nominal rate is regarded as too optimistic or liberal; coverage above the target rate as too conservative.

If the coverage rate is valid, i.e. close to the nominal rate, we prefer an \emph{average width of confidence intervals} as small as possible, which demonstrates statistical efficiency. We get the width by calculating the mean difference between the upper and lower bound of CI over the replications, i.e. $\mathrm{AW}(A_j, B_j)=\frac{1}{K} \sum_{k=1}^K \left( b_j^{(k)}-a_j^{(k)}\right)$. 

Finally, we calculate the \emph{root mean squared error (RMSE)} for a feature $j$ between the true regression coefficients $\beta_j$ and the ones resulting from the imputed datasets $\hat{\beta}_j^{(1)},\ldots,\hat{\beta}_j^{(K)}$ to assess how accurate and precise the coefficients can be estimated, i.e., %$\mathrm{RMSE}(\bm{\beta}, \hat{\bm{\beta}}) = \sqrt{(\bm{\beta} - \hat{\bm{\beta}})^2}$. 
\begin{align}
   \mathrm{RMSE}(\hat{\beta}_j^{(1)},\ldots,\hat{\beta}_j^{(K)}; \beta_j) = \sqrt{\frac{1}{K}\sum_{k=1}^K\left(\hat{\beta}_j^{(k)} - \beta_j\right)^2}. 
\end{align} 

\subsubsection*{Compared imputation methods}\label{sec:compared_methods}
For setting I, we choose random imputation and median imputation as baseline methods. We do not consider complete-case analysis because it does not produce an imputed dataset and would thus only be relevant for performance evaluation and, more importantly, often results in too few observations under high missingness rates.
Further, we consider MICE\footnote{Note that MICE and MissForest PMM are constructed for creating valid inference with multiple imputation, therefore not intended for a single imputation application with $m=1$.} with internal imputation methods predictive mean matching (PMM) and RFs. In our experiments, we refer to these as \emph{MICE PMM} and \emph{MICE RF}, respectively. We use the \texttt{mice} \texttt{R} package \cite{mice}.
Lastly, we consider MissForest \cite{stekhoven2012missforest}, a popular single imputation method based on RFs. For MissForest, we choose the \texttt{R} package \texttt{missRanger} \cite{missranger}, which is a faster implementation of the original \texttt{missForest} package. The \texttt{missRanger} package provides an additional option to combine MissForest with PMM and by that creates more diverse datasets for multiple imputation. We consider both the original version (\emph{MissForest}) and the version including PMM (\emph{MissForest PMM}).
For MissARF, we choose 100 trees and 10 as the minimum node size.\footnote{The number of trees is greater than the default setting in \texttt{arf} for data generation, as we found in smaller simulations that this improves imputation performance (Figures~\ref{fig: min node num trees coverage} and~\ref{fig: min node num trees NRMSE}). We chose 10 as the minimum node size because this performed well for generating counterfactual explanations in Dandl et al. \cite{dandl2024countarfactuals}.} 

For setting II, we chose the same methods as in the previous settings except for median imputation, which, by construction, cannot generate multiple datasets. We create 20 datasets for multiple imputation for simplicity and due to computational restrictions. As \emph{MissForest} (without PMM), we consider the naïve approach of running MissForest several times, without an additional PMM step. 

\subsection{Results}
In the following, we present the results for single and multiple imputation methods on simulated data. In the first setting, we compare different single imputation methods and measure the performance with NRMSE and the Brier Score. In the second setting, we compare different multiple imputation methods and evaluate them by the coverage rate, average width of confidence intervals, and the RMSE of regression coefficients. To give a better overview of the results, we describe common patterns and show representative example plots in the figures below. 

\begin{figure}
    \centering
    \includegraphics[width=0.92\linewidth]{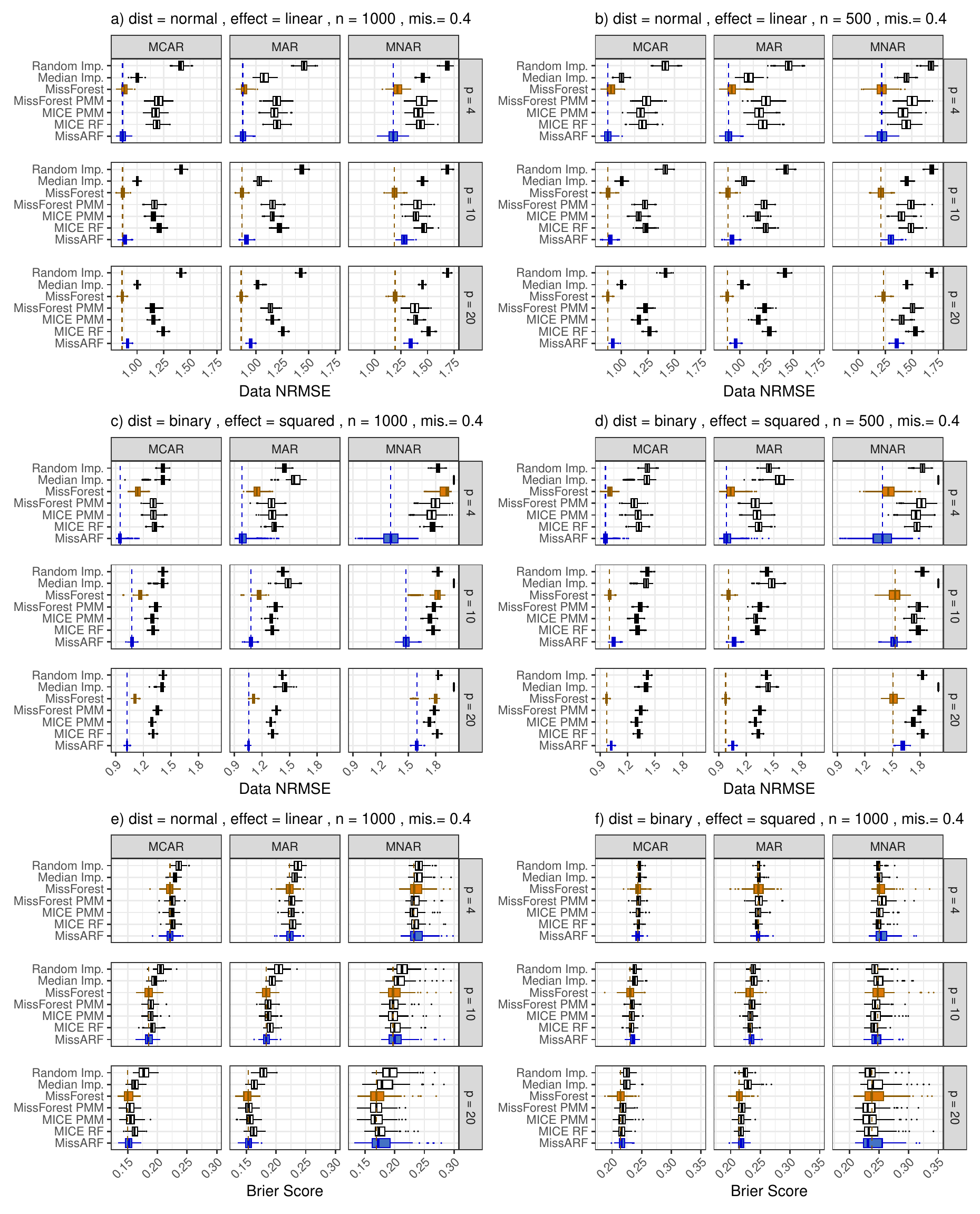}
    \caption{\textbf{Common patterns in simulation results of setting I} over different missingness patterns and dimensionality ($p$) for missingness rates (mis.) $0.4$. Common pattern for the NRMSE are exemplified for the normal distribution with a linear effect and a) $n=1000$ and b) $n=500$ and the binary distribution with a squared effect for c) $n=1000$ and d) $n=500$. The Brier score pattern is exemplified for e) normal distribution with a linear effect and $n=1000$ and f) the binary distribution with a squared effect and $n=1000$. For single imputation, MissARF (blue) and MissForest (orange) perform best in NRMSE (a-d) and Brier score (e). The boxplots are plotted over the replicates.}
    \label{fig: setting 1}
\end{figure}

\subsubsection*{Results of setting I: Single imputation}
\paragraph*{NRMSE}
For all distributions except the binary distribution, we notice a similar pattern, which is shown on an example plot in Figure~\ref{fig: setting 1}a), where the distribution is normal with a linear effect and $n= 1000$. Figure~\ref{fig: setting 1}a) shows the results for missingness proportion 0.4 and for the different number of features $p\in\{4,10,20 \}$. MissARF and MissForest perform the best depending on the number of features. While for $p=4$ MissARF dominates, for $p=10$ MissForest is slightly better or both perform equally well. For $p=20$, MissForest consistently attains the lowest error, with MissARF not far behind. 
In MCAR and MAR, median imputation performs next best after MissARF and MissForest, followed by MICE and MissForest PMM, which perform similarly well. For MNAR, median imputation is slightly worse than MICE and MissForest PMM. In all settings, random imputation performs the worst. This pattern is also seen across the missingness proportions, with the difference that median imputation is similar to MICE and MissForest PMM. Also for $n=500$ (Figure~\ref{fig: setting 1}b)) and $n=10\,000$ we notice a similar behaviour, with a slight advantage for MICE PMM compared to MICE RF and MissForest PMM for $n=500$. For the other distributions (except for binary), we observe a similar pattern. All the results can be found in the appendix in Figures~\ref{fig: logreg_nrmse_linear_normal_500}-\ref{fig: logreg_nrmse_linear_poisson} and Figures~\ref{fig: logreg_nrmse_squared_normal_500}-\ref{fig: logreg_nrmse_squared_poisson}. 

In contrast, for the binary distribution for both linear and squared effect, MissARF dominates in all settings for $n=1000$ and $n=10\,000$ (Figures~\ref{fig: logreg_nrmse_linear_binary} and~\ref{fig: logreg_nrmse_squared_binary}). This is exemplarily shown in Figure~\ref{fig: setting 1}c) for a squared effect, $n=1000$ and a fixed missingness proportion of 0.4 over the different $p$. Median imputation performs worse in this setting. For $n=500$ (Figure~\ref{fig: setting 1}d)) the difference between MissARF and MissForest is smaller, and as before MissARF is slightly better than MissForest for $p=4$ and MissForest is better otherwise (Figures~\ref{fig: logreg_nrmse_linear_binary_500} and~\ref{fig: logreg_nrmse_squared_binary_500}).%/(for higher $p$)

\paragraph*{Brier Score}
Generally, the differences in the Brier score are less pronounced than for the NRMSE. As a general trend, MissForest and MissARF perform the best, while random and median imputation perform the worst. MICE and MissForest PMM perform similarly and rank second-best. This pattern is exemplified in Figure~\ref{fig: setting 1}e) for the normal distribution with a linear effect and in Figure~\ref{fig: setting 1}f) for the binary distribution with a squared effect for $n=1000$ and $mis.=0.4$ over the different $p$.
This pattern can be observed across all distributions and all parameters, with all results shown in Appendix~\ref{sec:appx_A_Brier}.

\begin{figure}
    \centering
    \includegraphics[width=0.92\linewidth]{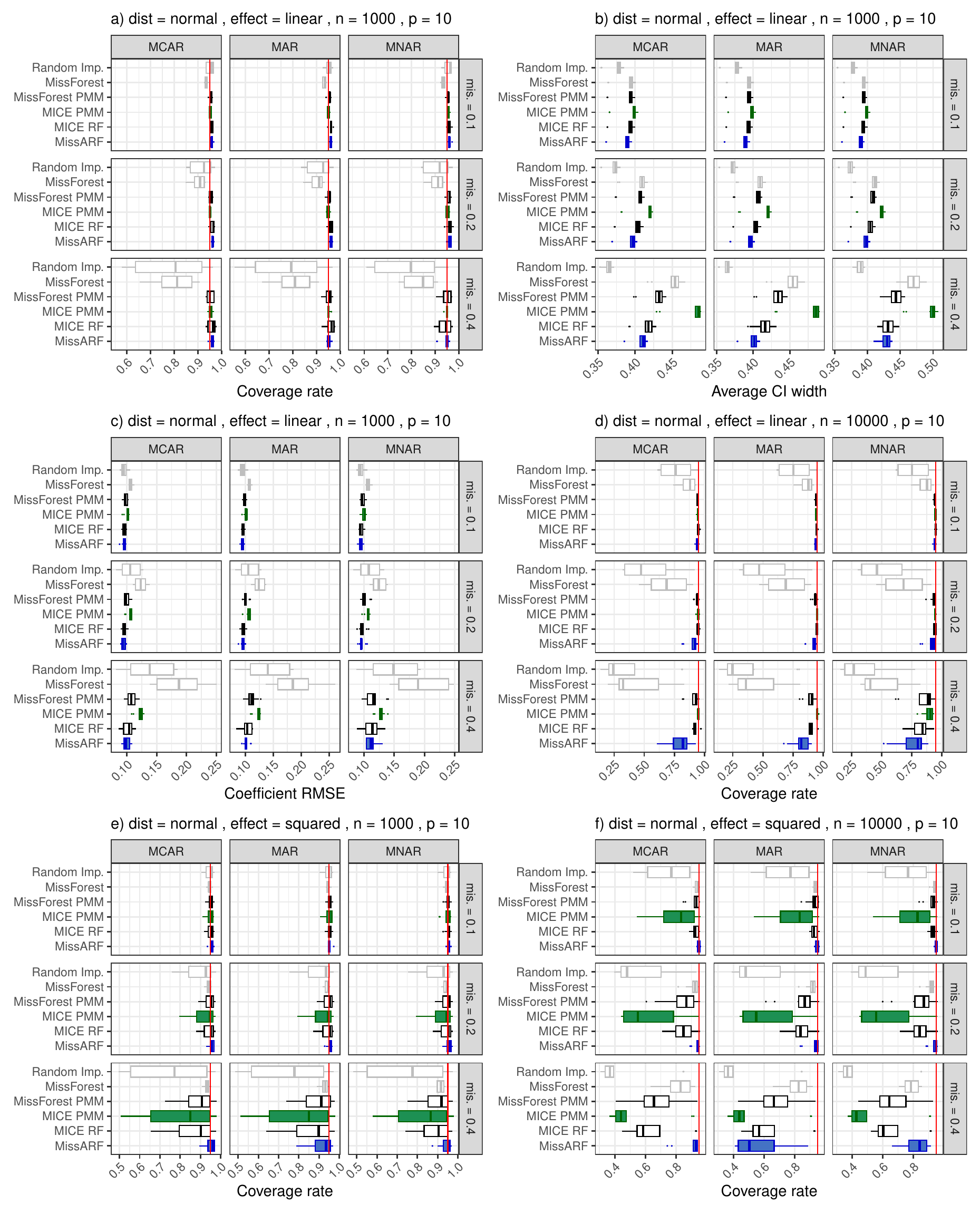}
    \caption{\textbf{Common patterns in simulation results of setting II} over different missingness patterns and rates (mis.) and dimensionality $p=10$. The coverage rate of the normal distribution is shown for a linear effect with a) $n=1000$ and d) $n=10\,000$ and a squared effect for e) $n=1000$ and f) $n=10\,000$. The red vertical line shows the nominal coverage level of 0.95. Random imputation and MissForest (gray) often show poor coverage. MICE PMM (green) works best for linear effects, whereas MissARF (blue) struggles with high missingness (d), but has the smallest average CI width (b) and RMSE (c), where MICE PMM has the largest. MissARF works best for squared effects, where MICE PMM struggles. The boxplots are plotted over the features.
    }
    \label{fig: setting 2}
\end{figure}

\subsubsection*{Results of setting II: Multiple imputation}
Unlike before, the results for the multiple imputation setting do not have a consistent pattern across settings. Nevertheless, we systematically grouped the results into three broader categories: \emph{1) Similar performance across all methods, MissARF with smallest average width. 2) PMM methods struggle, MissARF performs well. 3) All methods perform poorly}. While the behaviour for the linear effect is quite similar (category 1), we find more differences in the distributions for the squared effect (category 1-3). 
We first summarize the main results of categories 1 and 2 and illustrate them in Figure~\ref{fig: setting 2} with exemplary plots based on the normal distribution for $p=10$. 
For category 1, a common pattern for the coverage rate is illustrated in Figure~\ref{fig: setting 2}a). 
Here, MissForest and random imputation show the poorest performance. This is what we expect from the baseline methods and from MissForest because it is a single imputation method (naïvely repeated several times). In contrast, the other methods all show good results on a similar level. However, MissARF in some cases 
shows difficulties at higher levels of missingness, where MICE PMM typically performs best (Figure~\ref{fig: setting 2}d)). 
For the average CI width (Figure~\ref{fig: setting 2}b)) and the RMSE (Figure~\ref{fig: setting 2}c)) the methods perform similar, and differ more with growing missingness. Random imputation has the smallest width followed by MissARF, MICE RF and MissForest PMM. MICE PMM has a larger width in comparison with growing missingness proportions. Note that a smaller average width indicates better performance if coverage is close to 95\%. Regarding the RMSE with growing missingness, random imputation and MissForest get worse and MICE PMM is slightly larger than the other multiple imputation methods.
For category 2, a common pattern for the coverage is visualized for the normal distribution with squared effect in Figure~\ref{fig: setting 2}e-f). %(for $n=\{1000, 10\,000\}$). 
MissARF generally works best, while MICE PMM shows the poorest performance. Here, MICE RF and MissForest PMM perform better than MICE PMM but worse than MissARF. The average width remains similar across the settings (Figure~\ref{fig: logreg_aw_squared_normal}). The RMSE of the coefficients (Figure~\ref{fig: logreg_rmse_squared_normal}) often mirrors the results of the coverage rate: The closer a method is to 95\% in the coverage rate, the closer it is to 0 here, where small values indicate better performance.

In the following, we provide a more detailed description of the results in subcategories, where all plots are provided in the appendix.

\paragraph*{Category 1: Similar performance across all methods, MissARF with smallest average width}
\begin{description}
    \item[Gamma, normal, uniform, Poisson distribution with linear effect:]
    For gamma, normal and uniform distribution with linear effect and $n\in\{500, 1000\}$, the competing multiple imputation methods demonstrate similarly good coverage rates with a slight advantage for MICE PMM across the different missingness proportions and $p$. Random imputation and MissForest do not perform well with growing missingness. MICE PMM performs especially well when $p=4$ with a missingness proportion of 0.4---where the other methods perform poorer---as well as for $n=500$ in most distributions, and generally for the uniform distribution.  
    Regarding average width, MissARF often has the smallest along with random imputation and MICE RF not far behind, while MICE PMM has the largest. 
    At $n = 10\,000$ and considering the Poisson distribution for all $n$, we notice a poor performance of MissARF at high missingness rates. But also the other multiple imputation methods perform poorer, where often MICE PMM still maintains good coverage. For MNAR and Poisson and gamma distribution, all methods perform poorly. Notably, MissForest PMM performs poorly under $p=4$. (See figures in appendix: coverage: Figures~\ref{fig: logreg_coverage_linear_normal_500}-\ref{fig: logreg_coverage_linear_poisson}; average CI width: Figures~\ref{fig: logreg_aw_linear_normal_500}-\ref{fig: logreg_aw_linear_poisson}; RMSE: Figures~\ref{fig: logreg_rmse_linear_normal_500}-\ref{fig: logreg_rmse_linear_poisson}.)
    \item[Binary distribution with linear and squared effect:] 
    In most cases, all competing methods except MissForest generally perform well with MICE RF having an advantage for $p=4$ and MICE PMM otherwise. MissARF is often poorer for high missingness rates. For $p=4$, the PMM methods exhibit very high coverage rates, likely due to the large average width of the confidence intervals, where MICE RF and MissARF (except for mis.= 0.4) perform good. Surprisingly, random imputation shows strong performance, but with a sample size of $n=10\,000$ it cannot maintain its performance across all settings, leading to a drastic decrease in some cases. (See figures in appendix: coverage: Figures~\ref{fig: logreg_coverage_linear_binary_500}-\ref{fig: logreg_coverage_squared_binary}; average CI width: Figures~\ref{fig: logreg_aw_linear_binary_500}-\ref{fig: logreg_aw_squared_binary}; RMSE: Figures~\ref{fig: logreg_rmse_linear_binary_500}-\ref{fig: logreg_rmse_squared_binary}.)
\end{description}

\paragraph*{Category 2: PMM methods struggle, MissARF performs well}
\begin{description}
\item[Normal distribution with squared effect:] Among the methods, random imputation, MICE PMM, and MissForest PMM (in that order) exhibit poorer performance in the case of normal distribution with a squared effect. 
However, MissARF consistently performs well in this setting, in some cases along with MissForest and MICE RF. 
Notably, when most methods struggle (e.g., for $n\in \{500, 1000\}$, $p=4$ and mis.=$0.4$), MissARF often maintains a meaningful coverage rate. For $n=10\,000$ this pattern is more visible even for lower missingness rates. However, at 0.4 all methods perform poorly (except MissARF in one case in Figure \ref{fig: setting 2} f)).
The average width of the confidence intervals remains similar across methods, and the RMSE reflects the behavior of the coverage rate. Overall, we conclude that MissARF is the best-performing method in this setting. (See figures in appendix: coverage: Figures~\ref{fig: logreg_coverage_squared_normal_500} and~\ref{fig: logreg_coverage_squared_normal}; average CI width: Figures~\ref{fig: logreg_aw_squared_normal_500} and~\ref{fig: logreg_aw_squared_normal}; RMSE: Figures~\ref{fig: logreg_rmse_squared_normal_500} and~\ref{fig: logreg_rmse_squared_normal}.)

\item[Uniform distribution with squared effect:] 
For $n\in\{500, 1000\}$, all methods perform very well, where MissForest and MissARF have the best coverage rates. While MissForest tends to under-cover, the others mostly tend towards overly conservative confidence intervals. 
For $n=10\,000$, all methods have a similarly good coverage rate except for a missingness proportion of 0.4. Here, all methods tend towards under-coverage while random imputation and MICE PMM perform worst (similar to normal squared $n=500$). While for $n=10\,000$ the RMSE reflects the coverage rate results, the methods for $n\in\{500, 1000\}$ show similarly good RMSE values, with MissARF tending to be slightly larger, followed by MissForest at a greater distance. For all $n$, the average widths of the confidence intervals of the methods are very similar, except MissForest, which always has a smaller average width. (See figures in appendix: coverage: Figures~\ref{fig: logreg_coverage_squared_uniform_500} and~\ref{fig: logreg_coverage_squared_uniform}; average CI width: Figures~\ref{fig: logreg_aw_squared_uniform_500} and~\ref{fig: logreg_aw_squared_uniform}; RMSE: Figures~\ref{fig: logreg_rmse_squared_uniform_500} and~\ref{fig: logreg_rmse_squared_uniform}.)
\end{description}

\paragraph*{Category 3: All methods perform poorly}
\begin{description}
\item[Poisson and gamma distribution with squared effect:]
For Poisson and gamma distribution with a squared effect, all methods have difficulties, i.e., the coverage rate is far away from 95\%, sometimes even close to zero. The higher $n$, $p$, or missingness proportion, the worse the performance gets. In very few cases, the MICE methods and especially MissForest still attain meaningful coverage rates. A notable exception is the setting with $n\in\{500, 1000\}$, $p=4$ and missingness proportion 0.1, where the coverage rate for all methods except random imputation is relatively close to 95\%. For the cases, where the coverage is valid, the RMSE and the average width are very similar as well. (See figures in appendix: coverage: Figures~\ref{fig: logreg_coverage_squared_gamma_500}-\ref{fig: logreg_coverage_squared_poisson}; average CI width: Figures~\ref{fig: logreg_aw_squared_gamma_500}-\ref{fig: logreg_aw_squared_poisson}; RMSE: Figures~\ref{fig: logreg_rmse_squared_gamma_500}-\ref{fig: logreg_rmse_squared_poisson}.)
\end{description}

\subsubsection*{Aggregated simulation results and runtime}
Table~\ref{tab:aggr_results} provides aggregated results over all simulation settings, including single imputation (data NRMSE, Brier score) and multiple imputation (coverage rate, confidence interval width, RMSE of coefficients). 
For single imputation, MissARF and MissForest perform best in both the data NRMSE and the Brier score, followed by MissForest PMM, the two MICE methods and median imputation. Random imputation performs worst. As observed before, median imputation performs relatively well in terms of NRMSE but not as well regarding the Brier score. 

For multiple imputation, MissForest PMM, MICE PMM, MICE RF and MissARF all are relatively close to the nominal level of 95\%, ranging between 94.9\% (MICE PMM) and 95.3\% (MissForest PMM). MissForest (without PMM) and random imputation both have a poorer coverage rate. Random imputation, MissARF and  MICE RF show the smallest average confidence interval width. However, since the coverage rate of random imputation is generally far away from the nominal level, the results for random imputation should be interpreted with care. The coefficient RMSE is the smallest with MissForest PMM, MICE RF and MissARF and largest for random imputation and MissForest.

Table~\ref{tab:aggr_results} shows single-threaded runtime for simulated data with a normal distribution, linear effect and MAR, averaged over different numbers of features $p \in \{4, 10, 20\}$, $n\in\{500, 1000, 10\,000\}$ and missingness proportions $mis. \in \{0.1, 0.2, 0.4\}$. For single imputation, the simple baseline methods are the fastest, followed by the two MICE variants. MissForest, MissForest PMM and MissARF are slowest. For multiple imputation, MICE PMM is the fastest non-baseline method, followed by MissARF and, with a larger difference, MICE RF. While the MICE and MissForest methods are considerably slower for multiple imputation (as expected, approx. 20x slower for 20 multiple imputation), this is not the case for MissARF, which is similarly fast for both single and multiple imputation. 
More detailed runtime results over different $n$, $p$ and $mis.$ are presented in Appendix~\ref{sec:appx_runtime}. Generally, runtime tends to increase with growing $n$ and $p$, while the missingness proportion has little impact on the runtime. Using multithreading with 16 threads, MissARF is the fastest non-baseline method, followed by MICE PMM, MissForest and MICE RF (Table \ref{tab:runtime k16}). All methods except the baseline learners and MICE PMM are considerably faster when using multithreading, while MissForest and MissARF benefit most from it.

\begin{table}[ht]
\centering
\fontsize{8pt}{10pt}\selectfont
\begin{tabular}[t]{l|llr|lllr}
\toprule
%Method & NRMSE & Brier & Runtime\_single & Coverage & AvgWidth & RMSE & Runtime\_multi\\
& \multicolumn{3}{c}{Single imputation} & \multicolumn{4}{c}{Multiple imputation}\\
Method & Data NRMSE & Brier score & Runtime (s) & Coverage rate \% & CI width & Coef. RMSE& Runtime (s)\\
 & [mean (SD)] & [mean (SD)] & [mean] & [median (IQR)] & [median (IQR)] &  [median (IQR)] & [mean] \\
\midrule
Random Imp. & 1.47 (0.11) & 0.20 (0.05) & <0.1 & 92.0 (46.8) & 0.30 (0.47) & 0.13 (0.11) & <0.1\\
Median Imp. & 1.21 (0.29) & 0.19 (0.06) & <0.1 & - & - & - & -\\
MissForest & 0.98 (0.19) & 0.18 (0.06) & 16.1 & 90.9 (13.1) & 0.32 (0.47) & 0.12 (0.14) & 322.7\\
MissForest PMM & 1.26 (0.14) & 0.19 (0.06) & 16.5 & 95.3 (6.0) & 0.32 (0.48) & 0.10 (0.11) & 325.2\\
MICE PMM & 1.23 (0.13) & 0.19 (0.06) & 0.2 & 94.9 (3.3) & 0.34 (0.50) & 0.11 (0.13) & 3.9\\
MICE RF & 1.27 (0.13) & 0.19 (0.06) & 4.0 & 95.2 (6.2) & 0.31 (0.48) & 0.10 (0.11) & 78.5\\
MissARF & 0.97 (0.16) & 0.18 (0.06) & 14.4 & 95.0 (9.2) & 0.31 (0.48) & 0.10 (0.10) & 14.4\\
\bottomrule
\end{tabular}
    \caption{Aggregated results over all simulation settings for both single and multiple imputation. Data NRMSE and Brier score refer to performance in the single imputation setting, coverage rate, confidence interval (CI) width and coef. RMSE to performance in multiple imputation.}
    \label{tab:aggr_results}
\end{table}

\begin{figure}
    \centering
    \includegraphics[width=\linewidth]{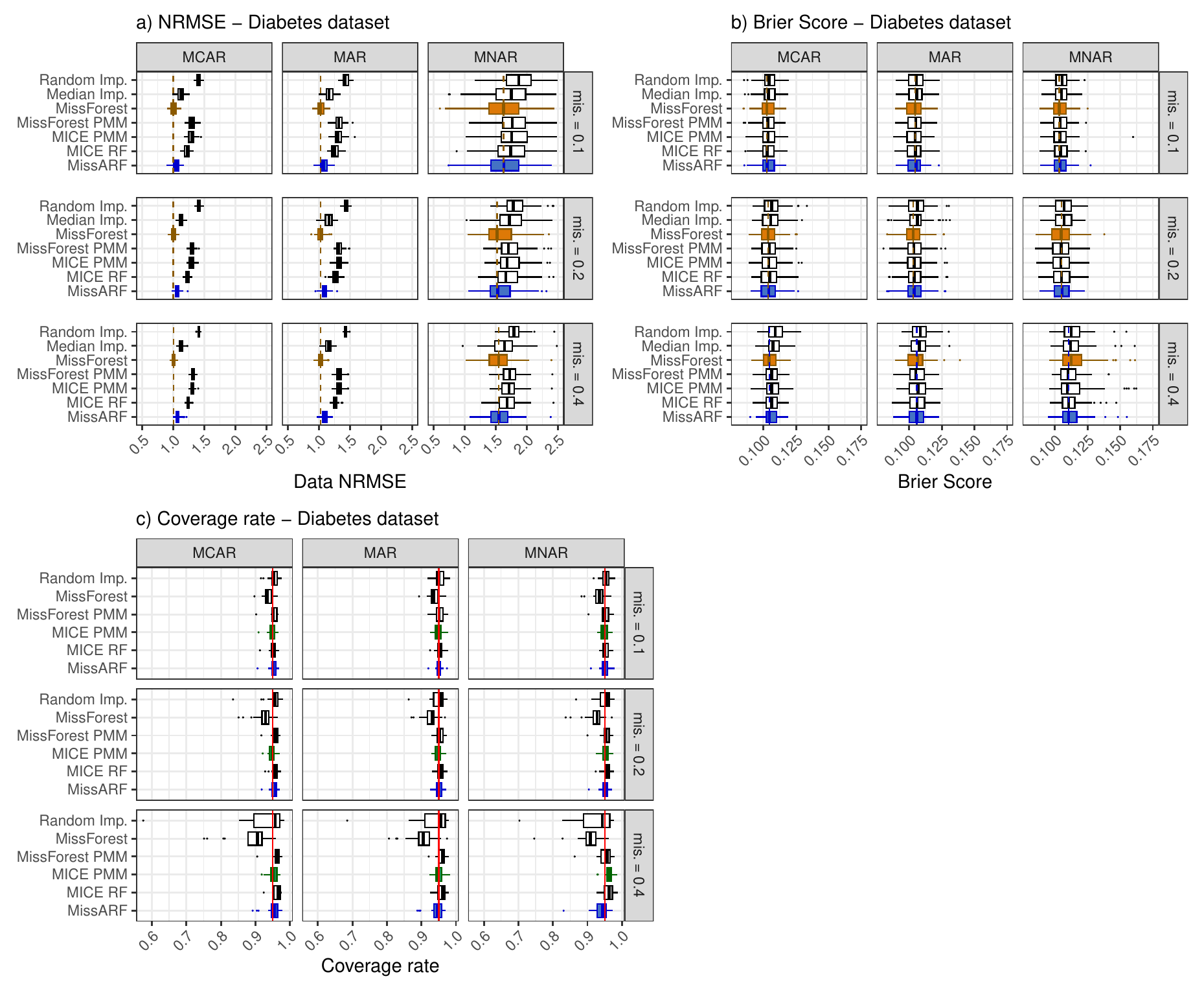}
    \caption{\textbf{Real data example results} over the different
missingness proportions and patterns. a) MissForest (orange) and
MissARF (blue) perform best in terms of NRMSE. b) The methods perform similarly regarding the Brier Score, with random and median imputation performing slightly worse. The boxplots are plotted over the features. c) MissARF, MICE PMM (green) and the other methods, except MissForest and random imputation, perform similarly well. The red vertical line shows the nominal coverage level of 0.95. The boxplots are plotted over the replicates. }
    \label{fig: real data results}
\end{figure}
\section{Real data example} \label{sec: 4 Real data example}
As a real data example, we use the diabetes health indicators dataset\cite{Diabetes-dataset}, where the aim is to predict diabetes from healthcare statistics and lifestyle survey information. The data was collected by the centers for disease control and prevention (CDC) in a health-related telephone survey behavioral risk factor surveillance system (BRFSS) in 2015. The dataset has 253,680 observations and 21 binary and integer features. The features include, e.g., the body mass index (BMI), smoking status, age and education, among others. The outcome is binary, indicating whether a person was ever diagnosed with diabetes or prediabetes ($Y=1$) or not ($Y=0$), according to a questionnaire. 

We use a similar approach as in the simulation studies described in Section~\ref{sec: 3 Experiments}: We simulate missing data according to MCAR, MAR and MNAR patterns as described in Section~\ref{sec: setup}\footnote{Note, that if the missingness rate exceeds the group size, the missingness rate is adjusted downwards.}, impute the datasets with different imputation methods for both single and multiple imputation, and compare those methods with the data NRMSE and Brier score for single imputation as well as coverage rate, average width and coefficient RMSE for multiple imputation. For multiple imputation, we fit a logistic regression model on the complete dataset and use the resulting estimated coefficients $\hat{\bm{\beta}}$ as ground-truth values to calculate, e.g., the coefficient RMSE and the coverage rate. In each replication, we sampled 1000 observations from the total 253,680 observations in the dataset and calculated the regression coefficients, confidence intervals, etc., only on these 1000 observations. However, we estimated the true values for the confidence interval coverage using all 253,680 observations. We deliberately choose such a huge dataset to minimize the variance and to approximate the population parameters $\bm{\beta}$ by $\hat{\bm{\beta}}$. We argue that the variance of the logistic regression on the full dataset with 253,680 observations is negligible in comparison to the variance induced by sampling 1000 observations and by the imputation. 

Figure~\ref{fig: real data results} shows the results for single imputation: a) for the data NRMSE and b) for the Brier score. Generally, the trends observed in the simulated data are also noticeable in the real dataset example. In terms of NRMSE, both MissARF and MissForest show the best performance. They are followed by median imputation and by MICE and MissForest PMM, while random imputation performs worst. Regarding the Brier score, all methods perform equally well. 
For multiple imputation, the results are shown in Figure~\ref{fig: real data results}c) for the coverage rate. Average width and coefficient RMSE results are shown in Figure~\ref{fig: real data rest} in the appendix. Generally, all methods perform quite well in terms of the median value in most settings. Concerning the coverage rate (Figure~\ref{fig: real data results}c)), MissForest is overly liberal compared to the other methods but the median is still above 90\% coverage. At a missingness rate of 0.4, random imputation has a wider interquartile range and therefore performs slightly worse. At a missingness rate of 0.4, MICE PMM, although it has a similar median, has an exceptionally wide range regarding the average width (Figure~\ref{fig: real data rest}b)) and the RMSE (Figure~\ref{fig: real data rest}c)), indicating that it performs poorly for some individual variables.

\section{Discussion} \label{sec: 5 Discussion}
In this paper, we proposed MissARF, a fast and easy-to-use imputation method based on generative modeling that offers both single and multiple imputation. We compared MissARF against state-of-the-art imputation methods on a real data example and on simulated data over different multivariate data distributions in different settings. We found that, in general, MissARF has comparable performance to MissForest in single imputation and to MICE in multiple imputation. In contrast to the competing methods, MissARF does not require any additional computational cost for multiple imputation. 

In more detail, we observed that, in single imputation, MissARF works best for binary data regardless of the dimensionality for $n\in \{1000, 10\,000\}$ and for other distributions MissARF works best in low or moderate dimensions, while MissForest performs better in higher dimensions.
In multiple imputation, MissARF generally has comparable performance to MICE PMM, the de facto standard method for multiple imputation. MissARF performs particularly well for squared effects. The rather poor performance of MICE PMM is expected from PMM methods in these cases, as it is documented that other built-in univariate imputation methods in MICE are recommended for non-linear effects \cite{doove2014recursive, van2018flexible}. Here, MICE RF is an alternative, which, however, does not perform as well as MissARF.

The exceptional performance of MissARF for binary data in single imputation could be explained by the general strength of ARF for categorical data, as it has a dedicated procedure for this and does not rely on encoding schemes or similar\cite{blesch2025conditional,dandl2024countarfactuals}. In general, generative approaches have difficulties with high-dimensional data \cite{watsonAdversarialRandomForests2023}, which may explain why MissARF currently performs best in low to moderate-dimensional settings. However, since tree-based methods are known to scale well with high dimensionality \cite{grinsztajn2022}, we see potential in MissARF to be extended and improved for high-dimensional applications in future work. In some settings of multiple imputation, MissARF has difficulties with higher proportions of missingness. Here, the average interval width of MissARF confidence intervals is comparably small, indicating that the multiple imputation would benefit from a larger diversity of imputed values. 
A possible solution is to tune ARF with the minimum node size. The larger the leaf node size, the more different variable values are considered for each imputed value, which could be helpful if the missingness rate is high (see additional experiments in Appendix~\ref{sec:appx_further}, where on a fixed example, we compared different minimum node sizes, number of trees and higher number of multiple imputations). That all methods perform poorly for Poisson and gamma distributions with squared effects is not surprising because it is a difficult setting to learn, since these distributions are skewed. With a squared effect, the values get more extreme, which makes imputation more difficult. The performance often gets worse for $n=10\,000$. One possible reason for this is that, when $n$ is large, confidence intervals get narrower, so bias induced by the imputation model could lead to non-coverage more often.

In future work, other methods to handle missing values within ARF training could be considered. Several such methods have been proposed \cite{tang2017random, josse2024consistency} for RF. However, it is still unclear which methods perform best in different settings. Further, good performance in supervised learning does not necessarily imply good performance in generative models, and a dedicated benchmark of such methods for the generative setting would be necessary. Another option is to handle the missing values before applying ARF, i.e., apply another imputation method first. However, by that, the performance of MissARF would depend on a successful pre-imputation and the imputation uncertainty of the pre-imputation would be ignored, leading to bias in imputation with ARF. Furthermore, the experiments have not been conducted on smaller sample sizes, which could arise in practice, e.g., in clinical research. The performance of all imputation methods decreases with smaller sample sizes and should be addressed in future work (see additional experiments in Appendix~\ref{sec:appx_further_small}, where we compared sample sizes between 50 and 500 on a fixed example).

In summary, MissARF offers an imputation solution based on generative modeling that is fast, easy to use, and performs well in both single and multiple imputation. Our method can be used as a general-purpose imputation approach in many settings. We recommend using MissARF in the following cases: (1) when both performance evaluation and inference are conducted on the same dataset and require a consistent imputation approach; (2) for single imputation in low-dimensional or binary data settings; and (3) for multiple imputation when missingness is low, squared effects are present, the data distribution is unknown, or a fast runtime is needed—e.g. when MICE PMM is unsuitable. 
Further research is required to improve MissARF's performance in very high dimensional settings and for high missingness proportions.

\section*{Author contributions}
PG and MNW performed the experiments. PG, JK, DSW and MNW implemented the software. PG wrote the first draft. All authors conceived the original
idea, discussed the results and contributed to the final manuscript.

\section*{Acknowledgments}
PG and MNW were supported by the German Research Foundation (DFG), Grant Number 437611051. 
MNW was supported by the German Research Foundation (DFG), Grant Number 459360854. 
MNW and JK were supported by the U Bremen Research Alliance/AI Center for Health Care, financially supported by the Federal State of Bremen. We thank Lukas Burk for running the 
experiments on the Beartooth Computing Environment \cite{beartooth} and Kristin Blesch for valuable discussions. Furthermore, we thank Stef van Buuren for valuable discussions and for giving good input for future work. Finally, we thank the anonymous reviewers of Statistics in Medicine for the thorough review, which we believe contributed positively to our work.

% \bmsection*{Financial disclosure}
% ??

% \bmsection*{Conflict of interest}
% ??
\bibliographystyle{unsrt}
\bibliography{missARF_sources.bib}

% \section*{Supporting information}
% The supplementary material includes additional results for the simulation studies and the real data example.

\clearpage

%\renewcommand{\appendixpagename}{Supplementary Material}
%\begin{appendices}
\newpage

\appendix
\label{ap}

\section{Setting I: Single imputation}
\subsection{NRMSE}
\subsubsection{Linear effect}
\begin{figure}[!ht]
\centering
\includegraphics[width=0.92\linewidth]{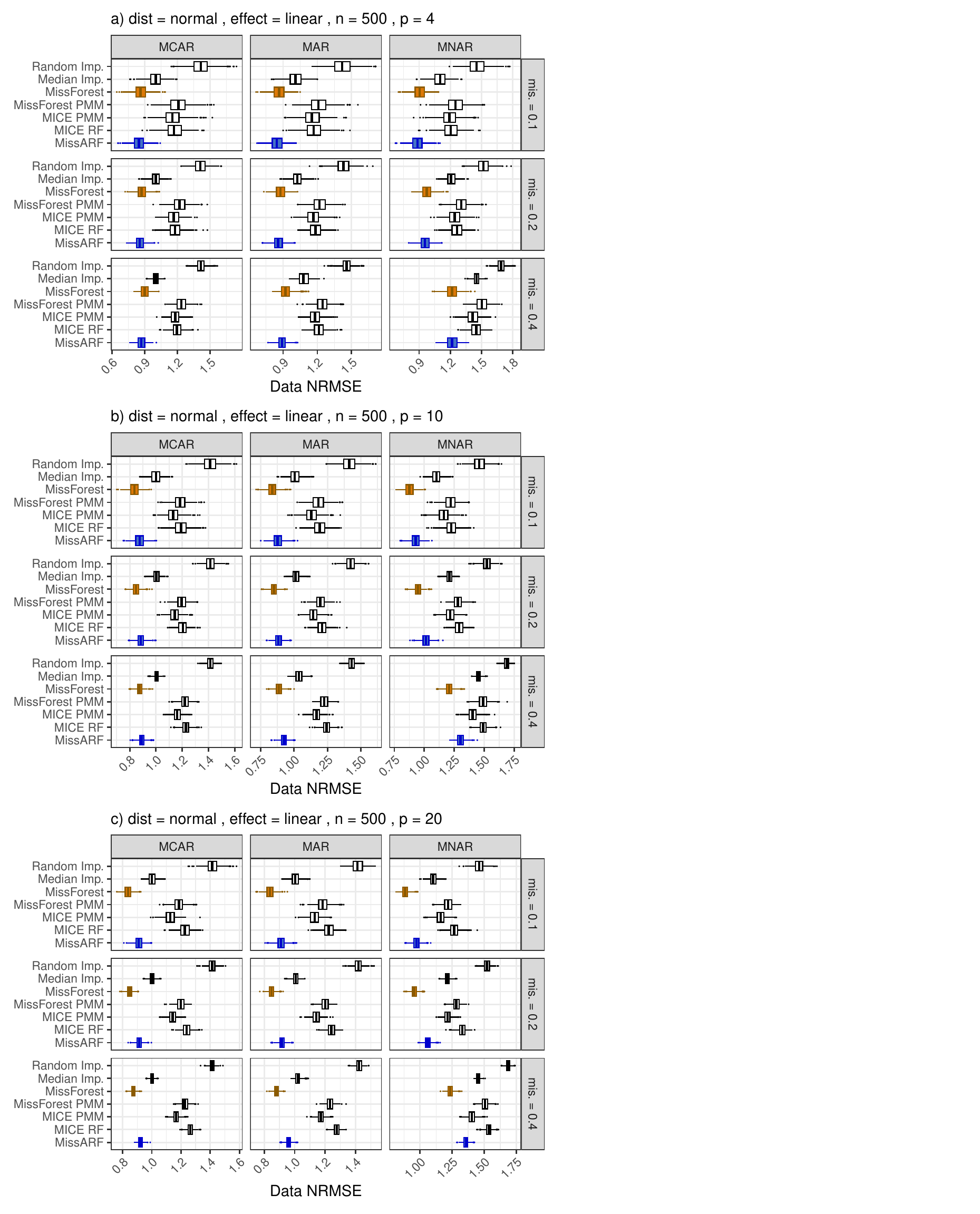}
\caption{\textbf{NRMSE} of the normal distribution setting with a linear effect over different missingness patterns, dimensionality ($p$) and missingness rates (mis.) with $n=500$. The boxplots are plotted over the replicates, with MissARF (blue) and MissForest (orange) highlighted.} \label{fig: logreg_nrmse_linear_normal_500}
\end{figure}

\begin{figure}[p]
\centering
\includegraphics[width=0.92\linewidth]{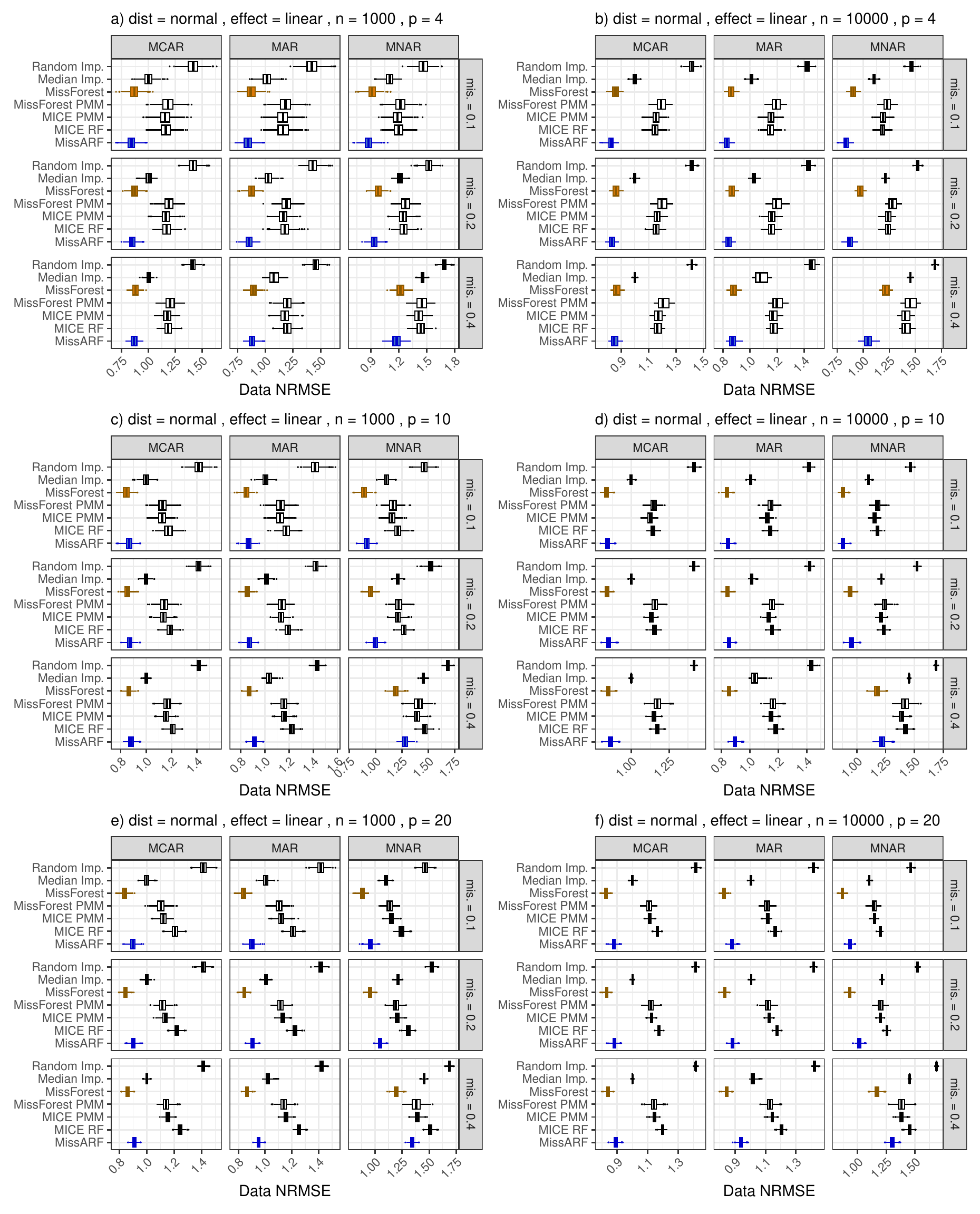}
\caption{\textbf{NRMSE} of the normal distribution setting with a linear effect over different missingness patterns, dimensionality ($p$) and missingness rates (mis.) with $n=1000$ (left) and $n= 10,000$ (right). The boxplots are plotted over the replicates, with MissARF (blue) and MissForest (orange) highlighted.} \label{fig: logreg_nrmse_linear_normal}
\end{figure}

\begin{figure}[p]
\centering
\includegraphics[width=0.92\linewidth]{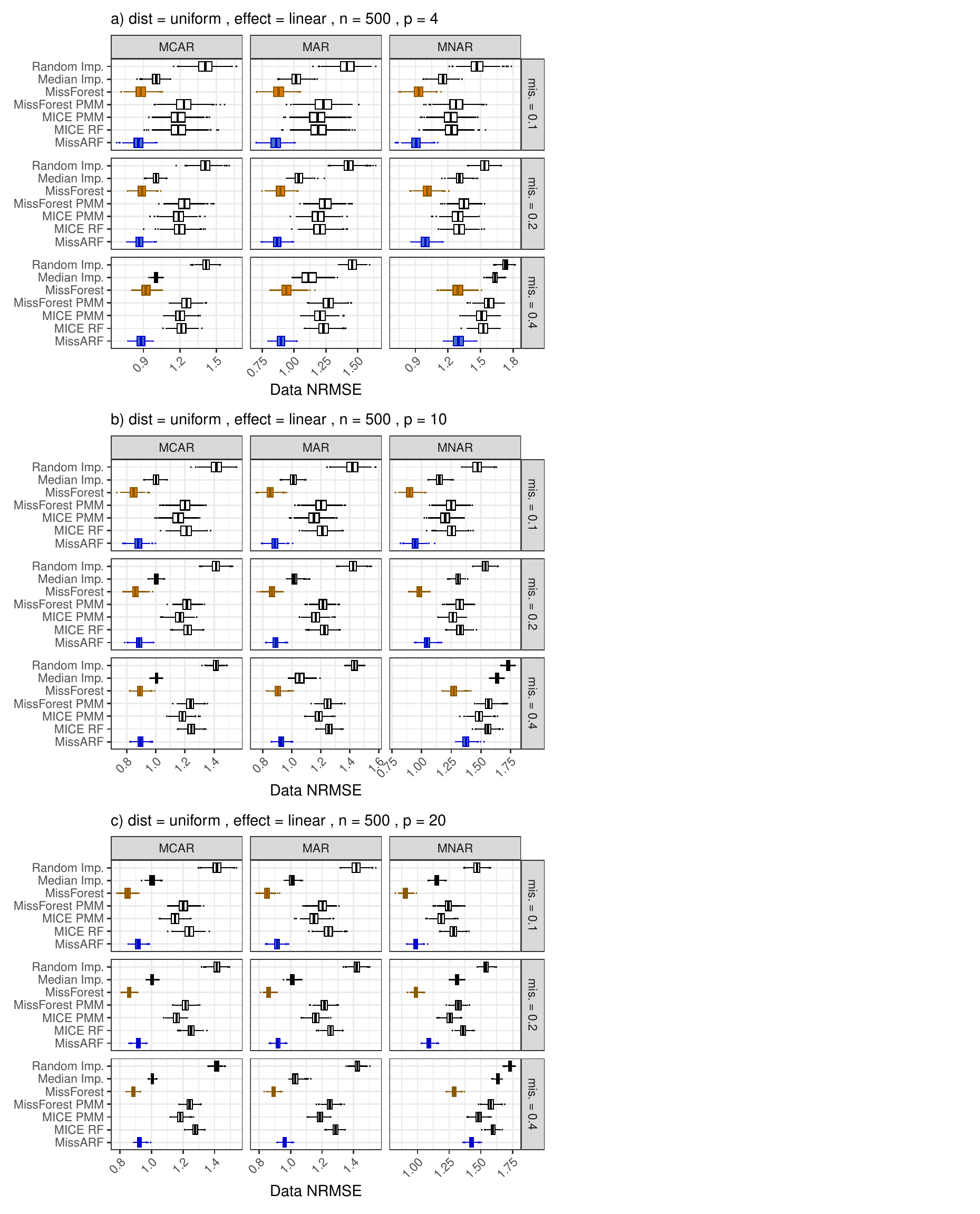}
\caption{\textbf{NRMSE} of the uniform distribution setting with a linear effect over different missingness patterns, dimensionality ($p$) and missingness rates (mis.) with $n=500$. The boxplots are plotted over the replicates, with MissARF (blue) and MissForest (orange) highlighted.} \label{fig: logreg_nrmse_linear_uniform_500}
\end{figure}

\begin{figure}[p]
\centering
\includegraphics[width=0.92\linewidth]{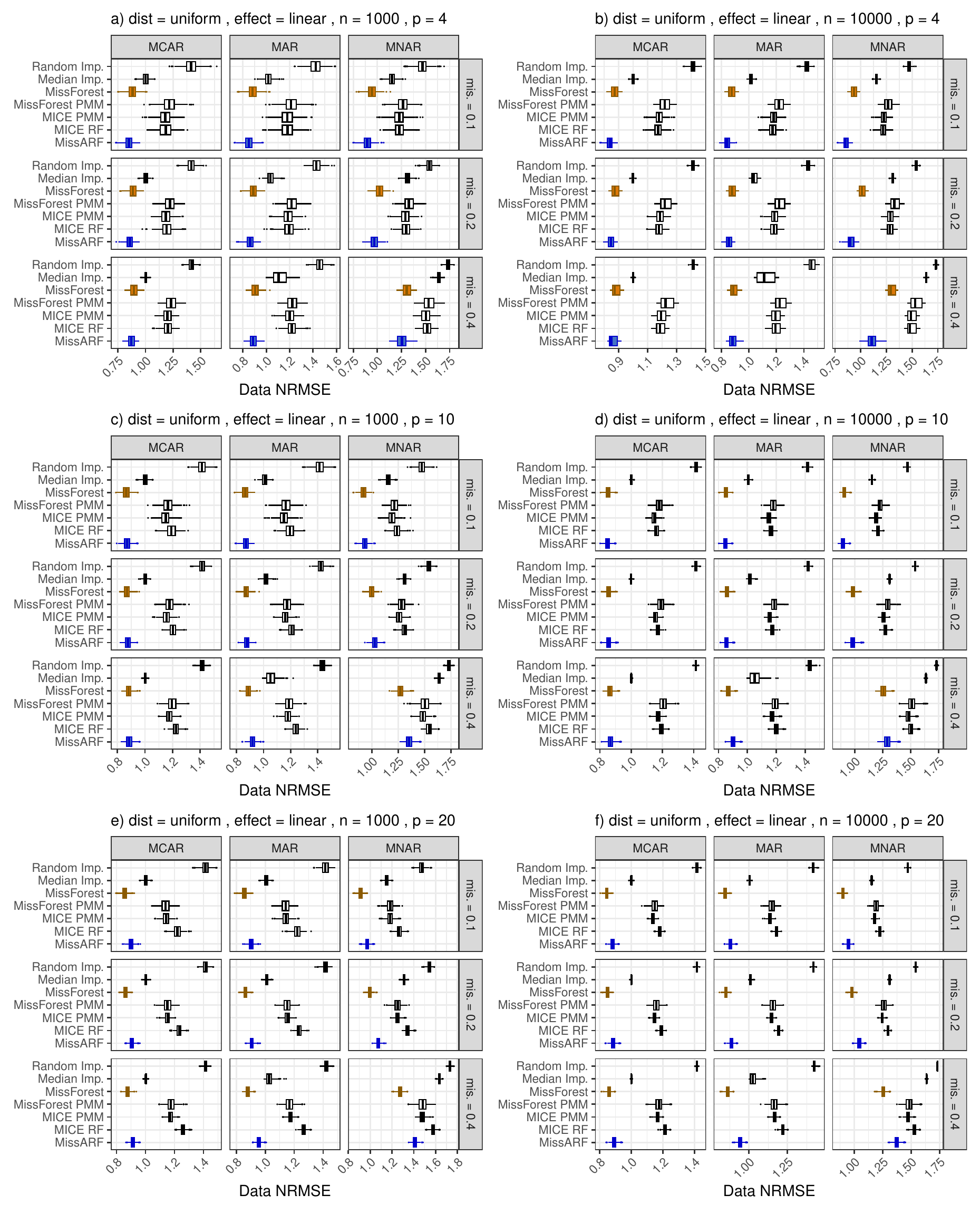}
\caption{\textbf{NRMSE} of the uniform distribution setting with a linear effect over different missingness patterns, dimensionality ($p$) and missingness rates (mis.) with $n=1000$ (left) and $n= 10,000$ (right). The boxplots are plotted over the replicates, with MissARF (blue) and MissForest (orange) highlighted.} \label{fig: logreg_nrmse_linear_uniform}
\end{figure}

\begin{figure}[p]
\centering
\includegraphics[width=0.92\linewidth]{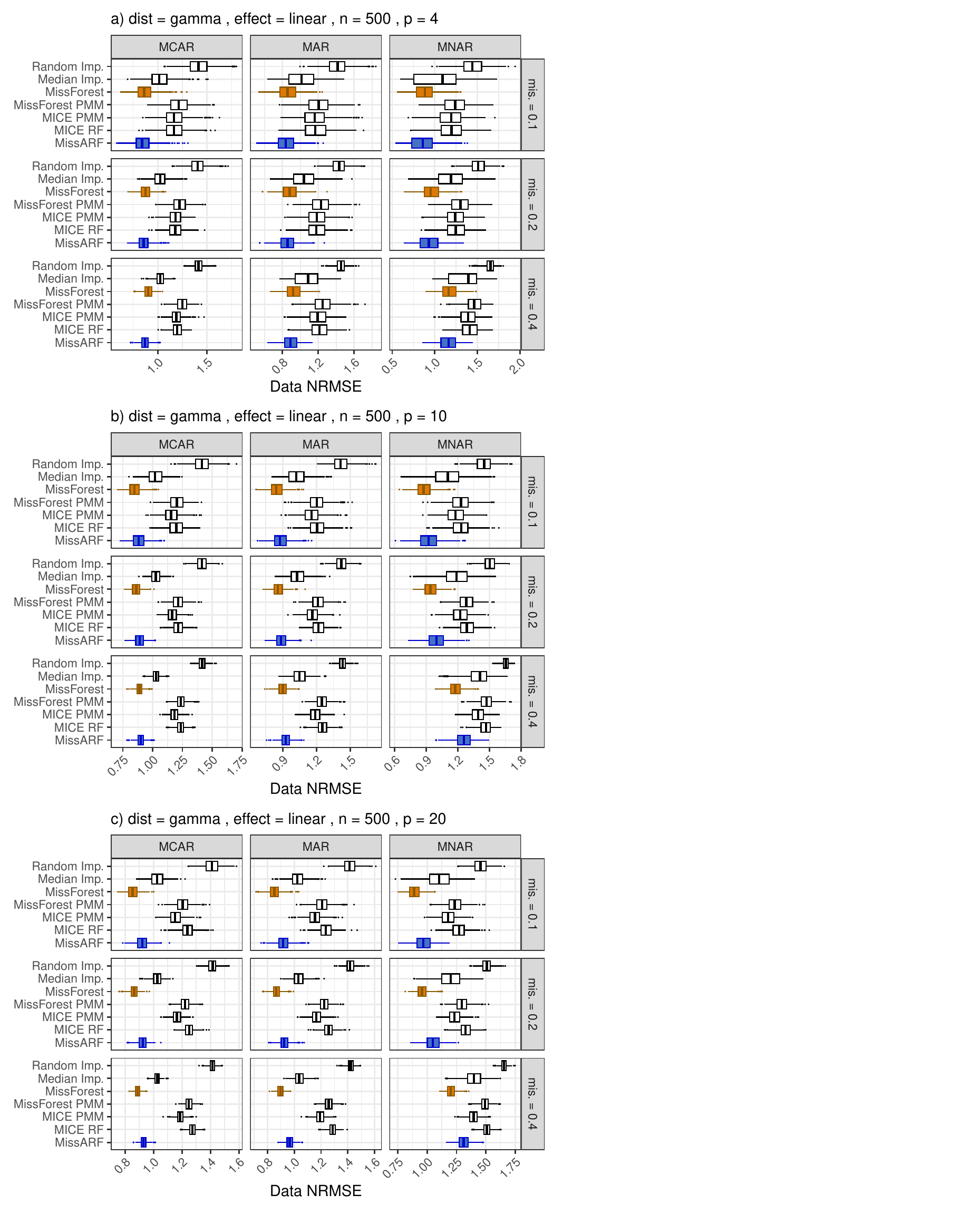}
\caption{ \textbf{NRMSE} of the gamma distribution setting with a linear effect over different missingness patterns, dimensionality ($p$) and missingness rates (mis.) with $n=500$. The boxplots are plotted over the replicates, with MissARF (blue) and MissForest (orange) highlighted.} \label{fig: logreg_nrmse_linear_gamma_500}
\end{figure}

\begin{figure}[p]
\centering
\includegraphics[width=0.92\linewidth]{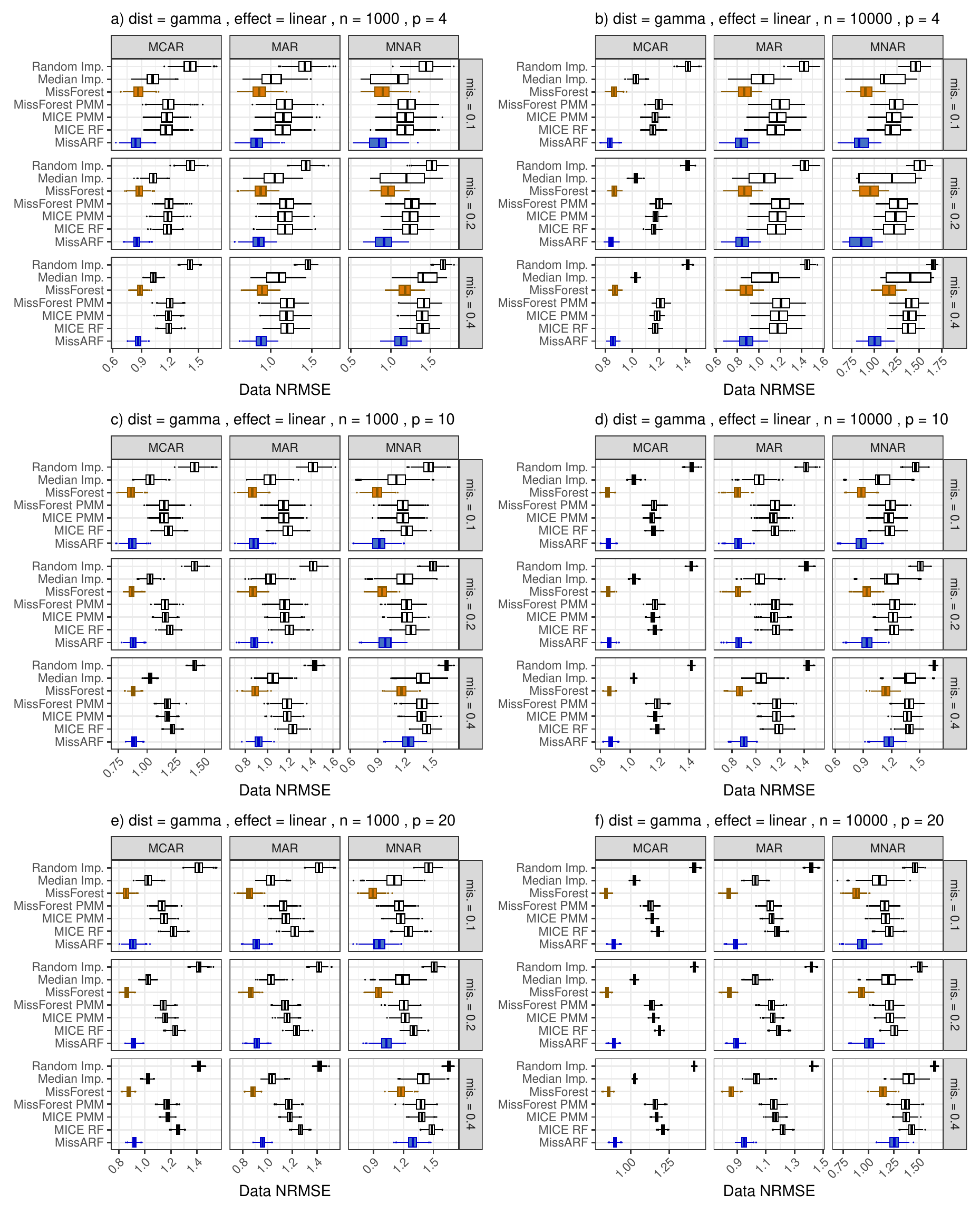}
\caption{ \textbf{NRMSE} of the gamma distribution setting with a linear effect over different missingness patterns, dimensionality ($p$) and missingness rates (mis.) with $n=1000$ (left) and $n= 10,000$ (right). The boxplots are plotted over the replicates, with MissARF (blue) and MissForest (orange) highlighted.} \label{fig: logreg_nrmse_linear_gamma}
\end{figure}

\begin{figure}[p]
\centering
\includegraphics[width=0.92\linewidth]{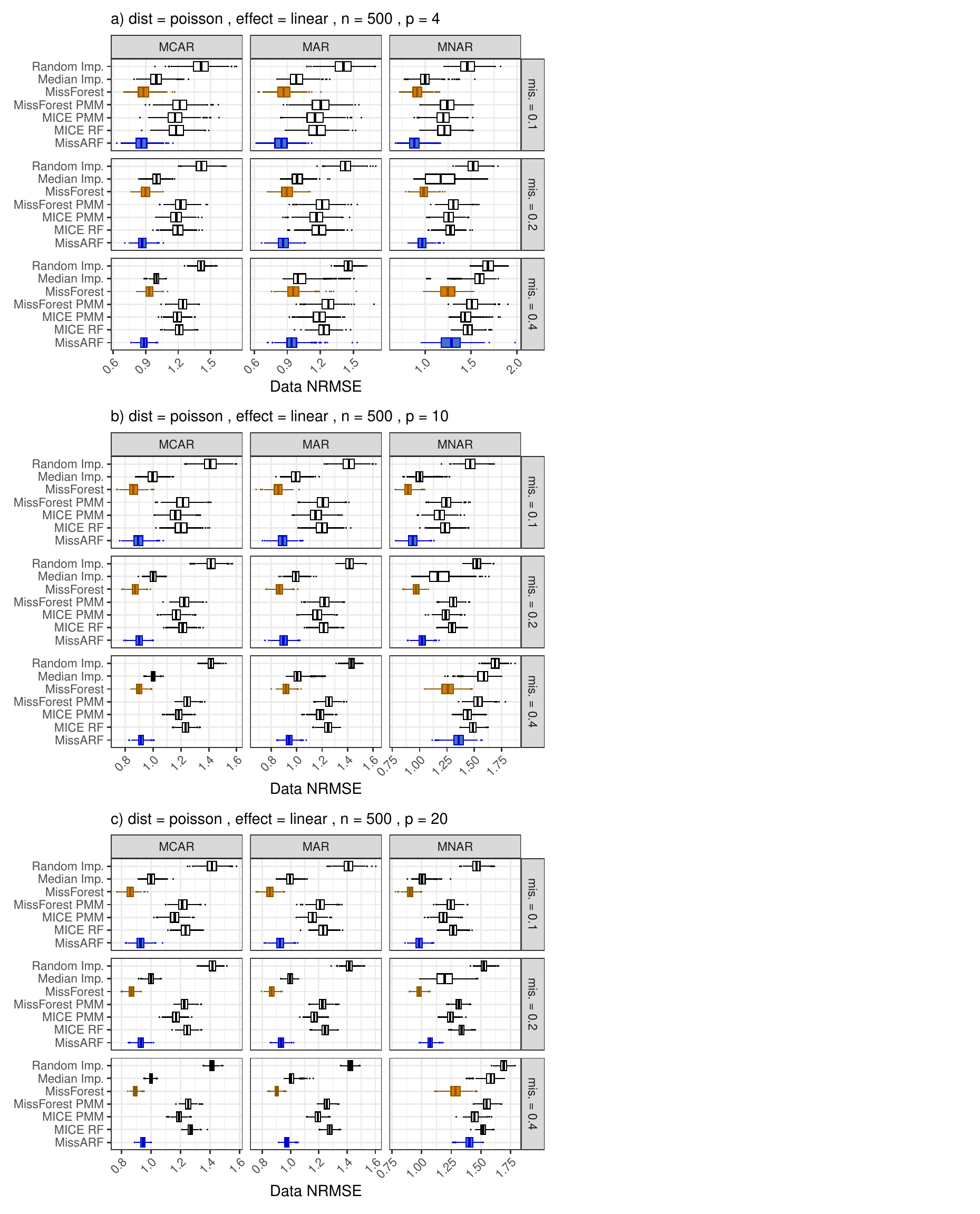}
\caption{ \textbf{NRMSE} of the Poisson distribution setting with a linear effect over different missingness patterns, dimensionality ($p$) and missingness rates (mis.) with $n=500$. The boxplots are plotted over the replicates, with MissARF (blue) and MissForest (orange) highlighted.} \label{fig: logreg_nrmse_linear_poisson_500}
\end{figure}

\begin{figure}[p]
\centering
\includegraphics[width=0.92\linewidth]{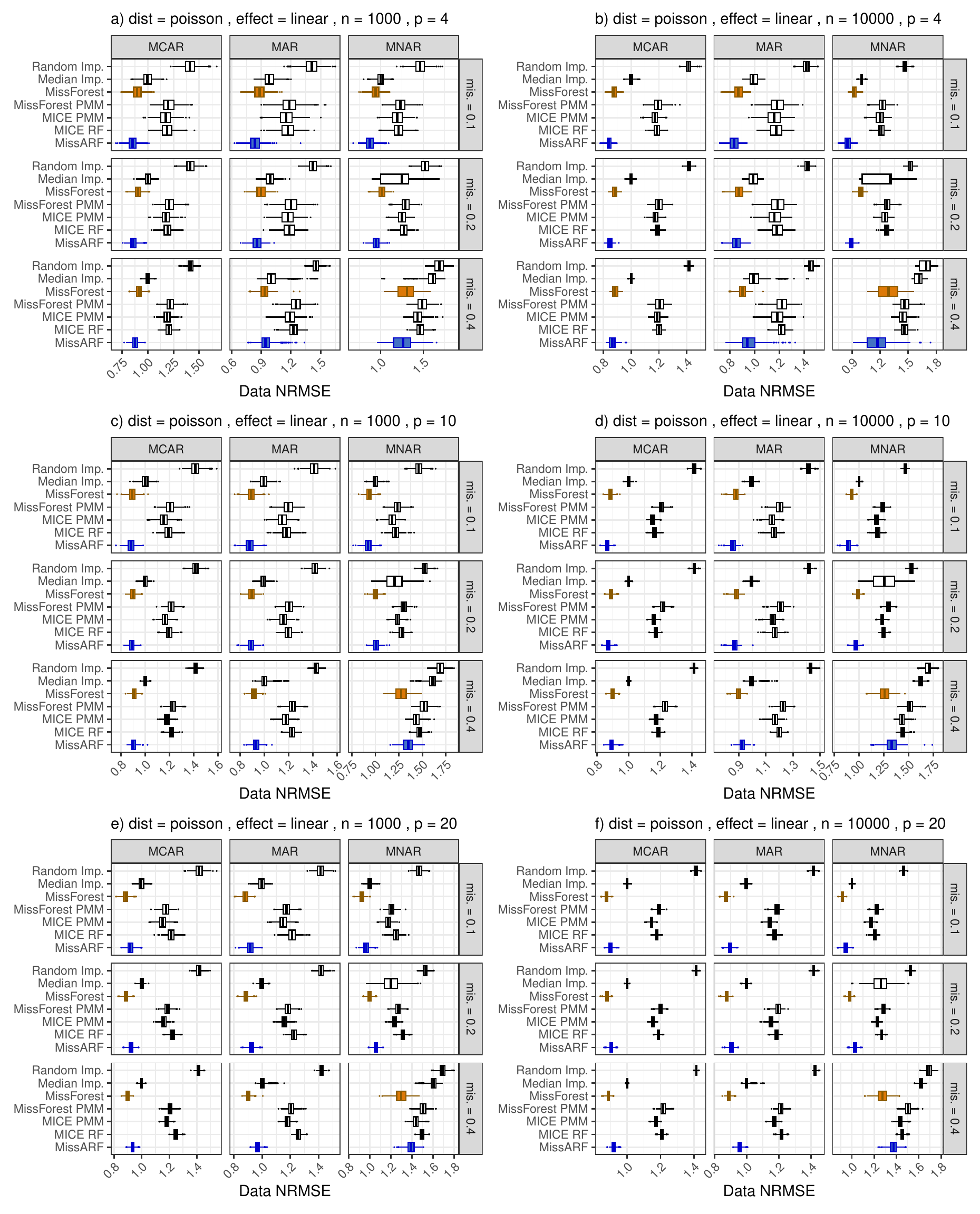}
\caption{ \textbf{NRMSE} of the Poisson distribution setting with a linear effect over different missingness patterns, dimensionality ($p$) and missingness rates (mis.) with $n=1000$ (left) and $n= 10,000$ (right). The boxplots are plotted over the replicates, with MissARF (blue) and MissForest (orange) highlighted.} \label{fig: logreg_nrmse_linear_poisson}
\end{figure}

\begin{figure}[p]
\centering
\includegraphics[width=0.92\linewidth]{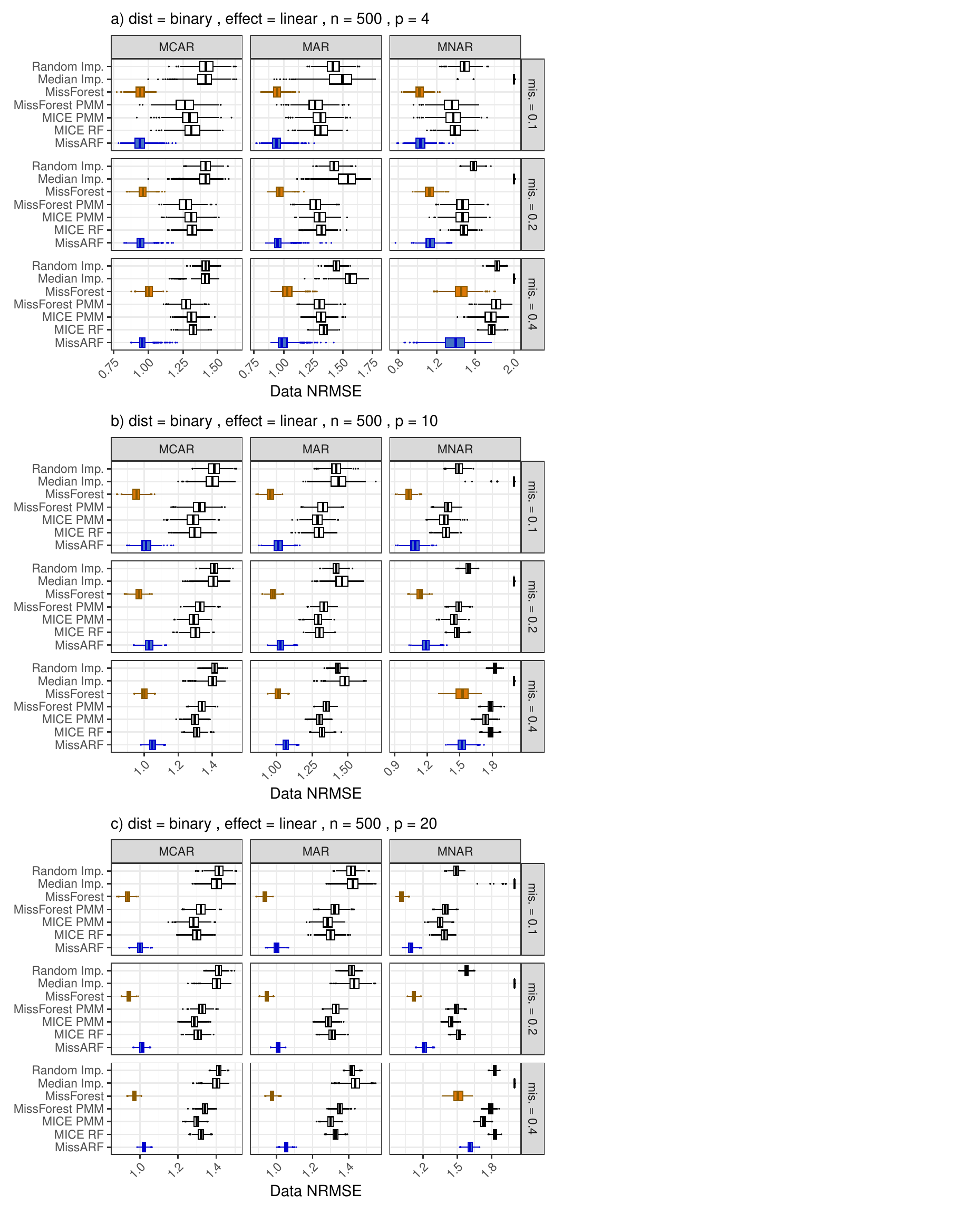}
\caption{ \textbf{NRMSE} of the binary distribution setting with a linear effect over different missingness patterns, dimensionality ($p$) and missingness rates (mis.) with $n=500$. The boxplots are plotted over the replicates, with MissARF (blue) and MissForest (orange) highlighted.} \label{fig: logreg_nrmse_linear_binary_500}
\end{figure}

\begin{figure}[p]
\centering
\includegraphics[width=0.92\linewidth]{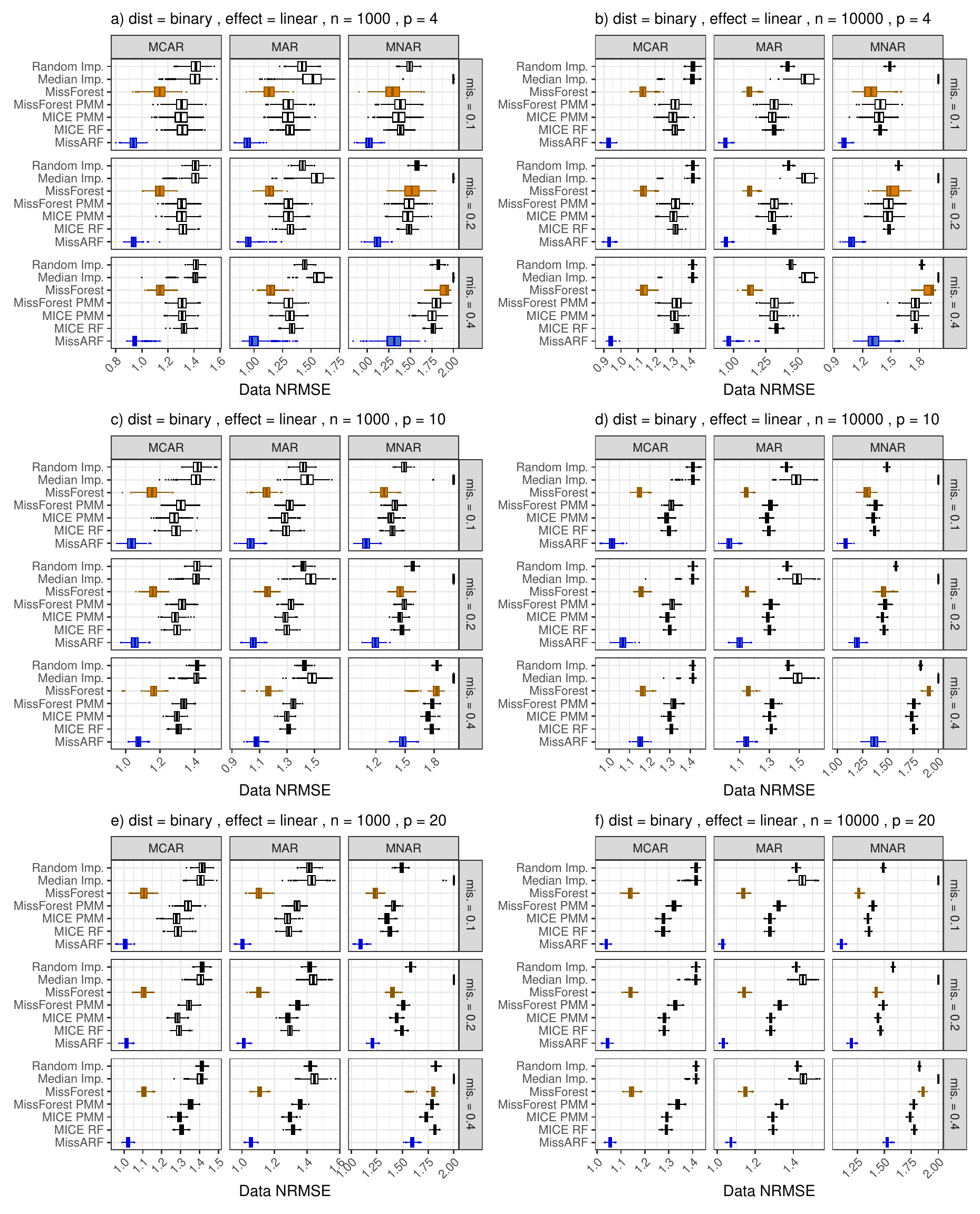}
\caption{ \textbf{NRMSE} of the binary distribution setting with a linear effect over different missingness patterns, dimensionality ($p$) and missingness rates (mis.) with $n=1000$ (left) and $n= 10,000$ (right). The boxplots are plotted over the replicates, with MissARF (blue) and MissForest (orange) highlighted.} \label{fig: logreg_nrmse_linear_binary}
\end{figure}

%squared
\clearpage
\subsubsection{Squared effect}
\begin{figure}[!ht]
\centering
\includegraphics[width=0.92\linewidth]{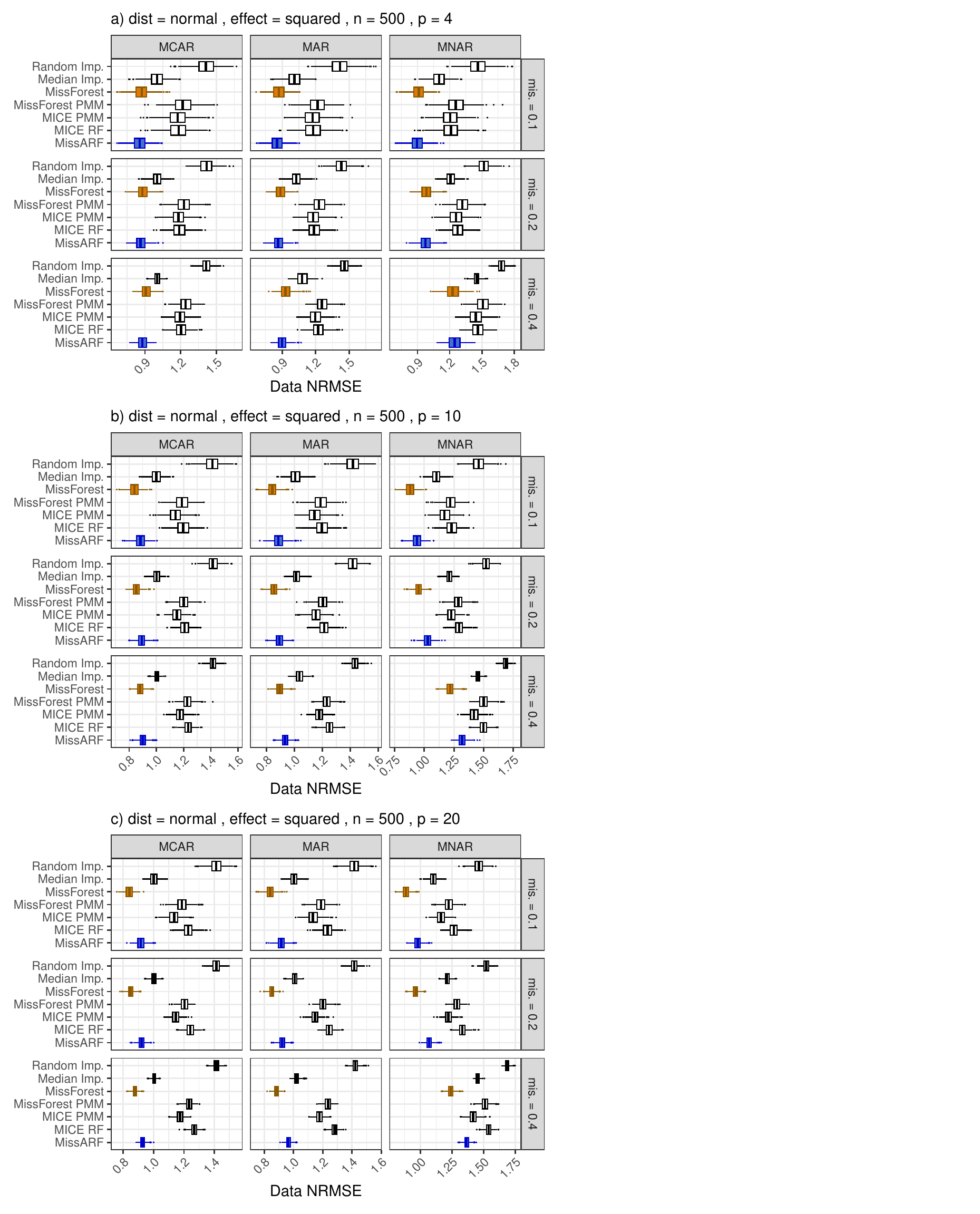}
\caption{ \textbf{NRMSE} of the normal distribution setting with a squared effect over different missingness patterns, dimensionality ($p$) and missingness rates (mis.) with $n=500$. The boxplots are plotted over the replicates, with MissARF (blue) and MissForest (orange) highlighted.} \label{fig: logreg_nrmse_squared_normal_500}
\end{figure}

\begin{figure}[p]
\centering
\includegraphics[width=0.92\linewidth]{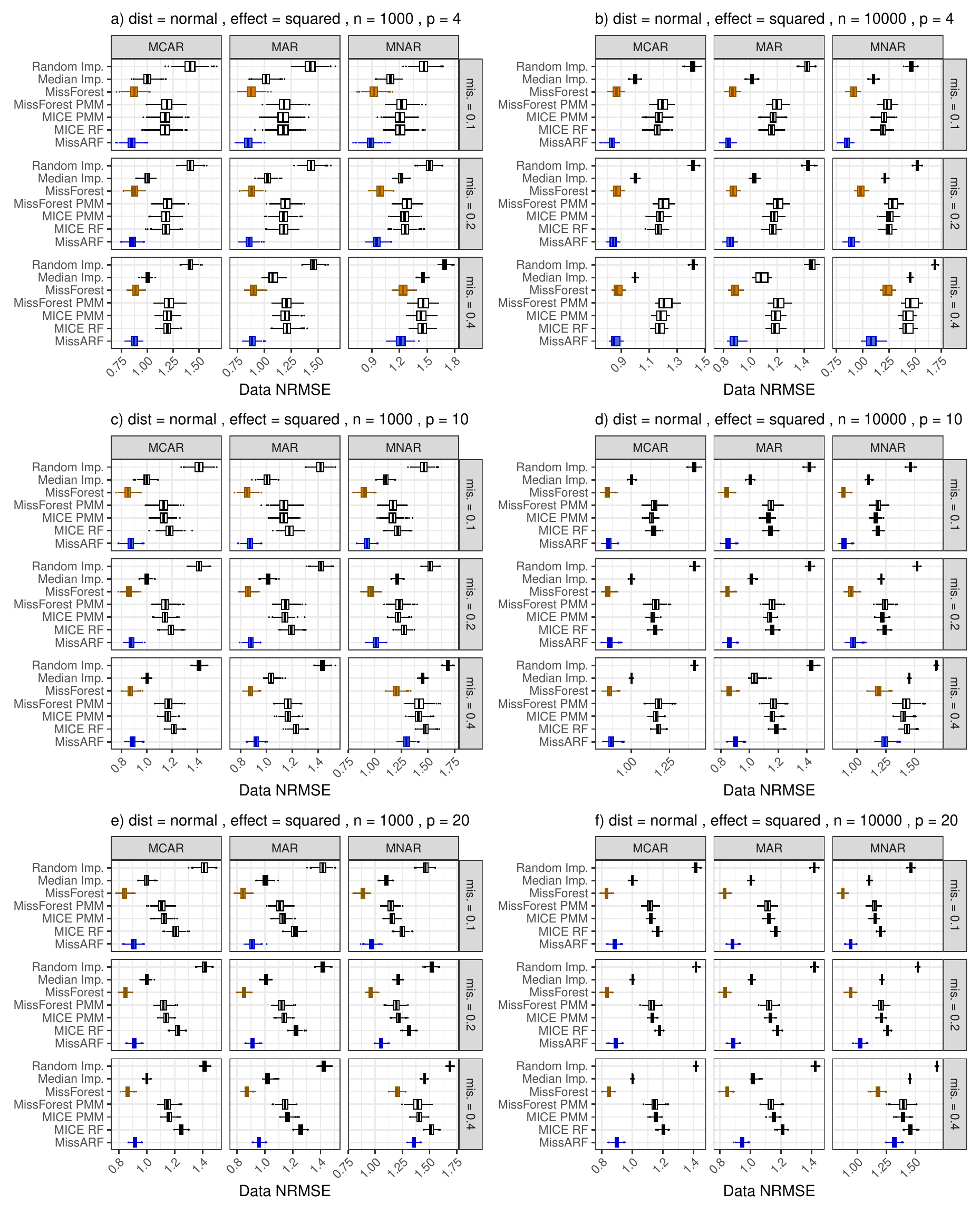}
\caption{ \textbf{NRMSE} of the normal distribution setting with a squared effect over different missingness patterns, dimensionality ($p$) and missingness rates (mis.) with $n=1000$ (left) and $n= 10,000$ (right). The boxplots are plotted over the replicates, with MissARF (blue) and MissForest (orange) highlighted.} \label{fig: logreg_nrmse_squared_normal}
\end{figure}

\begin{figure}[p]
\centering
\includegraphics[width=0.92\linewidth]{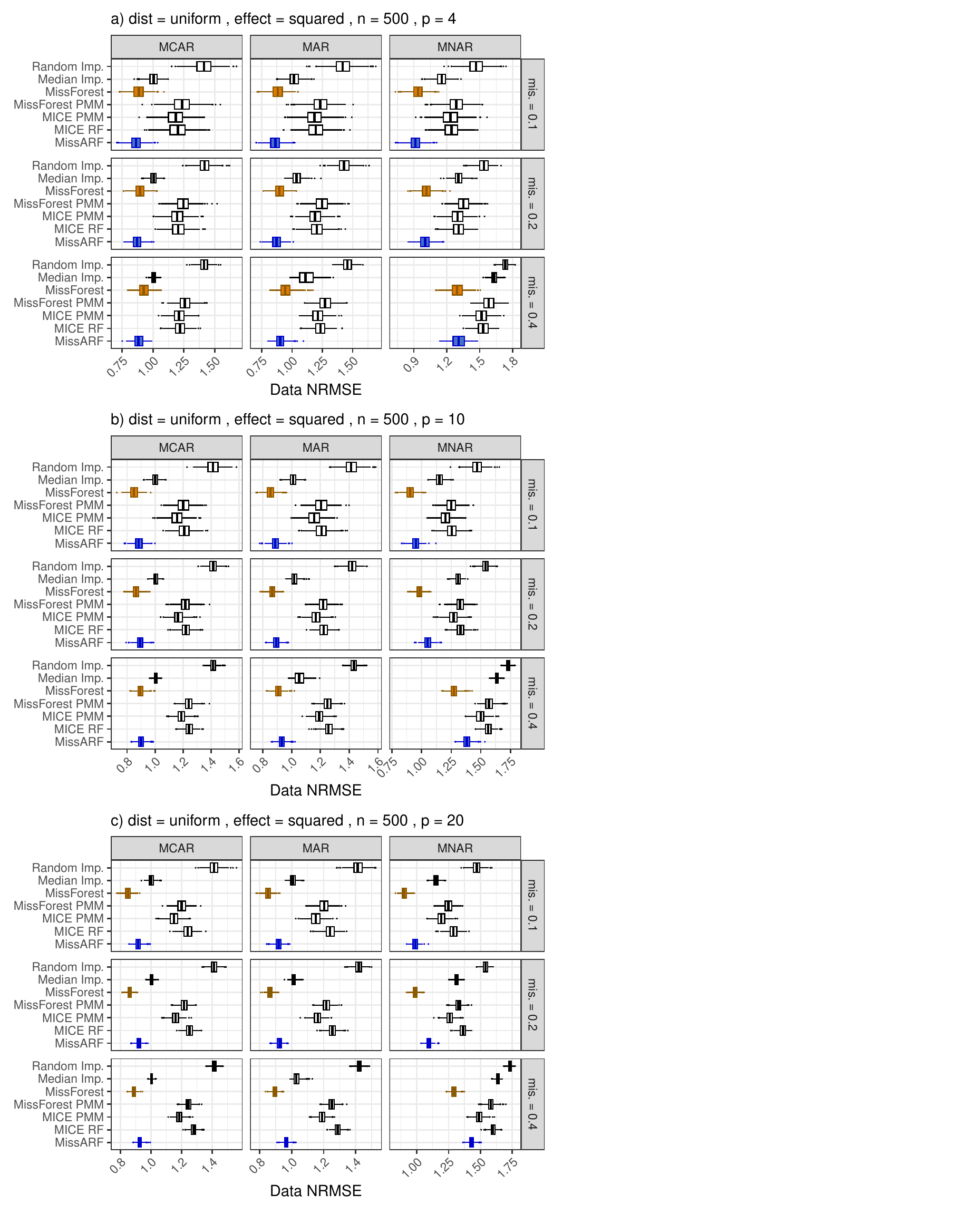}
\caption{ \textbf{NRMSE} of the uniform distribution setting with a squared effect over different missingness patterns, dimensionality ($p$) and missingness rates (mis.) with $n=500$. The boxplots are plotted over the replicates, with MissARF (blue) and MissForest (orange) highlighted.} \label{fig: logreg_nrmse_squared_uniform_500}
\end{figure}

\begin{figure}[p]
\centering
\includegraphics[width=0.92\linewidth]{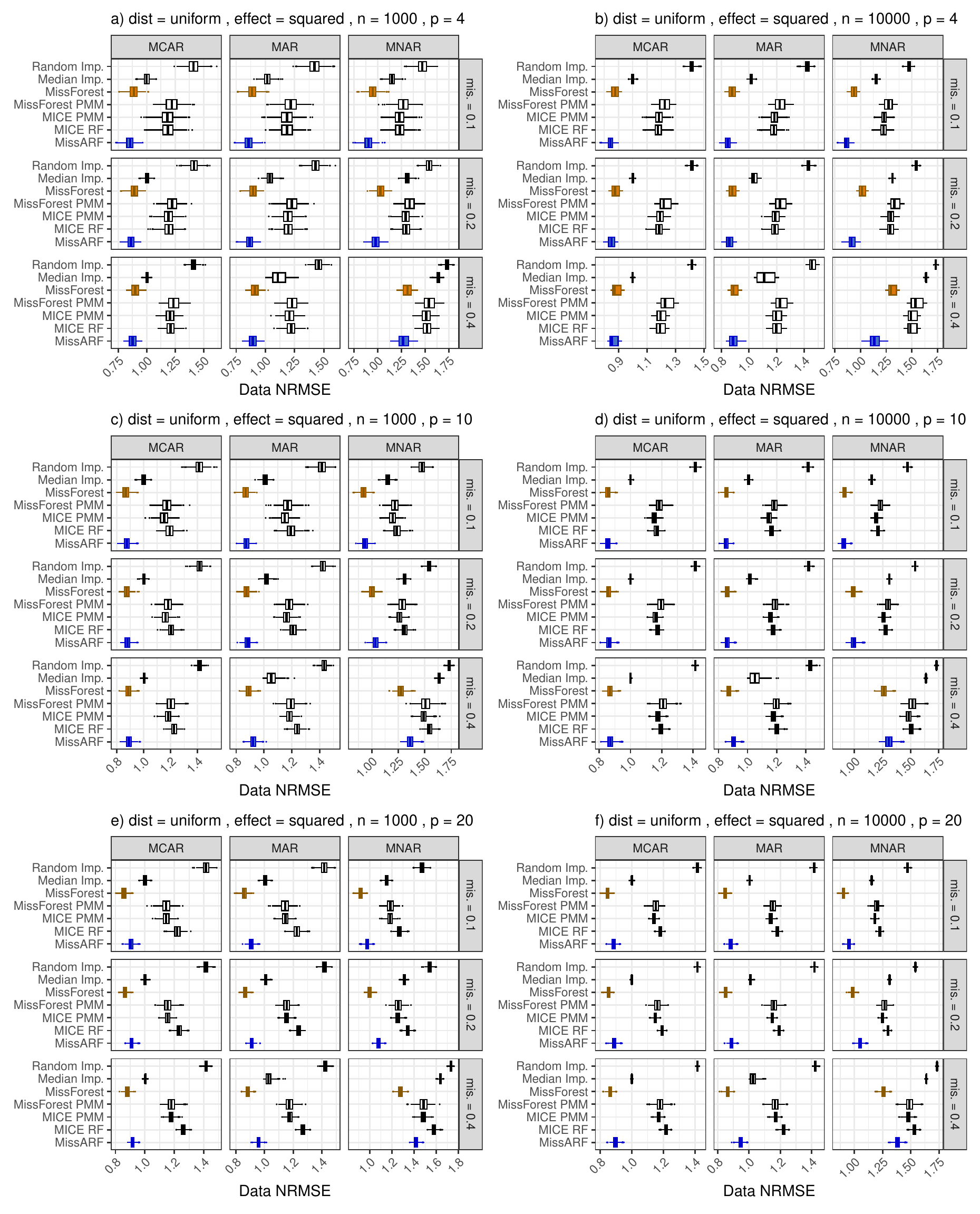}
\caption{ \textbf{NRMSE} of the uniform distribution setting with a squared effect over different missingness patterns, dimensionality ($p$) and missingness rates (mis.) with $n=1000$ (left) and $n= 10,000$ (right). The boxplots are plotted over the replicates, with MissARF (blue) and MissForest (orange) highlighted.} \label{fig: logreg_nrmse_squared_uniform}
\end{figure}

\begin{figure}[p]
\centering
\includegraphics[width=0.92\linewidth]{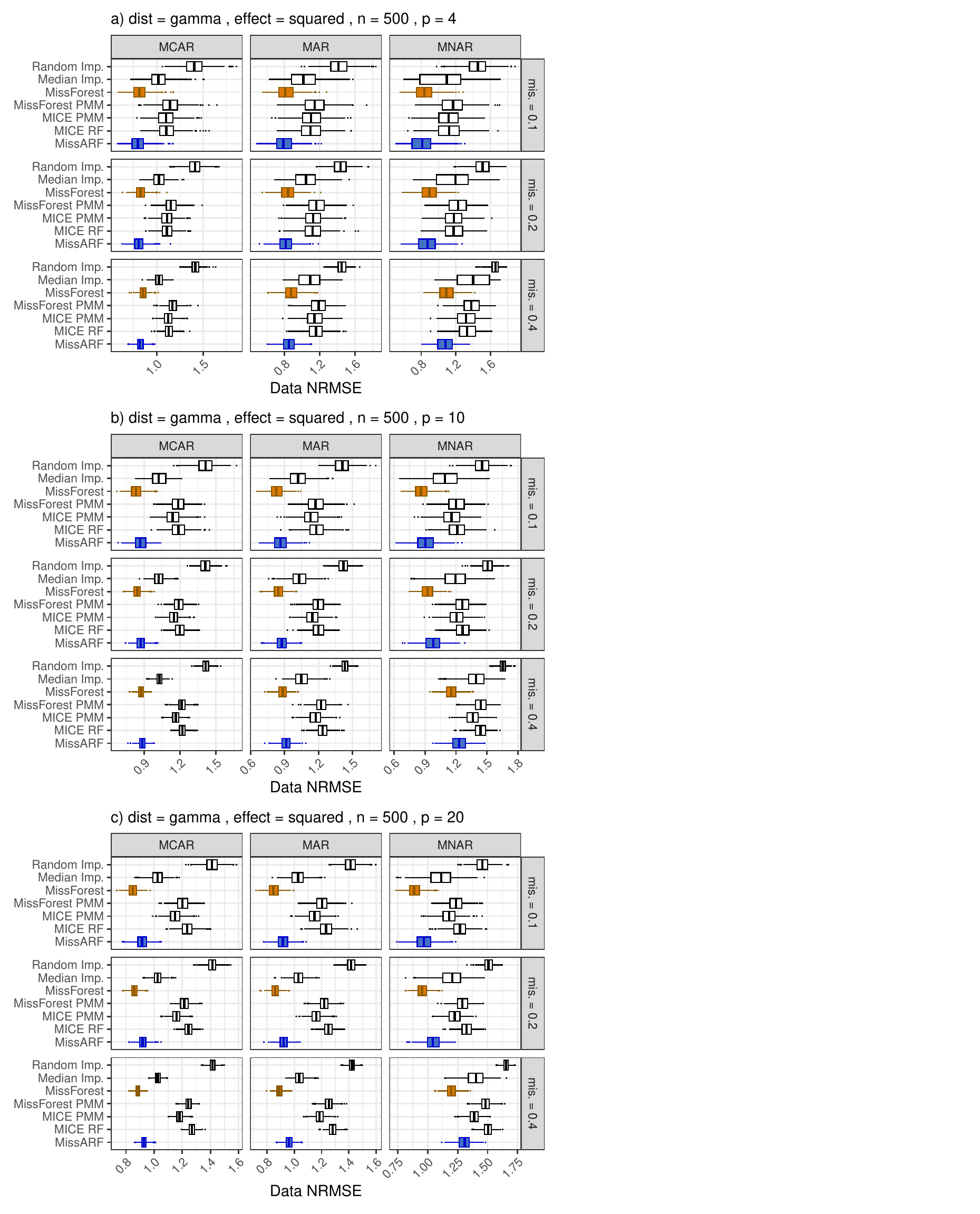}
\caption{ \textbf{NRMSE} of the gamma distribution setting with a squared effect over different missingness patterns, dimensionality ($p$) and missingness rates (mis.) with $n=500$. The boxplots are plotted over the replicates, with MissARF (blue) and MissForest (orange) highlighted.} \label{fig: logreg_nrmse_squared_gamma_500}
\end{figure}

\begin{figure}[p]
\centering
\includegraphics[width=0.92\linewidth]{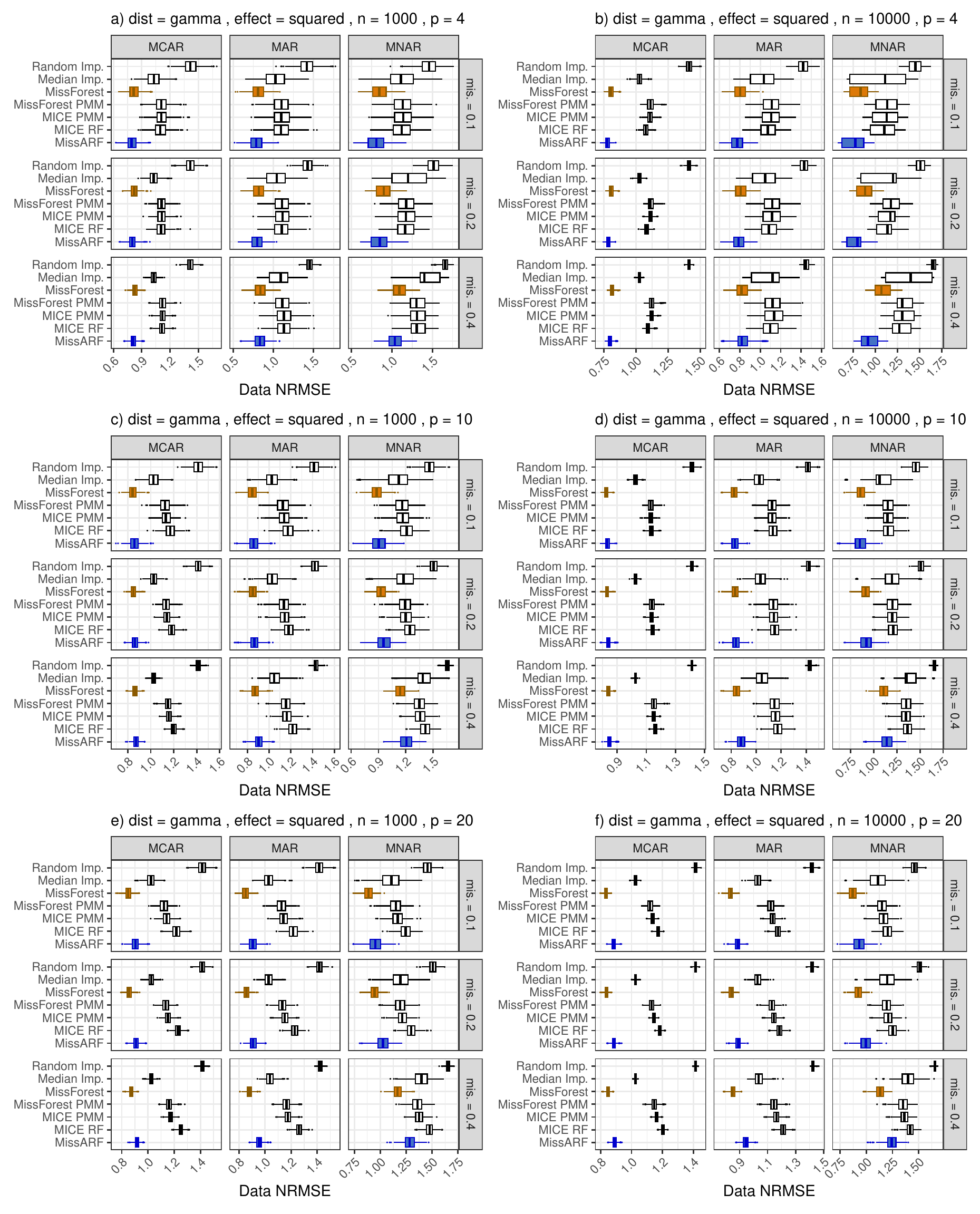}
\caption{ \textbf{NRMSE} of the gamma distribution setting with a squared effect over different missingness patterns, dimensionality ($p$) and missingness rates (mis.) with $n=1000$ (left) and $n= 10,000$ (right). The boxplots are plotted over the replicates, with MissARF (blue) and MissForest (orange) highlighted.} \label{fig: logreg_nrmse_squared_gamma}
\end{figure}

\begin{figure}[p]
\centering
\includegraphics[width=0.92\linewidth]{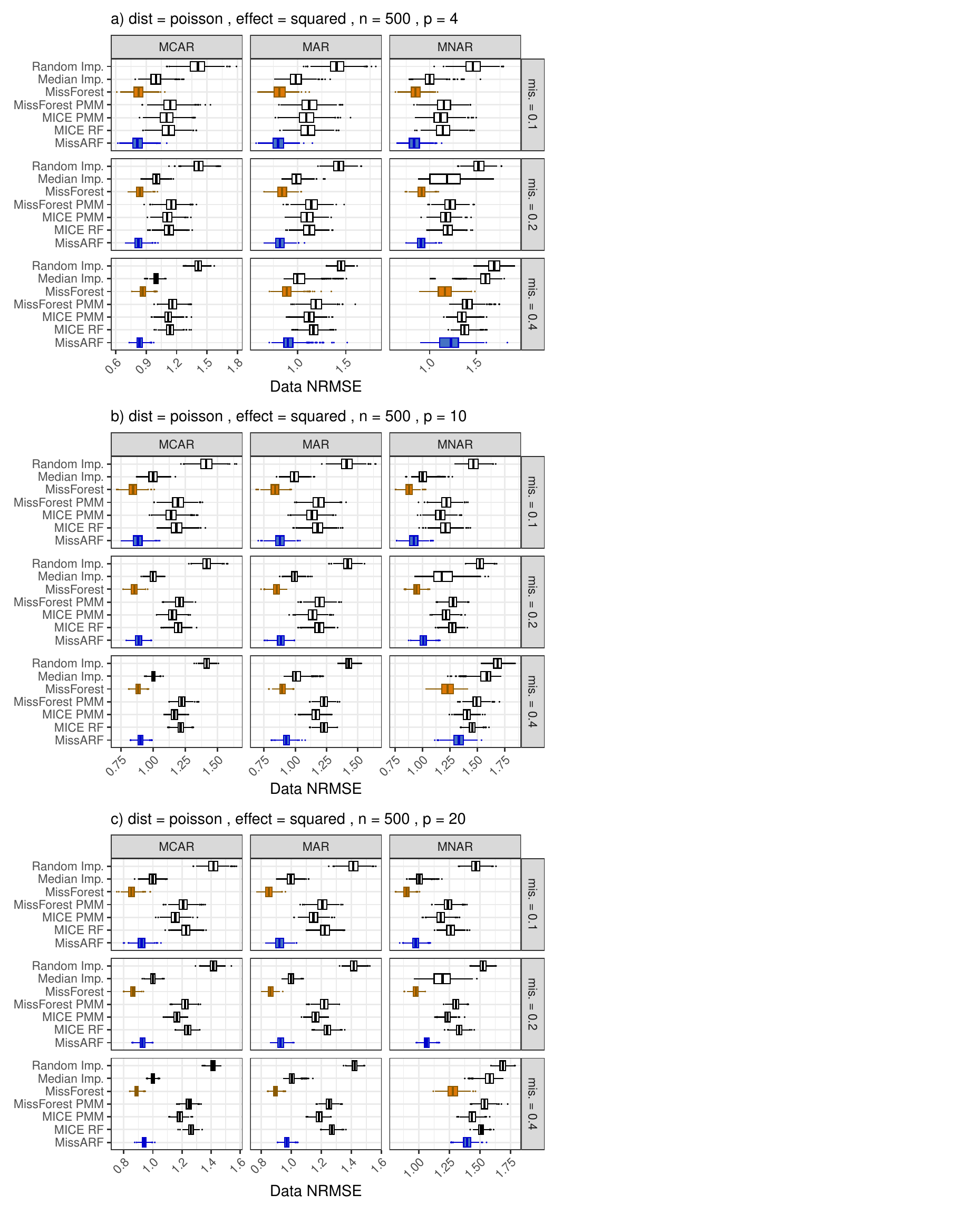}
\caption{ \textbf{NRMSE} of the Poisson distribution setting with a squared effect over different missingness patterns, dimensionality ($p$) and missingness rates (mis.) with $n=500$. The boxplots are plotted over the replicates, with MissARF (blue) and MissForest (orange) highlighted.} \label{fig: logreg_nrmse_squared_poisson_500}
\end{figure}

\begin{figure}[p]
\centering
\includegraphics[width=0.92\linewidth]{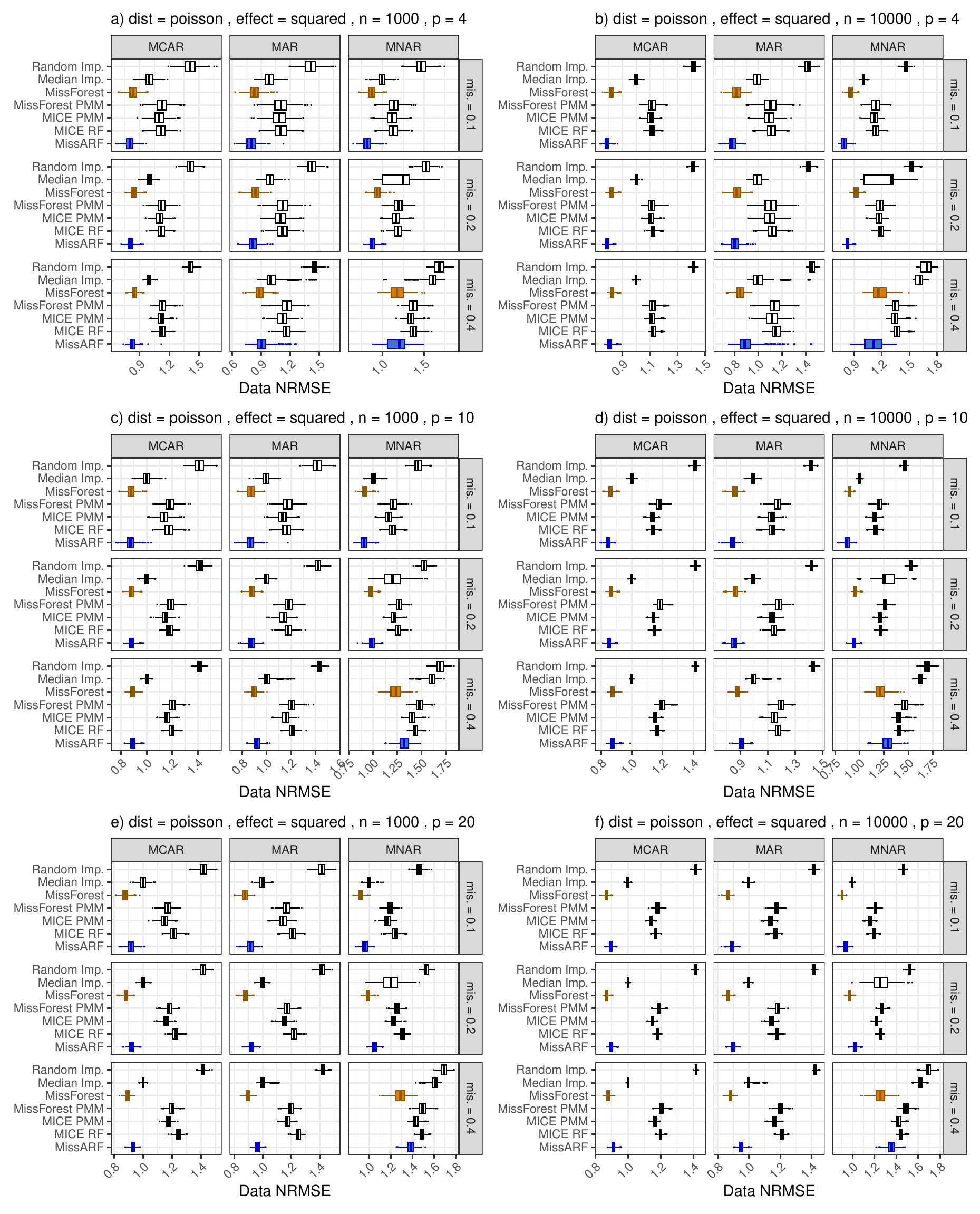}
\caption{ \textbf{NRMSE} of the Poisson distribution setting with a squared effect over different missingness patterns, dimensionality ($p$) and missingness rates (mis.) with $n=1000$ (left) and $n= 10,000$ (right). The boxplots are plotted over the replicates, with MissARF (blue) and MissForest (orange) highlighted.} \label{fig: logreg_nrmse_squared_poisson}
\end{figure}

\begin{figure}[p]
\centering
\includegraphics[width=0.92\linewidth]{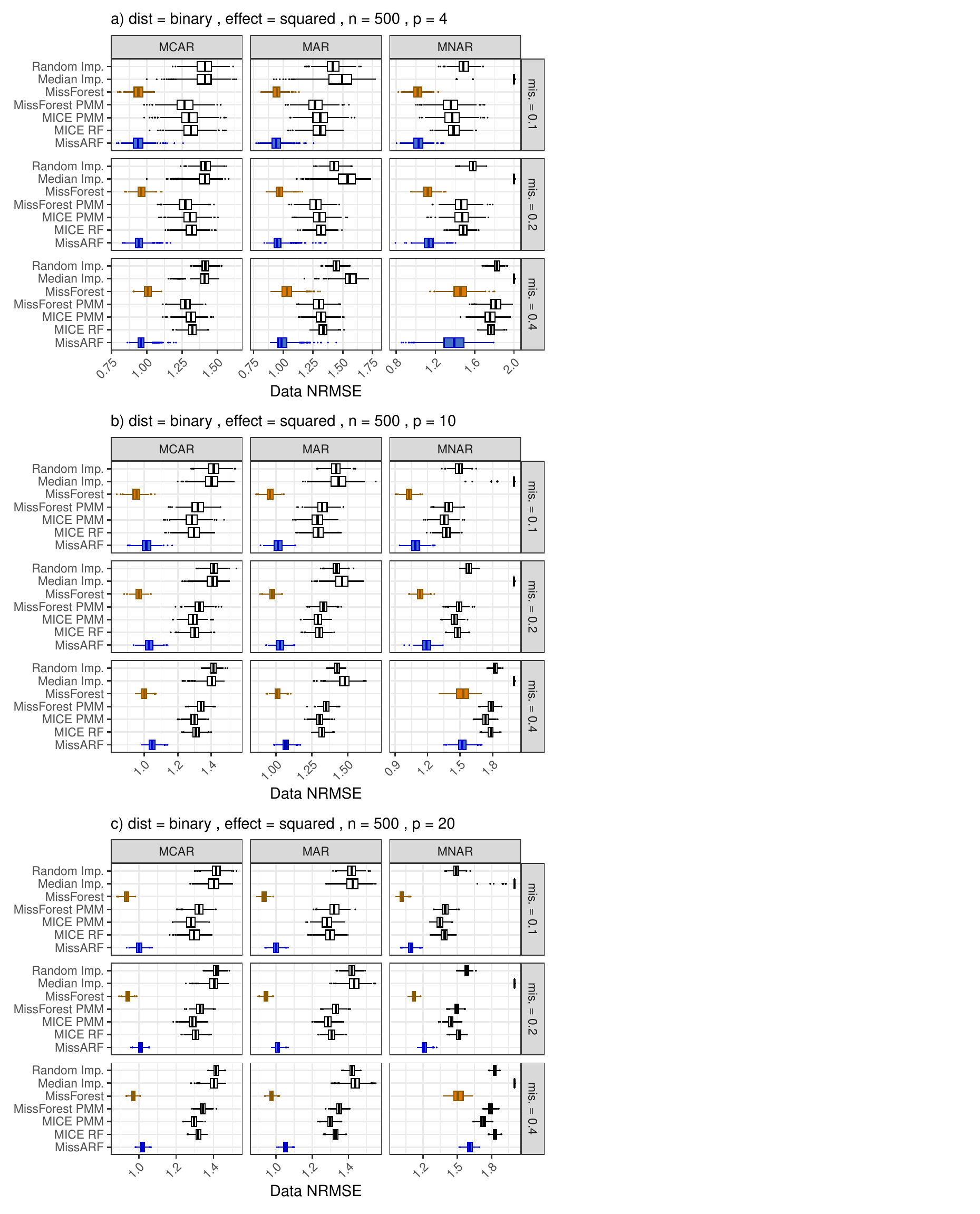}
\caption{ \textbf{NRMSE} of the binary distribution setting with a squared effect over different missingness patterns, dimensionality ($p$) and missingness rates (mis.) with $n=500$. The boxplots are plotted over the replicates, with MissARF (blue) and MissForest (orange) highlighted.} \label{fig: logreg_nrmse_squared_binary_500}
\end{figure}

\begin{figure}[p]
\centering
\includegraphics[width=0.92\linewidth]{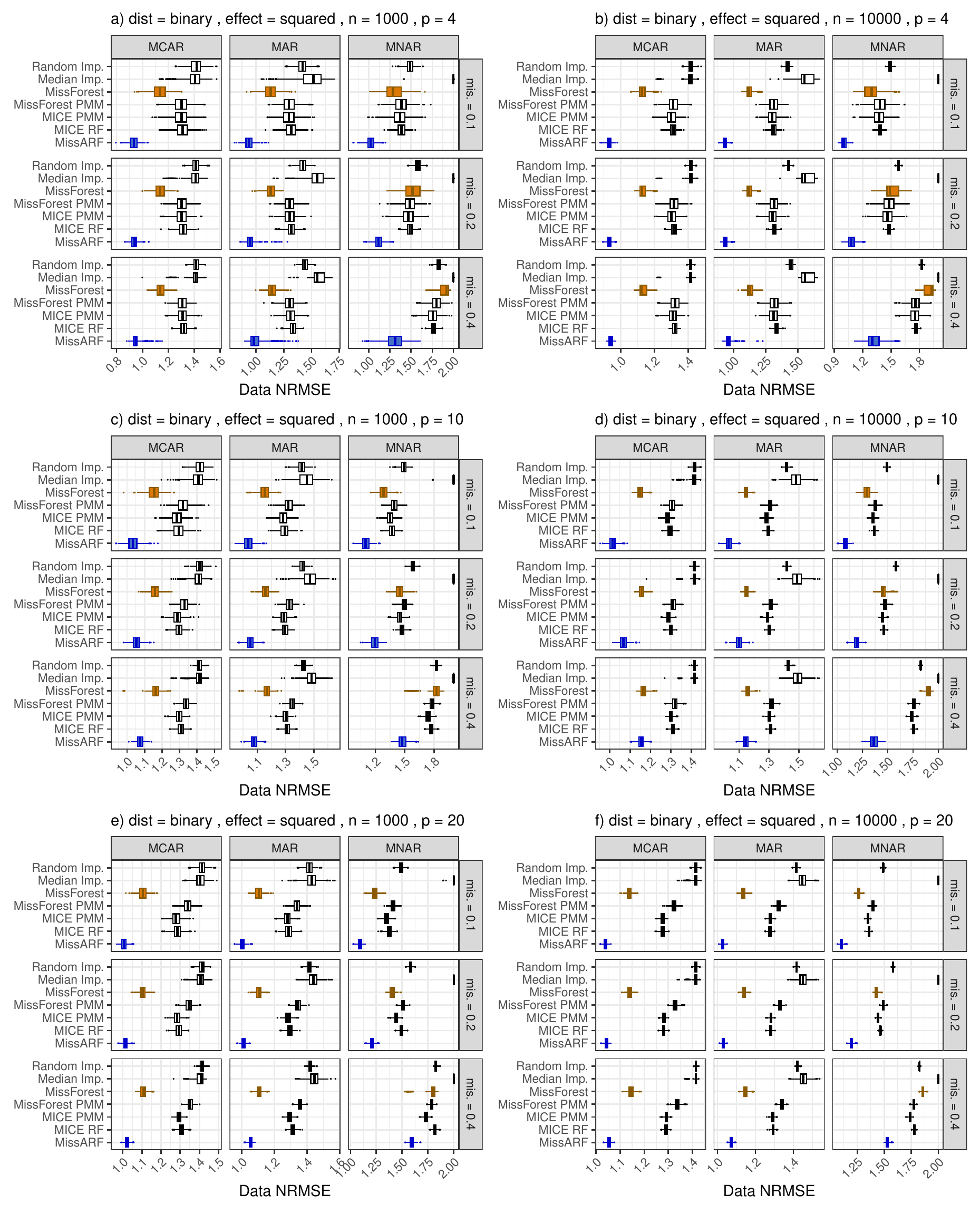}
\caption{ \textbf{NRMSE} of the binary distribution setting with a squared effect over different missingness patterns, dimensionality ($p$) and missingness rates (mis.) with $n=1000$ (left) and $n= 10,000$ (right). The boxplots are plotted over the replicates, with MissARF (blue) and MissForest (orange) highlighted.} \label{fig: logreg_nrmse_squared_binary}
\end{figure}

\clearpage
\subsection{Brier Score}
\label{sec:appx_A_Brier}
\subsubsection{Linear effect}
%% Brier Score
\begin{figure}[!ht]
\centering
\includegraphics[width=0.92\linewidth]{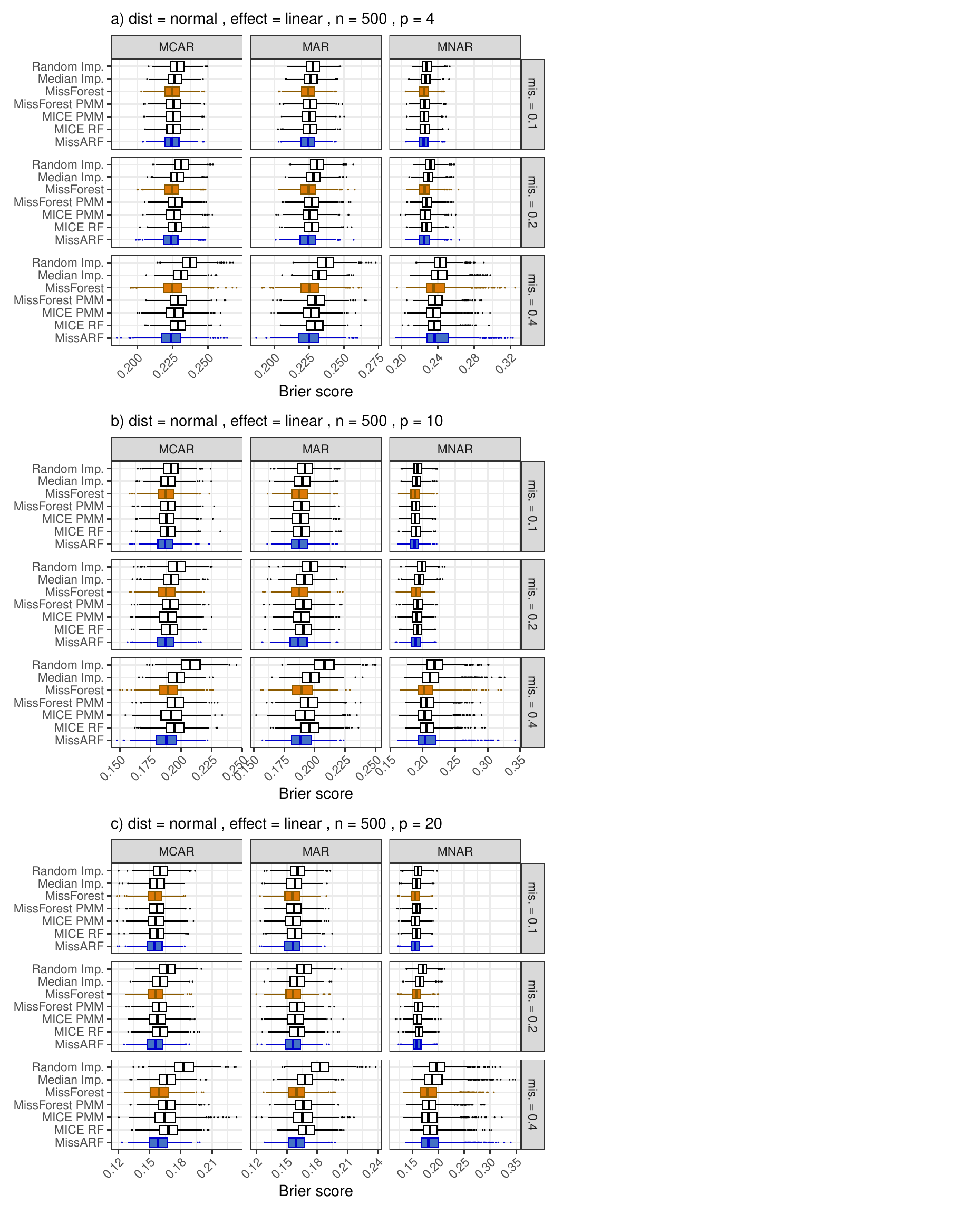}
\caption{\textbf{Brier Score} of the normal distribution setting with a linear effect over different missingness patterns, dimensionality ($p$) and missingness rates (mis.) with $n=500$. The boxplots are plotted over the replicates, with MissARF (blue) and MissForest (orange) highlighted.} \label{fig: logreg_pred_linear_normal_500}
\end{figure}

\begin{figure}[p]
\centering
\includegraphics[width=0.92\linewidth]{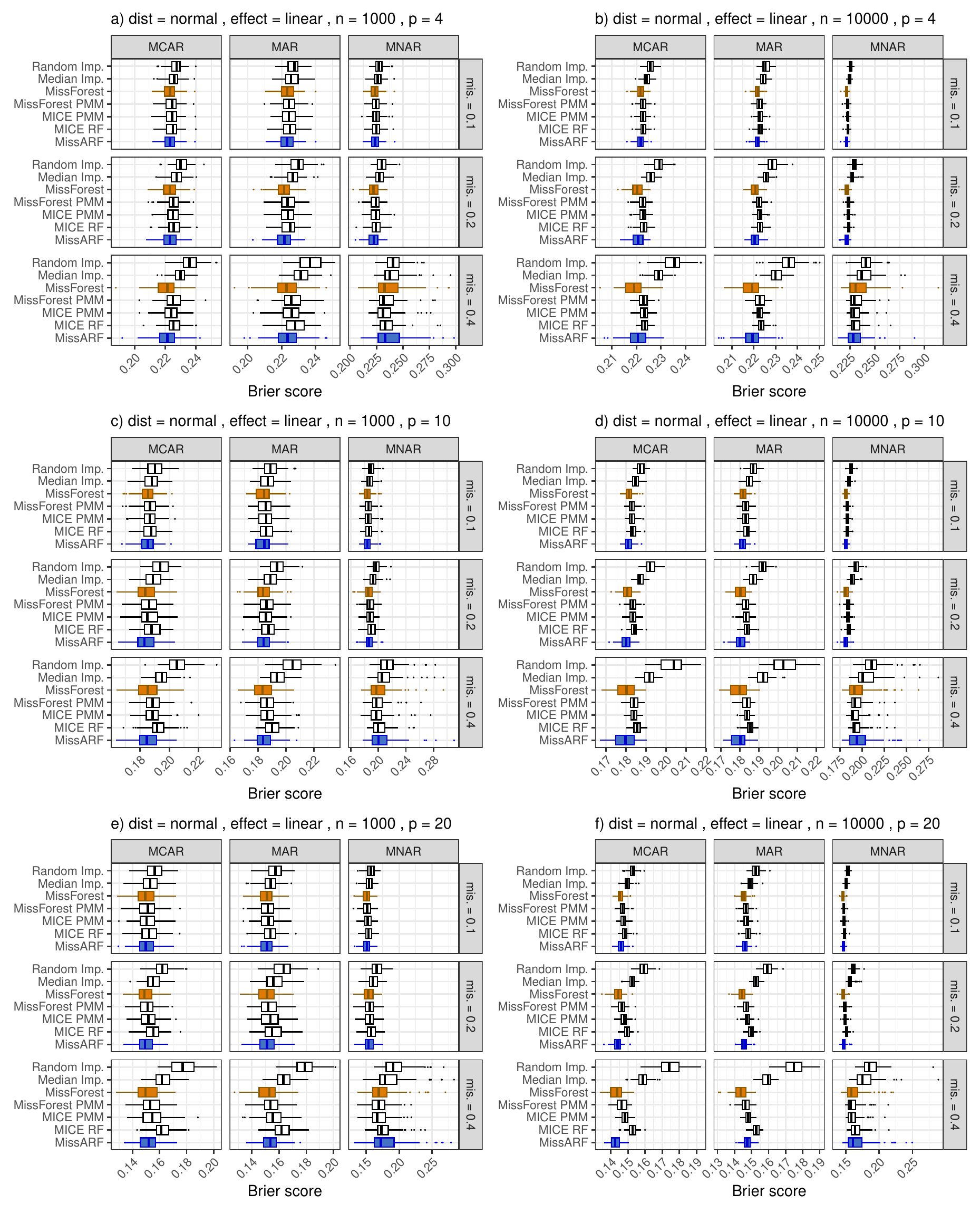}
\caption{\textbf{Brier Score} of the normal distribution setting with a linear effect over different missingness patterns, dimensionality ($p$) and missingness rates (mis.) with $n=1000$ (left) and $n= 10,000$ (right). The boxplots are plotted over the replicates, with MissARF (blue) and MissForest (orange) highlighted.} \label{fig: logreg_pred_linear_normal}
\end{figure}

\begin{figure}[p]
\centering
\includegraphics[width=0.92\linewidth]{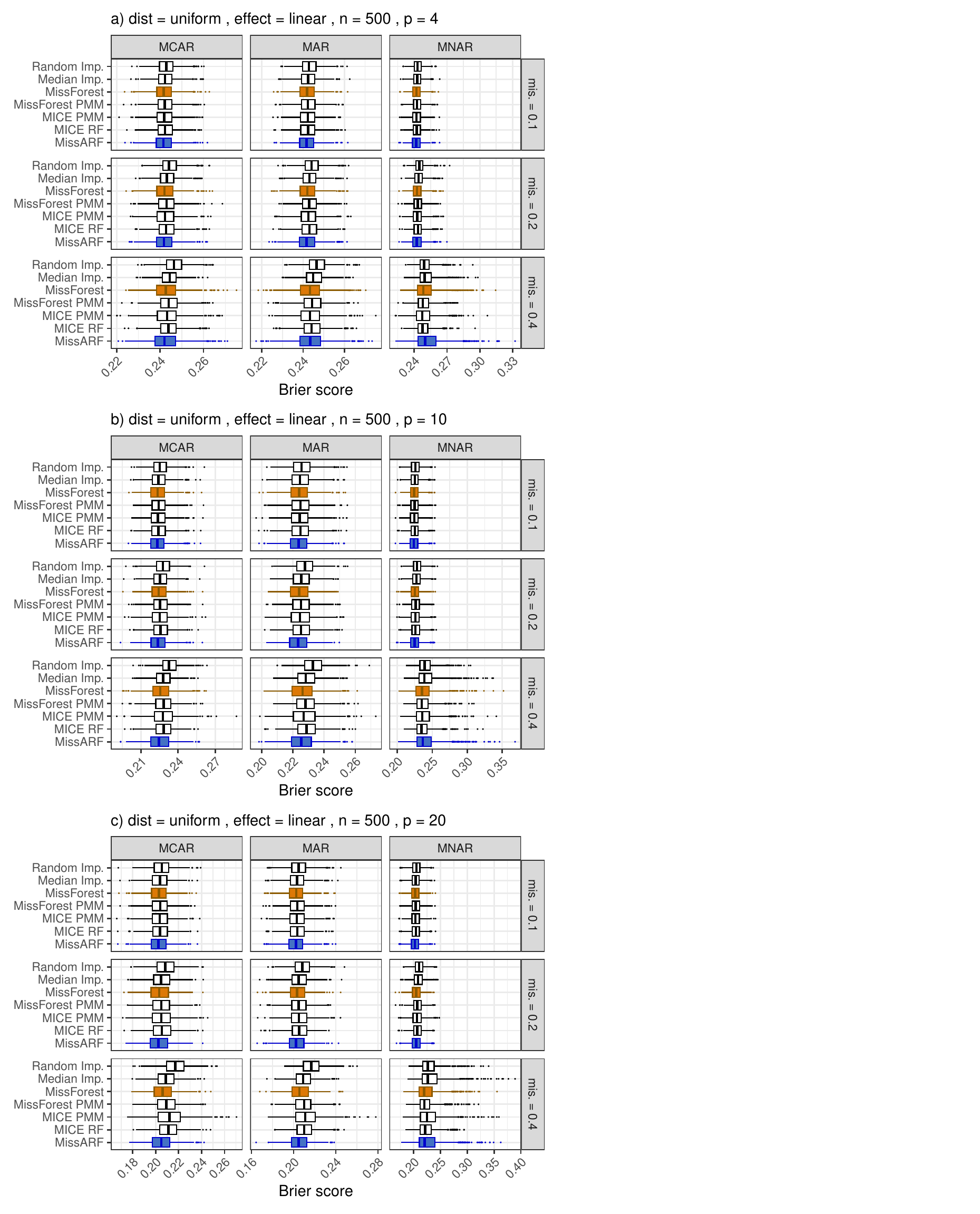}
\caption{\textbf{Brier Score} of the uniform distribution setting with a linear effect over different missingness patterns, dimensionality ($p$) and missingness rates (mis.) with $n=500$. The boxplots are plotted over the replicates, with MissARF (blue) and MissForest (orange) highlighted.} \label{fig: logreg_pred_linear_uniform_500}
\end{figure}

\begin{figure}[p]
\centering
\includegraphics[width=0.92\linewidth]{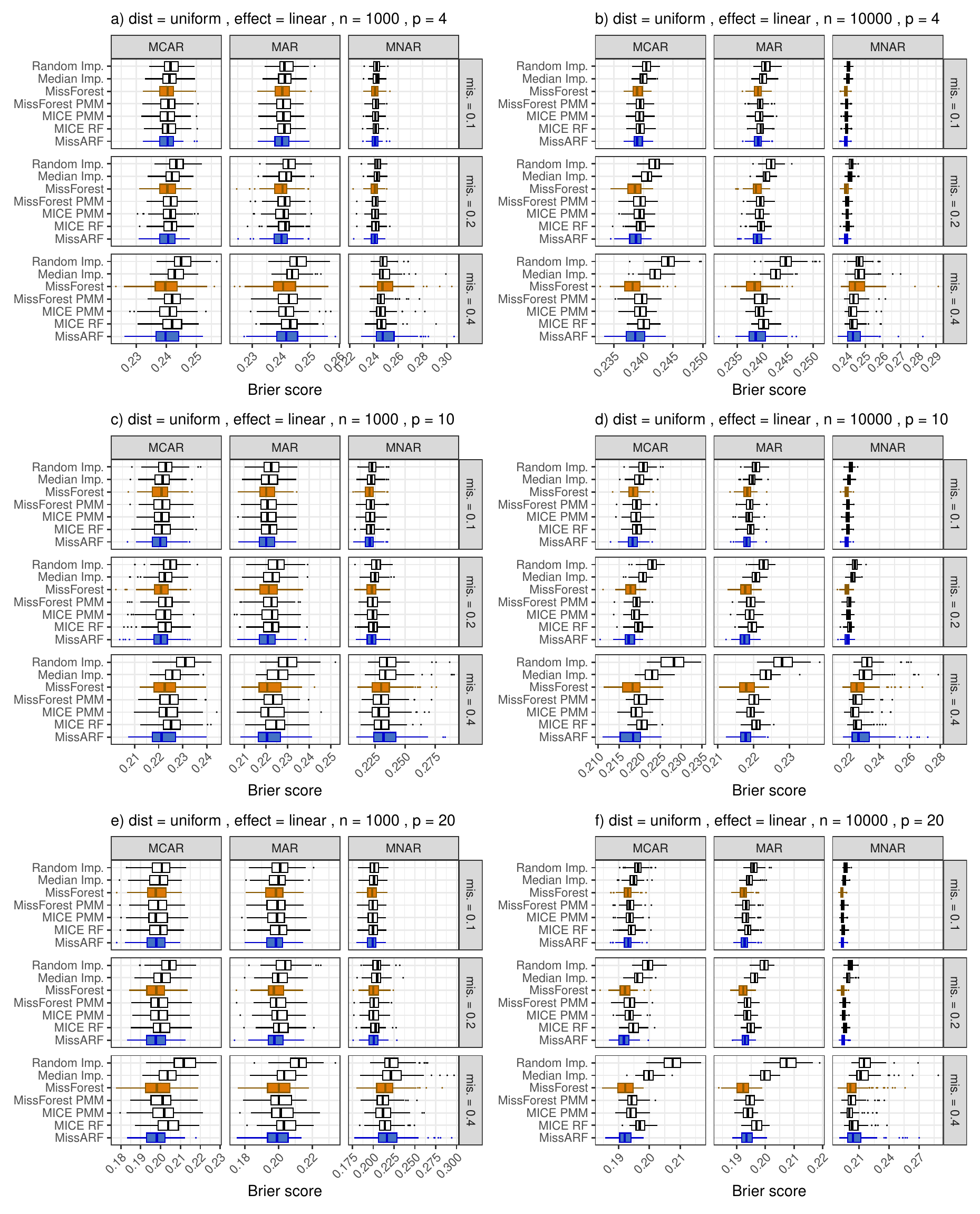}
\caption{\textbf{Brier Score} of the uniform distribution setting with a linear effect over different missingness patterns, dimensionality ($p$) and missingness rates (mis.) with $n=1000$ (left) and $n= 10,000$ (right). The boxplots are plotted over the replicates, with MissARF (blue) and MissForest (orange) highlighted.} \label{fig: logreg_pred_linear_uniform}
\end{figure}

\begin{figure}[p]
\centering
\includegraphics[width=0.92\linewidth]{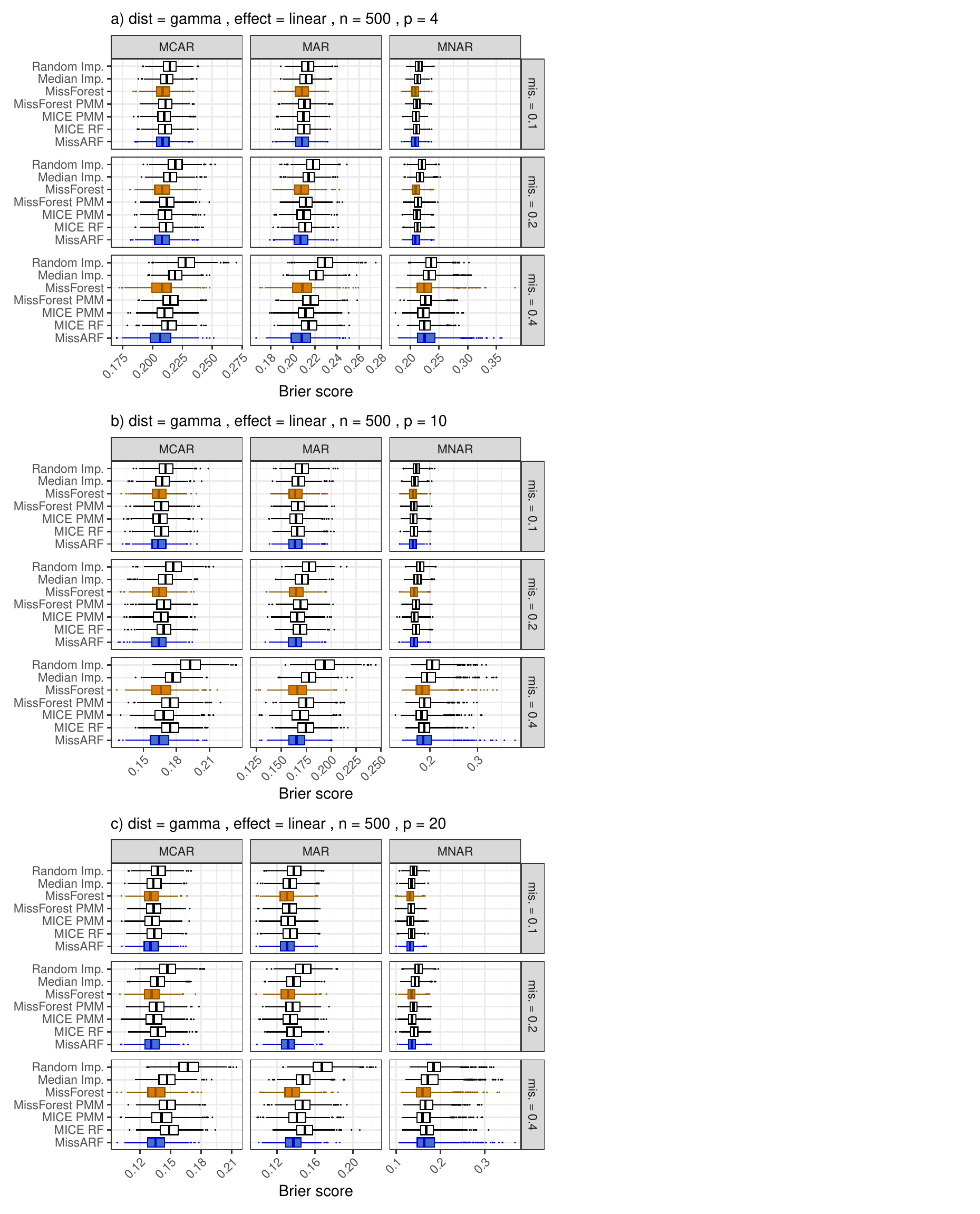}
\caption{ \textbf{Brier Score} of the gamma distribution setting with a linear effect over different missingness patterns, dimensionality ($p$) and missingness rates (mis.) with $n=500$. The boxplots are plotted over the replicates, with MissARF (blue) and MissForest (orange) highlighted.} \label{fig: logreg_pred_linear_gamma_500}
\end{figure}

\begin{figure}[p]
\centering
\includegraphics[width=0.92\linewidth]{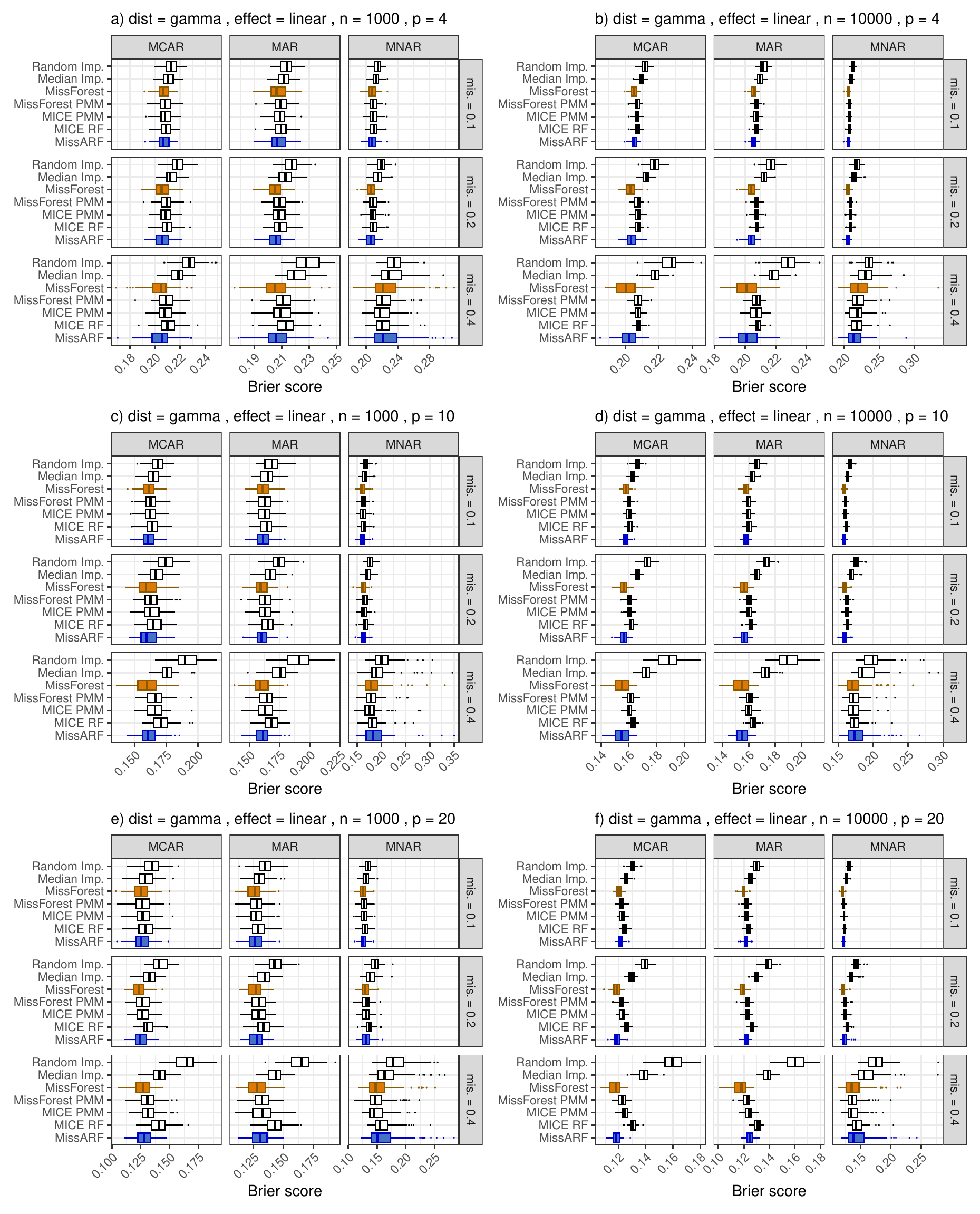}
\caption{ \textbf{Brier Score} of the gamma distribution setting with a linear effect over different missingness patterns, dimensionality ($p$) and missingness rates (mis.) with $n=1000$ (left) and $n= 10,000$ (right). The boxplots are plotted over the replicates, with MissARF (blue) and MissForest (orange) highlighted.} \label{fig: logreg_pred_linear_gamma}
\end{figure}

\begin{figure}[p]
\centering
\includegraphics[width=0.92\linewidth]{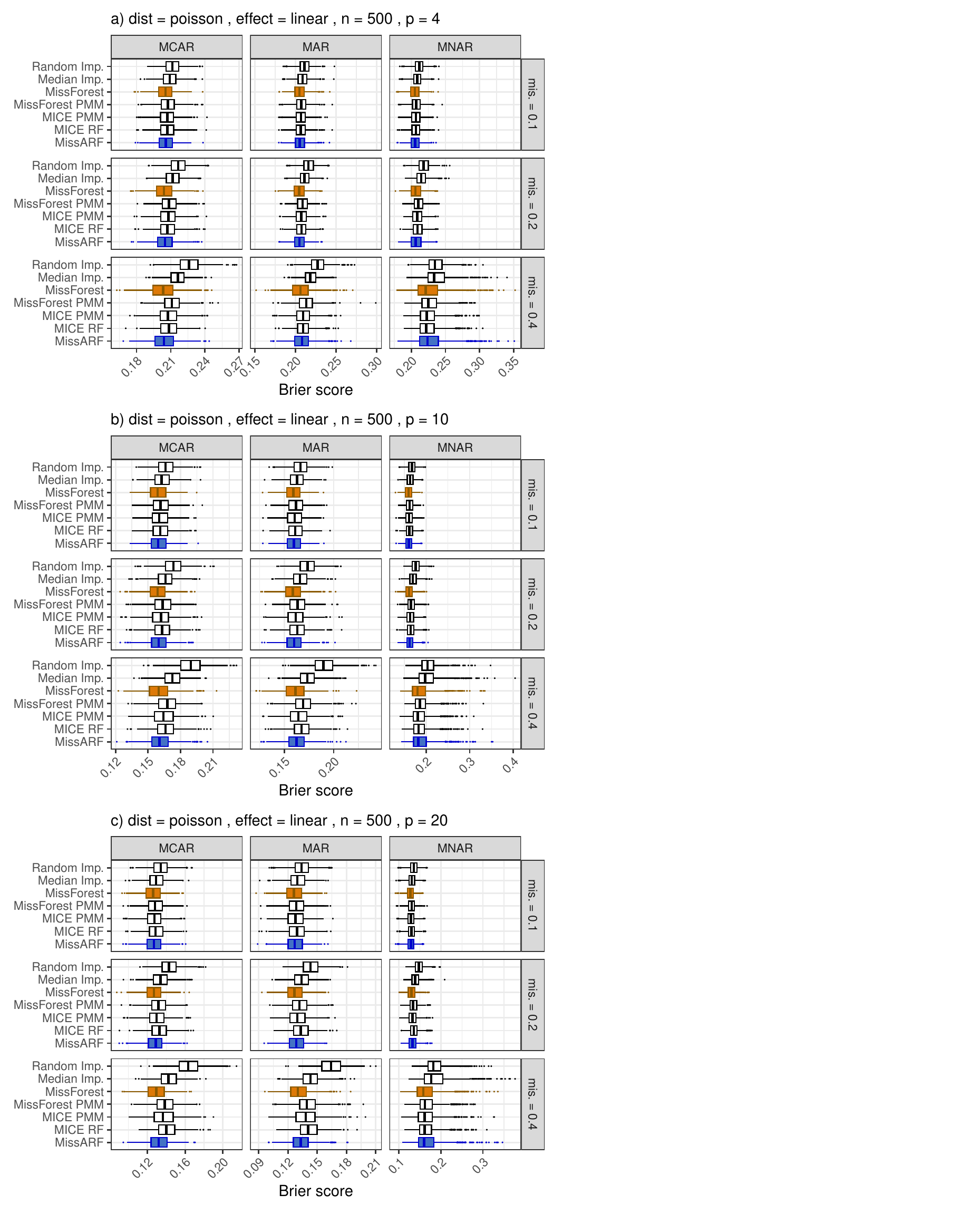}
\caption{ \textbf{Brier Score} of the Poisson distribution setting with a linear effect over different missingness patterns, dimensionality ($p$) and missingness rates (mis.) with $n=500$. The boxplots are plotted over the replicates, with MissARF (blue) and MissForest (orange) highlighted.} \label{fig: logreg_pred_linear_poisson_500}
\end{figure}

\begin{figure}[p]
\centering
\includegraphics[width=0.92\linewidth]{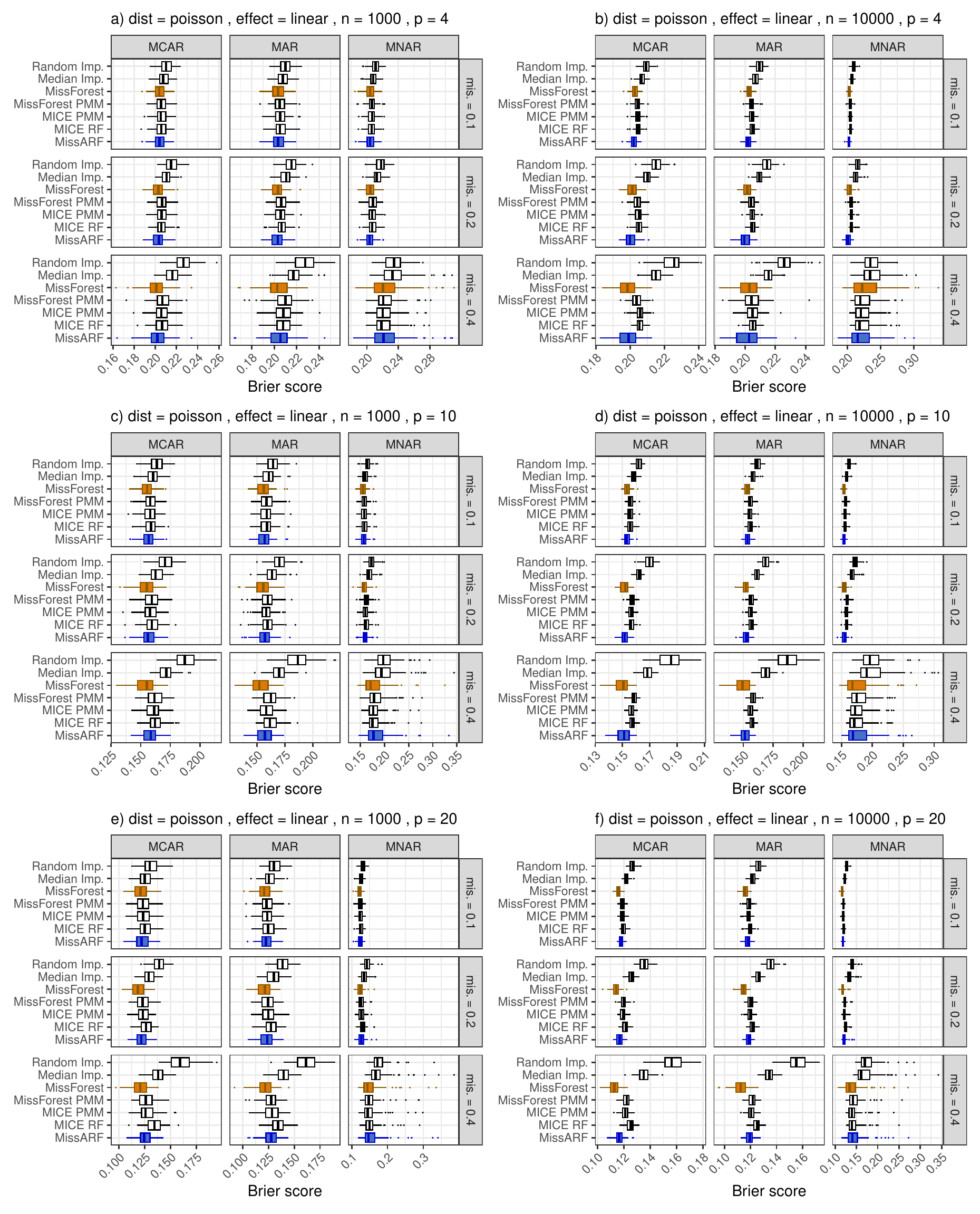}
\caption{ \textbf{Brier Score} of the Poisson distribution setting with a linear effect over different missingness patterns, dimensionality ($p$) and missingness rates (mis.) with $n=1000$ (left) and $n= 10,000$ (right). The boxplots are plotted over the replicates, with MissARF (blue) and MissForest (orange) highlighted.} \label{fig: logreg_pred_linear_poisson}
\end{figure}

\begin{figure}[p]
\centering
\includegraphics[width=0.92\linewidth]{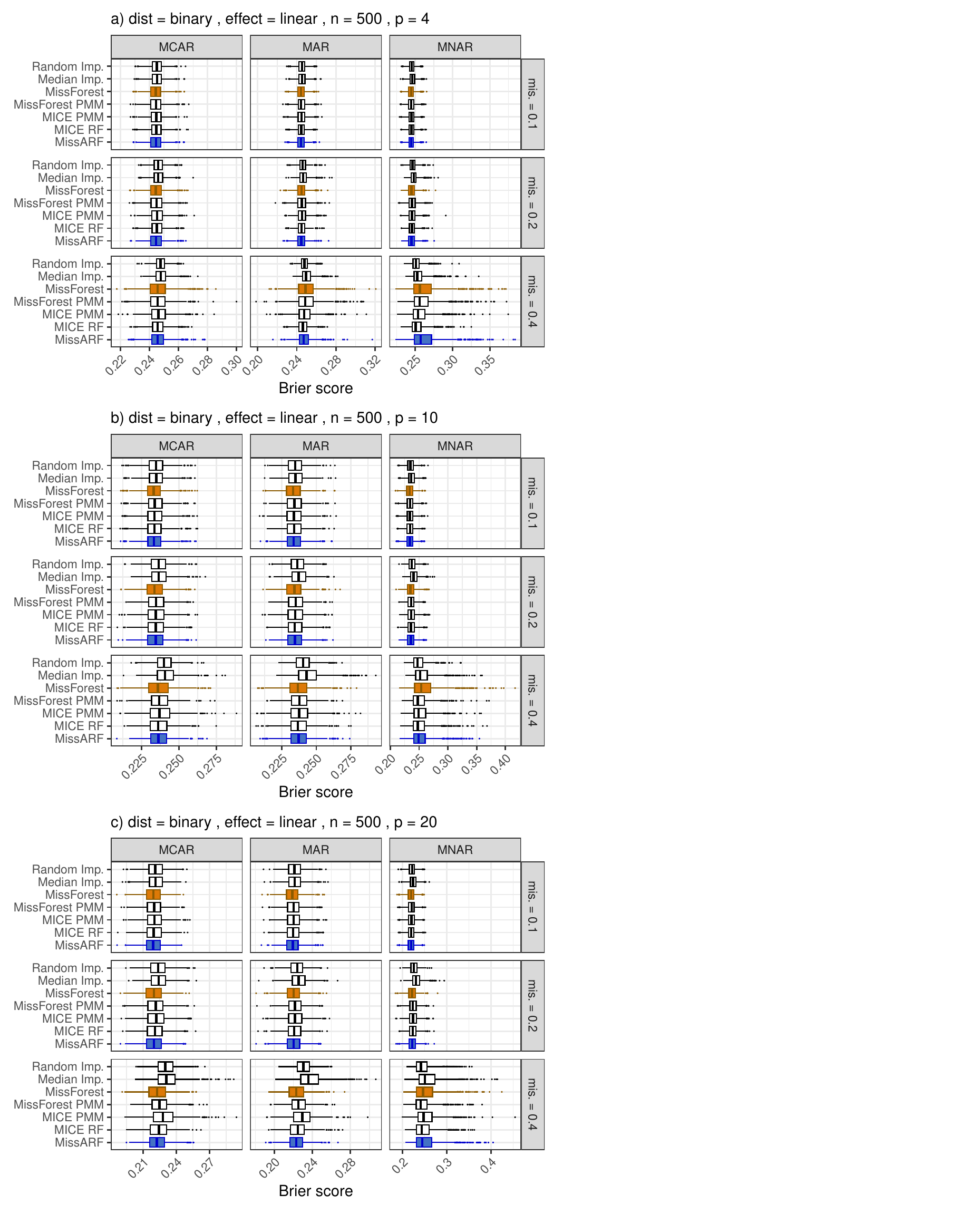}
\caption{ \textbf{Brier Score} of the binary distribution setting with a linear effect over different missingness patterns, dimensionality ($p$) and missingness rates (mis.) with $n=500$. The boxplots are plotted over the replicates, with MissARF (blue) and MissForest (orange) highlighted.} \label{fig: logreg_pred_linear_binary_500}
\end{figure}

\begin{figure}[p]
\centering
\includegraphics[width=0.92\linewidth]{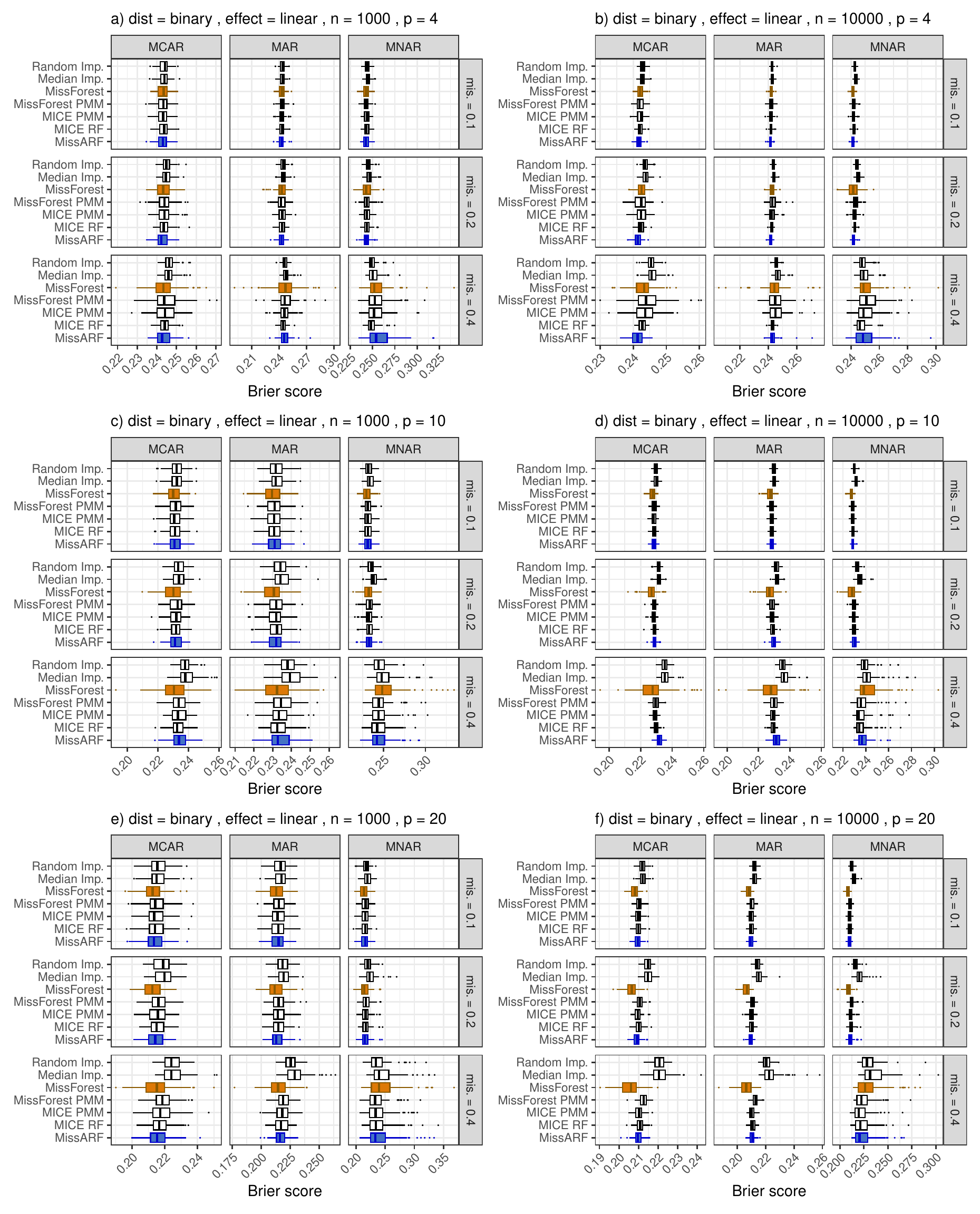}
\caption{ \textbf{Brier Score} of the binary distribution setting with a linear effect over different missingness patterns, dimensionality ($p$) and missingness rates (mis.) with $n=1000$ (left) and $n= 10,000$ (right). The boxplots are plotted over the replicates, with MissARF (blue) and MissForest (orange) highlighted.} \label{fig: logreg_pred_linear_binary}
\end{figure}

\clearpage
%squared
\subsubsection{Squared effect}
\begin{figure}[!ht]
\centering
\includegraphics[width=0.92\linewidth]{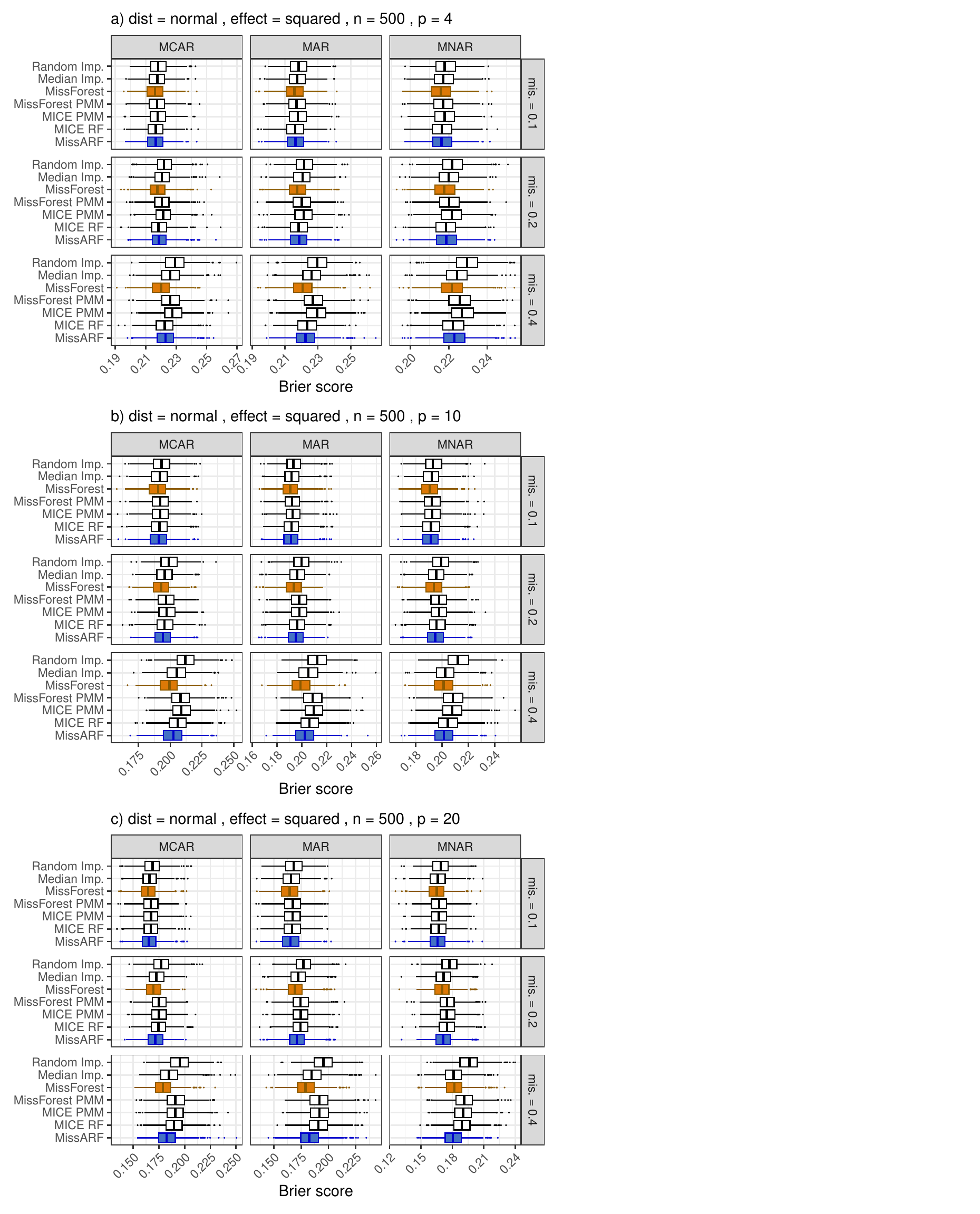}
\caption{ \textbf{Brier Score} of the normal distribution setting with a squared effect over different missingness patterns, dimensionality ($p$) and missingness rates (mis.) with $n=500$. The boxplots are plotted over the replicates, with MissARF (blue) and MissForest (orange) highlighted.} \label{fig: logreg_pred_squared_normal_500}
\end{figure}

\begin{figure}[p]
\centering
\includegraphics[width=0.92\linewidth]{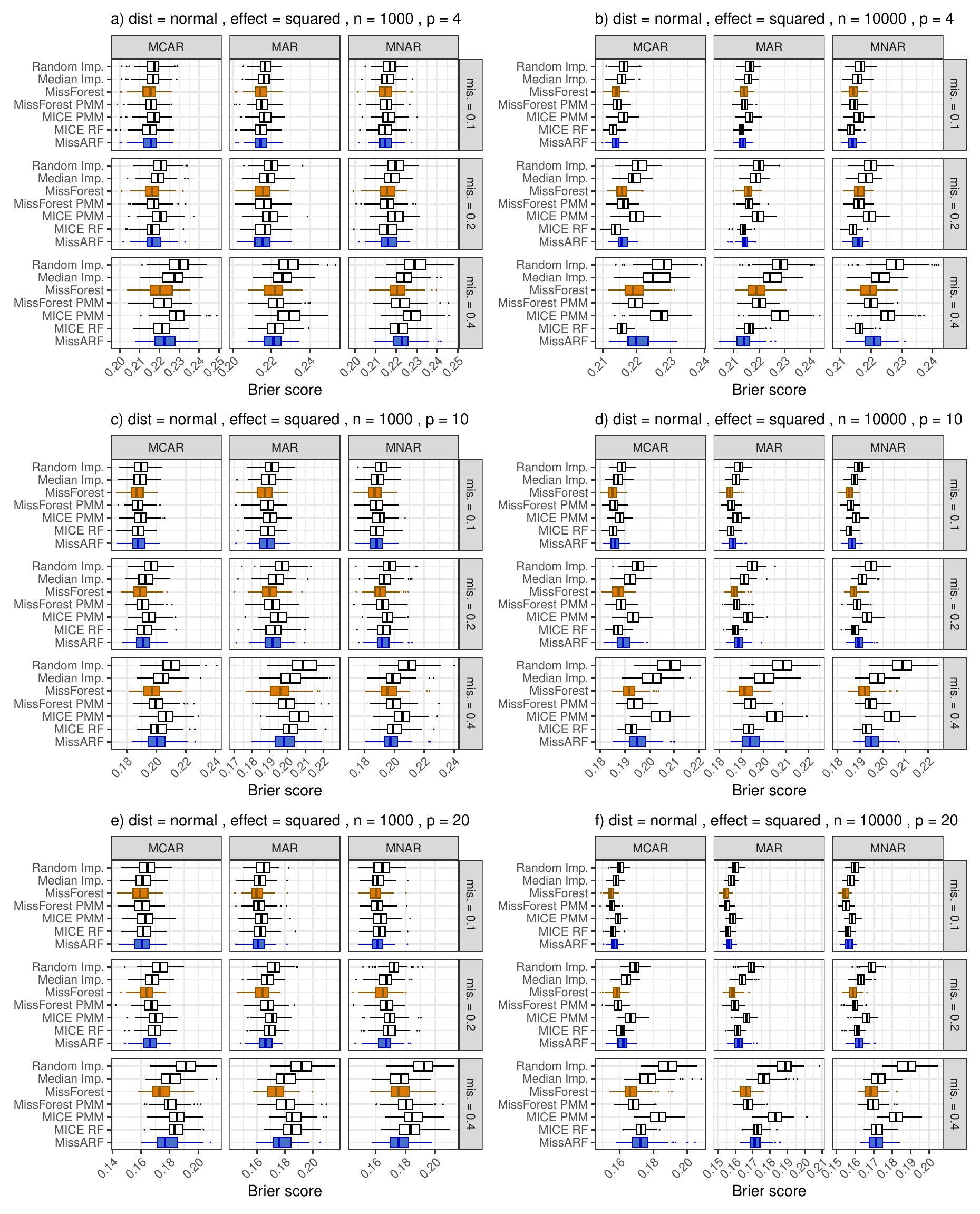}
\caption{ \textbf{Brier Score} of the normal distribution setting with a squared effect over different missingness patterns, dimensionality ($p$) and missingness rates (mis.) with $n=1000$ (left) and $n= 10,000$ (right). The boxplots are plotted over the replicates, with MissARF (blue) and MissForest (orange) highlighted.} \label{fig: logreg_pred_squared_normal}
\end{figure}

\begin{figure}[p]
\centering
\includegraphics[width=0.92\linewidth]{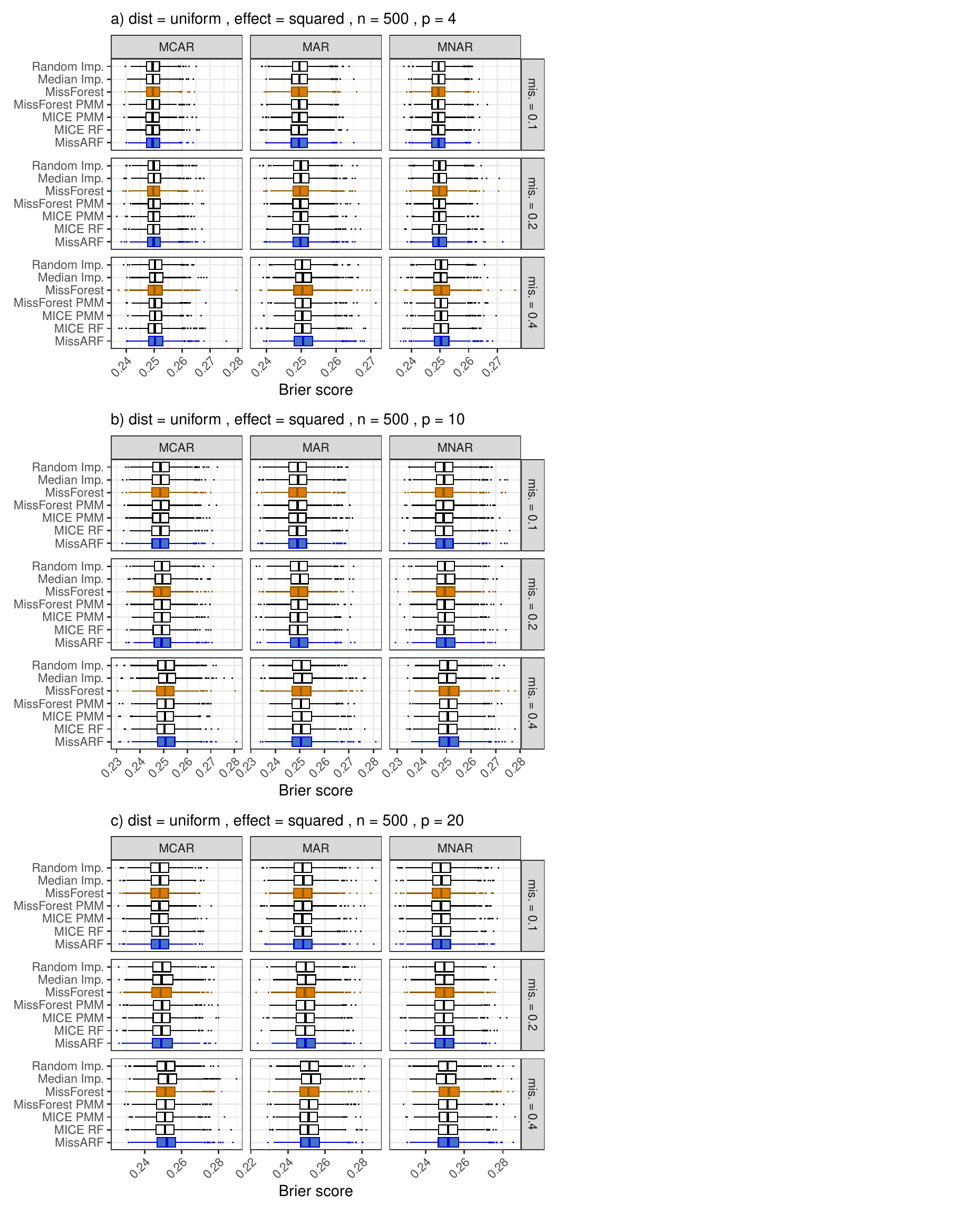}
\caption{ \textbf{Brier Score} of the uniform distribution setting with a squared effect over different missingness patterns, dimensionality ($p$) and missingness rates (mis.) with $n=500$. The boxplots are plotted over the replicates, with MissARF (blue) and MissForest (orange) highlighted.} \label{fig: logreg_pred_squared_uniform_500}
\end{figure}

\begin{figure}[p]
\centering
\includegraphics[width=0.92\linewidth]{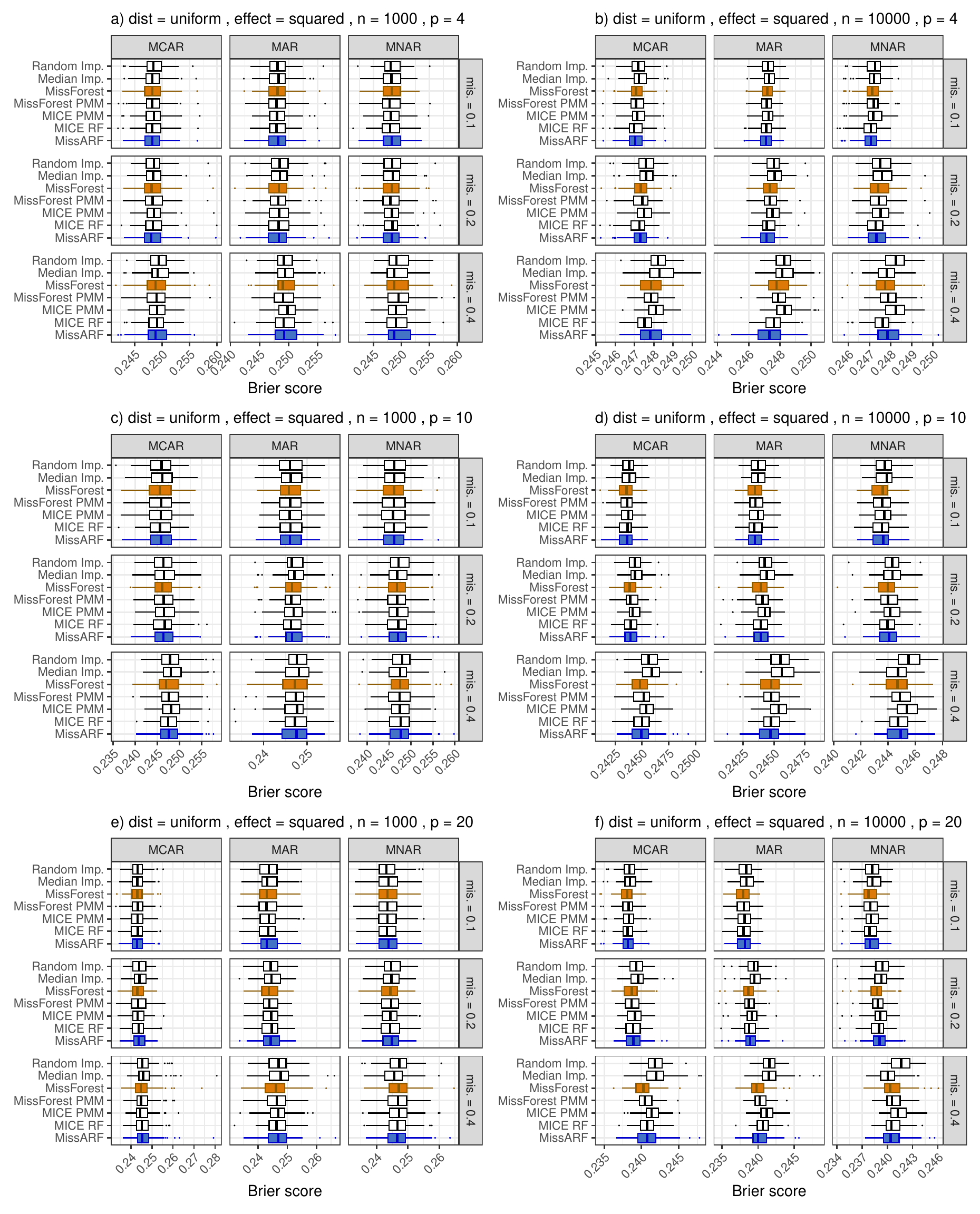}
\caption{ \textbf{Brier Score} of the uniform distribution setting with a squared effect over different missingness patterns, dimensionality ($p$) and missingness rates (mis.) with $n=1000$ (left) and $n= 10,000$ (right). The boxplots are plotted over the replicates, with MissARF (blue) and MissForest (orange) highlighted.} \label{fig: logreg_pred_squared_uniform}
\end{figure}

\begin{figure}[p]
\centering
\includegraphics[width=0.92\linewidth]{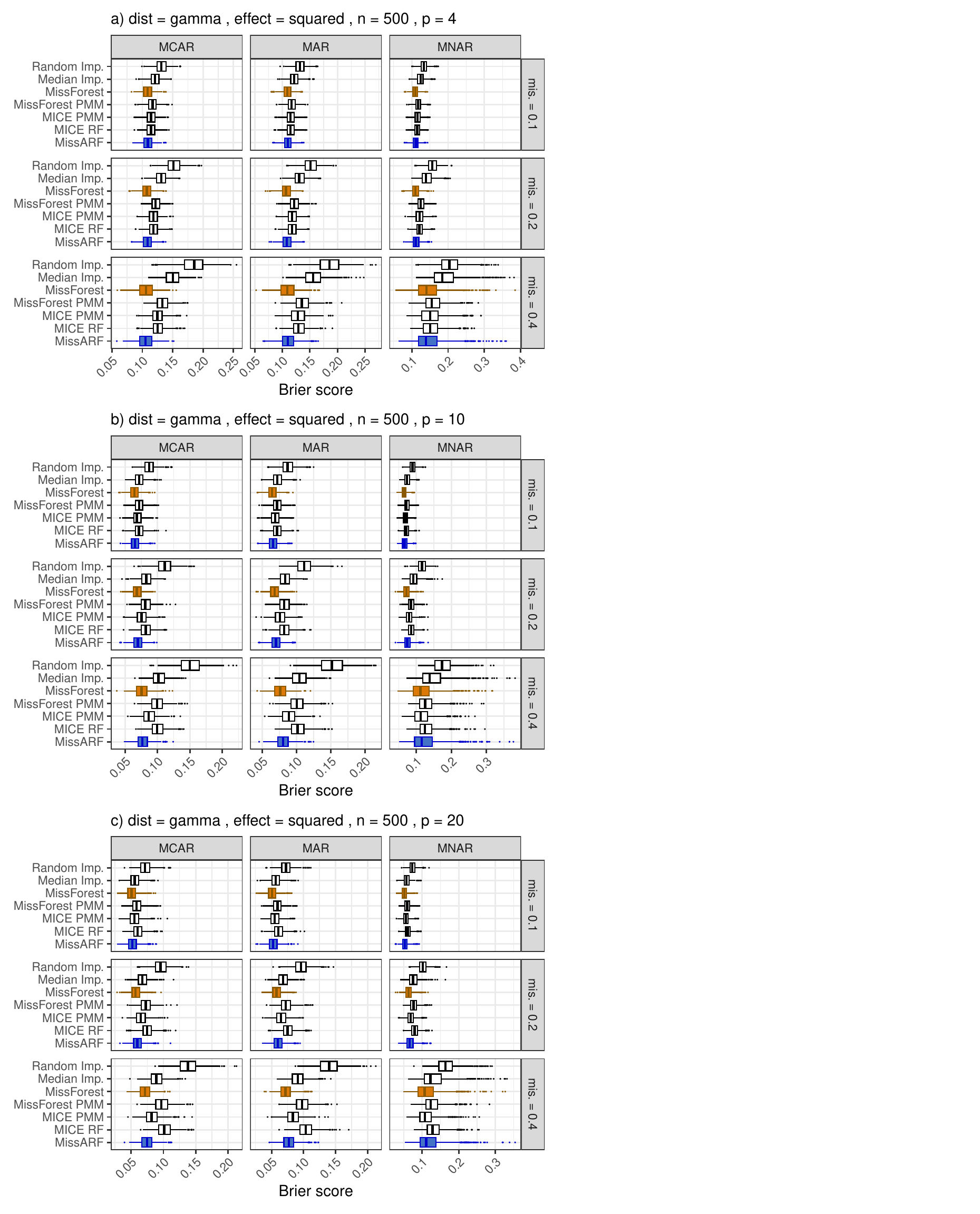}
\caption{ \textbf{Brier Score} of the gamma distribution setting with a squared effect over different missingness patterns, dimensionality ($p$) and missingness rates (mis.) with $n=500$. The boxplots are plotted over the replicates, with MissARF (blue) and MissForest (orange) highlighted.} \label{fig: logreg_pred_squared_gamma_500}
\end{figure}

\begin{figure}[p]
\centering
\includegraphics[width=0.92\linewidth]{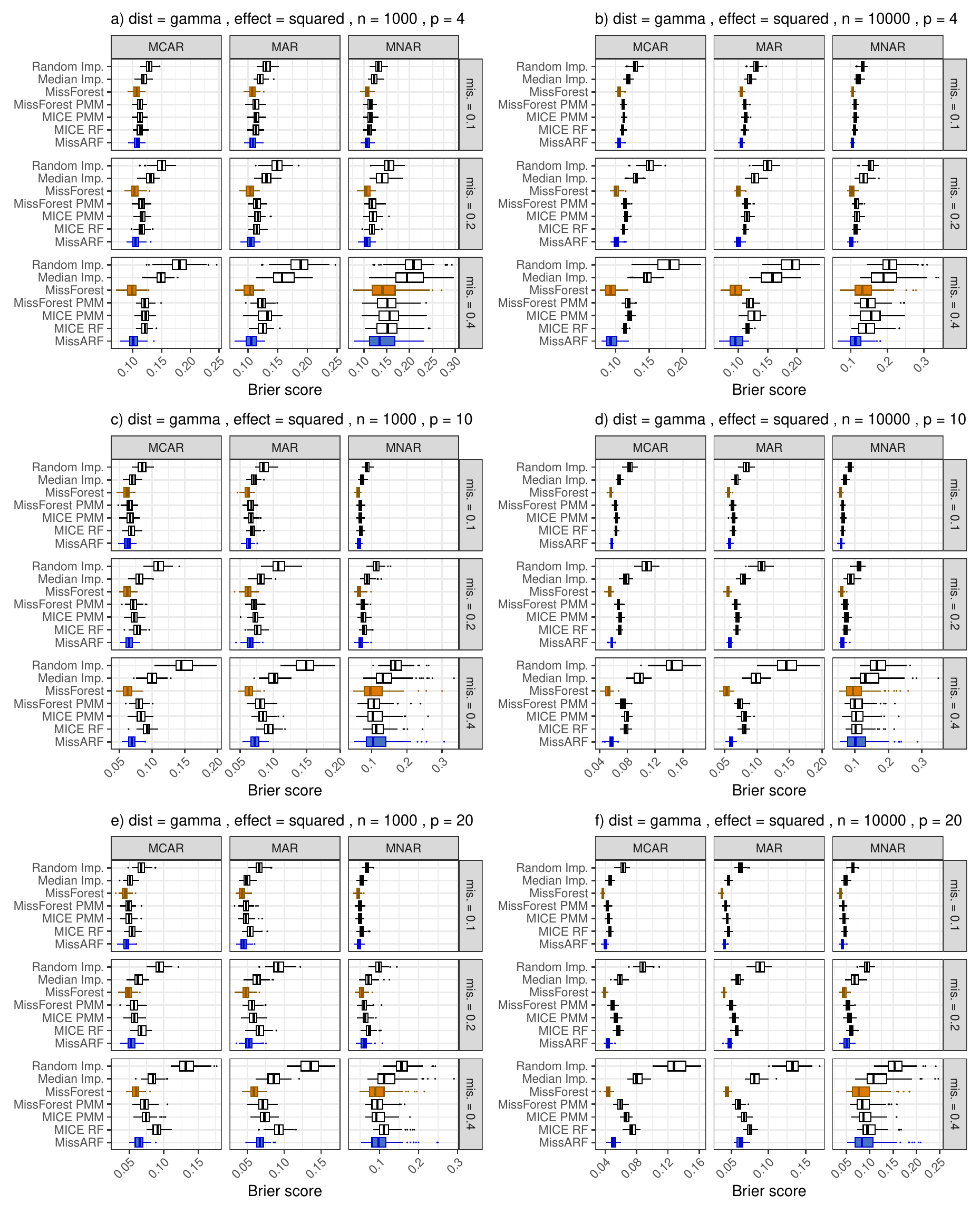}
\caption{ \textbf{Brier Score} of the gamma distribution setting with a squared effect over different missingness patterns, dimensionality ($p$) and missingness rates (mis.) with $n=1000$ (left) and $n= 10,000$ (right). The boxplots are plotted over the replicates, with MissARF (blue) and MissForest (orange) highlighted.} \label{fig: logreg_pred_squared_gamma}
\end{figure}

\begin{figure}[p]
\centering
\includegraphics[width=0.92\linewidth]{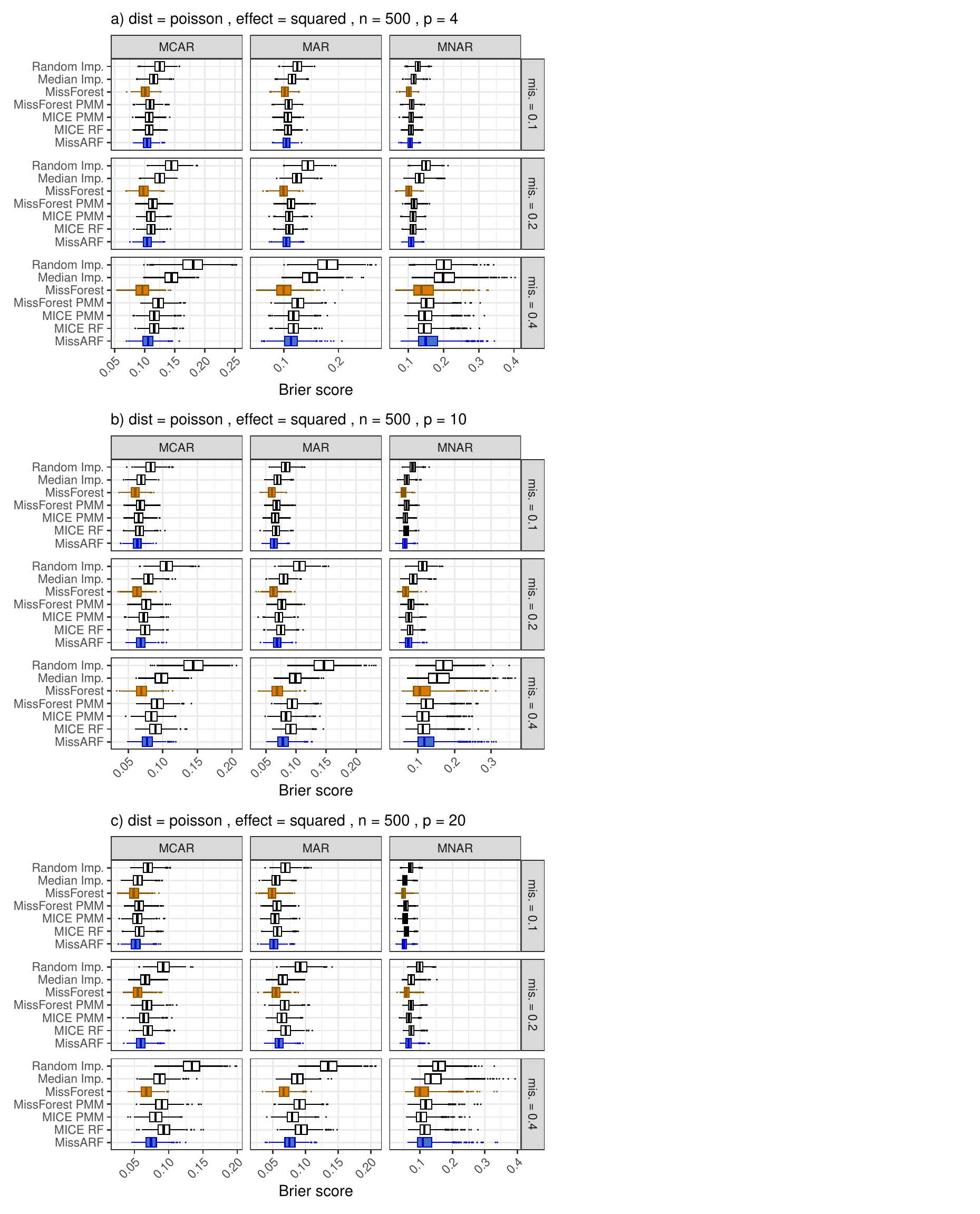}
\caption{ \textbf{Brier Score} of the Poisson distribution setting with a squared effect over different missingness patterns, dimensionality ($p$) and missingness rates (mis.) with $n=500$. The boxplots are plotted over the replicates, with MissARF (blue) and MissForest (orange) highlighted.} \label{fig: logreg_pred_squared_poisson_500}
\end{figure}

\begin{figure}[p]
\centering
\includegraphics[width=0.92\linewidth]{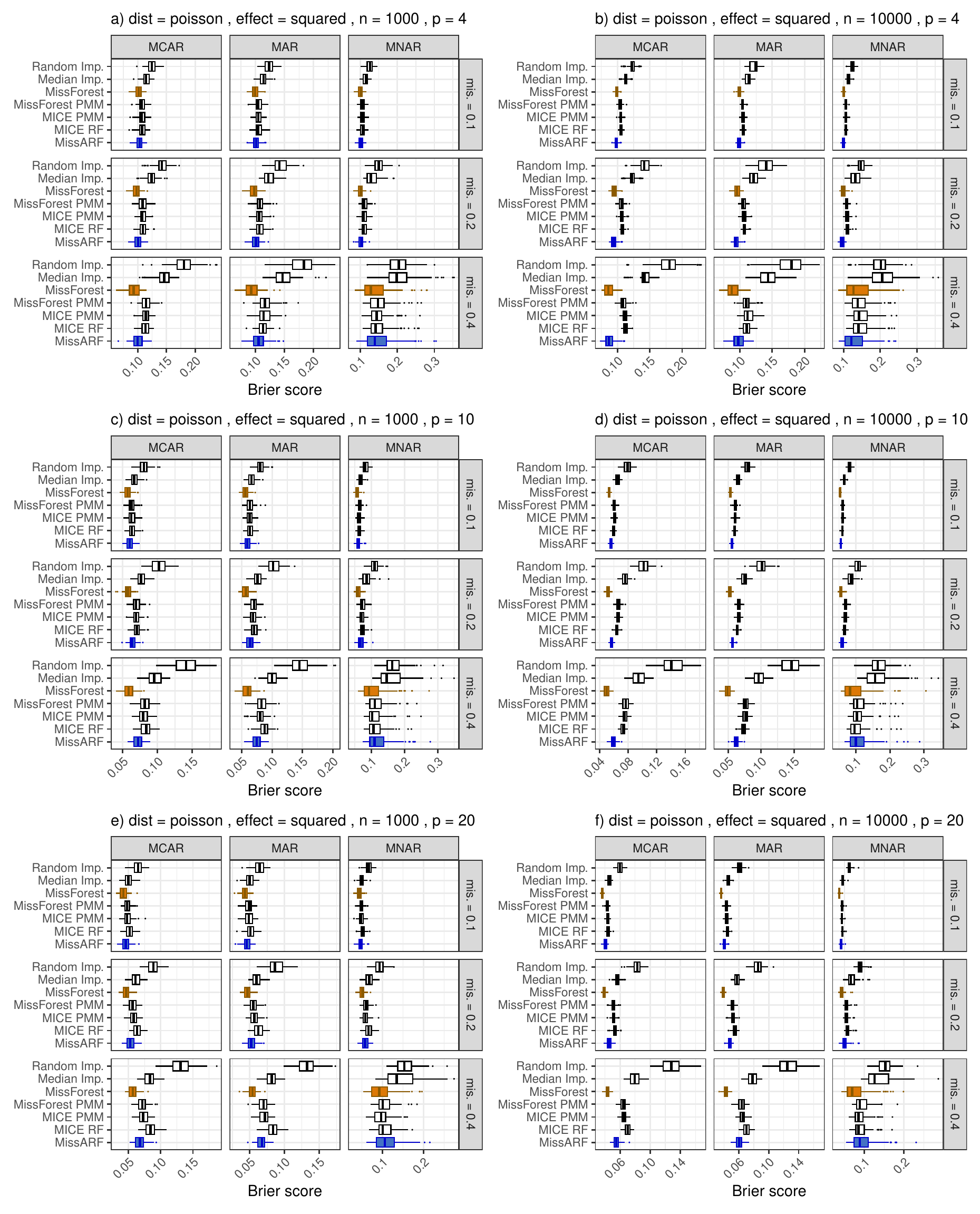}
\caption{ \textbf{Brier Score} of the Poisson distribution setting with a squared effect over different missingness patterns, dimensionality ($p$) and missingness rates (mis.) with $n=1000$ (left) and $n= 10,000$ (right). The boxplots are plotted over the replicates, with MissARF (blue) and MissForest (orange) highlighted.} \label{fig: logreg_pred_squared_poisson}
\end{figure}

\begin{figure}[p]
\centering
\includegraphics[width=0.92\linewidth]{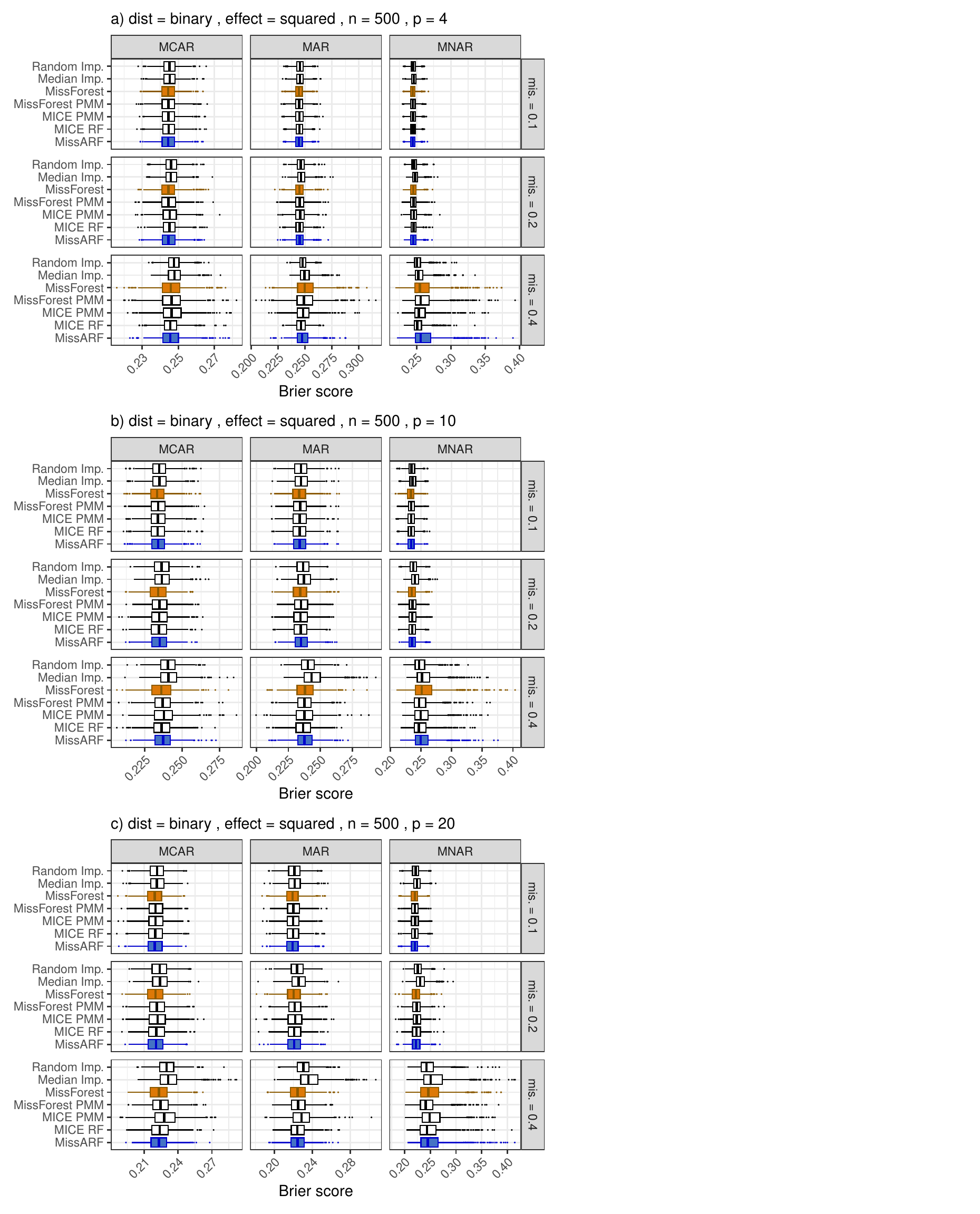}
\caption{ \textbf{Brier Score} of the binary distribution setting with a squared effect over different missingness patterns, dimensionality ($p$) and missingness rates (mis.) with $n=500$. The boxplots are plotted over the replicates, with MissARF (blue) and MissForest (orange) highlighted.} \label{fig: logreg_pred_squared_binary_500}
\end{figure}

\begin{figure}[p]
\centering
\includegraphics[width=0.92\linewidth]{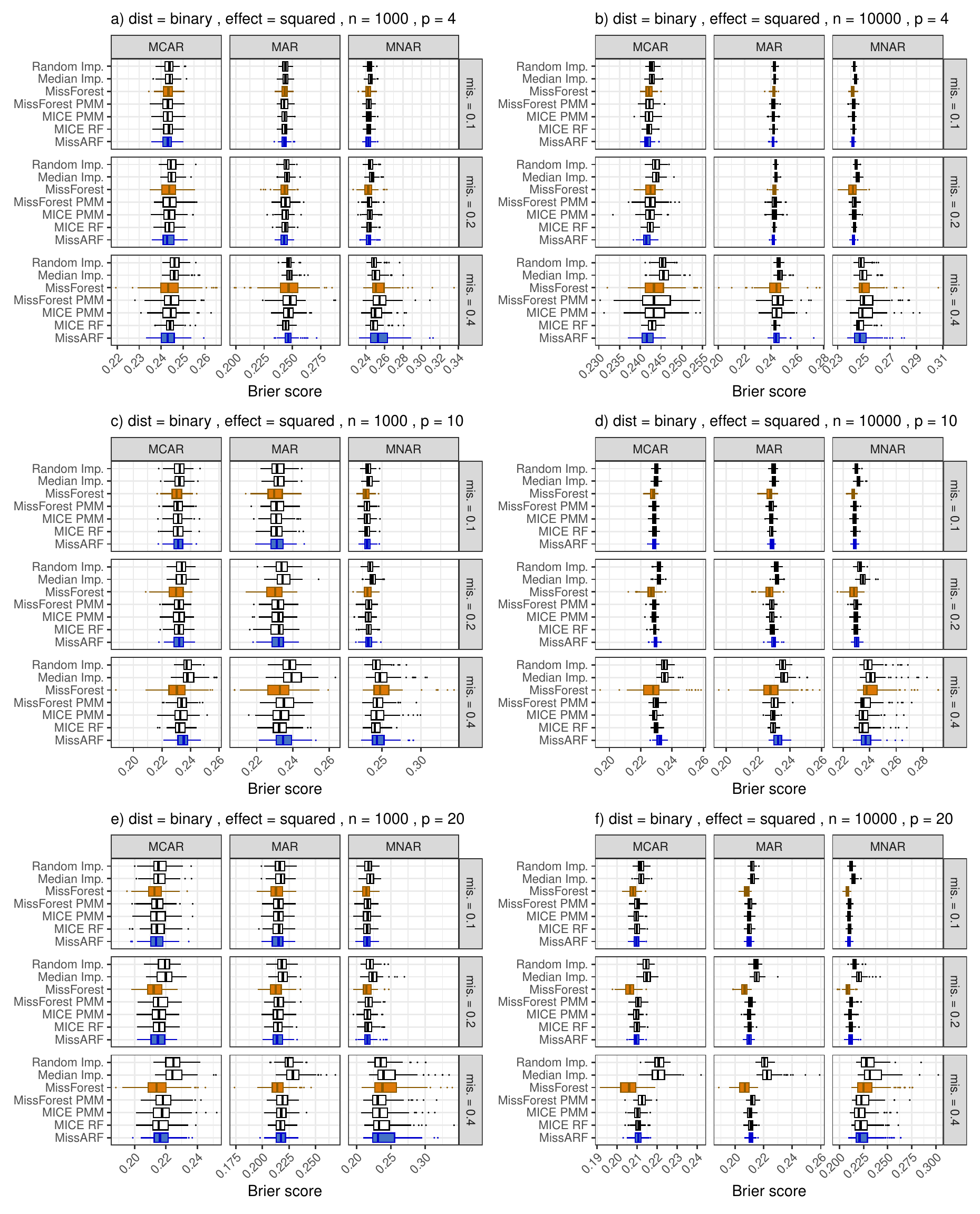}
\caption{ \textbf{Brier Score} of the binary distribution setting with a squared effect over different missingness patterns, dimensionality ($p$) and missingness rates (mis.) with $n=1000$ (left) and $n= 10,000$ (right). The boxplots are plotted over the replicates, with MissARF (blue) and MissForest (orange) highlighted.} \label{fig: logreg_pred_squared_binary}
\end{figure}

\clearpage
\section{Setting II: Multiple imputation}
\subsection{Coverage rate}
%% Coverage rate
\subsubsection{Category 1: Similar performance across all methods, MissARF with smallest average width}

\begin{figure}[!ht]
\centering
\includegraphics[width=0.92\linewidth]{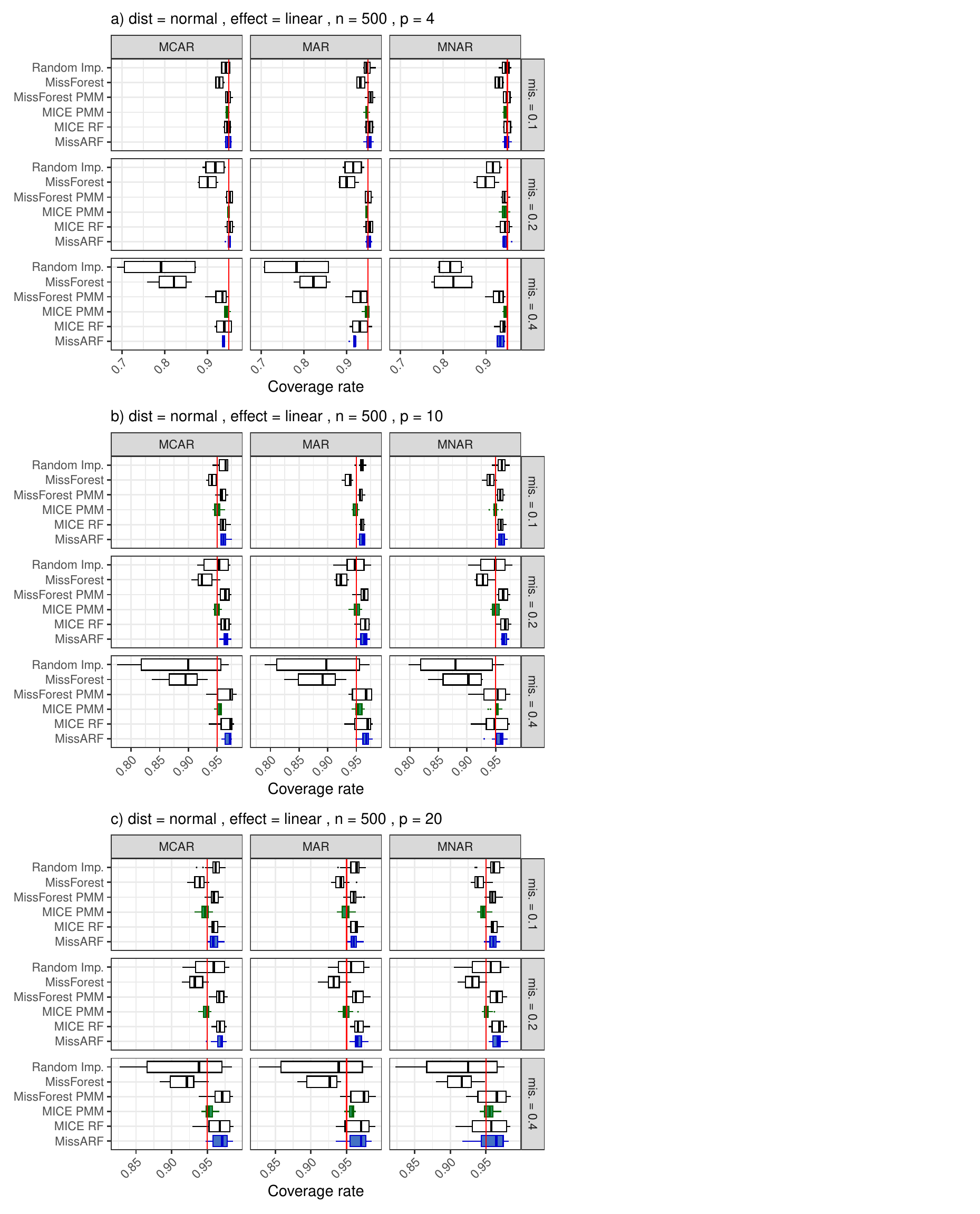}
\caption{ \textbf{Coverage rate} of the normal distribution setting with a linear effect over different missingness patterns, dimensionality ($p$) and missingness rates (mis.) with $n=500$. The red vertical line shows the nominal coverage level of 0.95. Boxplots are plotted over features, with MissARF (blue) and MICE PMM (green).} \label{fig: logreg_coverage_linear_normal_500}
\end{figure}

\begin{figure}[!ht]
\centering
\includegraphics[width=0.92\linewidth]{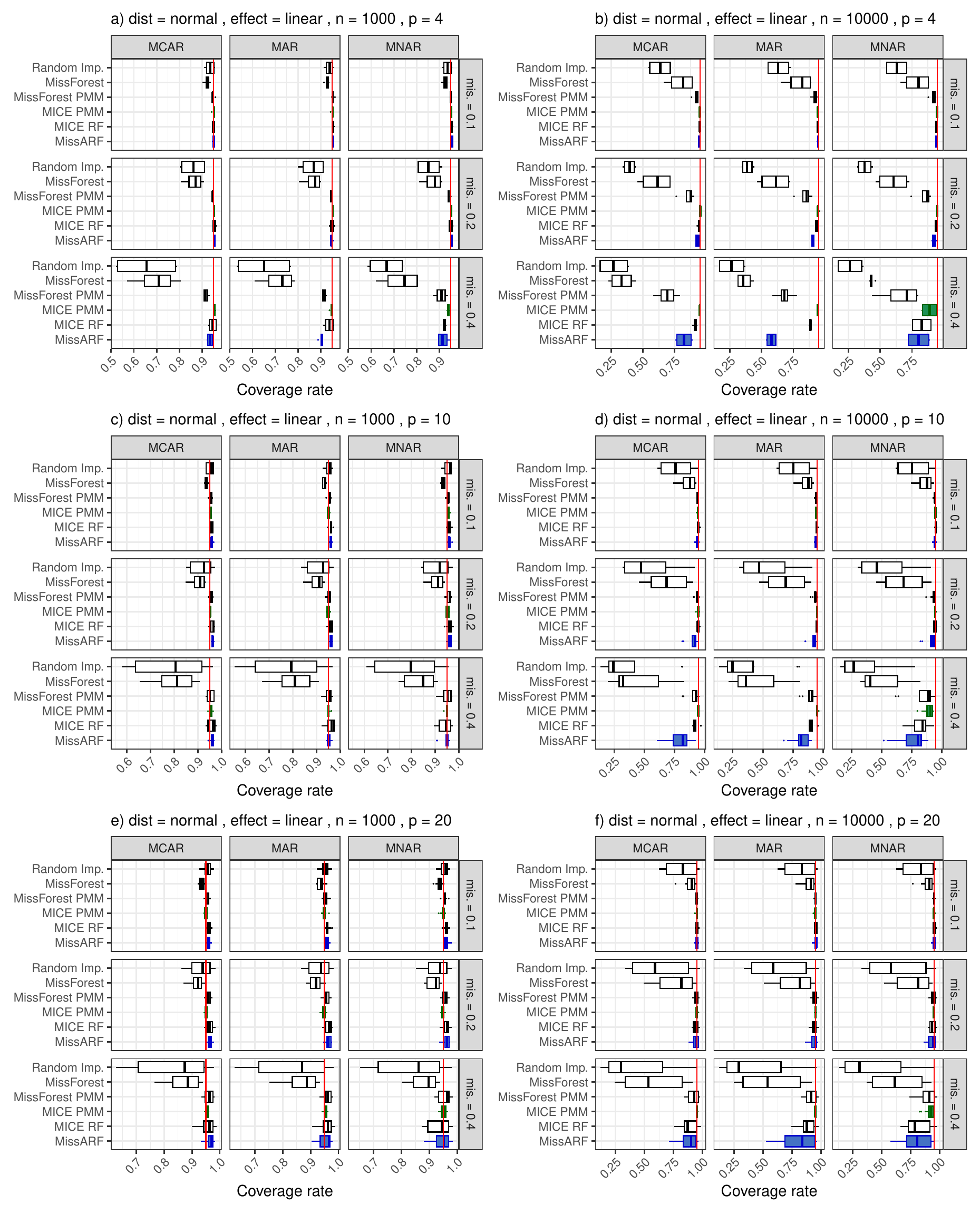}
\caption{ \textbf{Coverage rate} of the normal distribution setting with a linear effect over different missingness patterns, dimensionality ($p$) and missingness rates (mis.) with $n=1000$ (left) and $n= 10,000$ (right). The red vertical line shows the nominal coverage level of 0.95. Boxplots are plotted over features, with MissARF (blue) and MICE PMM (green).} \label{fig: logreg_coverage_linear_normal}
\end{figure}

\begin{figure}[p]
\centering
\includegraphics[width=0.92\linewidth]{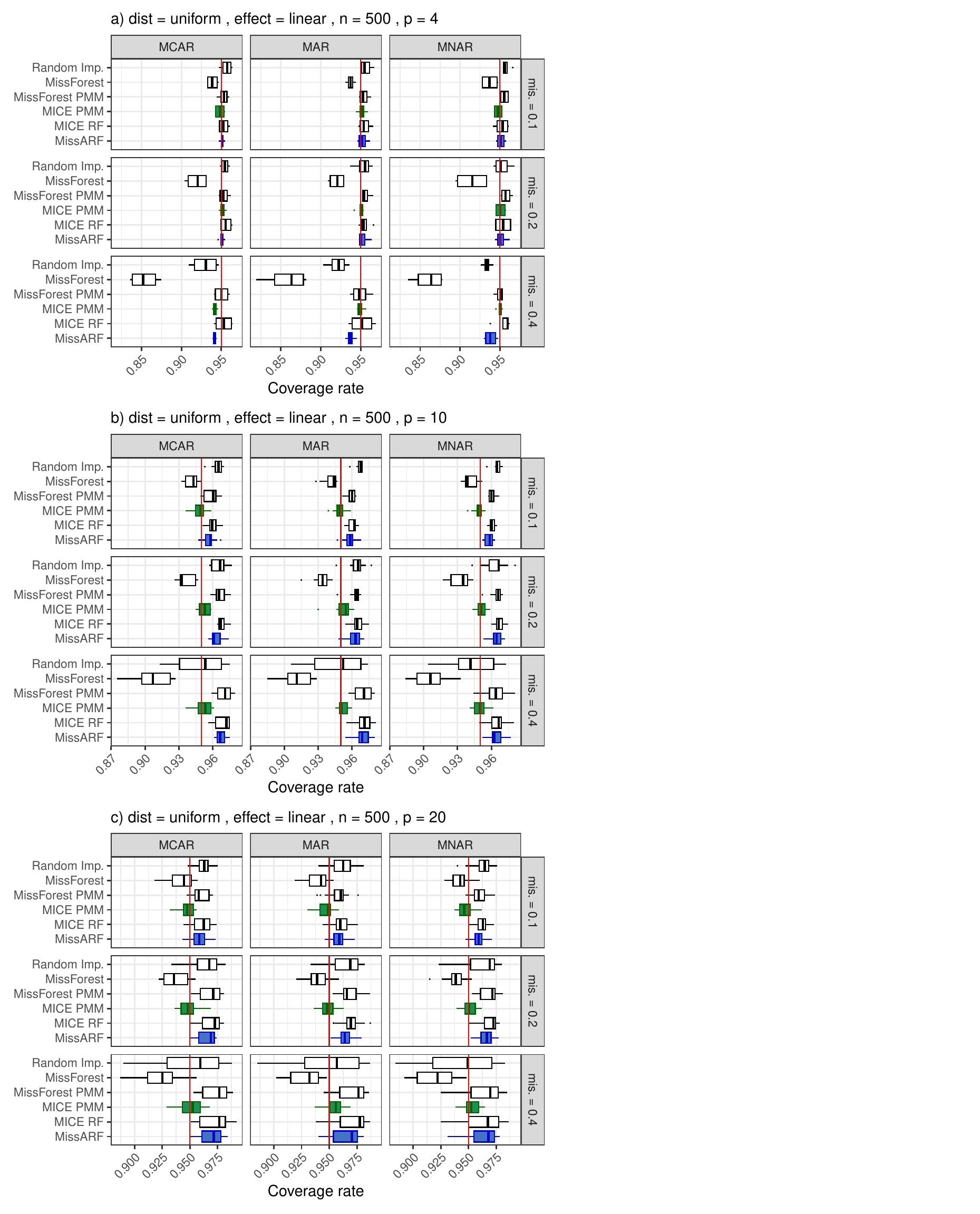}
\caption{ \textbf{Coverage rate} of the uniform distribution setting with a linear effect over different missingness patterns, dimensionality ($p$) and missingness rates (mis.) with $n=500$. The red vertical line shows the nominal coverage level of 0.95. Boxplots are plotted over features, with MissARF (blue) and MICE PMM (green).} \label{fig: logreg_coverage_linear_uniform_500}
\end{figure}

\begin{figure}[p]
\centering
\includegraphics[width=0.92\linewidth]{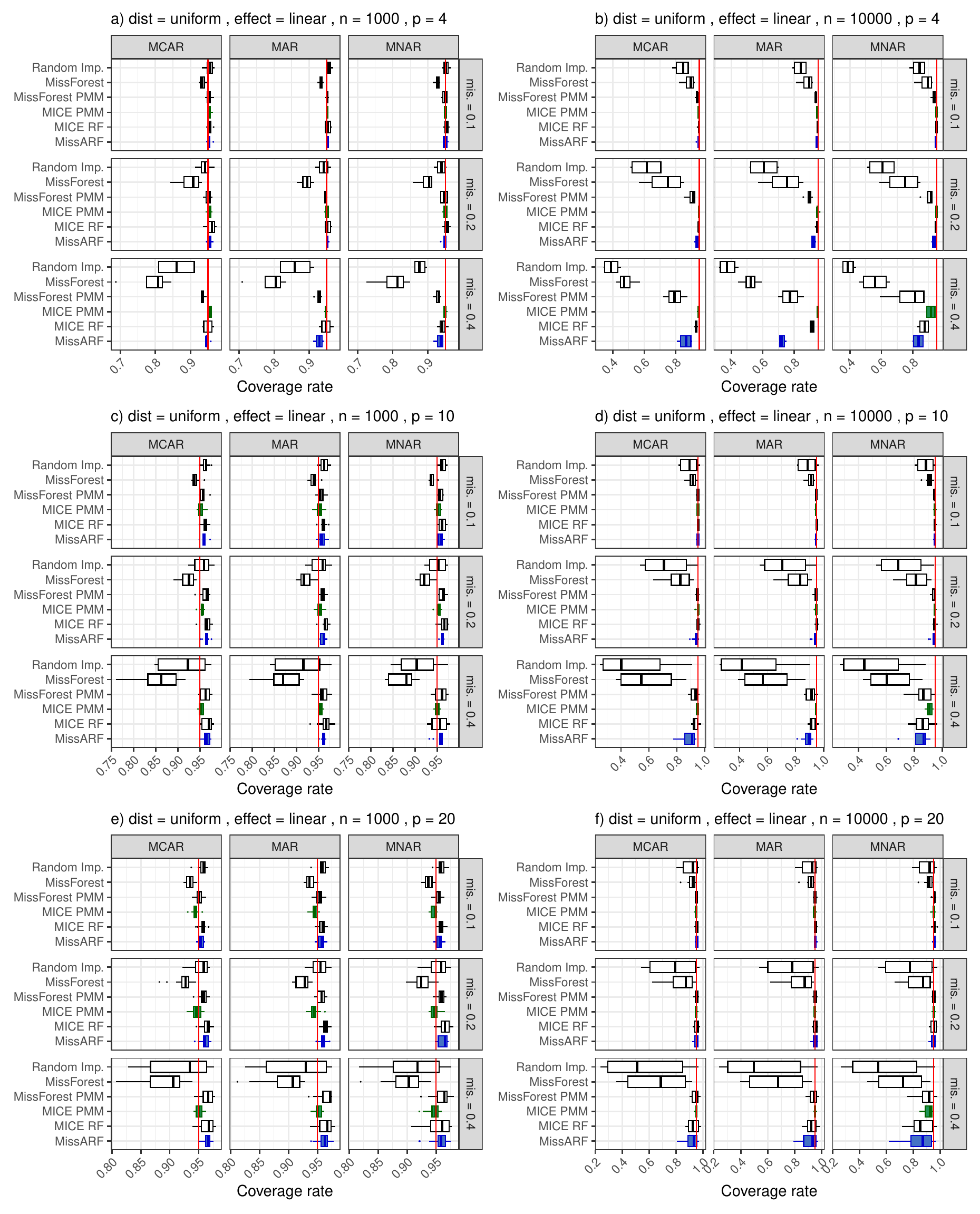}
\caption{ \textbf{Coverage rate} of the uniform distribution setting with a linear effect over different missingness patterns, dimensionality ($p$) and missingness rates (mis.) with $n=1000$ (left) and $n= 10,000$ (right). The red vertical line shows the nominal coverage level of 0.95. Boxplots are plotted over features, with MissARF (blue) and MICE PMM (green).} \label{fig: logreg_coverage_linear_uniform}
\end{figure}

\begin{figure}[p]
\centering
\includegraphics[width=0.92\linewidth]{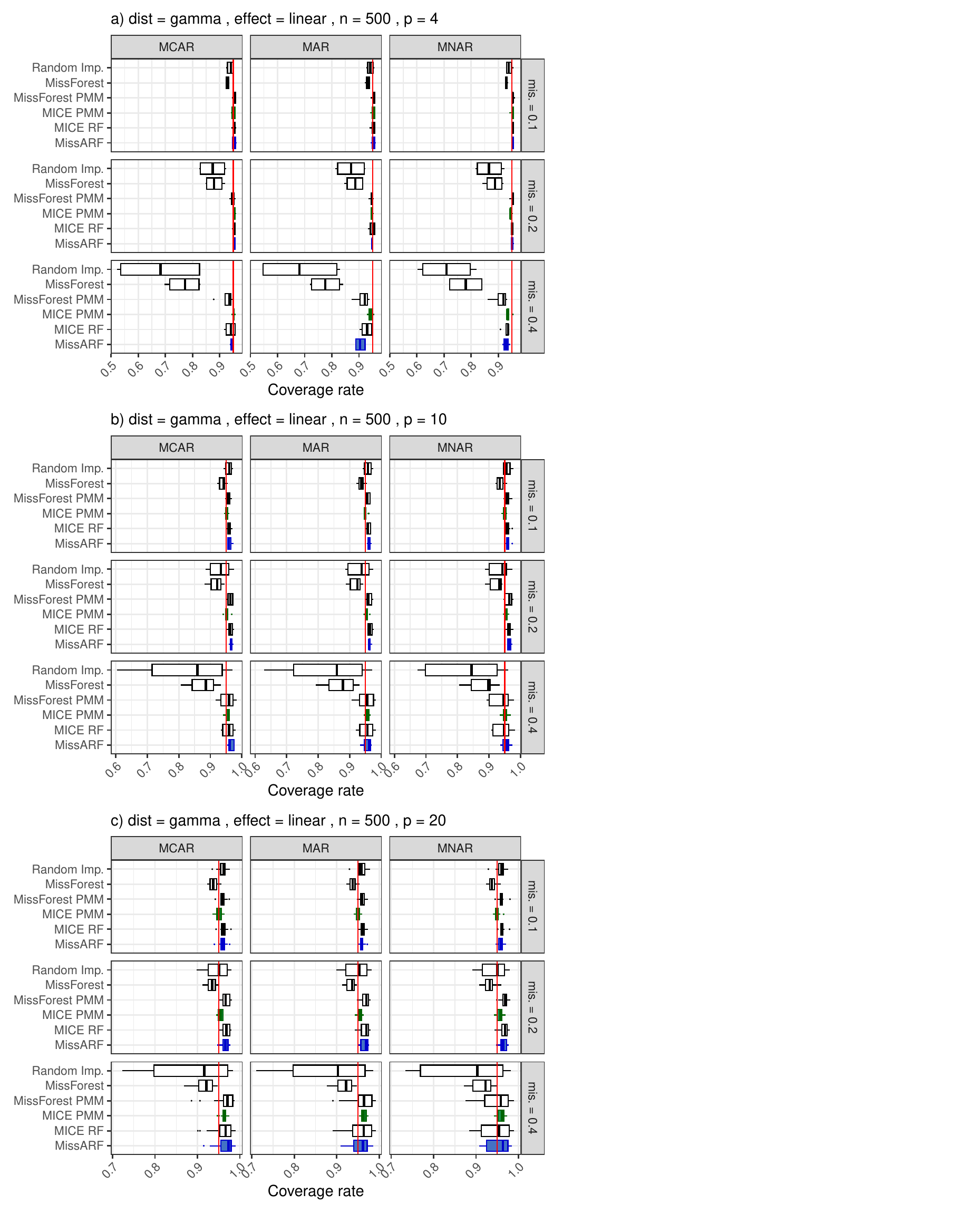}
\caption{ \textbf{Coverage rate} of the gamma distribution setting with a linear effect over different missingness patterns, dimensionality ($p$) and missingness rates (mis.) with $n=500$. The red vertical line shows the nominal coverage level of 0.95. Boxplots are plotted over features, with MissARF (blue) and MICE PMM (green).} \label{fig: logreg_coverage_linear_gamma_500}
\end{figure}

\begin{figure}[p]
\centering
\includegraphics[width=0.92\linewidth]{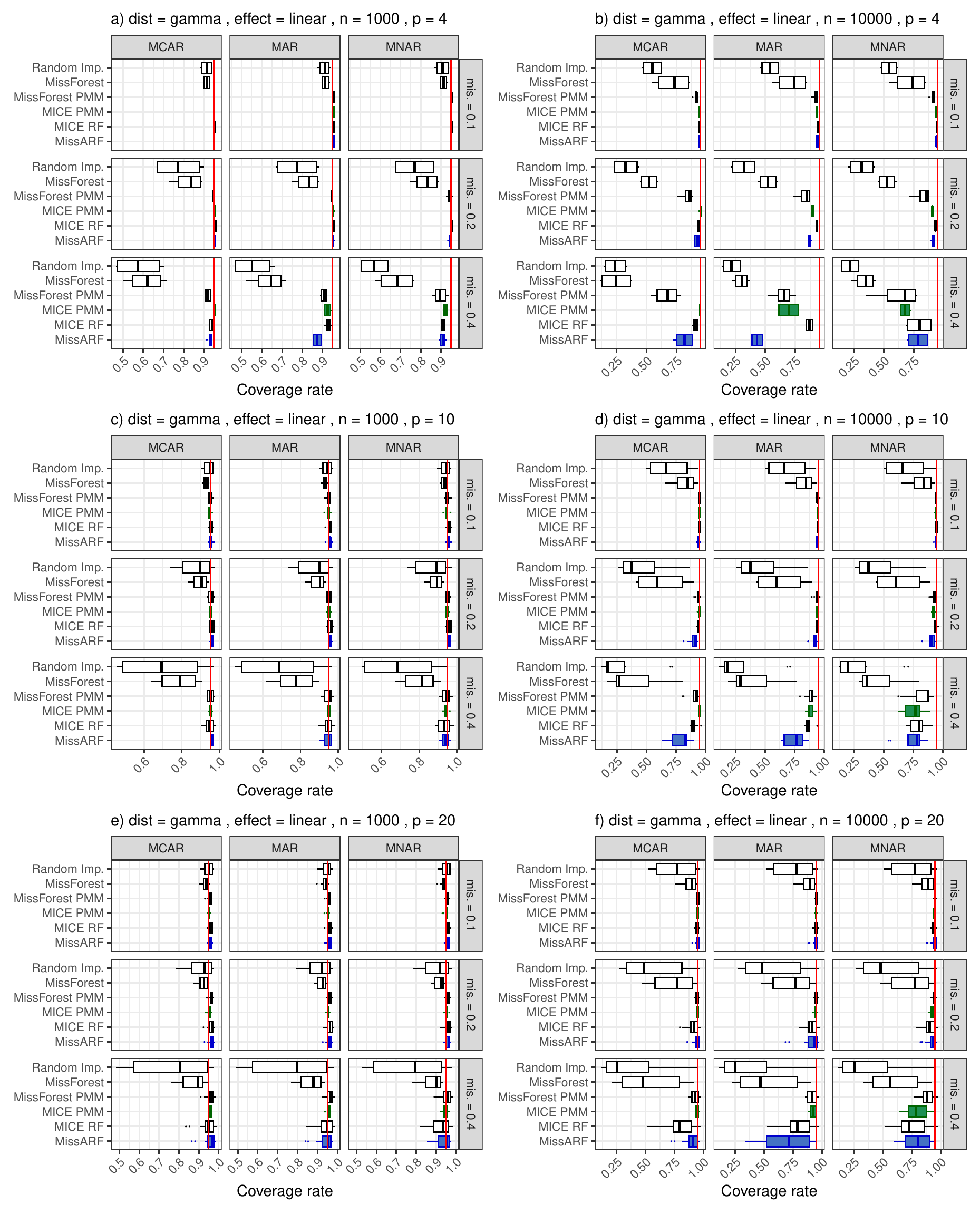}
\caption{ \textbf{Coverage rate} of the gamma distribution setting with a linear effect over different missingness patterns, dimensionality ($p$) and missingness rates (mis.) with $n=1000$ (left) and $n= 10,000$ (right). The red vertical line shows the nominal coverage level of 0.95. Boxplots are plotted over features, with MissARF (blue) and MICE PMM (green).} \label{fig: logreg_coverage_linear_gamma}
\end{figure}

\begin{figure}[p]
\centering
\includegraphics[width=0.92\linewidth]{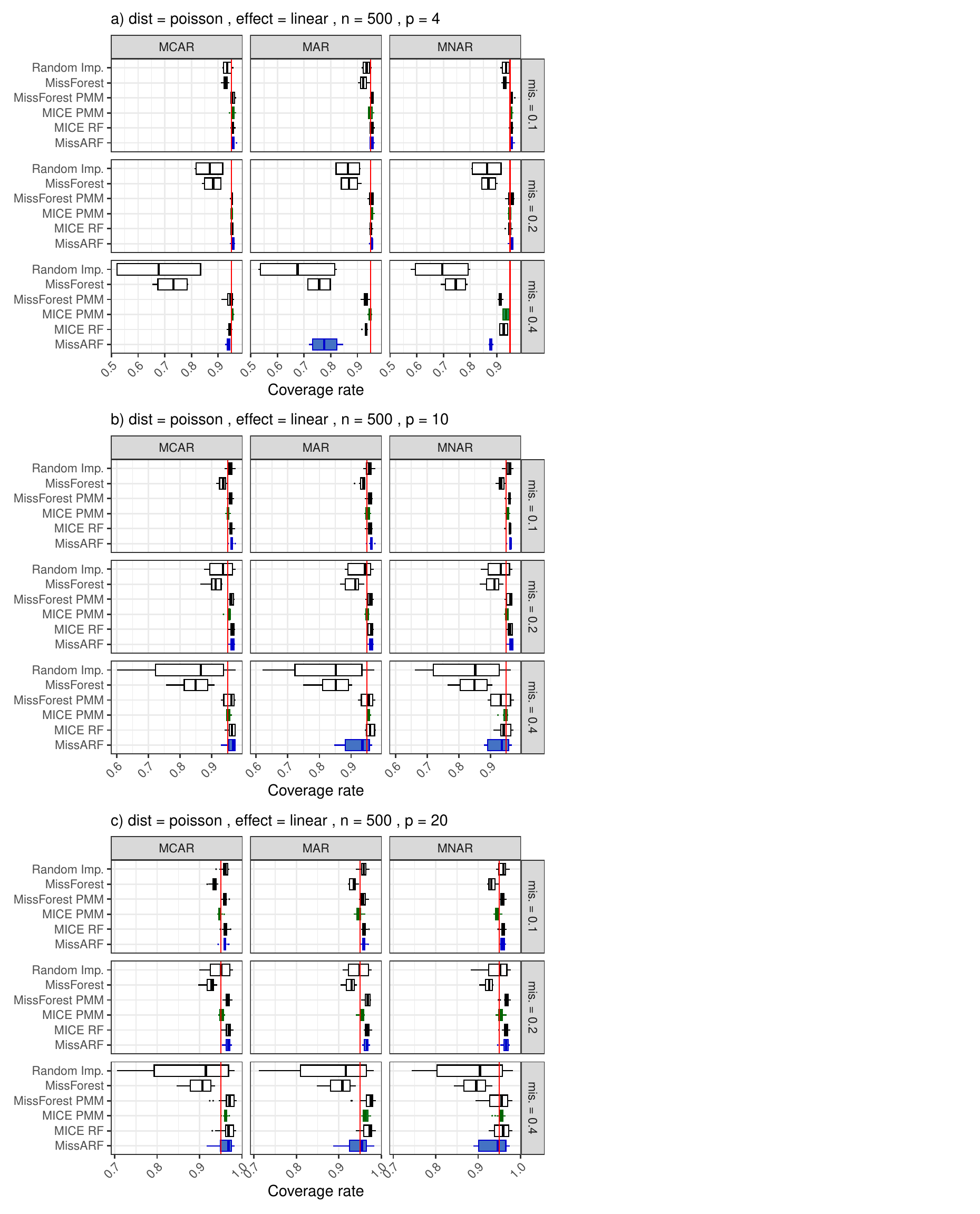}
\caption{ \textbf{Coverage rate} of the Poisson distribution setting with a linear effect over different missingness patterns, dimensionality ($p$) and missingness rates (mis.) with $n=500$. The red vertical line shows the nominal coverage level of 0.95. Boxplots are plotted over features, with MissARF (blue) and MICE PMM (green).} \label{fig: logreg_coverage_linear_poisson_500}
\end{figure}

\begin{figure}[p]
\centering
\includegraphics[width=0.92\linewidth]{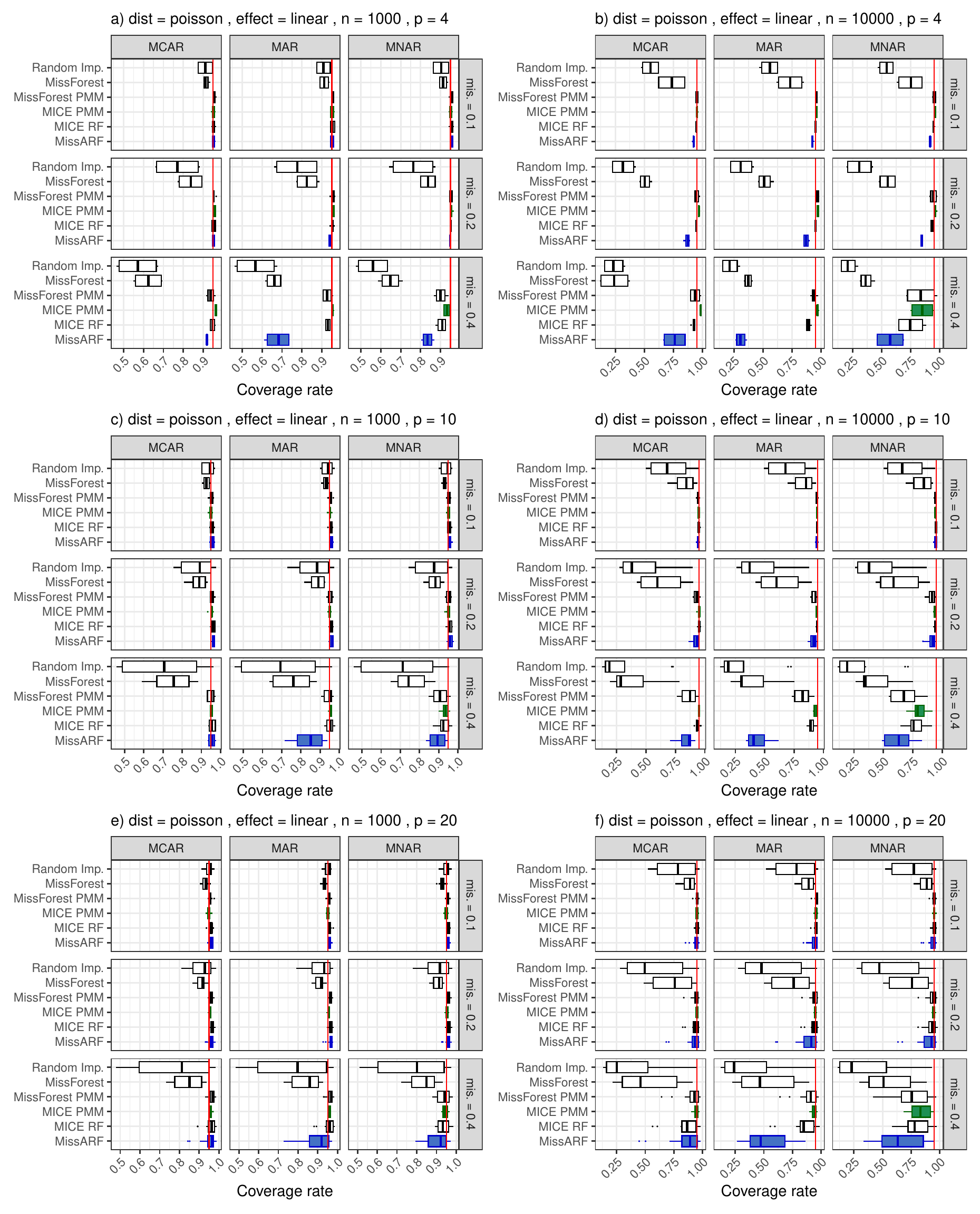}
\caption{ \textbf{Coverage rate} of the Poisson distribution setting with a linear effect over different missingness patterns, dimensionality ($p$) and missingness rates (mis.) with $n=1000$ (left) and $n= 10,000$ (right). The red vertical line shows the nominal coverage level of 0.95. Boxplots are plotted over features, with MissARF (blue) and MICE PMM (green).} \label{fig: logreg_coverage_linear_poisson}
\end{figure}

\begin{figure}[p]
\centering
\includegraphics[width=0.92\linewidth]{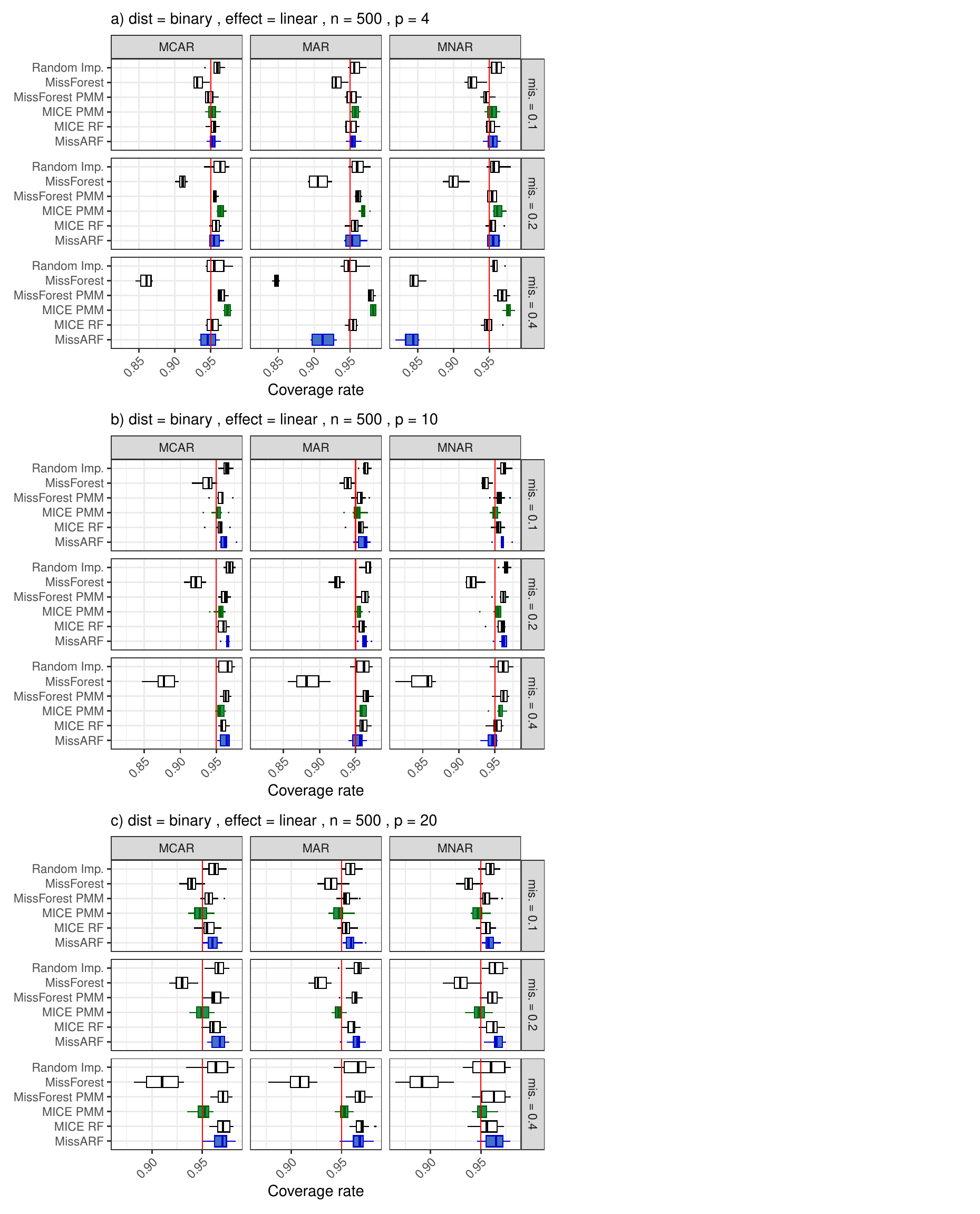}
\caption{ \textbf{Coverage rate} of the binary distribution setting with a linear effect over different missingness patterns, dimensionality ($p$) and missingness rates (mis.) with $n=500$. The red vertical line shows the nominal coverage level of 0.95. Boxplots are plotted over features, with MissARF (blue) and MICE PMM (green).} \label{fig: logreg_coverage_linear_binary_500}
\end{figure}

\begin{figure}[p]
\centering
\includegraphics[width=0.92\linewidth]{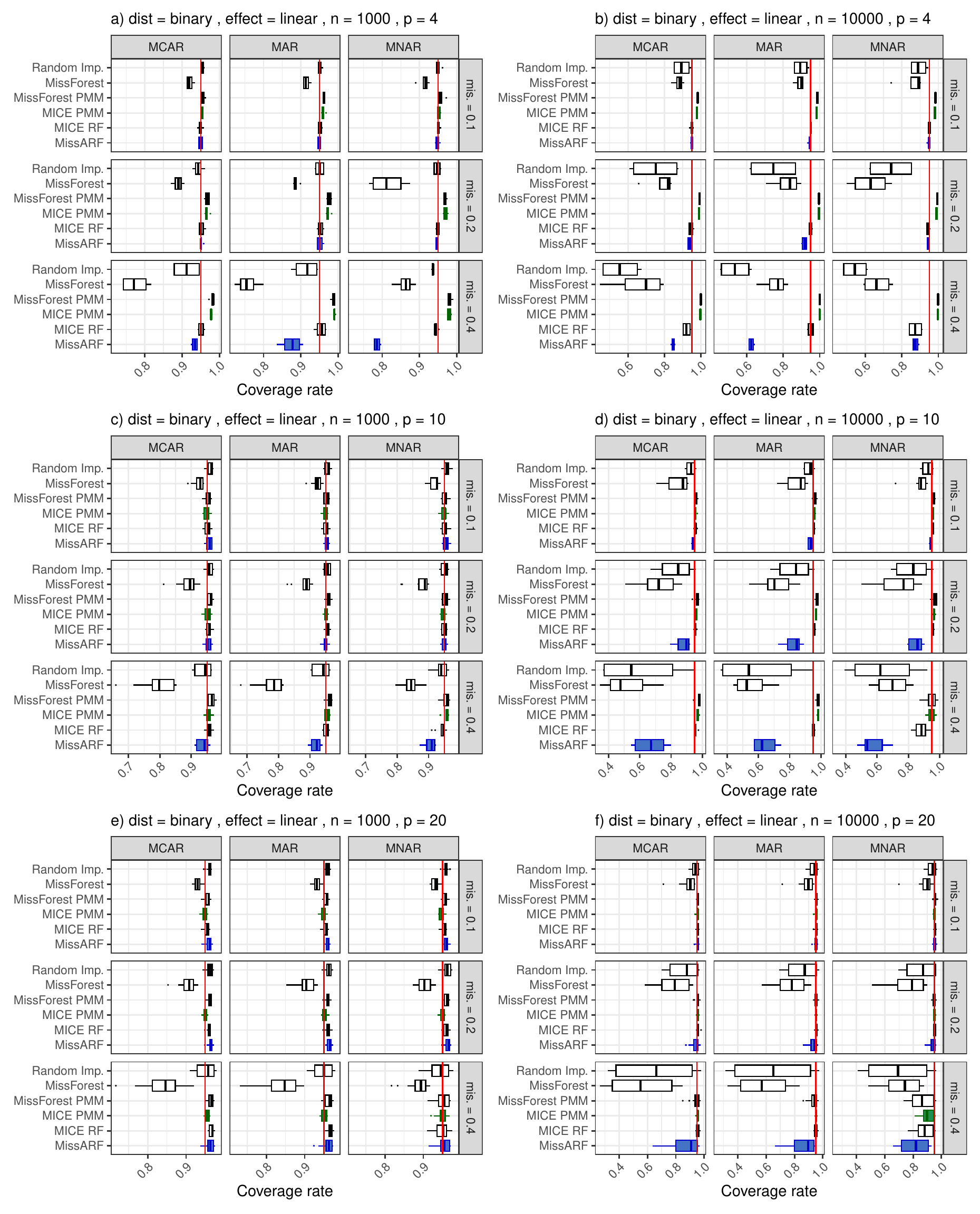}
\caption{ \textbf{Coverage rate} of the binary distribution setting with a linear effect over different missingness patterns, dimensionality ($p$) and missingness rates (mis.) with $n=1000$ (left) and $n= 10,000$ (right). The red vertical line shows the nominal coverage level of 0.95. Boxplots are plotted over features, with MissARF (blue) and MICE PMM (green).} \label{fig: logreg_coverage_linear_binary}
\end{figure}

\begin{figure}[p]
\centering
\includegraphics[width=0.92\linewidth]{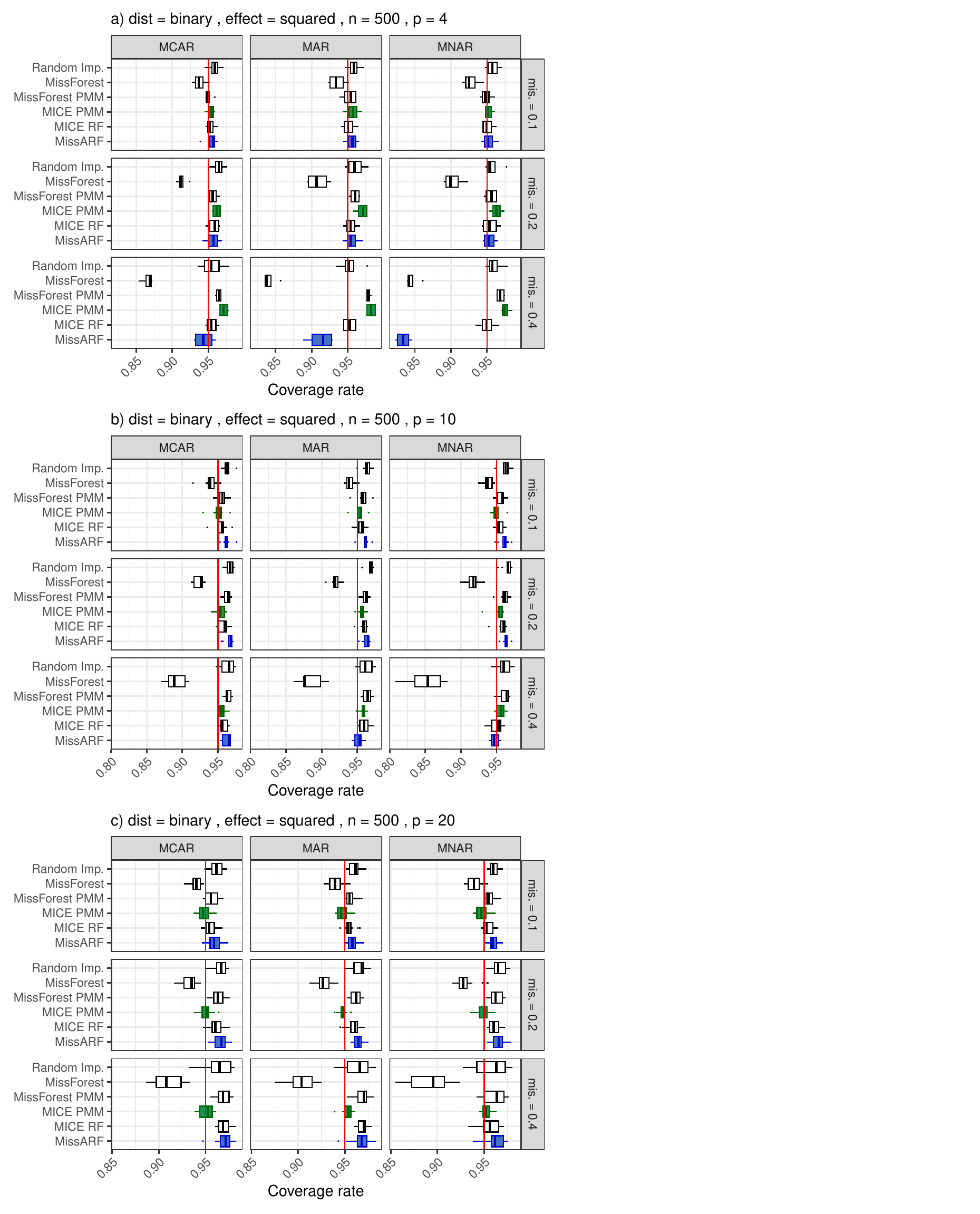}
\caption{ \textbf{Coverage rate} of the binary distribution setting with a squared effect over different missingness patterns, dimensionality ($p$) and missingness rates (mis.) with $n=500$. The red vertical line shows the nominal coverage level of 0.95. Boxplots are plotted over features, with MissARF (blue) and MICE PMM (green).} \label{fig: logreg_coverage_squared_binary_500}
\end{figure}

\begin{figure}[p]
\centering
\includegraphics[width=0.92\linewidth]{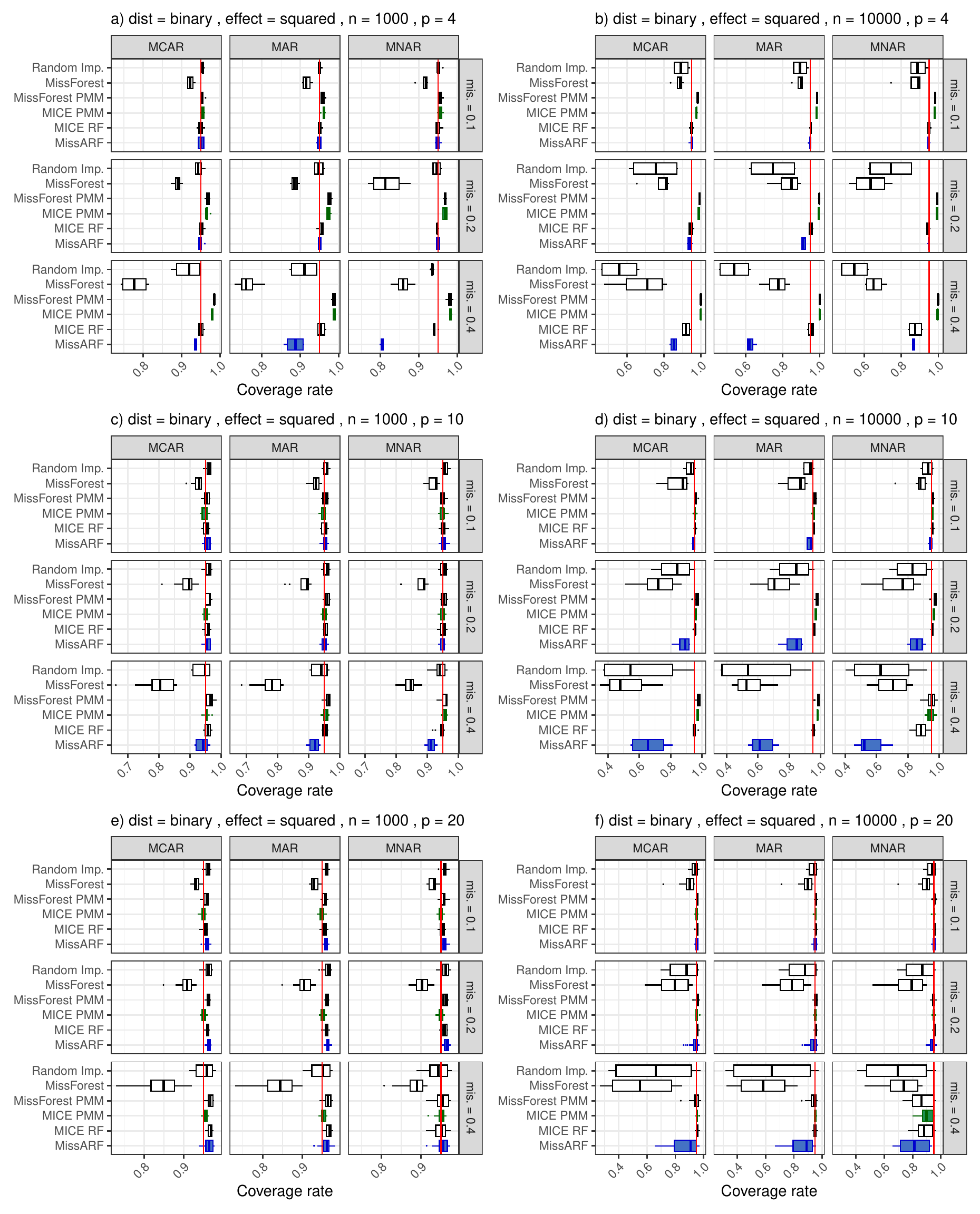}
\caption{ \textbf{Coverage rate} of the binary distribution setting with a squared effect over different missingness patterns, dimensionality ($p$) and missingness rates (mis.) with $n=1000$ (left) and $n= 10,000$ (right). The red vertical line shows the nominal coverage level of 0.95. Boxplots are plotted over features, with MissARF (blue) and MICE PMM (green).} \label{fig: logreg_coverage_squared_binary}
\end{figure}

\clearpage
\subsubsection{Category 2: PMM methods struggle, MissARF performs well}
\begin{figure}[!ht]
\centering
\includegraphics[width=0.92\linewidth]{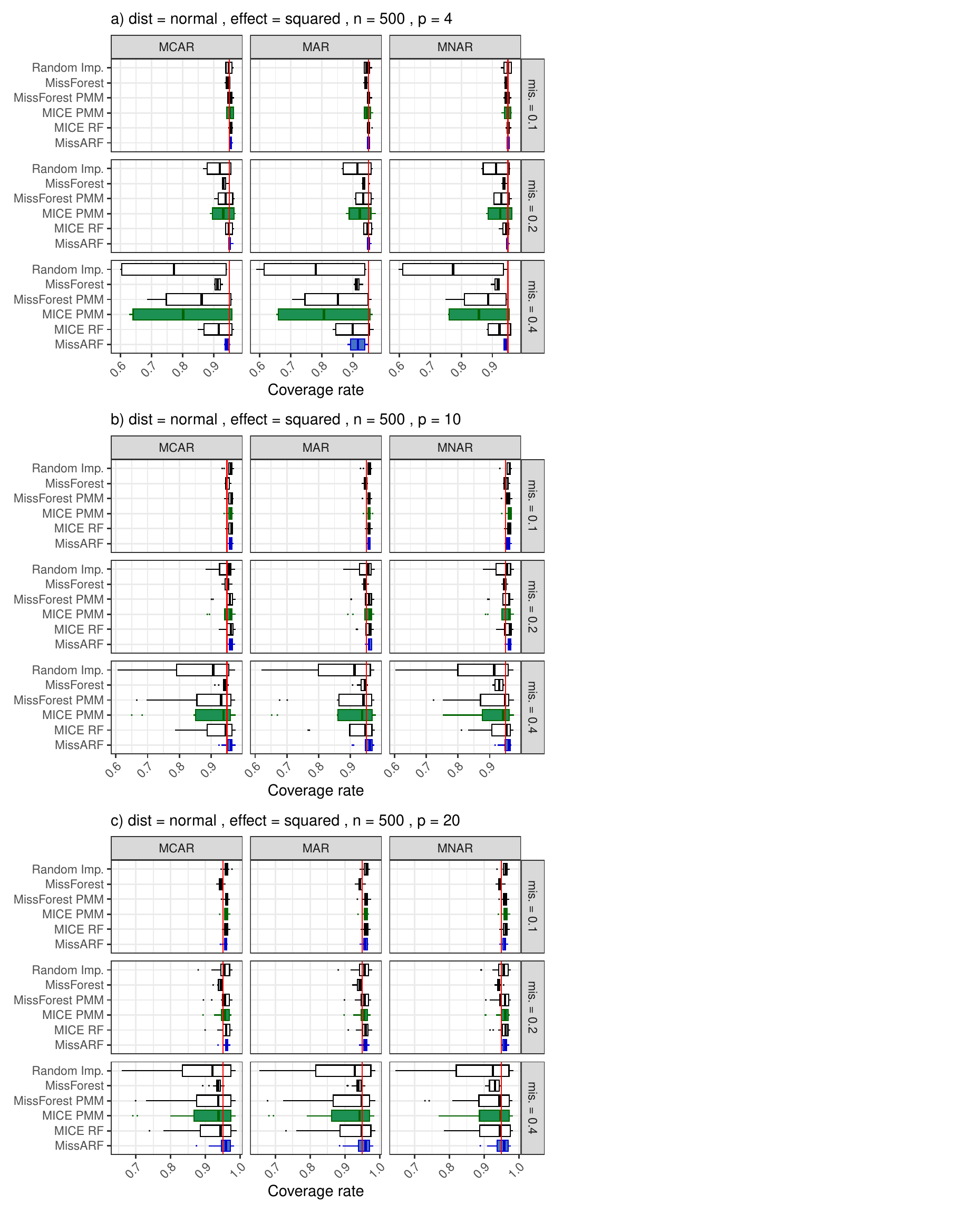}
\caption{ \textbf{Coverage rate} of the normal distribution setting with a squared effect over different missingness patterns, dimensionality ($p$) and missingness rates (mis.) with $n=500$. The red vertical line shows the nominal coverage level of 0.95. Boxplots are plotted over features, with MissARF (blue) and MICE PMM (green).} \label{fig: logreg_coverage_squared_normal_500}
\end{figure}

\begin{figure}[p]
\centering
\includegraphics[width=0.92\linewidth]{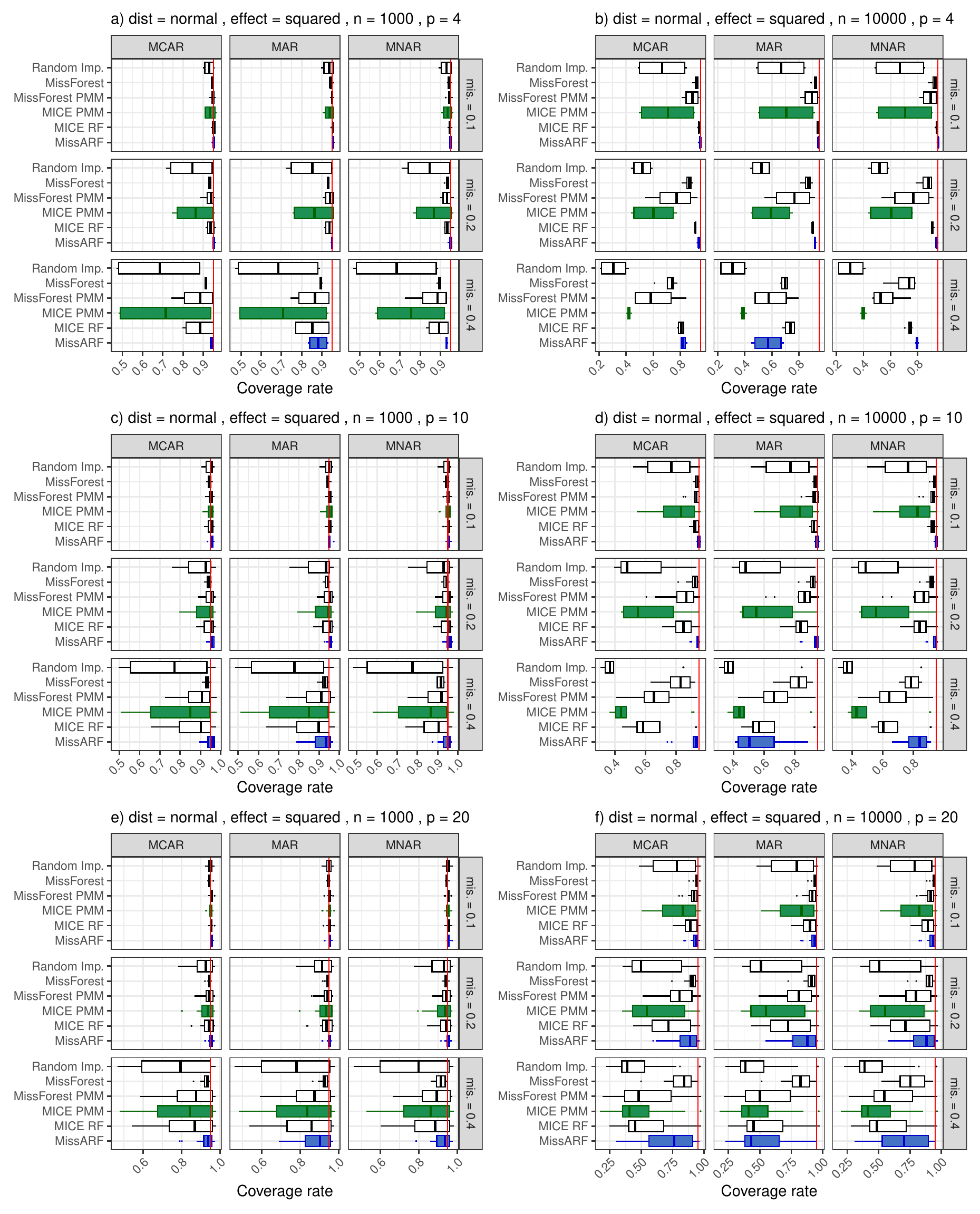}
\caption{ \textbf{Coverage rate} of the normal distribution setting with a squared effect over different missingness patterns, dimensionality ($p$) and missingness rates (mis.) with $n=1000$ (left) and $n= 10,000$ (right). The red vertical line shows the nominal coverage level of 0.95. Boxplots are plotted over features, with MissARF (blue) and MICE PMM (green).} \label{fig: logreg_coverage_squared_normal}
\end{figure}

\begin{figure}[p]
\centering
\includegraphics[width=0.92\linewidth]{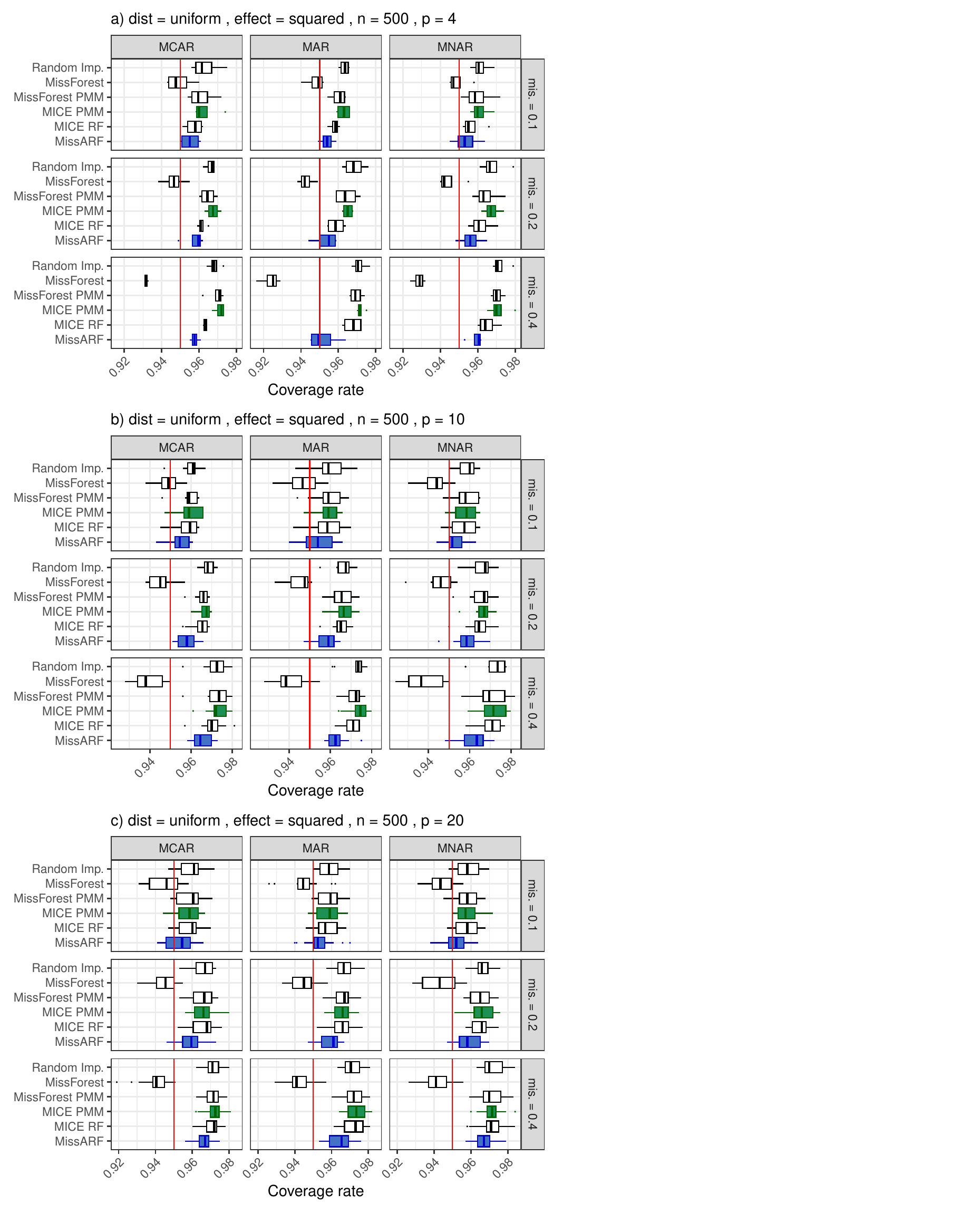}
\caption{ \textbf{Coverage rate} of the uniform distribution setting with a squared effect over different missingness patterns, dimensionality ($p$) and missingness rates (mis.) with $n=500$. The red vertical line shows the nominal coverage level of 0.95. Boxplots are plotted over features, with MissARF (blue) and MICE PMM (green).} \label{fig: logreg_coverage_squared_uniform_500}
\end{figure}

\begin{figure}[p]
\centering
\includegraphics[width=0.92\linewidth]{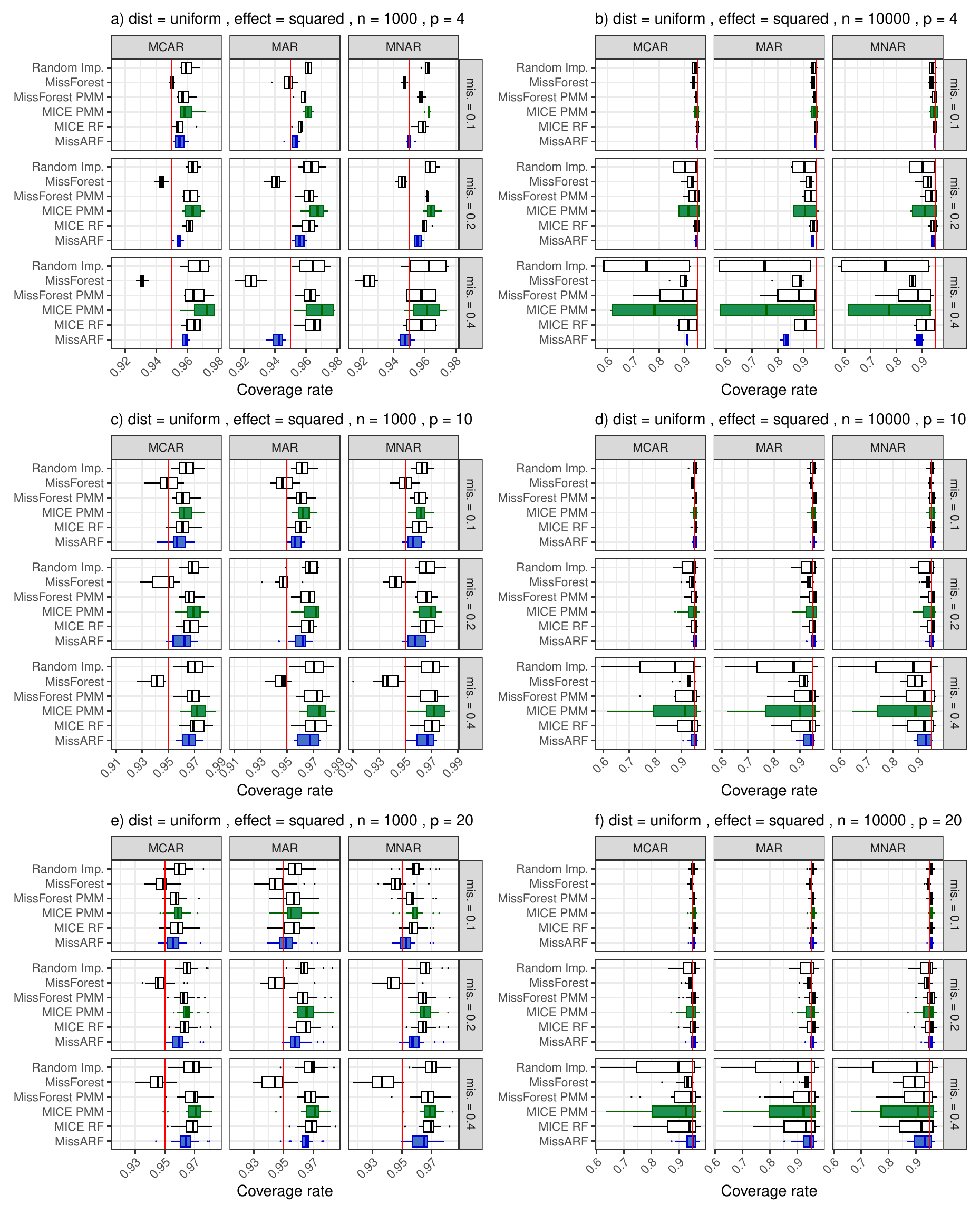}
\caption{ \textbf{Coverage rate} of the uniform distribution setting with a squared effect over different missingness patterns, dimensionality ($p$) and missingness rates (mis.) with $n=1000$ (left) and $n= 10,000$ (right). The red vertical line shows the nominal coverage level of 0.95. Boxplots are plotted over features, with MissARF (blue) and MICE PMM (green).} \label{fig: logreg_coverage_squared_uniform}
\end{figure}

\clearpage
\subsubsection{Category 3: All methods perform poorly}
\label{sec:appx_B13}
\begin{figure}[!ht]
\centering
\includegraphics[width=0.92\linewidth]{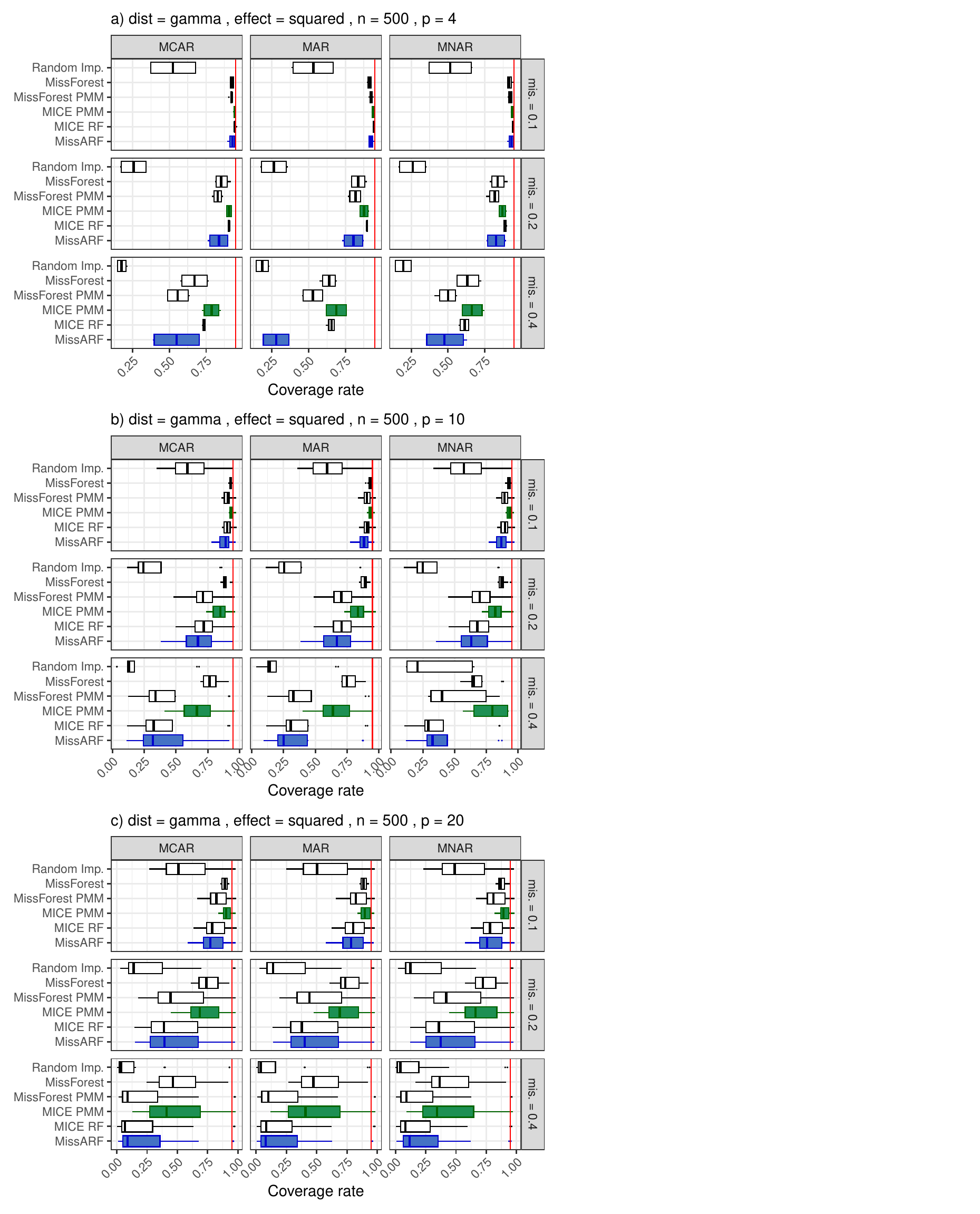}
\caption{ \textbf{Coverage rate} of the gamma distribution setting with a squared effect over different missingness patterns, dimensionality ($p$) and missingness rates (mis.) with $n=500$. The red vertical line shows the nominal coverage level of 0.95. Boxplots are plotted over features, with MissARF (blue) and MICE PMM (green).} \label{fig: logreg_coverage_squared_gamma_500}
\end{figure}

\begin{figure}[p]
\centering
\includegraphics[width=0.92\linewidth]{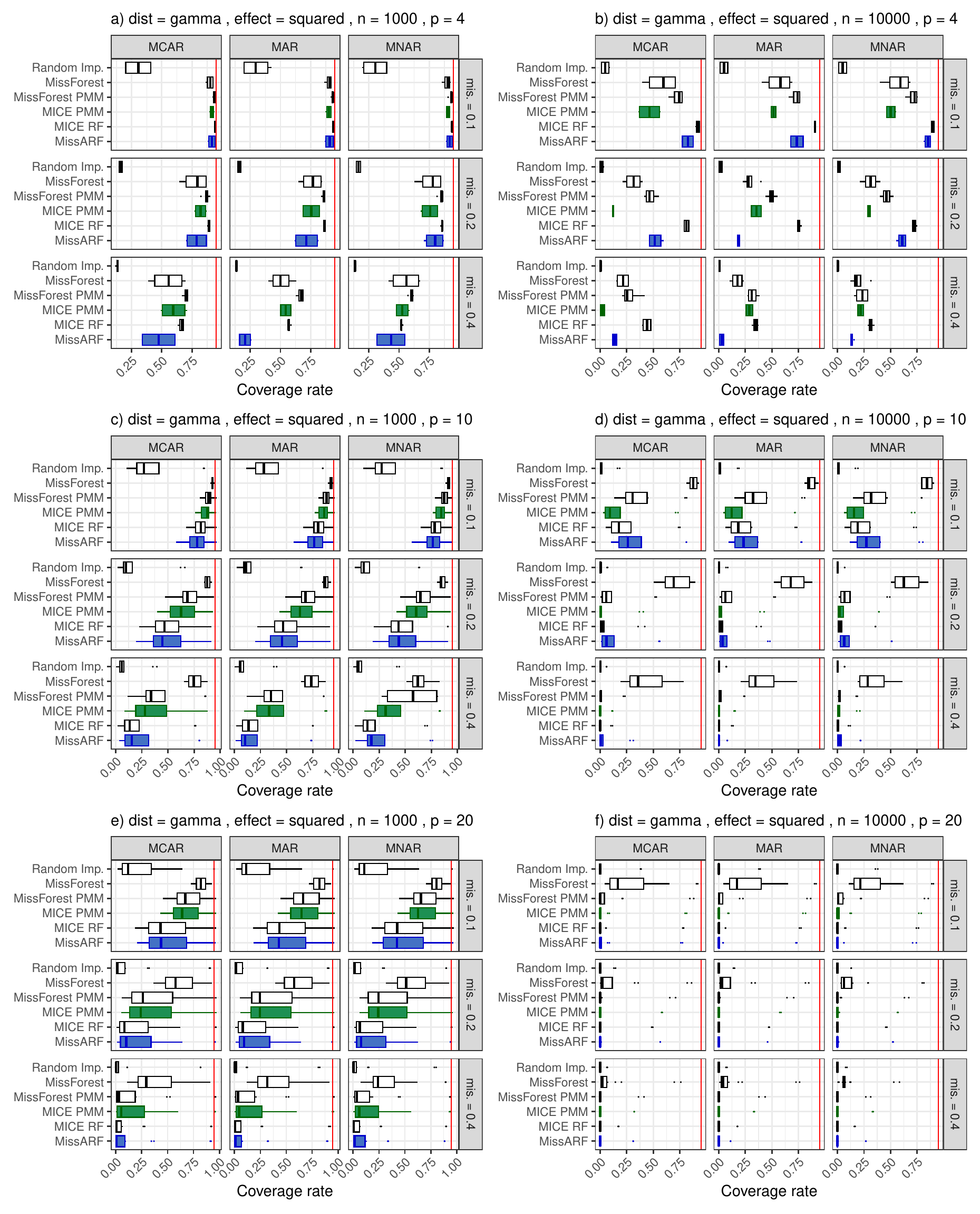}
\caption{ \textbf{Coverage rate} of the gamma distribution setting with a squared effect over different missingness patterns, dimensionality ($p$) and missingness rates (mis.) with $n=1000$ (left) and $n= 10,000$ (right). The red vertical line shows the nominal coverage level of 0.95. Boxplots are plotted over features, with MissARF (blue) and MICE PMM (green).} \label{fig: logreg_coverage_squared_gamma}
\end{figure}

\begin{figure}[p]
\centering
\includegraphics[width=0.92\linewidth]{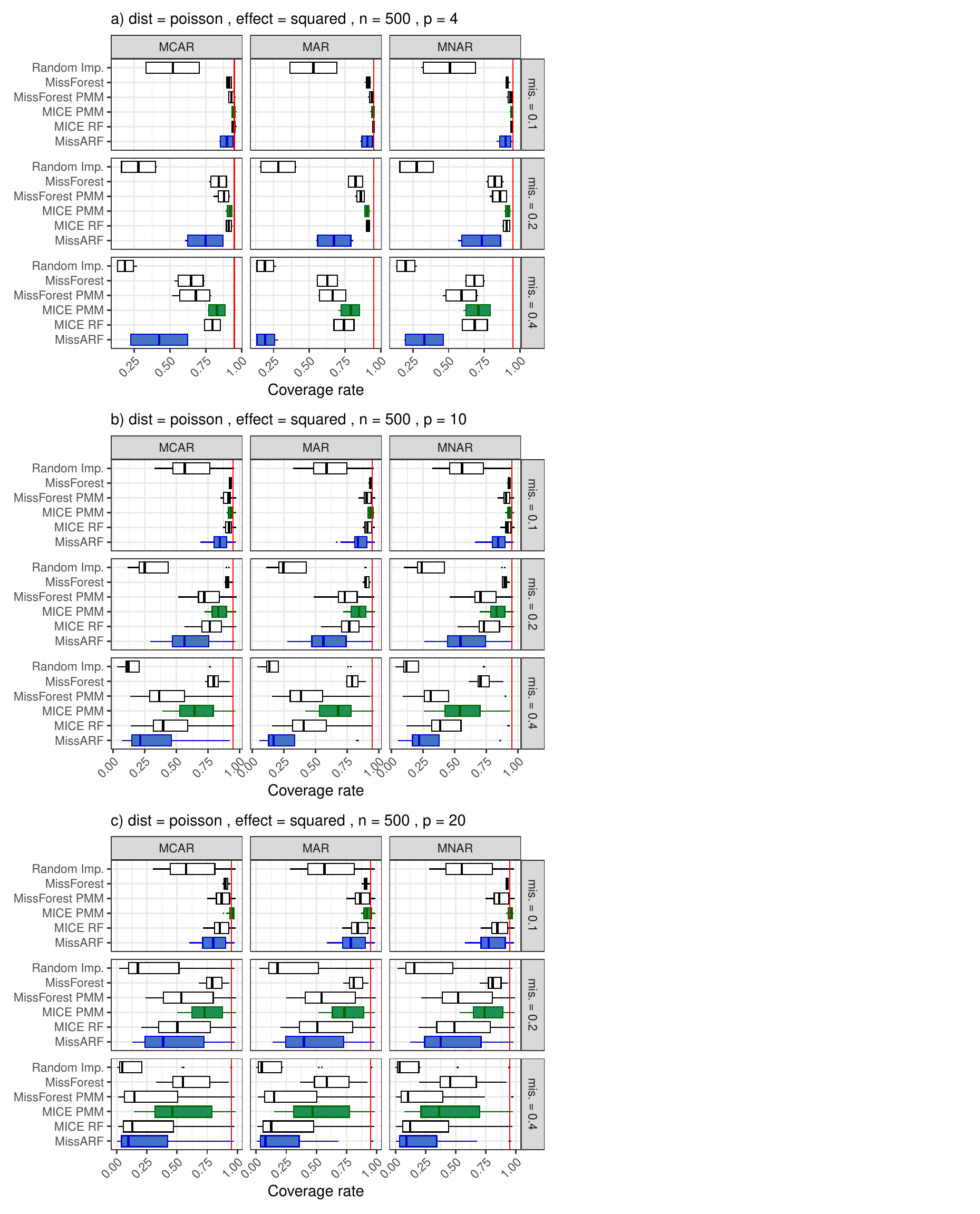}
\caption{ \textbf{Coverage rate} of the Poisson distribution setting with a squared effect over different missingness patterns, dimensionality ($p$) and missingness rates (mis.) with $n=500$. The red vertical line shows the nominal coverage level of 0.95. Boxplots are plotted over features, with MissARF (blue) and MICE PMM (green).} \label{fig: logreg_coverage_squared_poisson_500}
\end{figure}

\begin{figure}[p]
\centering
\includegraphics[width=0.92\linewidth]{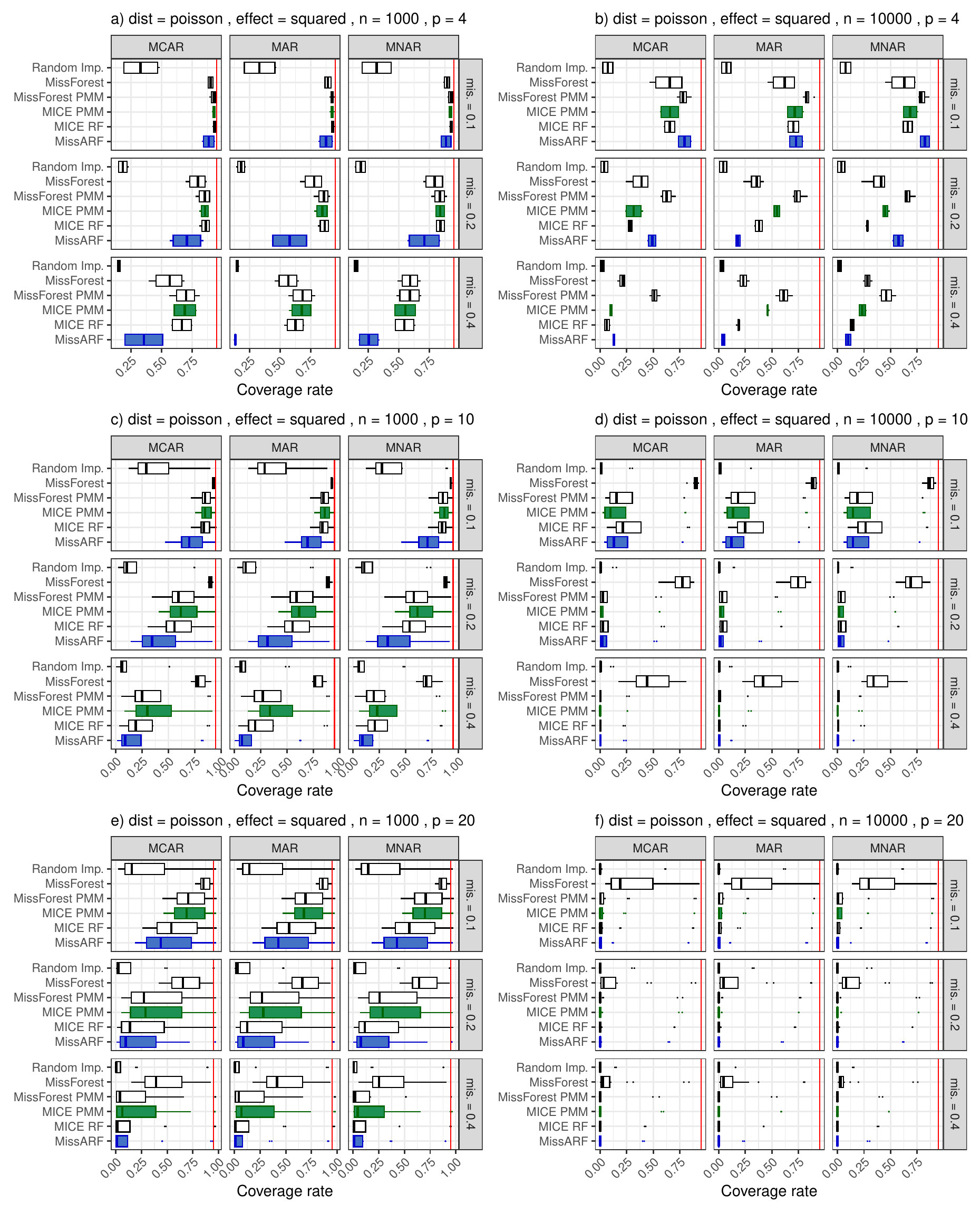}
\caption{ \textbf{Coverage rate} of the Poisson distribution setting with a squared effect over different missingness patterns, dimensionality ($p$) and missingness rates (mis.) with $n=1000$ (left) and $n= 10,000$ (right). The red vertical line shows the nominal coverage level of 0.95. Boxplots are plotted over features, with MissARF (blue) and MICE PMM (green).} \label{fig: logreg_coverage_squared_poisson}
\end{figure}

\clearpage
\subsection{Average CI Width}
\subsubsection{Category 1: Similar performance across all methods, MissARF with smallest average width}

\begin{figure}[!ht]
\centering
\includegraphics[width=0.92\linewidth]{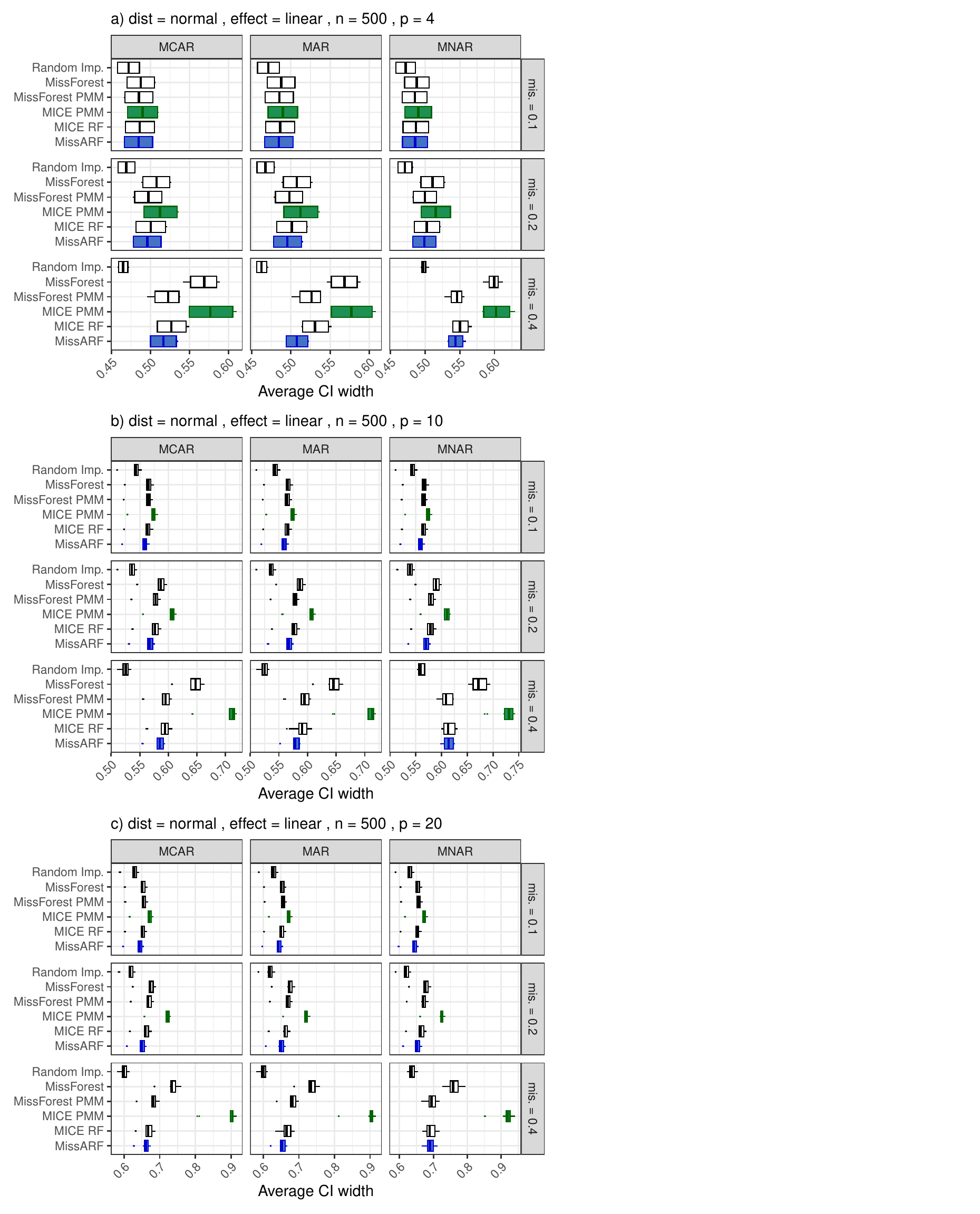}
\caption{\textbf{Average width of the confidence intervals} of the normal distribution setting with a linear effect over different missingness patterns, dimensionality ($p$) and missingness rates (mis.) with $n=500$. The boxplots are plotted over the features, with MissARF (blue) and MICE PMM (green) highlighted.} \label{fig: logreg_aw_linear_normal_500}
\end{figure}

\begin{figure}[p]
\centering
\includegraphics[width=0.92\linewidth]{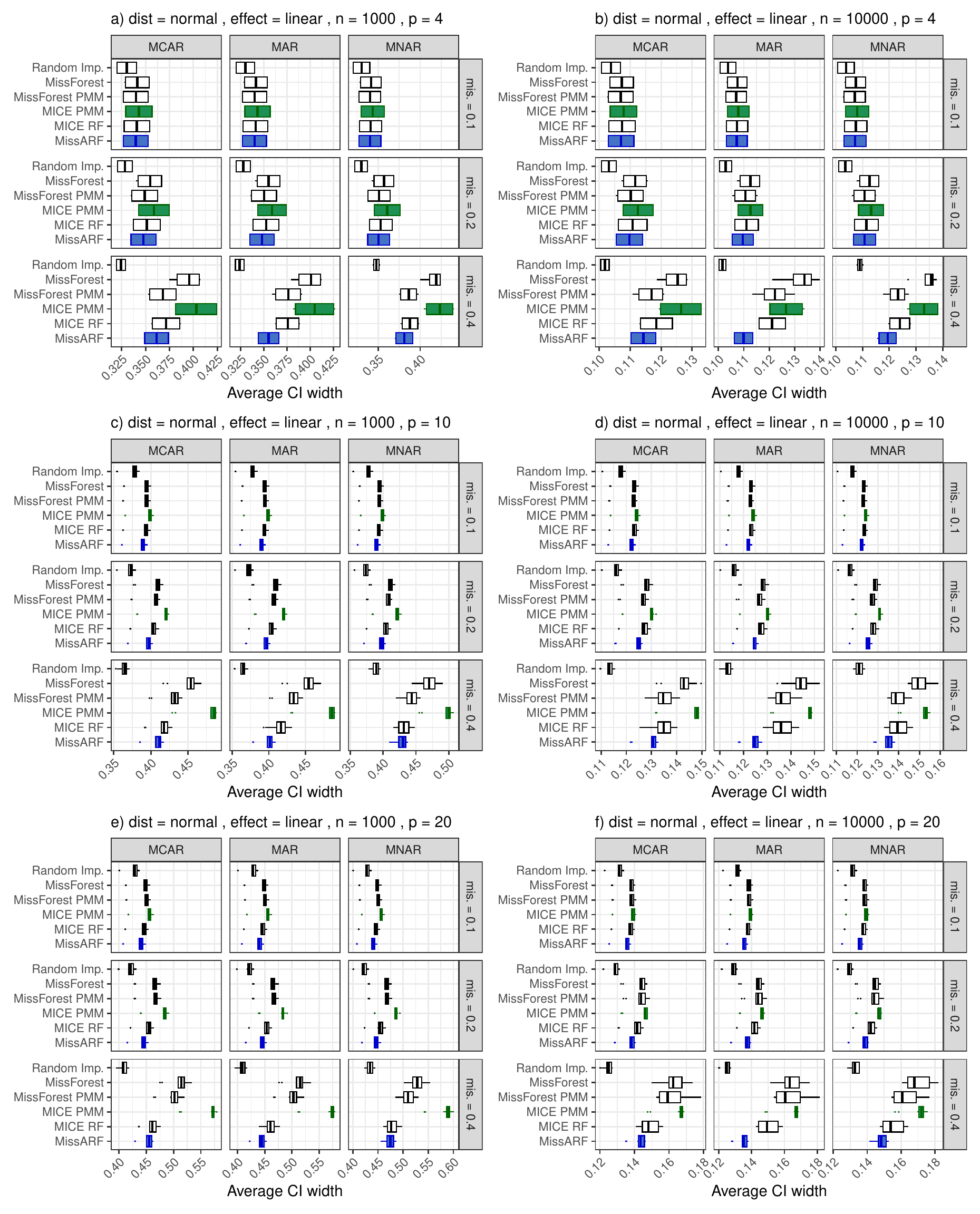}
\caption{\textbf{Average width of the confidence intervals} of the normal distribution setting with a linear effect over different missingness patterns, dimensionality ($p$) and missingness rates (mis.) with $n=1000$ (left) and $n= 10,000$ (right). The boxplots are plotted over the features, with MissARF (blue) and MICE PMM (green) highlighted.} \label{fig: logreg_aw_linear_normal}
\end{figure}

\begin{figure}[p]
\centering
\includegraphics[width=0.92\linewidth]{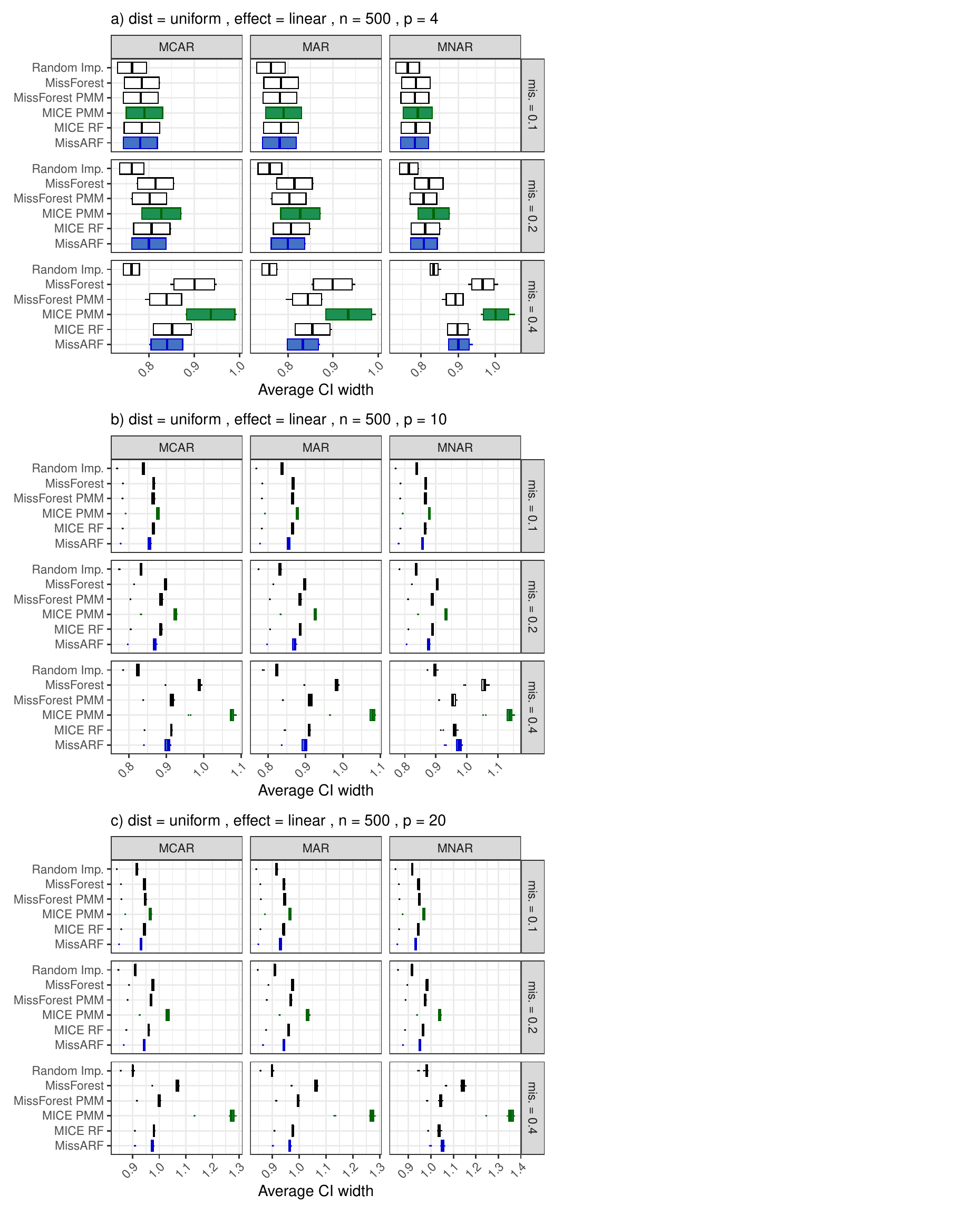}
\caption{\textbf{Average width of the confidence intervals} of the uniform distribution setting with a linear effect over different missingness patterns, dimensionality ($p$) and missingness rates (mis.) with $n=500$. The boxplots are plotted over the features, with MissARF (blue) and MICE PMM (green) highlighted.} \label{fig: logreg_aw_linear_uniform_500}
\end{figure}

\begin{figure}[p]
\centering
\includegraphics[width=0.92\linewidth]{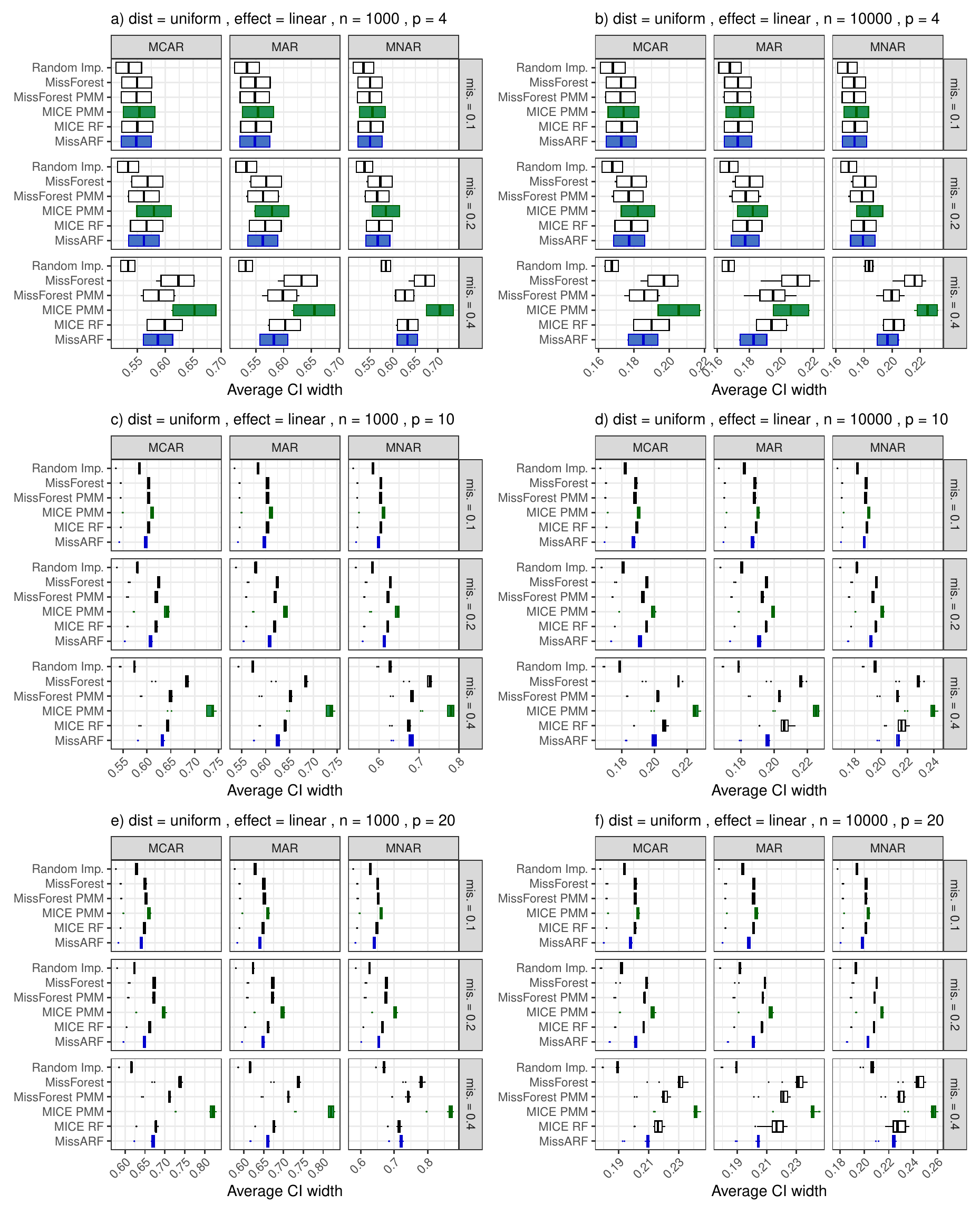}
\caption{\textbf{Average width of the confidence intervals} of the uniform distribution setting with a linear effect over different missingness patterns, dimensionality ($p$) and missingness rates (mis.) with $n=1000$ (left) and $n= 10,000$ (right). The boxplots are plotted over the features, with MissARF (blue) and MICE PMM (green) highlighted.} \label{fig: logreg_aw_linear_uniform}
\end{figure}

\begin{figure}[p]
\centering
\includegraphics[width=0.92\linewidth]{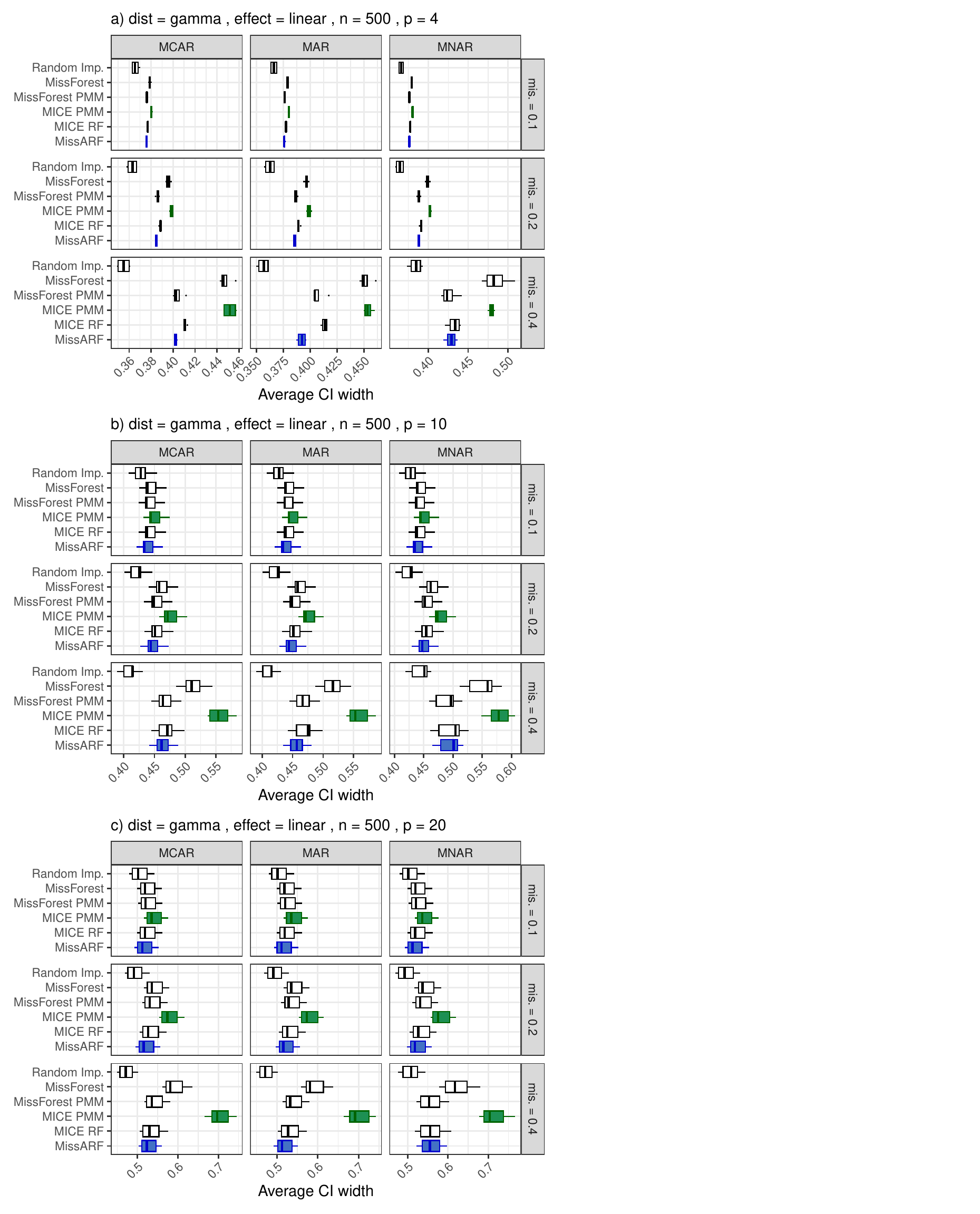}
\caption{ \textbf{Average width of the confidence intervals} of the gamma distribution setting with a linear effect over different missingness patterns, dimensionality ($p$) and missingness rates (mis.) with $n=500$. The boxplots are plotted over the features, with MissARF (blue) and MICE PMM (green) highlighted.} \label{fig: logreg_aw_linear_gamma_500}
\end{figure}

\begin{figure}[p]
\centering
\includegraphics[width=0.92\linewidth]{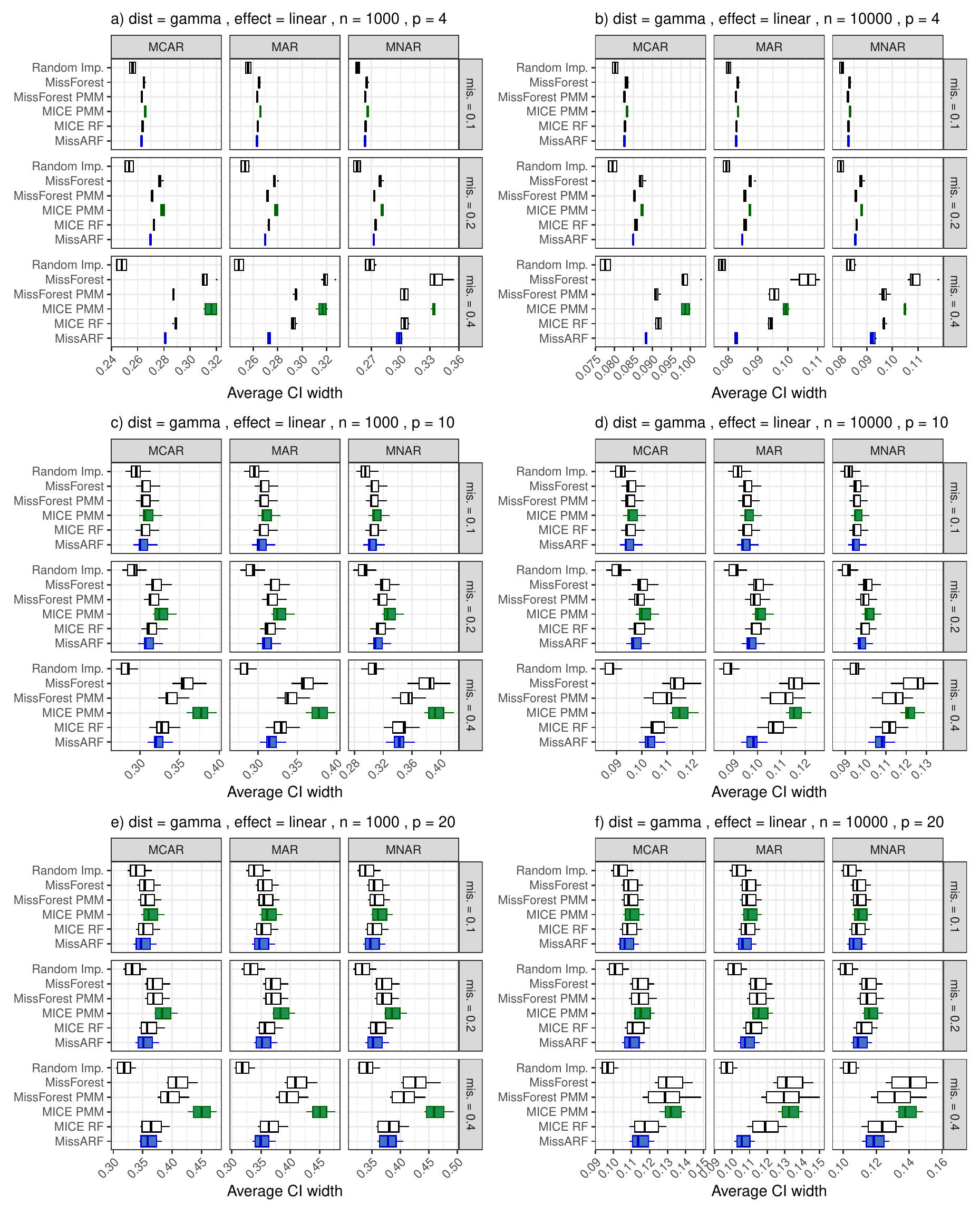}
\caption{ \textbf{Average width of the confidence intervals} of the gamma distribution setting with a linear effect over different missingness patterns, dimensionality ($p$) and missingness rates (mis.) with $n=1000$ (left) and $n= 10,000$ (right). The boxplots are plotted over the features, with MissARF (blue) and MICE PMM (green) highlighted.} \label{fig: logreg_aw_linear_gamma}
\end{figure}

\begin{figure}[p]
\centering
\includegraphics[width=0.92\linewidth]{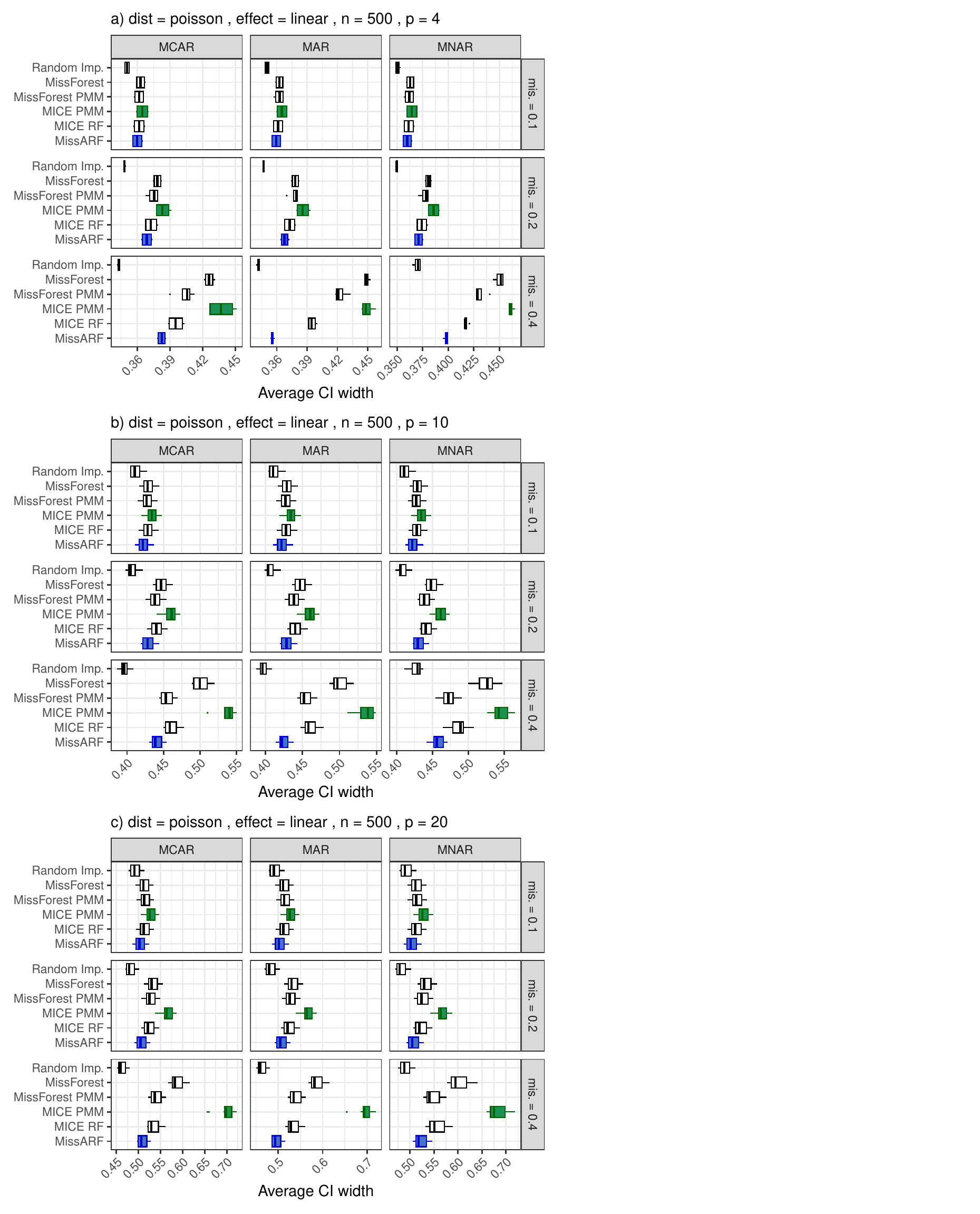}
\caption{ \textbf{Average width of the confidence intervals} of the Poisson distribution setting with a linear effect over different missingness patterns, dimensionality ($p$) and missingness rates (mis.) with $n=500$. The boxplots are plotted over the features, with MissARF (blue) and MICE PMM (green) highlighted.} \label{fig: logreg_aw_linear_poisson_500}
\end{figure}

\begin{figure}[p]
\centering
\includegraphics[width=0.92\linewidth]{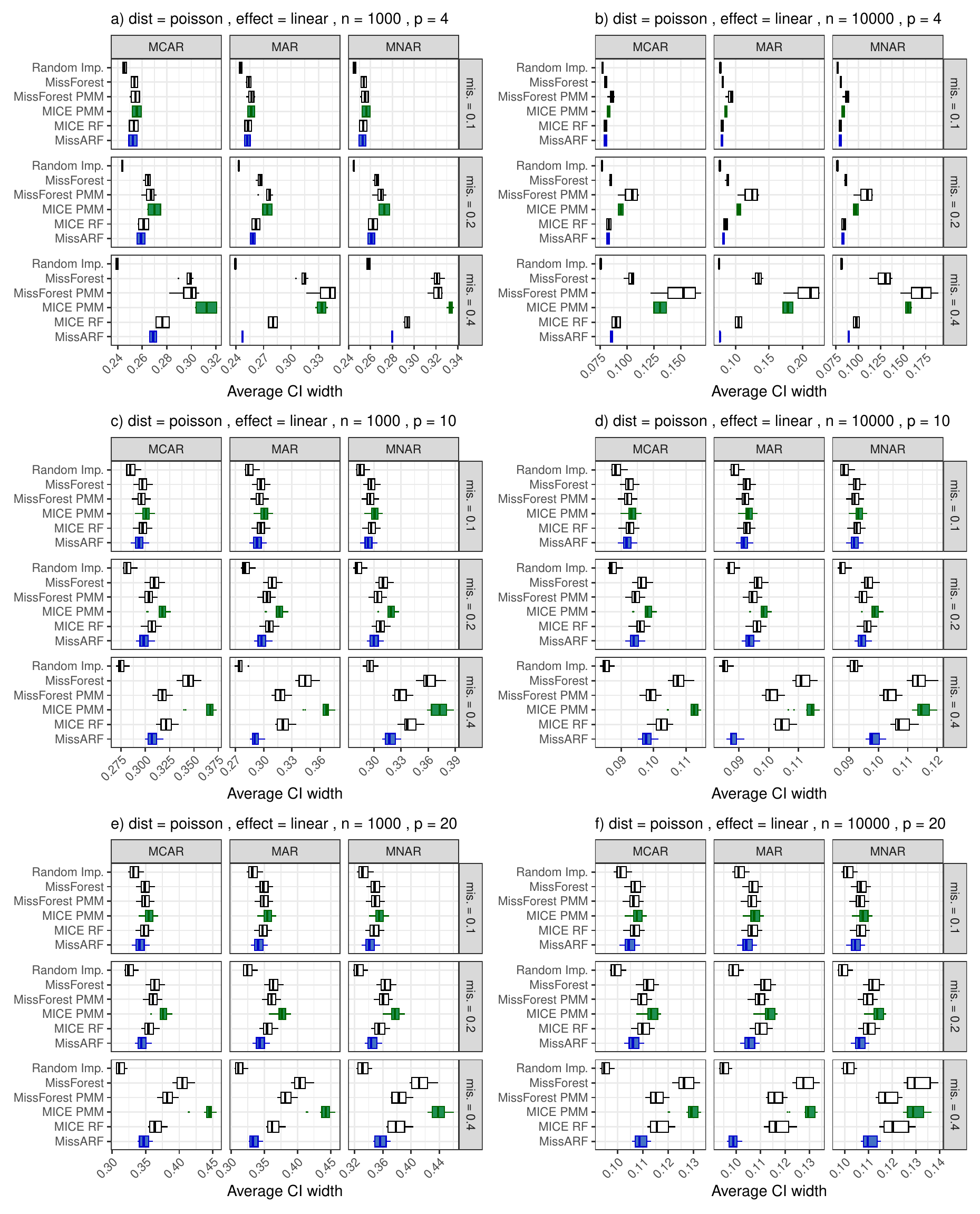}
\caption{ \textbf{Average width of the confidence intervals} of the Poisson distribution setting with a linear effect over different missingness patterns, dimensionality ($p$) and missingness rates (mis.) with $n=1000$ (left) and $n= 10,000$ (right). The boxplots are plotted over the features, with MissARF (blue) and MICE PMM (green) highlighted.} \label{fig: logreg_aw_linear_poisson}
\end{figure}

\begin{figure}[p]
\centering
\includegraphics[width=0.92\linewidth]{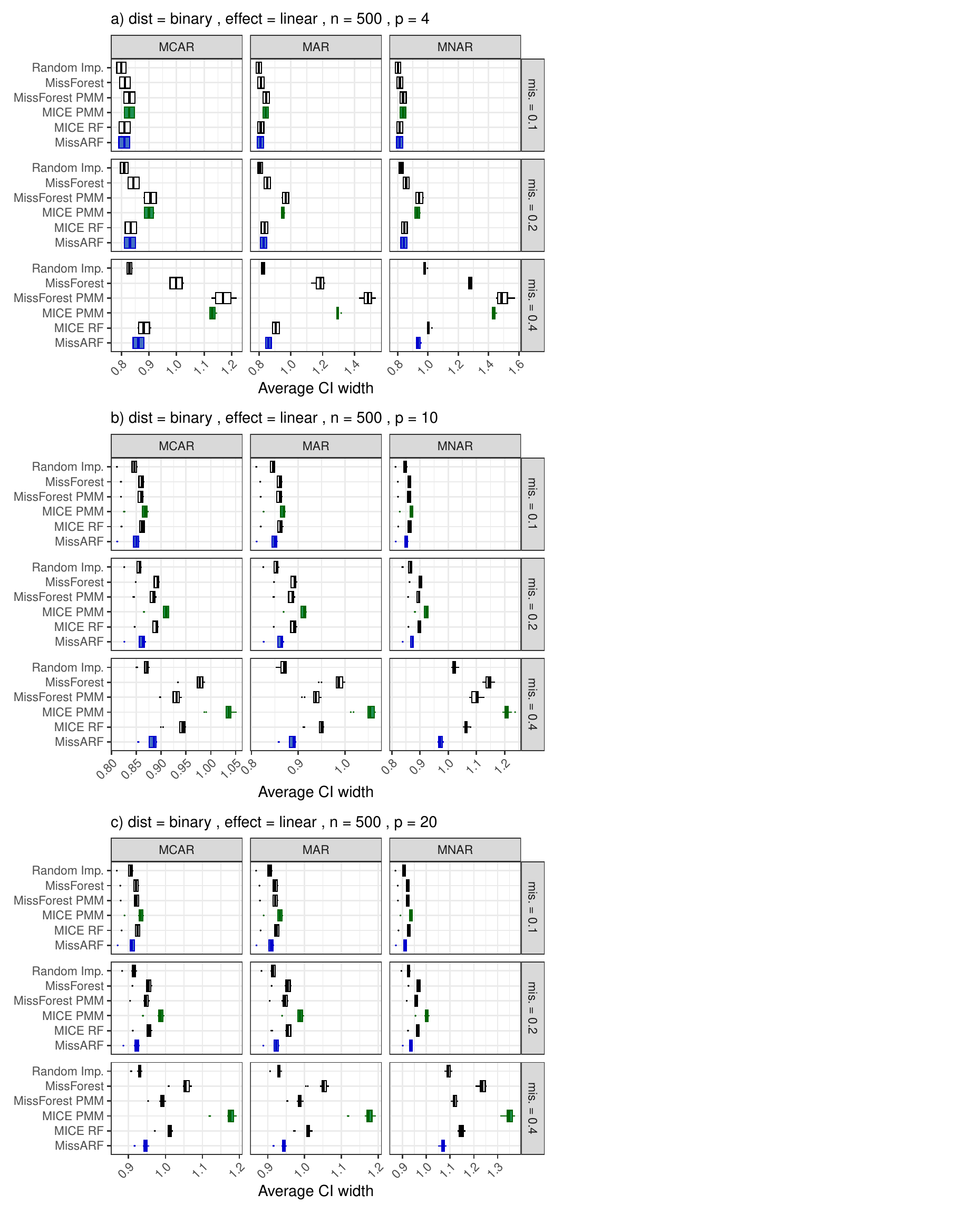}
\caption{ \textbf{Average width of the confidence intervals} of the binary distribution setting with a linear effect over different missingness patterns, dimensionality ($p$) and missingness rates (mis.) with $n=500$. The boxplots are plotted over the features, with MissARF (blue) and MICE PMM (green) highlighted.} \label{fig: logreg_aw_linear_binary_500}
\end{figure}

\begin{figure}[p]
\centering
\includegraphics[width=0.92\linewidth]{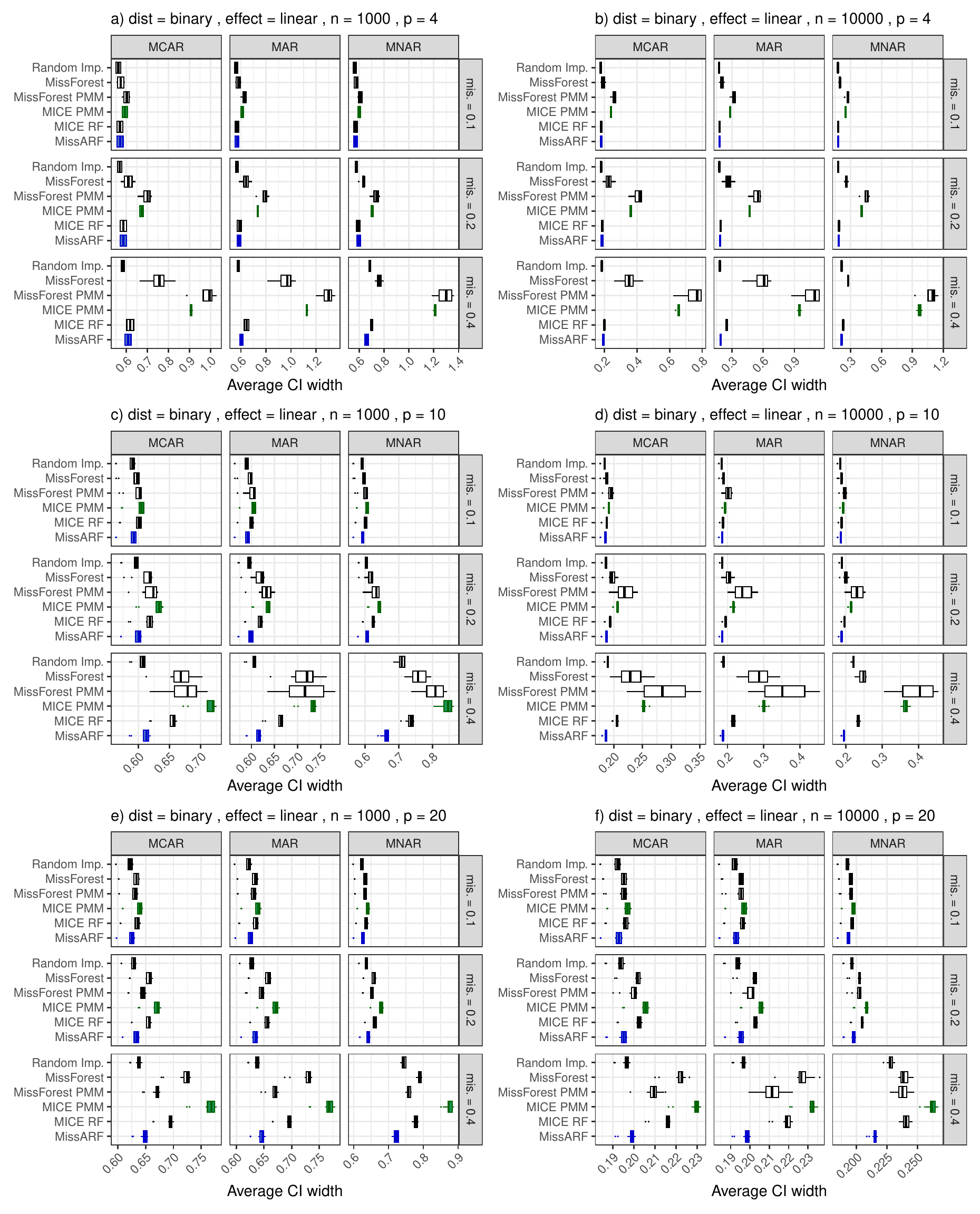}
\caption{ \textbf{Average width of the confidence intervals} of the binary distribution setting with a linear effect over different missingness patterns, dimensionality ($p$) and missingness rates (mis.) with $n=1000$ (left) and $n= 10,000$ (right). The boxplots are plotted over the features, with MissARF (blue) and MICE PMM (green) highlighted.} \label{fig: logreg_aw_linear_binary}
\end{figure}

\begin{figure}[p]
\centering
\includegraphics[width=0.92\linewidth]{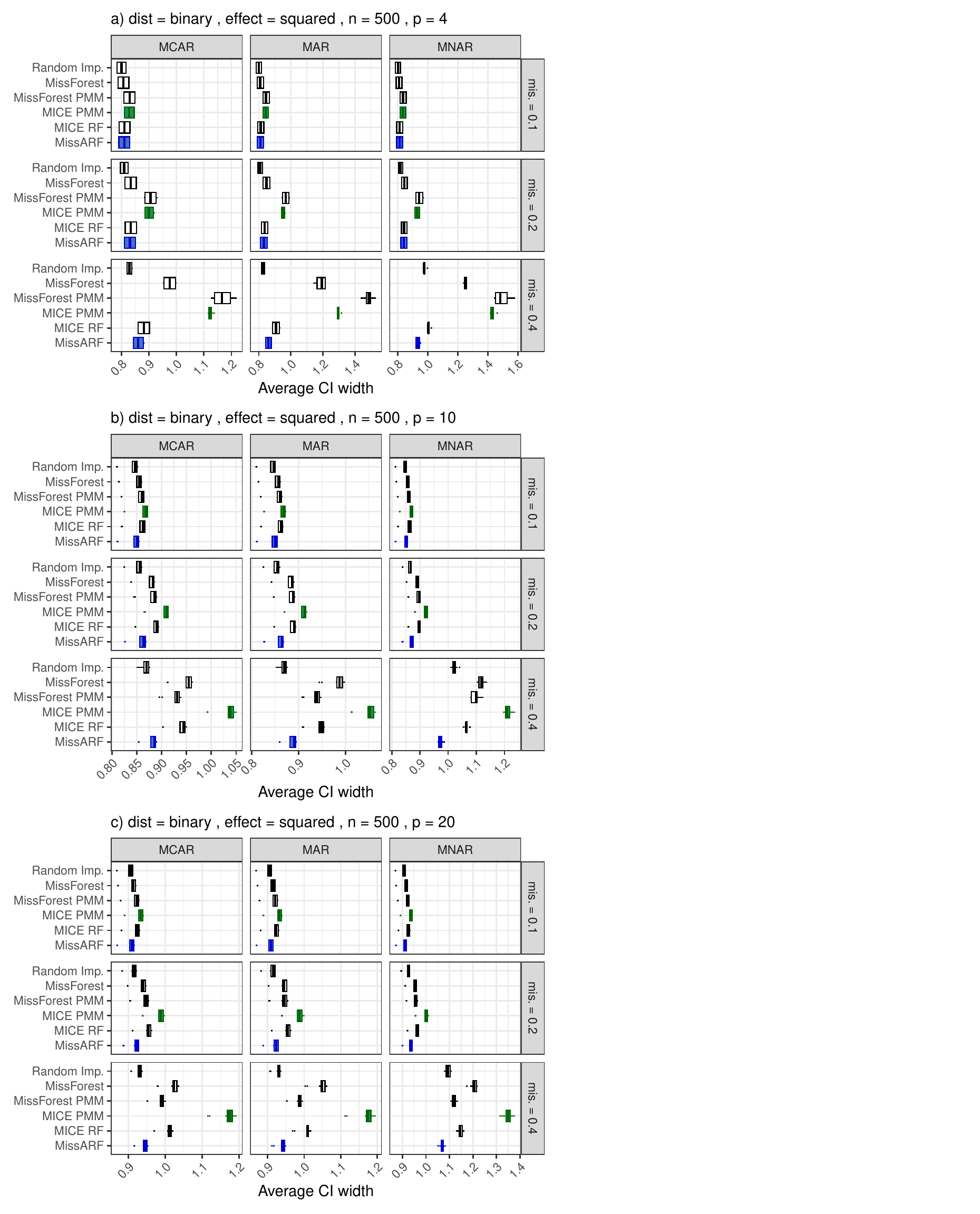}
\caption{ \textbf{Average width of the confidence intervals} of the binary distribution setting with a squared effect over different missingness patterns, dimensionality ($p$) and missingness rates (mis.) with $n=500$. The boxplots are plotted over the features, with MissARF (blue) and MICE PMM (green) highlighted.} \label{fig: logreg_aw_squared_binary_500}
\end{figure}

\begin{figure}[p]
\centering
\includegraphics[width=0.92\linewidth]{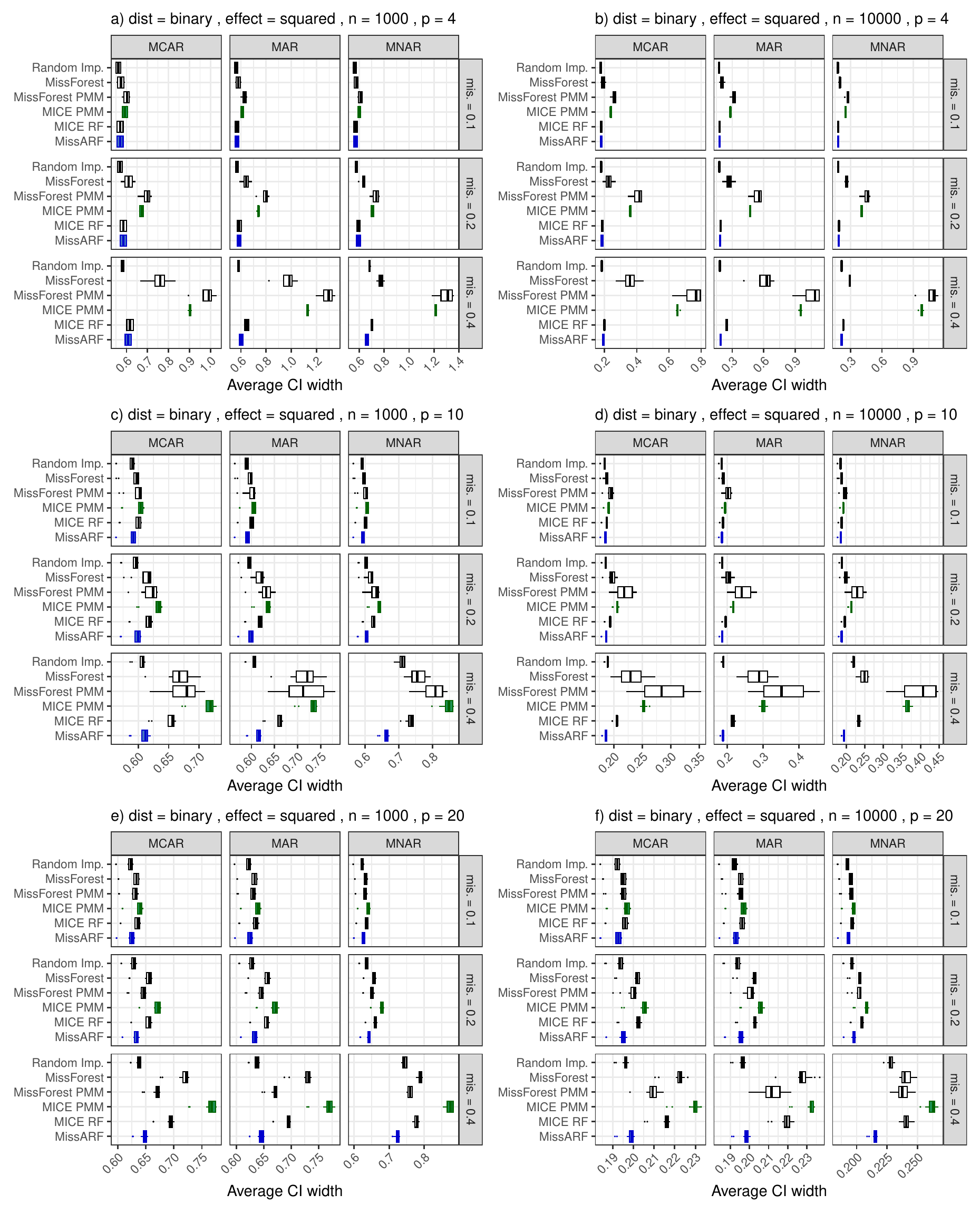}
\caption{ \textbf{Average width of the confidence intervals} of the binary distribution setting with a squared effect over different missingness patterns, dimensionality ($p$) and missingness rates (mis.) with $n=1000$ (left) and $n= 10,000$ (right). The boxplots are plotted over the features, with MissARF (blue) and MICE PMM (green) highlighted.} \label{fig: logreg_aw_squared_binary}
\end{figure}

\clearpage
\subsubsection{Category 2: PMM methods struggle, MissARF performs well}

\begin{figure}[!ht]
\centering
\includegraphics[width=0.92\linewidth]{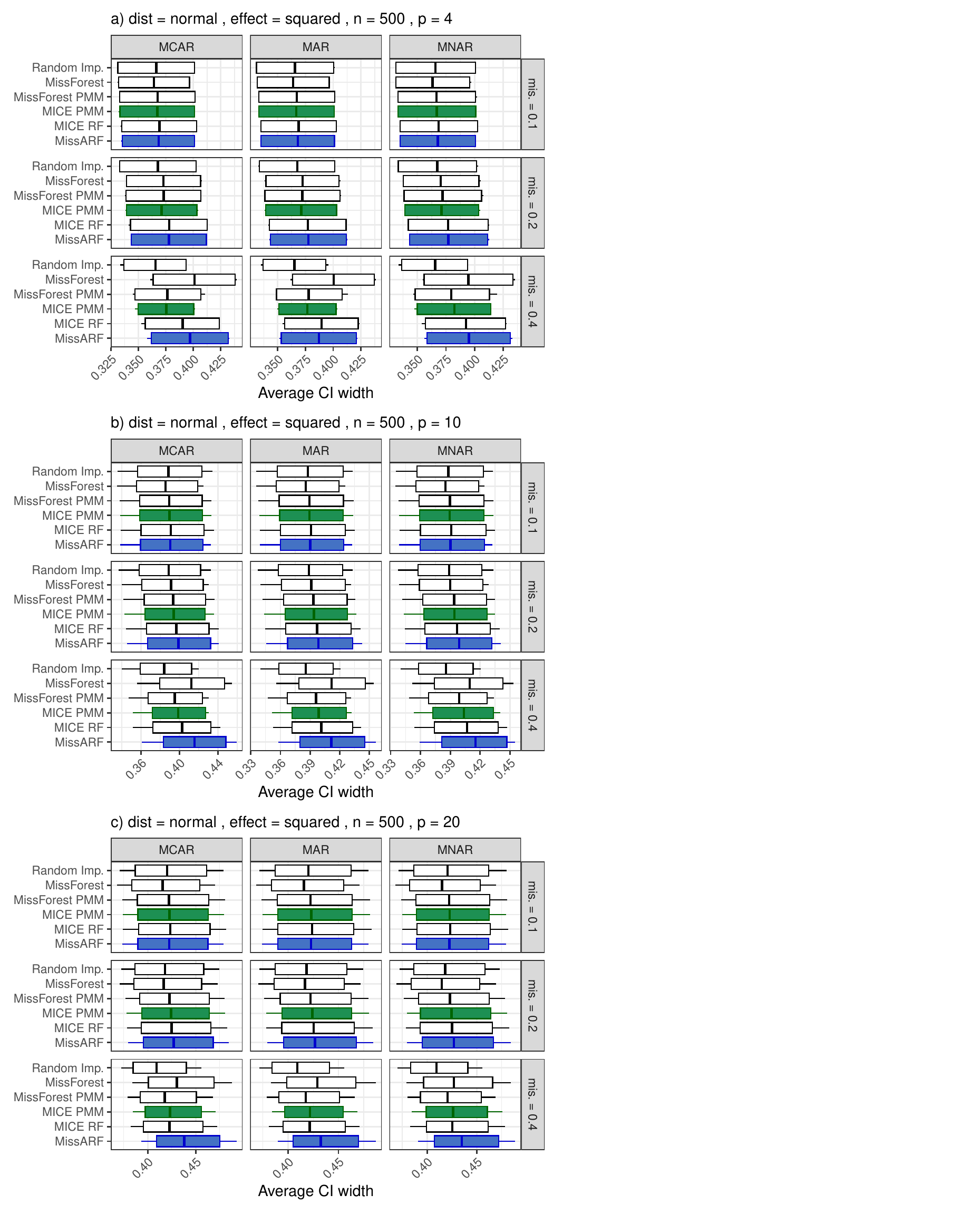}
\caption{ \textbf{Average width of the confidence intervals} of the normal distribution setting with a squared effect over different missingness patterns, dimensionality ($p$) and missingness rates (mis.) with $n=500$. The boxplots are plotted over the features, with MissARF (blue) and MICE PMM (green) highlighted.} \label{fig: logreg_aw_squared_normal_500}
\end{figure}

\begin{figure}[p]
\centering
\includegraphics[width=0.92\linewidth]{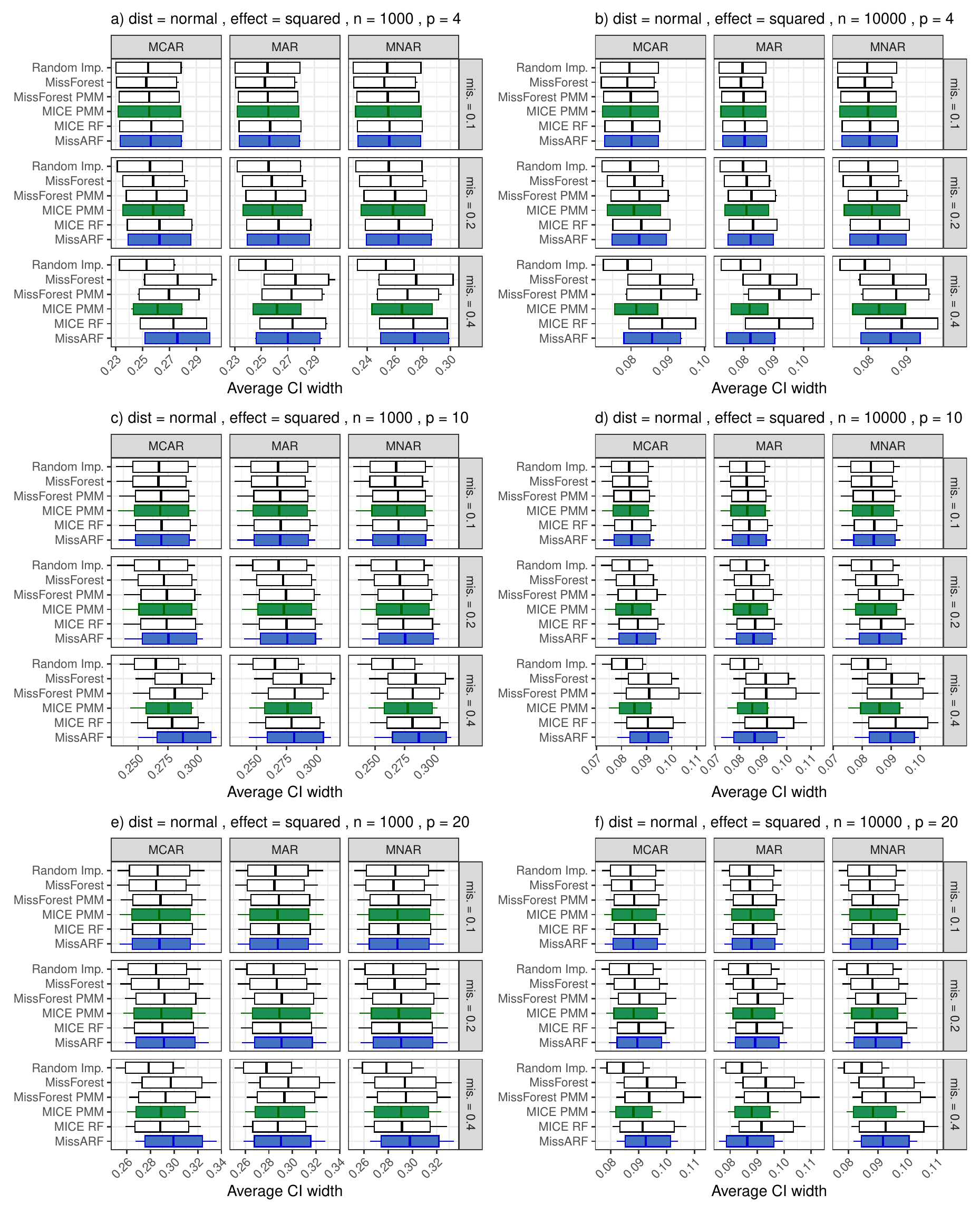}
\caption{ \textbf{Average width of the confidence intervals} of the normal distribution setting with a squared effect over different missingness patterns, dimensionality ($p$) and missingness rates (mis.) with $n=1000$ (left) and $n= 10,000$ (right). The boxplots are plotted over the features, with MissARF (blue) and MICE PMM (green) highlighted.} \label{fig: logreg_aw_squared_normal}
\end{figure}

\begin{figure}[p]
\centering
\includegraphics[width=0.92\linewidth]{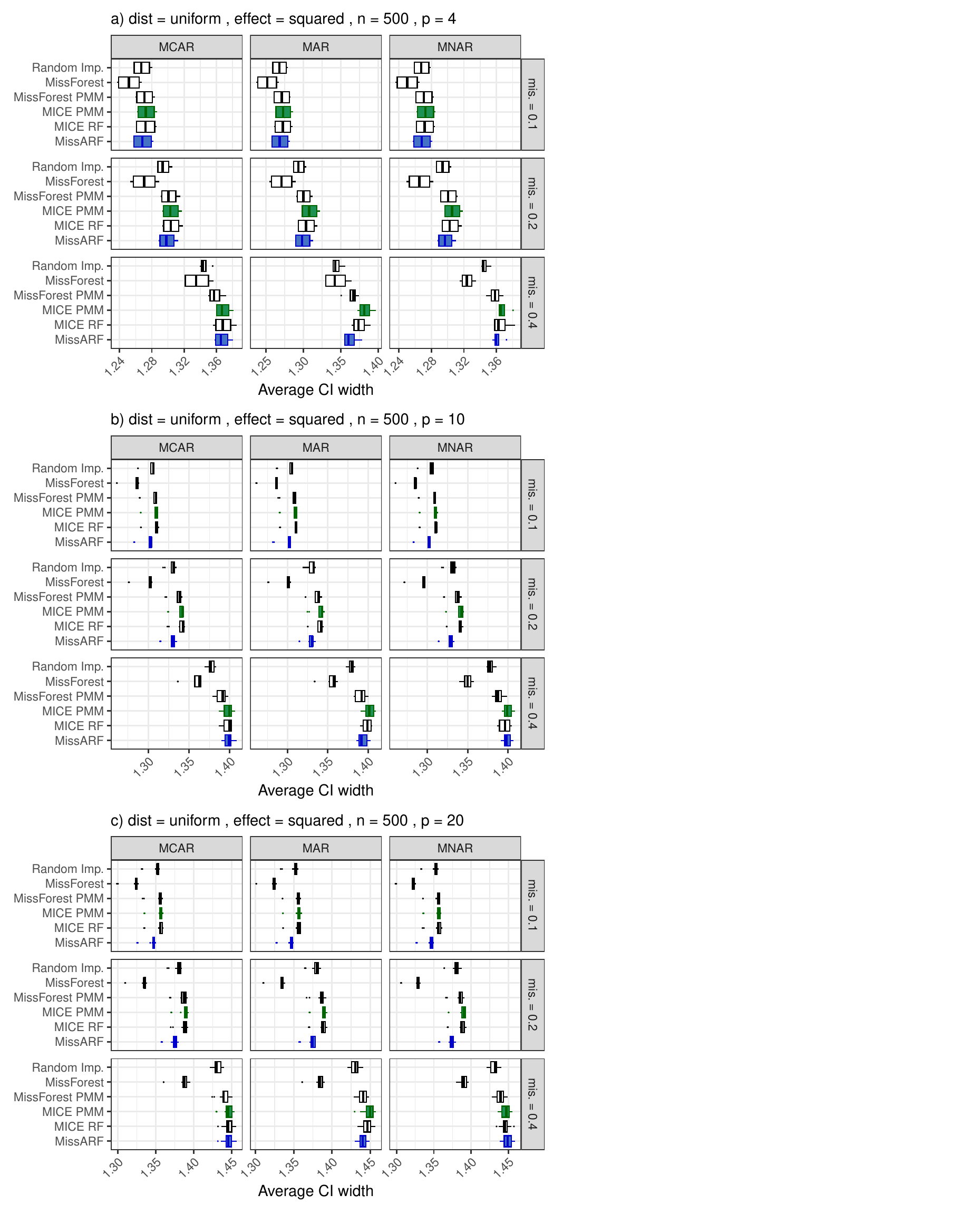}
\caption{ \textbf{Average width of the confidence intervals} of the uniform distribution setting with a squared effect over different missingness patterns, dimensionality ($p$) and missingness rates (mis.) with $n=500$. The boxplots are plotted over the features, with MissARF (blue) and MICE PMM (green) highlighted.} \label{fig: logreg_aw_squared_uniform_500}
\end{figure}

\begin{figure}[p]
\centering
\includegraphics[width=0.92\linewidth]{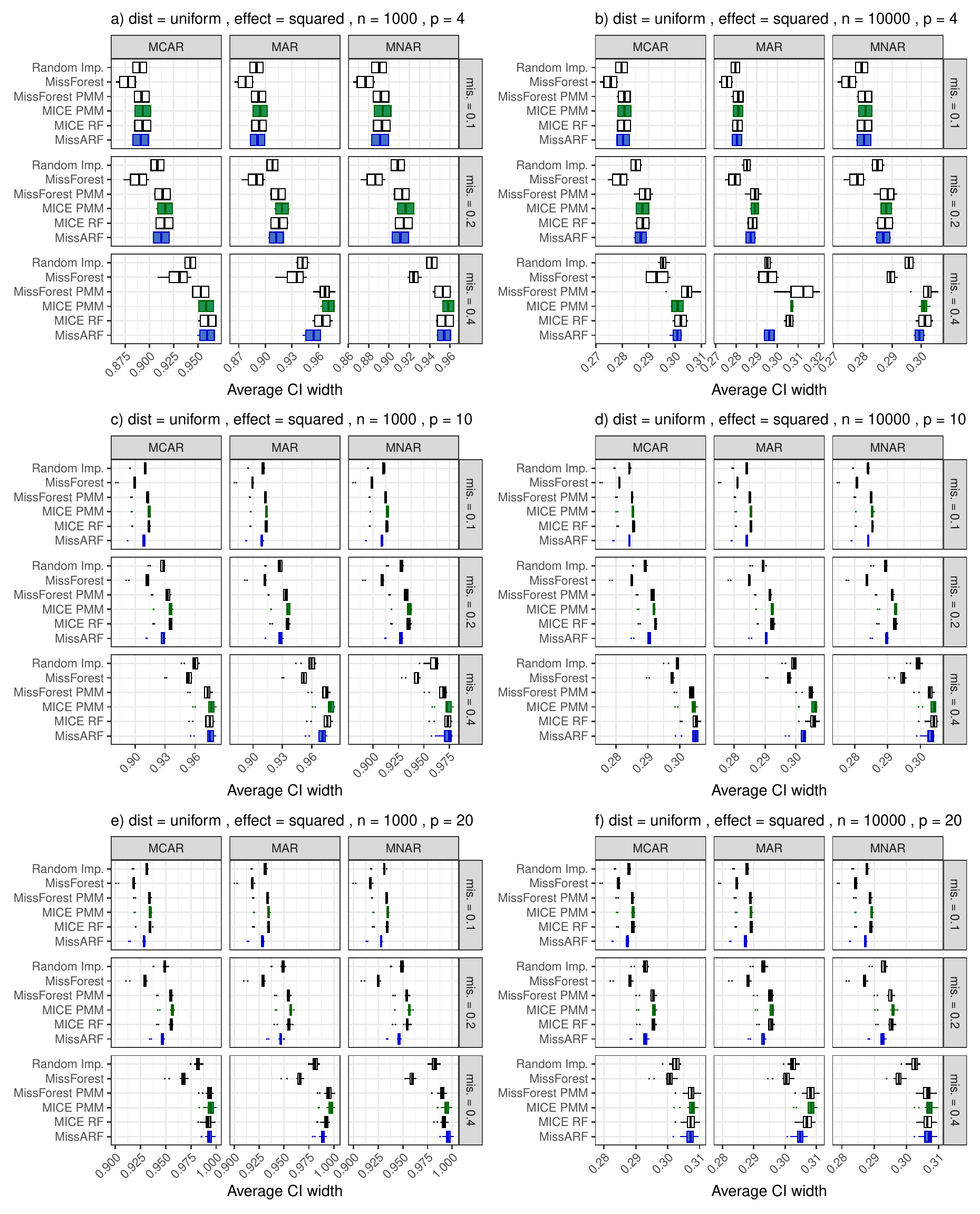}
\caption{ \textbf{Average width of the confidence intervals} of the uniform distribution setting with a squared effect over different missingness patterns, dimensionality ($p$) and missingness rates (mis.) with $n=1000$ (left) and $n= 10,000$ (right). The boxplots are plotted over the features, with MissARF (blue) and MICE PMM (green) highlighted.} \label{fig: logreg_aw_squared_uniform}
\end{figure}

\clearpage

\subsubsection{Category 3: All methods perform poorly}
\begin{figure}[!ht]
\centering
\includegraphics[width=0.92\linewidth]{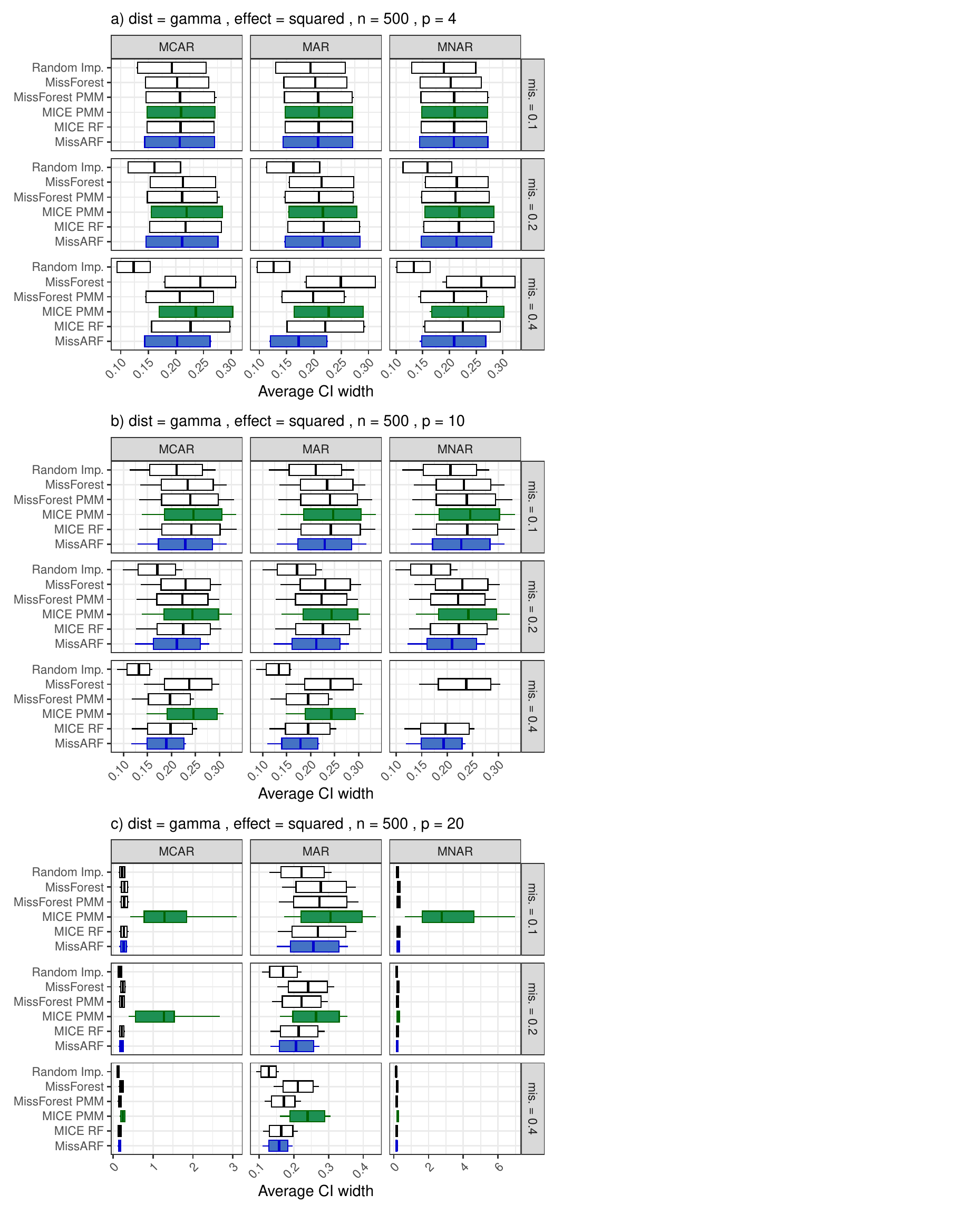}
\caption{ \textbf{Average width of the confidence intervals} of the gamma distribution setting with a squared effect over different missingness patterns, dimensionality ($p$) and missingness rates (mis.) with $n=500$. The boxplots are plotted over the features, with MissARF (blue) and MICE PMM (green) highlighted.} \label{fig: logreg_aw_squared_gamma_500}
\end{figure}

\begin{figure}[p]
\centering
\includegraphics[width=0.92\linewidth]{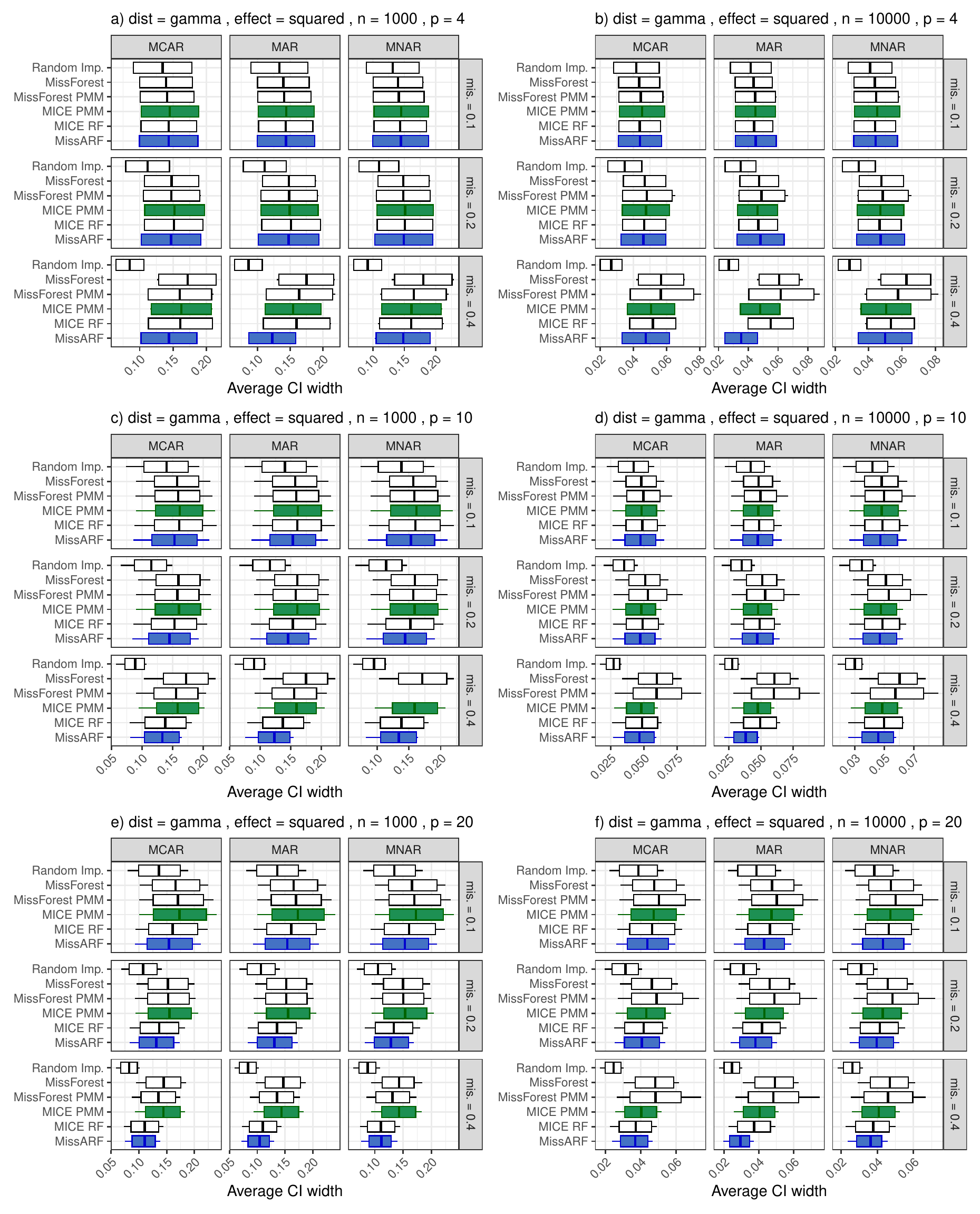}
\caption{ \textbf{Average width of the confidence intervals} of the gamma distribution setting with a squared effect over different missingness patterns, dimensionality ($p$) and missingness rates (mis.) with $n=1000$ (left) and $n= 10,000$ (right). The boxplots are plotted over the features, with MissARF (blue) and MICE PMM (green) highlighted.} \label{fig: logreg_aw_squared_gamma}
\end{figure}

\begin{figure}[p]
\centering
\includegraphics[width=0.92\linewidth]{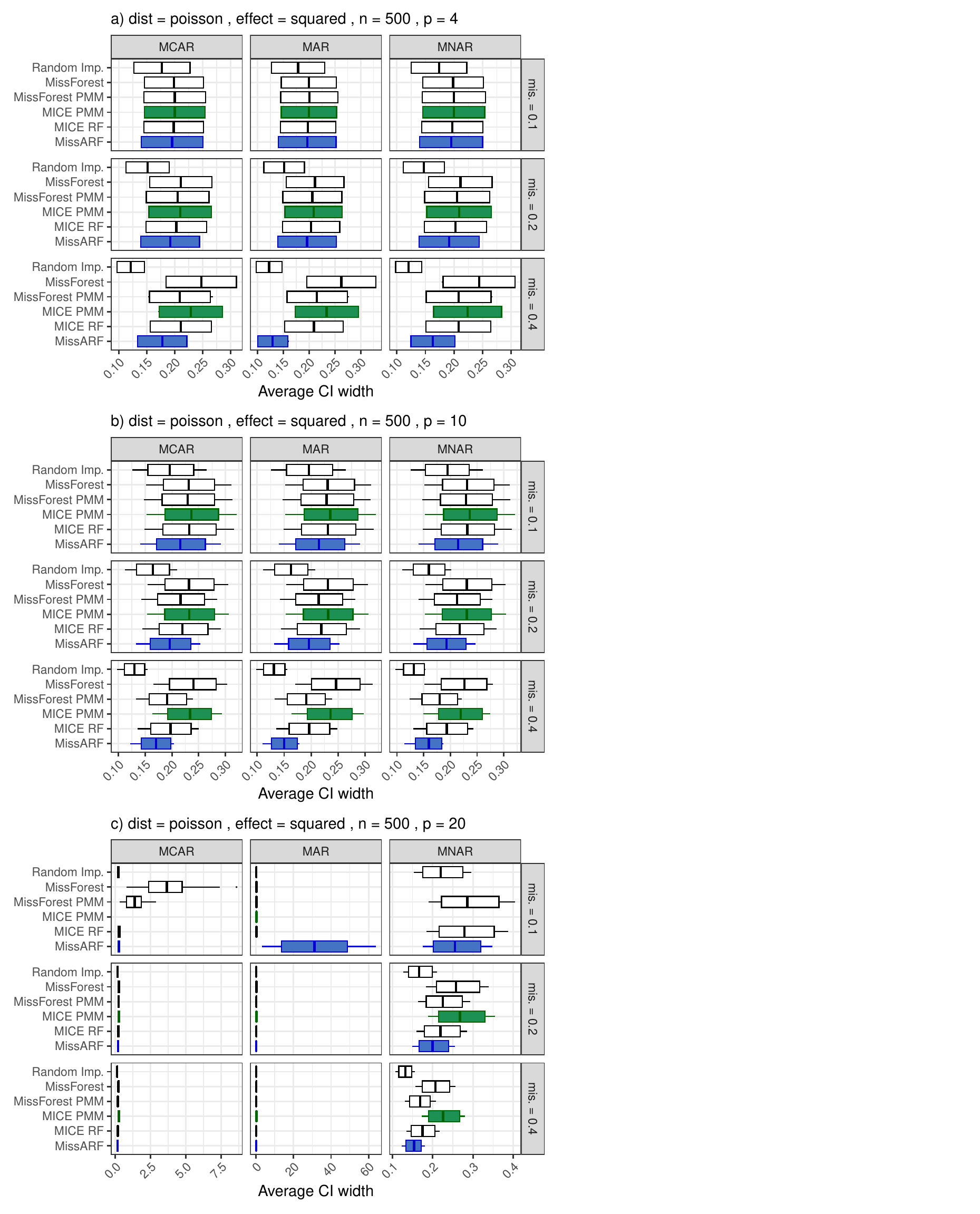}
\caption{ \textbf{Average width of the confidence intervals} of the Poisson distribution setting with a squared effect over different missingness patterns, dimensionality ($p$) and missingness rates (mis.) with $n=500$. The boxplots are plotted over the features, with MissARF (blue) and MICE PMM (green) highlighted.} \label{fig: logreg_aw_squared_poisson_500}
\end{figure}

\begin{figure}[p]
\centering
\includegraphics[width=0.92\linewidth]{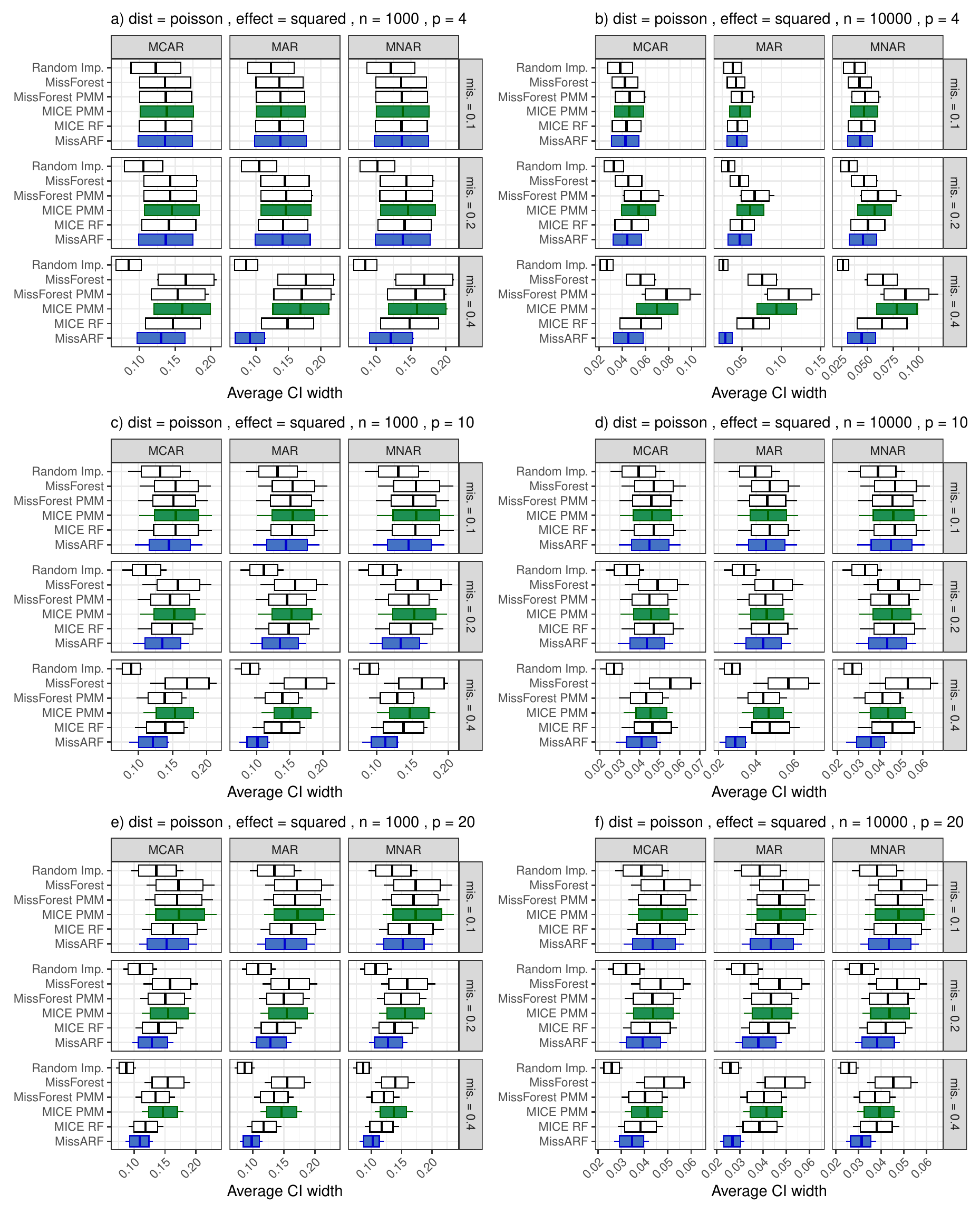}
\caption{ \textbf{Average width of the confidence intervals} of the Poisson distribution setting with a squared effect over different missingness patterns, dimensionality ($p$) and missingness rates (mis.) with $n=1000$ (left) and $n= 10,000$ (right). The boxplots are plotted over the features, with MissARF (blue) and MICE PMM (green) highlighted.} \label{fig: logreg_aw_squared_poisson}
\end{figure}

\clearpage
\subsection{RMSE}
\subsubsection{Category 1: Similar performance across all methods, MissARF with smallest average width}
\begin{figure}[!ht]
\centering
\includegraphics[width=0.92\linewidth]{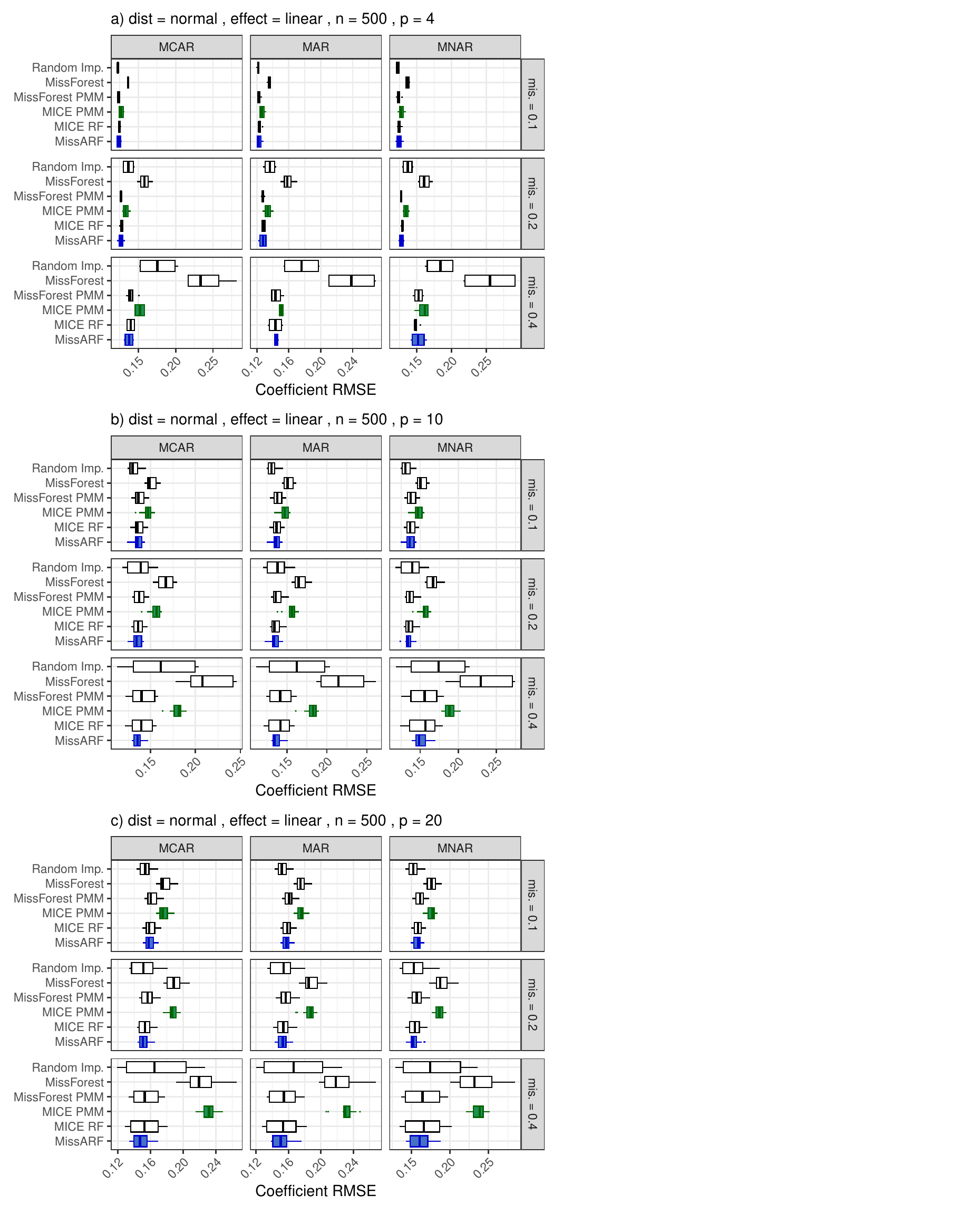}
\caption{\textbf{RMSE of the regression coefficients} of the normal distribution setting with a linear effect over different missingness patterns, dimensionality ($p$) and missingness rates (mis.) with $n=500$. The boxplots are plotted over the features, with MissARF (blue) and MICE PMM (green) highlighted.} \label{fig: logreg_rmse_linear_normal_500}
\end{figure}

\begin{figure}[p]
\centering
\includegraphics[width=0.92\linewidth]{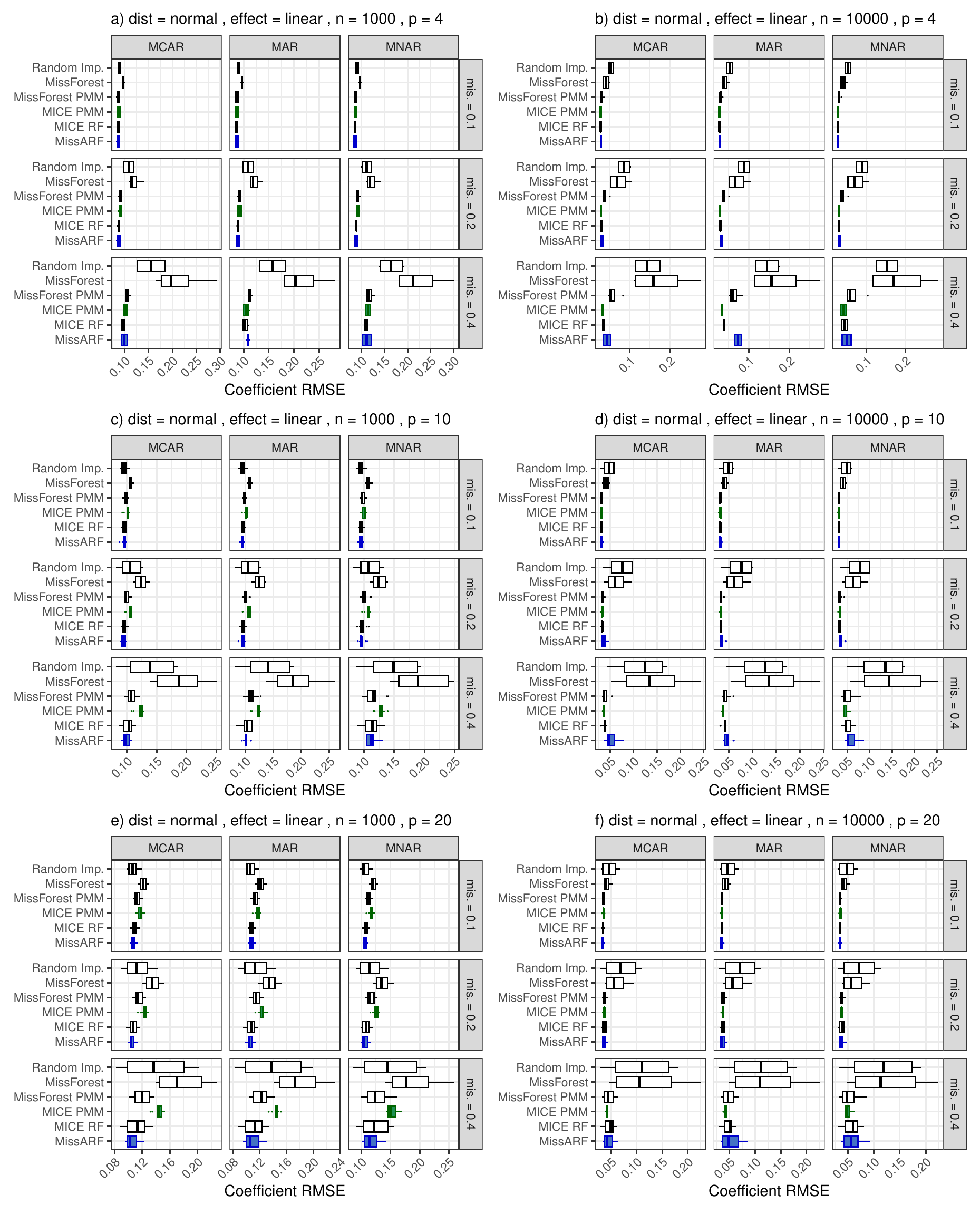}
\caption{\textbf{RMSE of the regression coefficients} of the normal distribution setting with a linear effect over different missingness patterns, dimensionality ($p$) and missingness rates (mis.) with $n=1000$ (left) and $n= 10,000$ (right). The boxplots are plotted over the features, with MissARF (blue) and MICE PMM (green) highlighted.} \label{fig: logreg_rmse_linear_normal}
\end{figure}

\begin{figure}[p]
\centering
\includegraphics[width=0.92\linewidth]{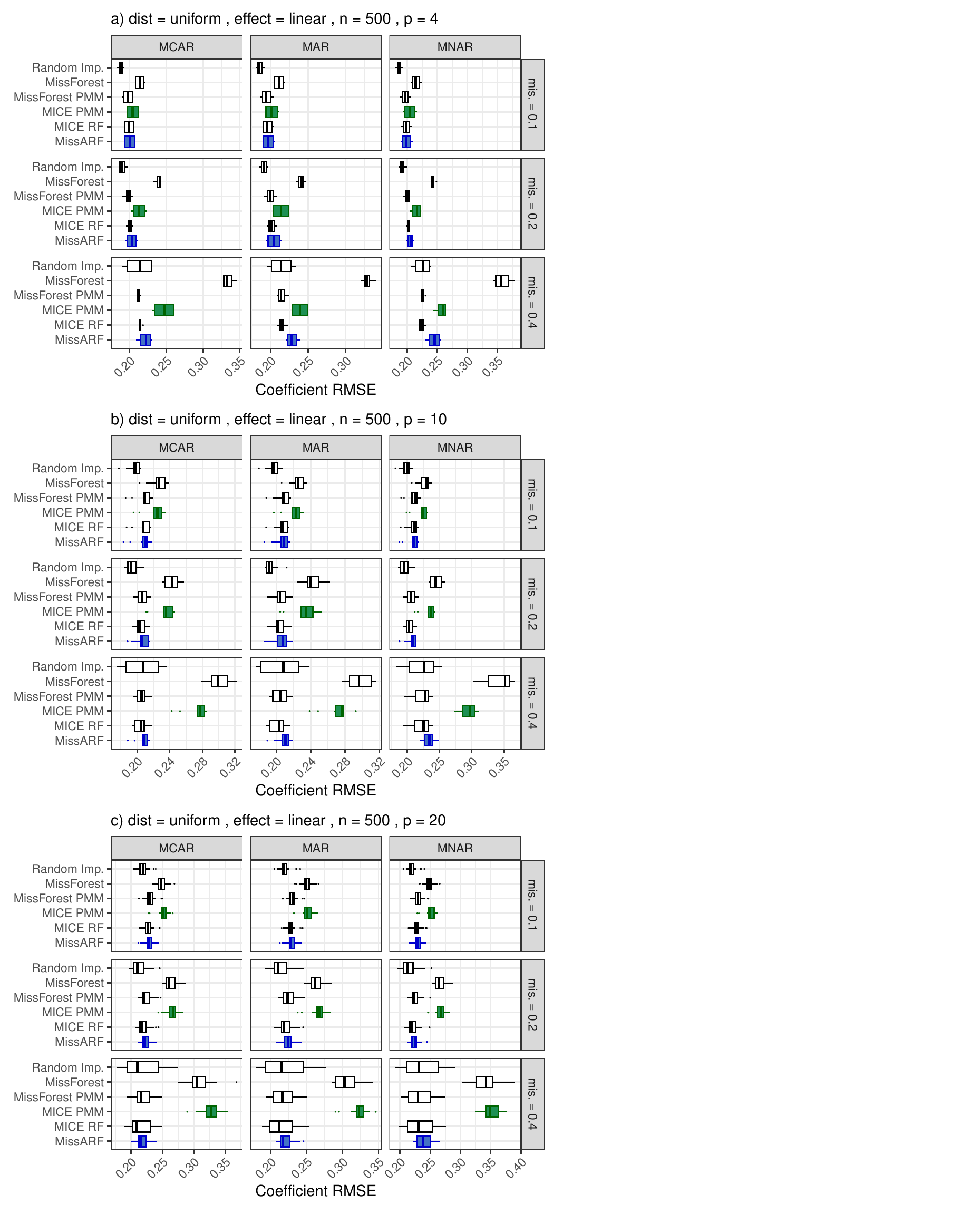}
\caption{\textbf{RMSE of the regression coefficients} of the uniform distribution setting with a linear effect over different missingness patterns, dimensionality ($p$) and missingness rates (mis.) with $n=500$. The boxplots are plotted over the features, with MissARF (blue) and MICE PMM (green) highlighted.} \label{fig: logreg_rmse_linear_uniform_500}
\end{figure}

\begin{figure}[p]
\centering
\includegraphics[width=0.92\linewidth]{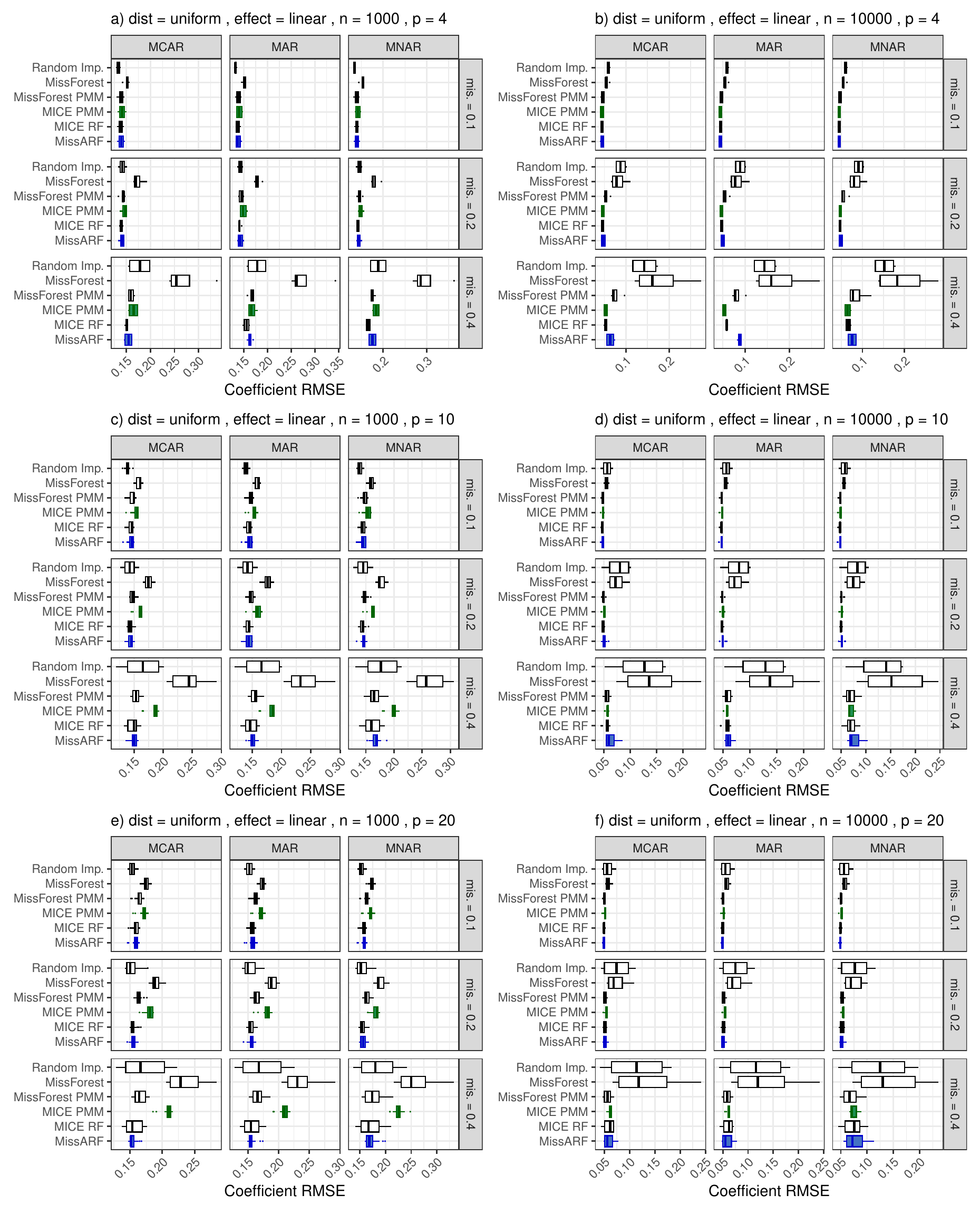}
\caption{\textbf{RMSE of the regression coefficients} of the uniform distribution setting with a linear effect over different missingness patterns, dimensionality ($p$) and missingness rates (mis.) with $n=1000$ (left) and $n= 10,000$ (right). The boxplots are plotted over the features, with MissARF (blue) and MICE PMM (green) highlighted.} \label{fig: logreg_rmse_linear_uniform}
\end{figure}

\begin{figure}[p]
\centering
\includegraphics[width=0.92\linewidth]{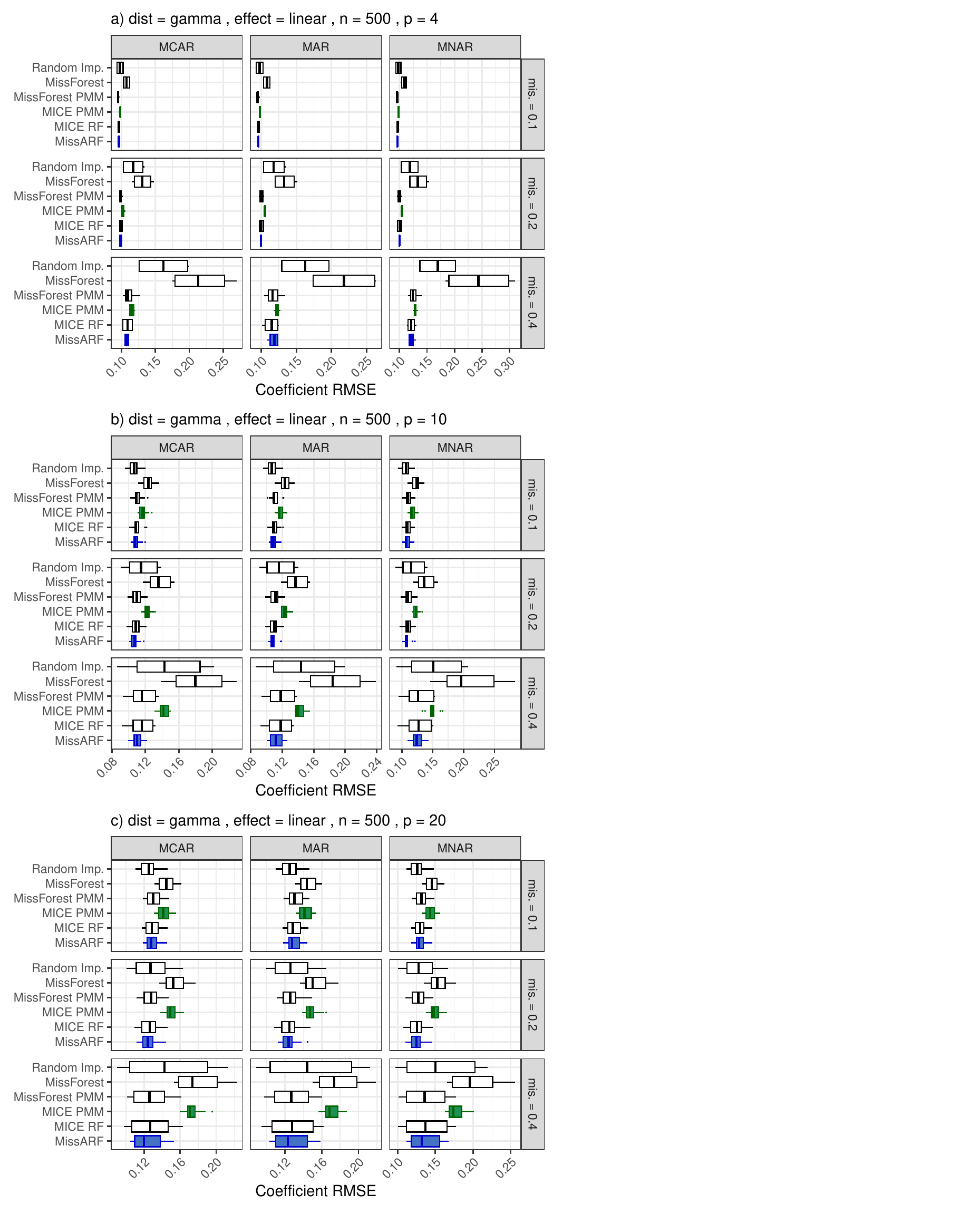}
\caption{ \textbf{RMSE of the regression coefficients} of the gamma distribution setting with a linear effect over different missingness patterns, dimensionality ($p$) and missingness rates (mis.) with $n=500$. The boxplots are plotted over the features, with MissARF (blue) and MICE PMM (green) highlighted.} \label{fig: logreg_rmse_linear_gamma_500}
\end{figure}

\begin{figure}[p]
\centering
\includegraphics[width=0.92\linewidth]{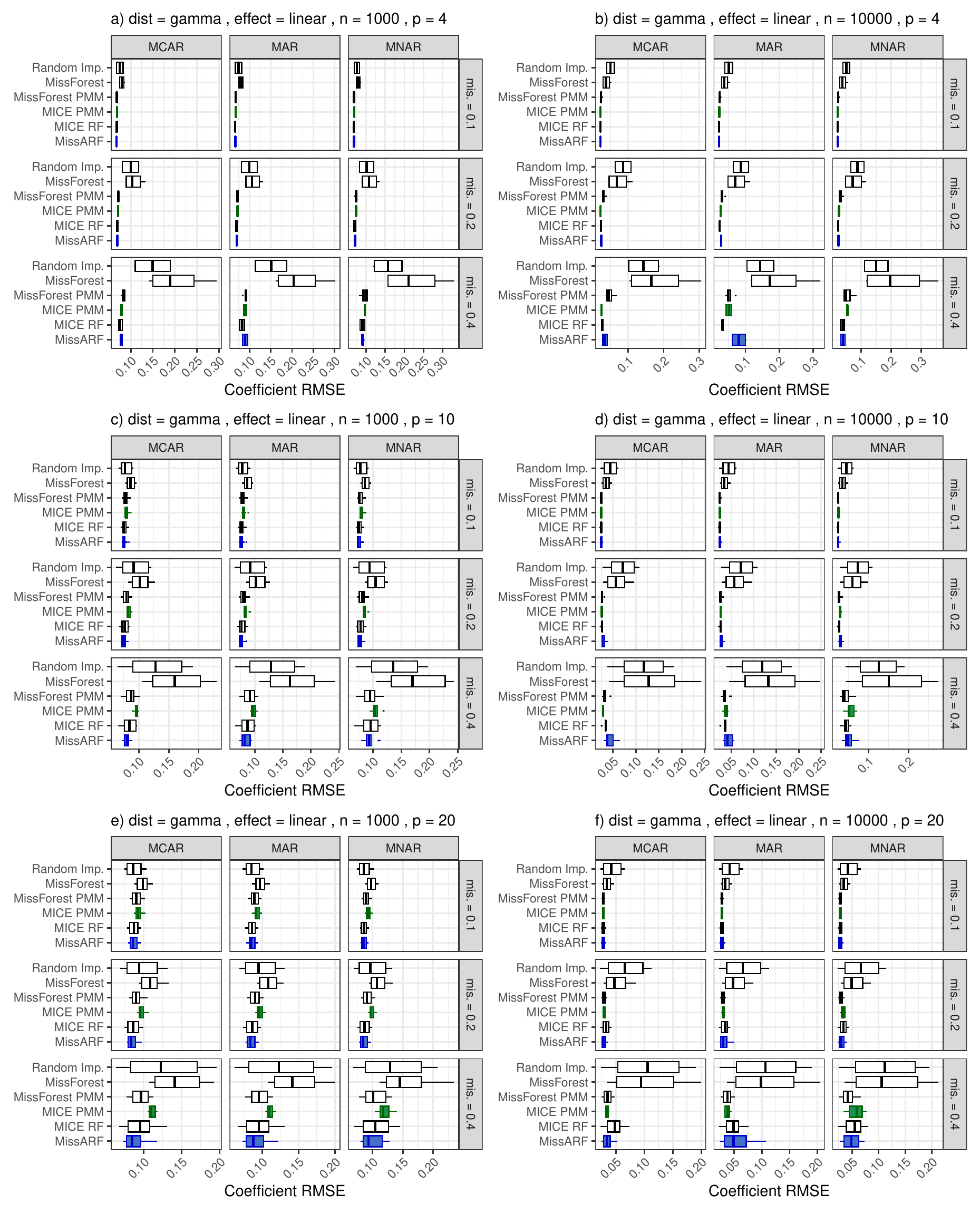}
\caption{ \textbf{RMSE of the regression coefficients} of the gamma distribution setting with a linear effect over different missingness patterns, dimensionality ($p$) and missingness rates (mis.) with $n=1000$ (left) and $n= 10,000$ (right). The boxplots are plotted over the features, with MissARF (blue) and MICE PMM (green) highlighted.} \label{fig: logreg_rmse_linear_gamma}
\end{figure}

\begin{figure}[p]
\centering
\includegraphics[width=0.92\linewidth]{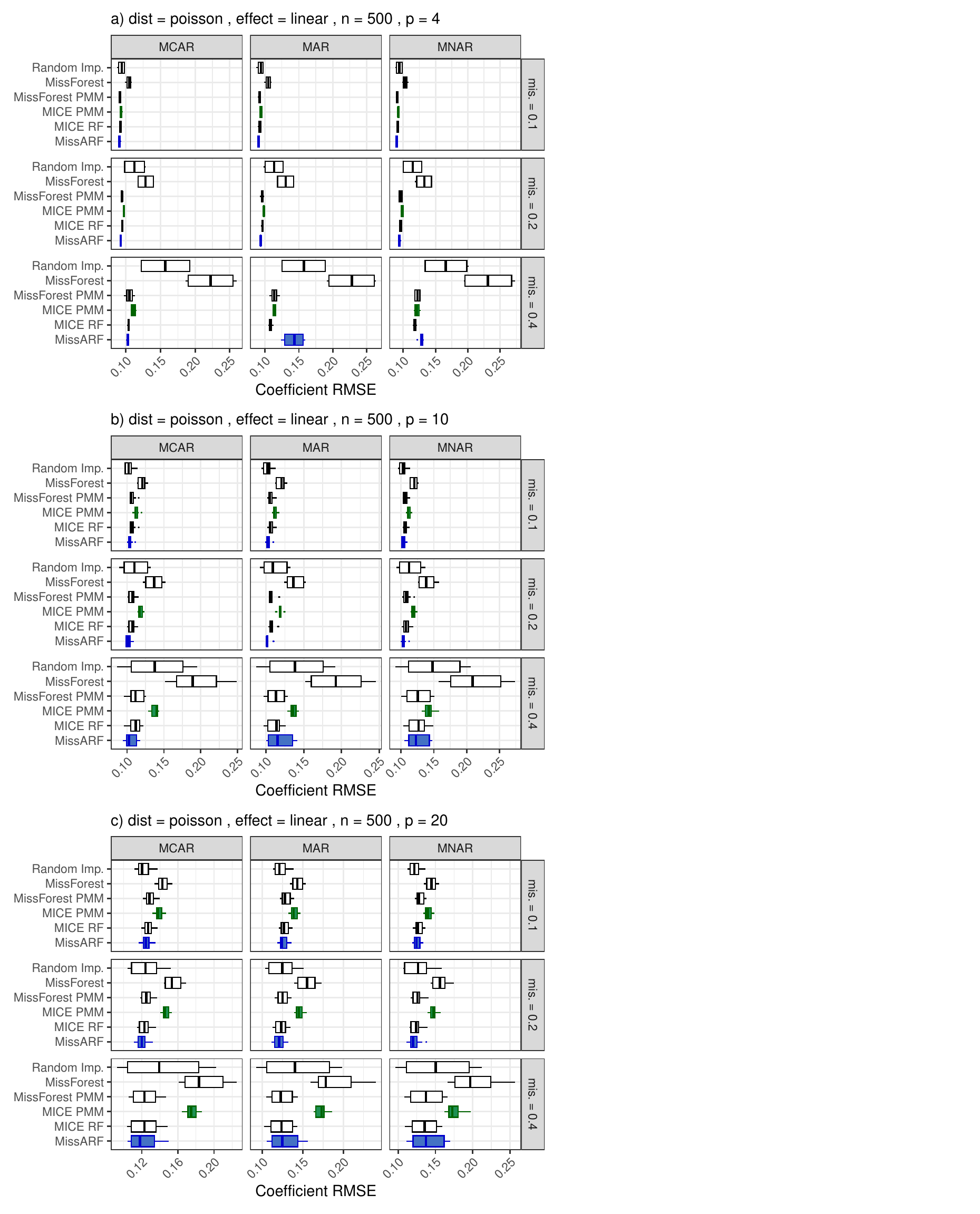}
\caption{ \textbf{RMSE of the regression coefficients} of the Poisson distribution setting with a linear effect over different missingness patterns, dimensionality ($p$) and missingness rates (mis.) with $n=500$. The boxplots are plotted over the features, with MissARF (blue) and MICE PMM (green) highlighted.} \label{fig: logreg_rmse_linear_poisson_500}
\end{figure}

\begin{figure}[p]
\centering
\includegraphics[width=0.92\linewidth]{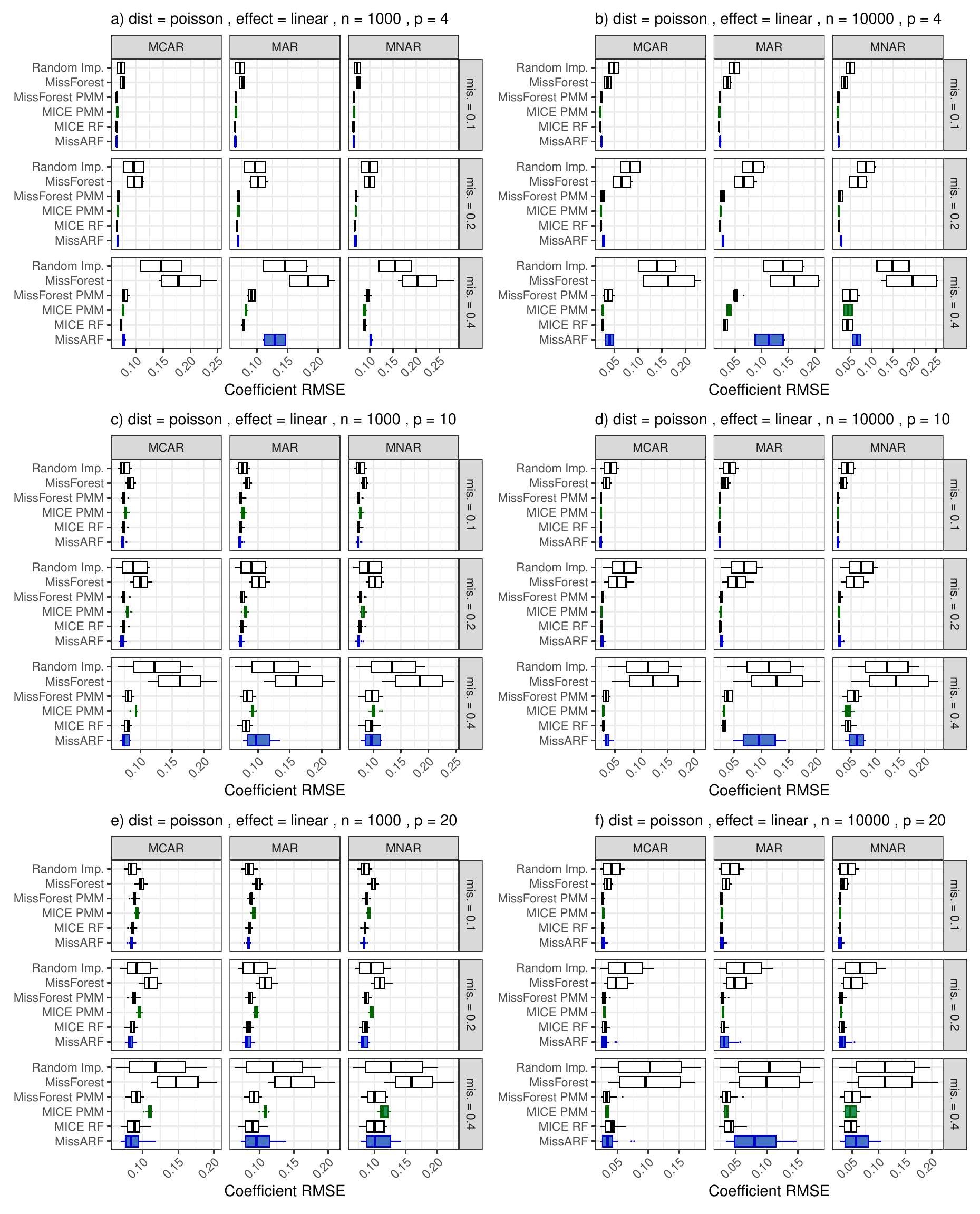}
\caption{ \textbf{RMSE of the regression coefficients} of the Poisson distribution setting with a linear effect over different missingness patterns, dimensionality ($p$) and missingness rates (mis.) with $n=1000$ (left) and $n= 10,000$ (right). The boxplots are plotted over the features, with MissARF (blue) and MICE PMM (green) highlighted.} \label{fig: logreg_rmse_linear_poisson}
\end{figure}

\begin{figure}[p]
\centering
\includegraphics[width=0.92\linewidth]{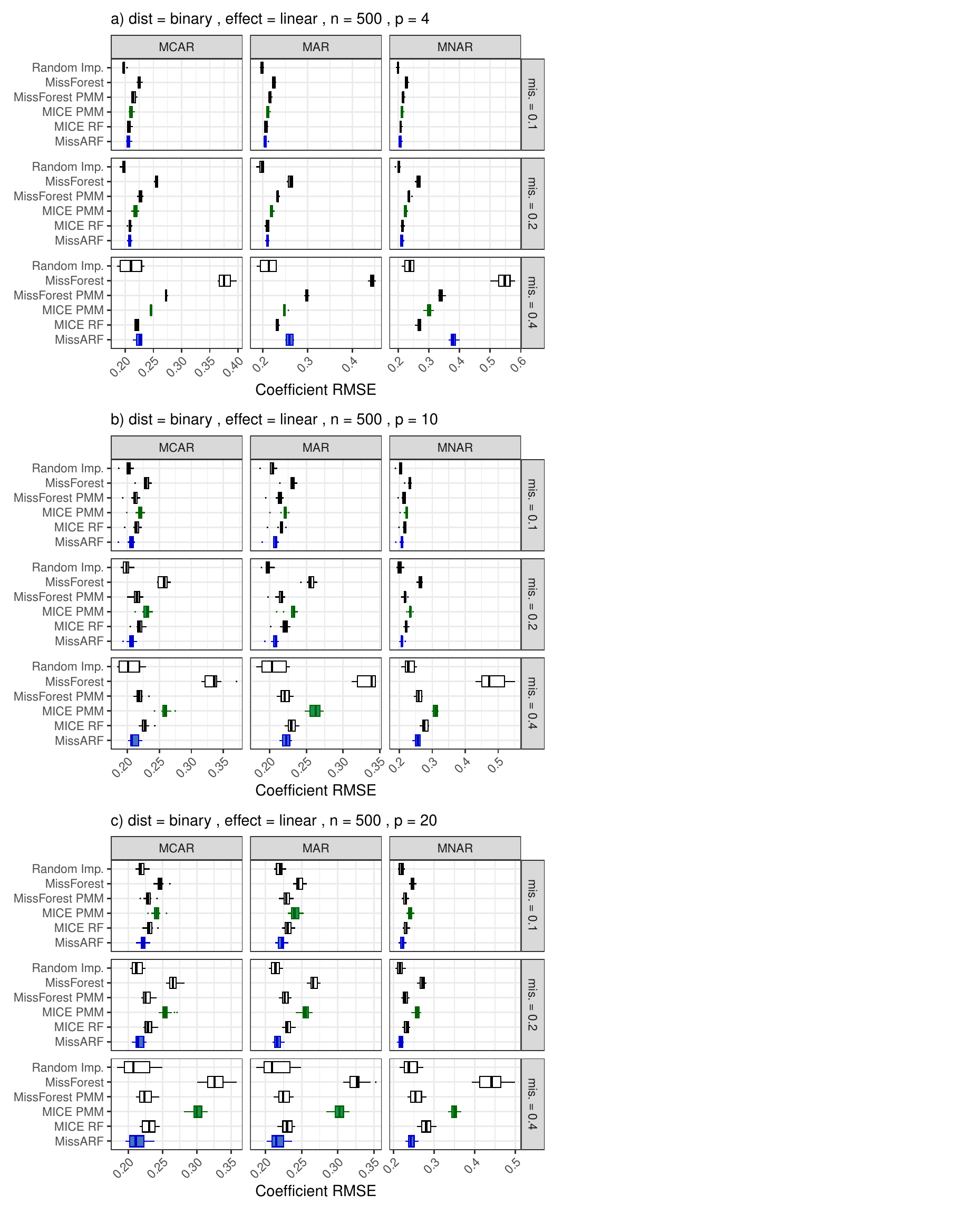}
\caption{ \textbf{RMSE of the regression coefficients} of the binary distribution setting with a linear effect over different missingness patterns, dimensionality ($p$) and missingness rates (mis.) with $n=500$. The boxplots are plotted over the features, with MissARF (blue) and MICE PMM (green) highlighted.} \label{fig: logreg_rmse_linear_binary_500}
\end{figure}

\begin{figure}[p]
\centering
\includegraphics[width=0.92\linewidth]{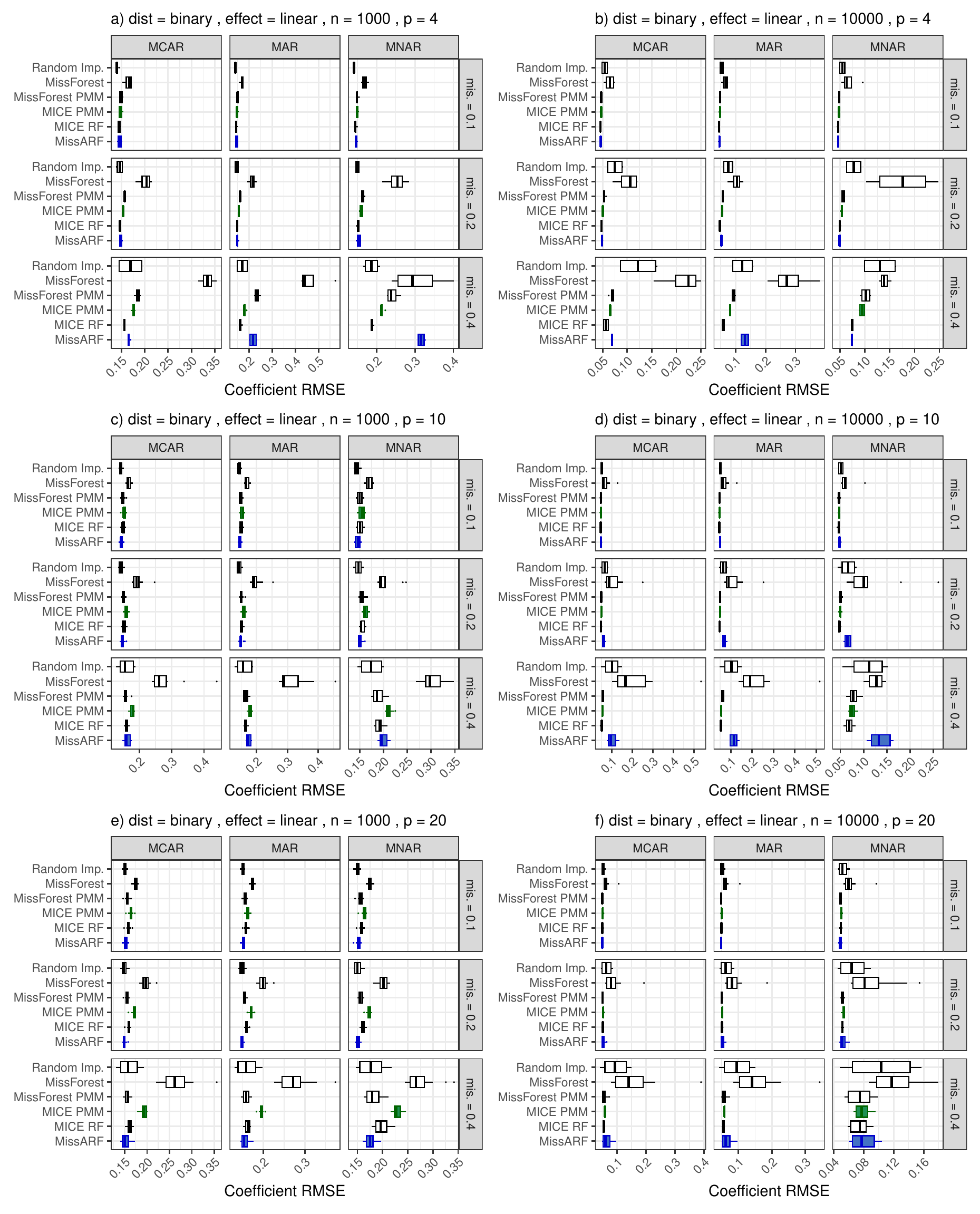}
\caption{ \textbf{RMSE of the regression coefficients} of the binary distribution setting with a linear effect over different missingness patterns, dimensionality ($p$) and missingness rates (mis.) with $n=1000$ (left) and $n= 10,000$ (right). The boxplots are plotted over the features, with MissARF (blue) and MICE PMM (green) highlighted.} \label{fig: logreg_rmse_linear_binary}
\end{figure}

\begin{figure}[p]
\centering
\includegraphics[width=0.92\linewidth]{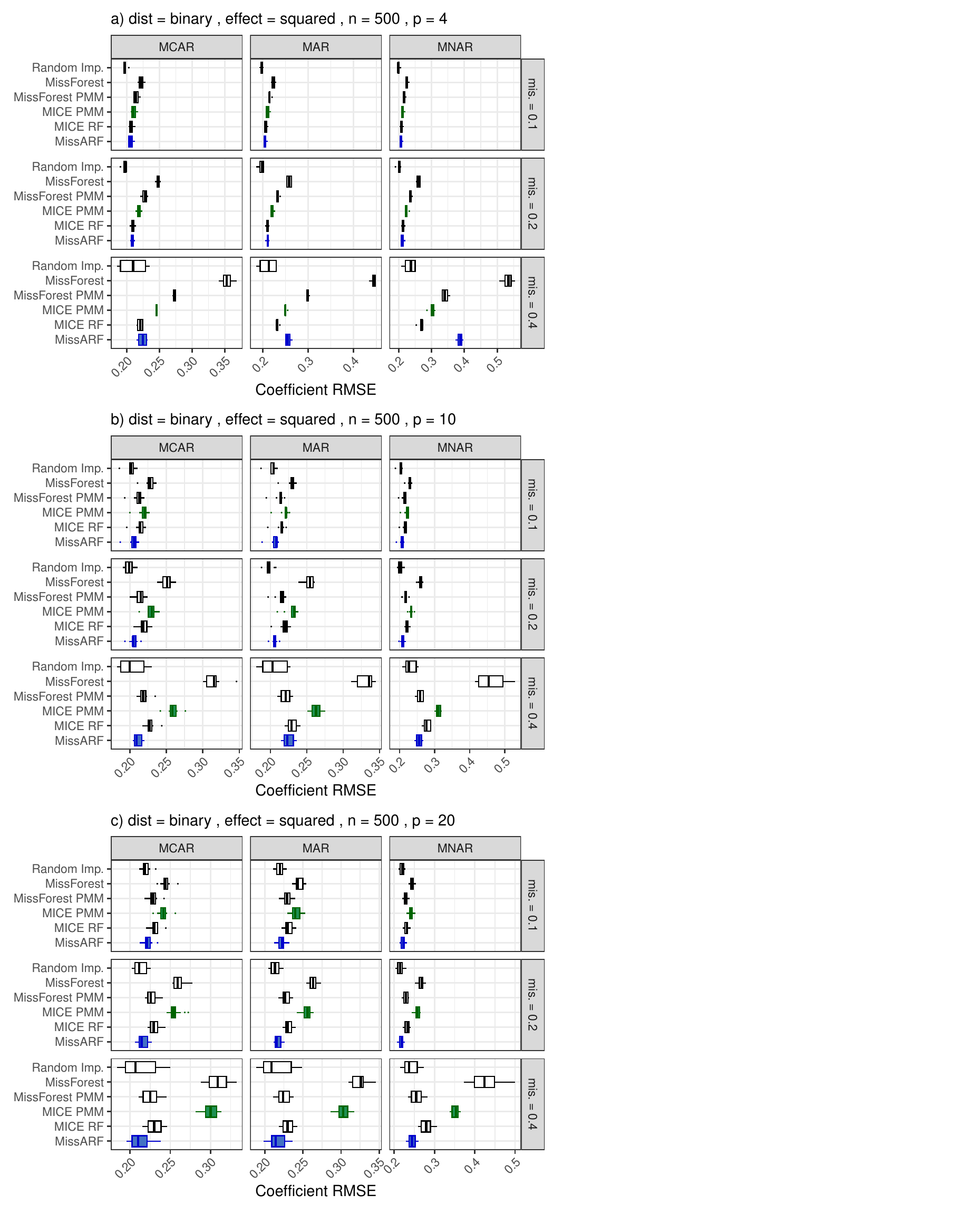}
\caption{ \textbf{RMSE of the regression coefficients} of the binary distribution setting with a squared effect over different missingness patterns, dimensionality ($p$) and missingness rates (mis.) with $n=500$. The boxplots are plotted over the features, with MissARF (blue) and MICE PMM (green) highlighted.} \label{fig: logreg_rmse_squared_binary_500}
\end{figure}

\begin{figure}[p]
\centering
\includegraphics[width=0.92\linewidth]{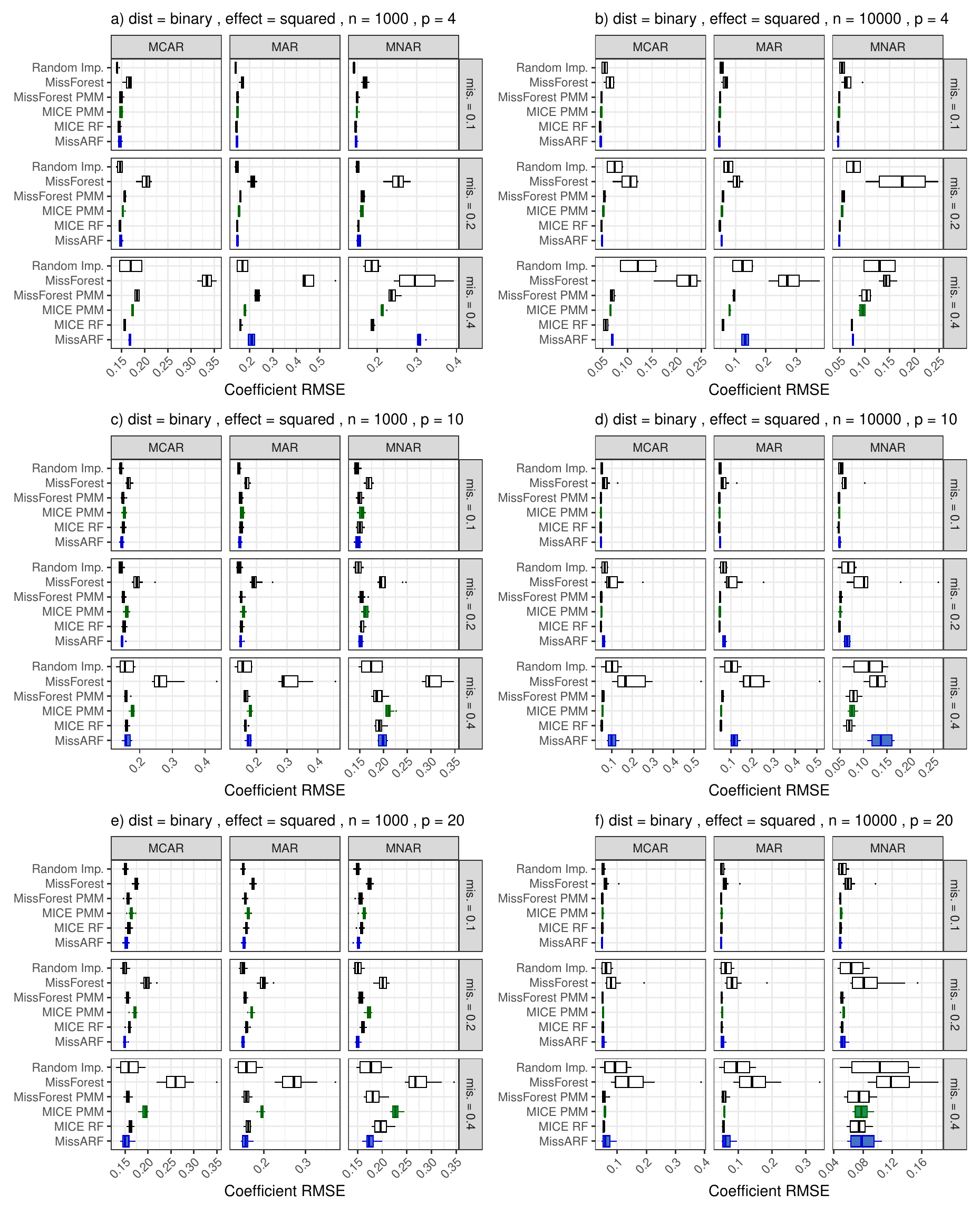}
\caption{ \textbf{RMSE of the regression coefficients} of the binary distribution setting with a squared effect over different missingness patterns, dimensionality ($p$) and missingness rates (mis.) with $n=1000$ (left) and $n= 10,000$ (right). The boxplots are plotted over the features, with MissARF (blue) and MICE PMM (green) highlighted.} \label{fig: logreg_rmse_squared_binary}
\end{figure}

\clearpage

\subsubsection{Category 2: PMM methods struggle, MissARF performs well}
\begin{figure}[!ht]
\centering
\includegraphics[width=0.92\linewidth]{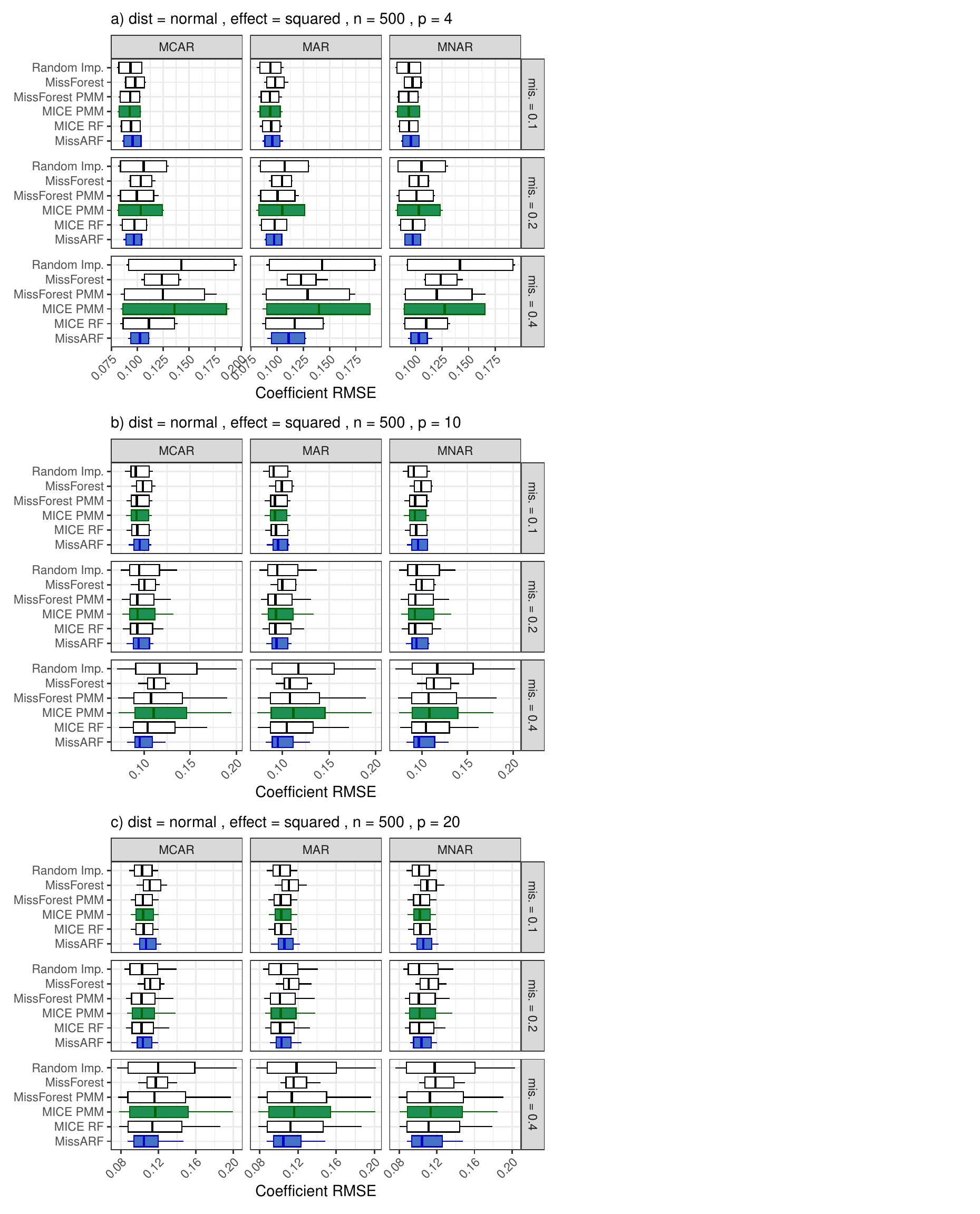}
\caption{ \textbf{RMSE of the regression coefficients} of the normal distribution setting with a squared effect over different missingness patterns, dimensionality ($p$) and missingness rates (mis.) with $n=500$. The boxplots are plotted over the features, with MissARF (blue) and MICE PMM (green) highlighted.} \label{fig: logreg_rmse_squared_normal_500}
\end{figure}

\begin{figure}[p]
\centering
\includegraphics[width=0.92\linewidth]{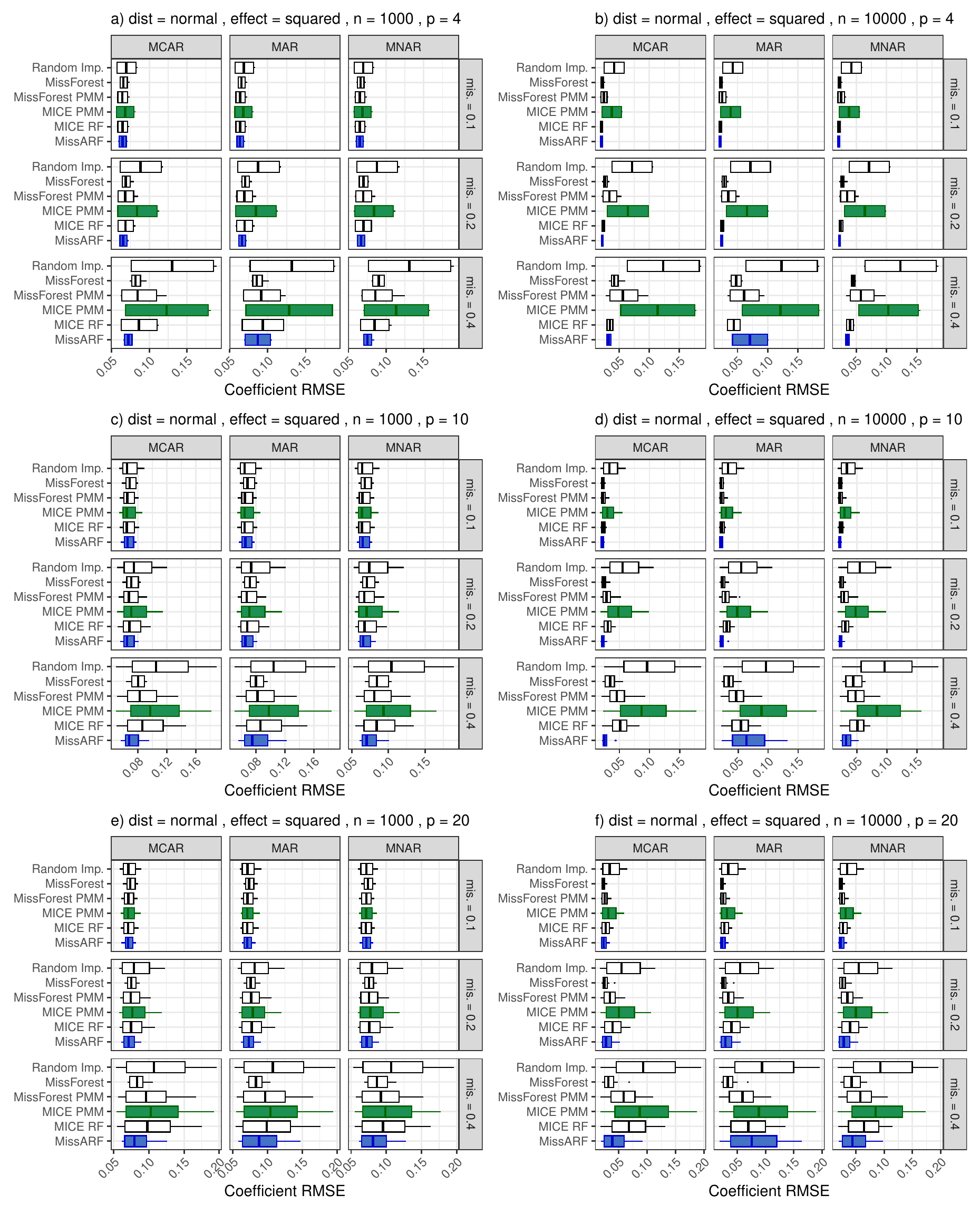}
\caption{ \textbf{RMSE of the regression coefficients} of the normal distribution setting with a squared effect over different missingness patterns, dimensionality ($p$) and missingness rates (mis.) with $n=1000$ (left) and $n= 10,000$ (right). The boxplots are plotted over the features, with MissARF (blue) and MICE PMM (green) highlighted.} \label{fig: logreg_rmse_squared_normal}
\end{figure}

\begin{figure}[p]
\centering
\includegraphics[width=0.92\linewidth]{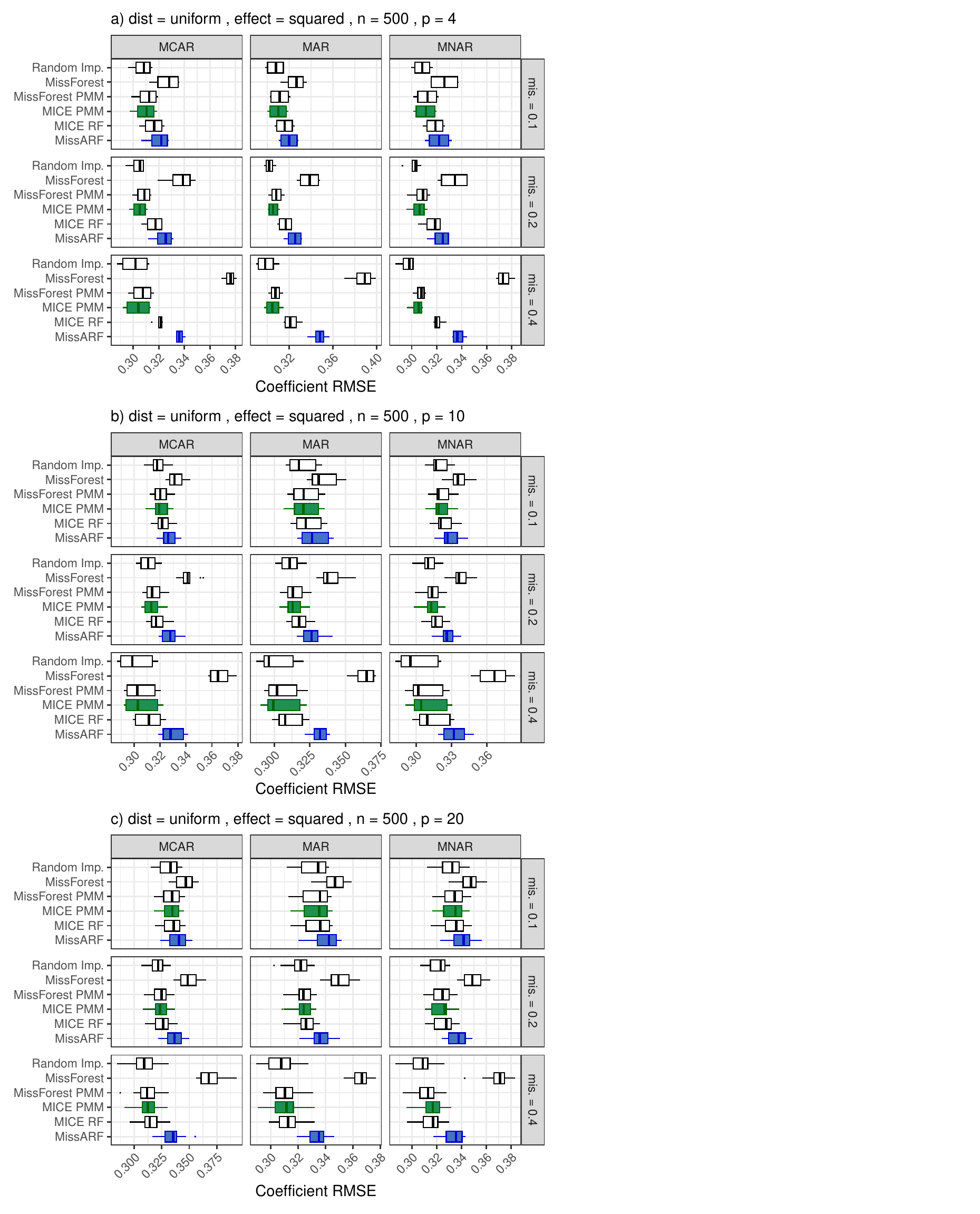}
\caption{ \textbf{RMSE of the regression coefficients} of the uniform distribution setting with a squared effect over different missingness patterns, dimensionality ($p$) and missingness rates (mis.) with $n=500$. The boxplots are plotted over the features, with MissARF (blue) and MICE PMM (green) highlighted.} \label{fig: logreg_rmse_squared_uniform_500}
\end{figure}

\begin{figure}[p]
\centering
\includegraphics[width=0.92\linewidth]{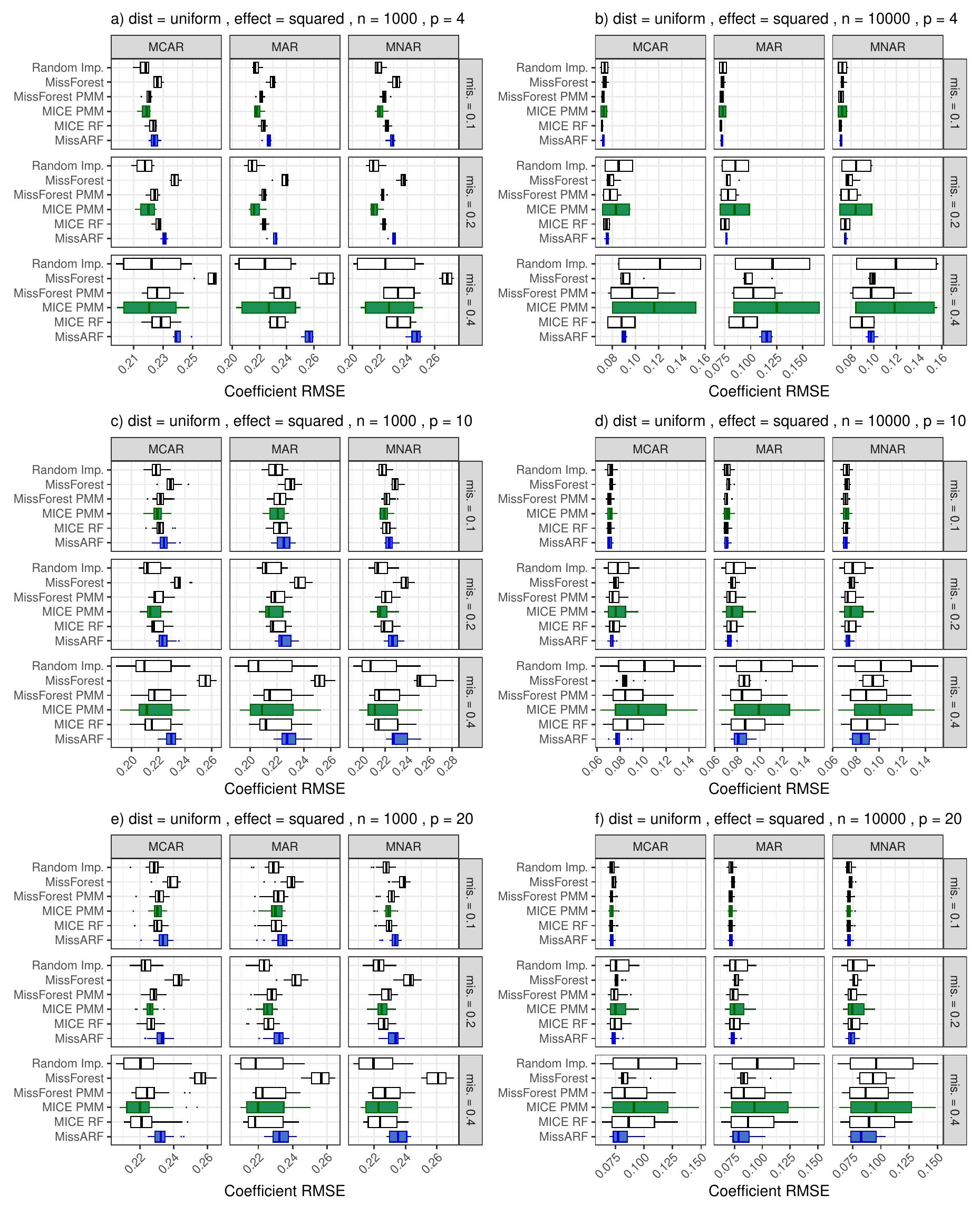}
\caption{ \textbf{RMSE of the regression coefficients} of the uniform distribution setting with a squared effect over different missingness patterns, dimensionality ($p$) and missingness rates (mis.) with $n=1000$ (left) and $n= 10,000$ (right). The boxplots are plotted over the features, with MissARF (blue) and MICE PMM (green) highlighted.} \label{fig: logreg_rmse_squared_uniform}
\end{figure}

\clearpage

\subsubsection{Category 3: All methods perform poorly}
\begin{figure}[!ht]
\centering
\includegraphics[width=0.92\linewidth]{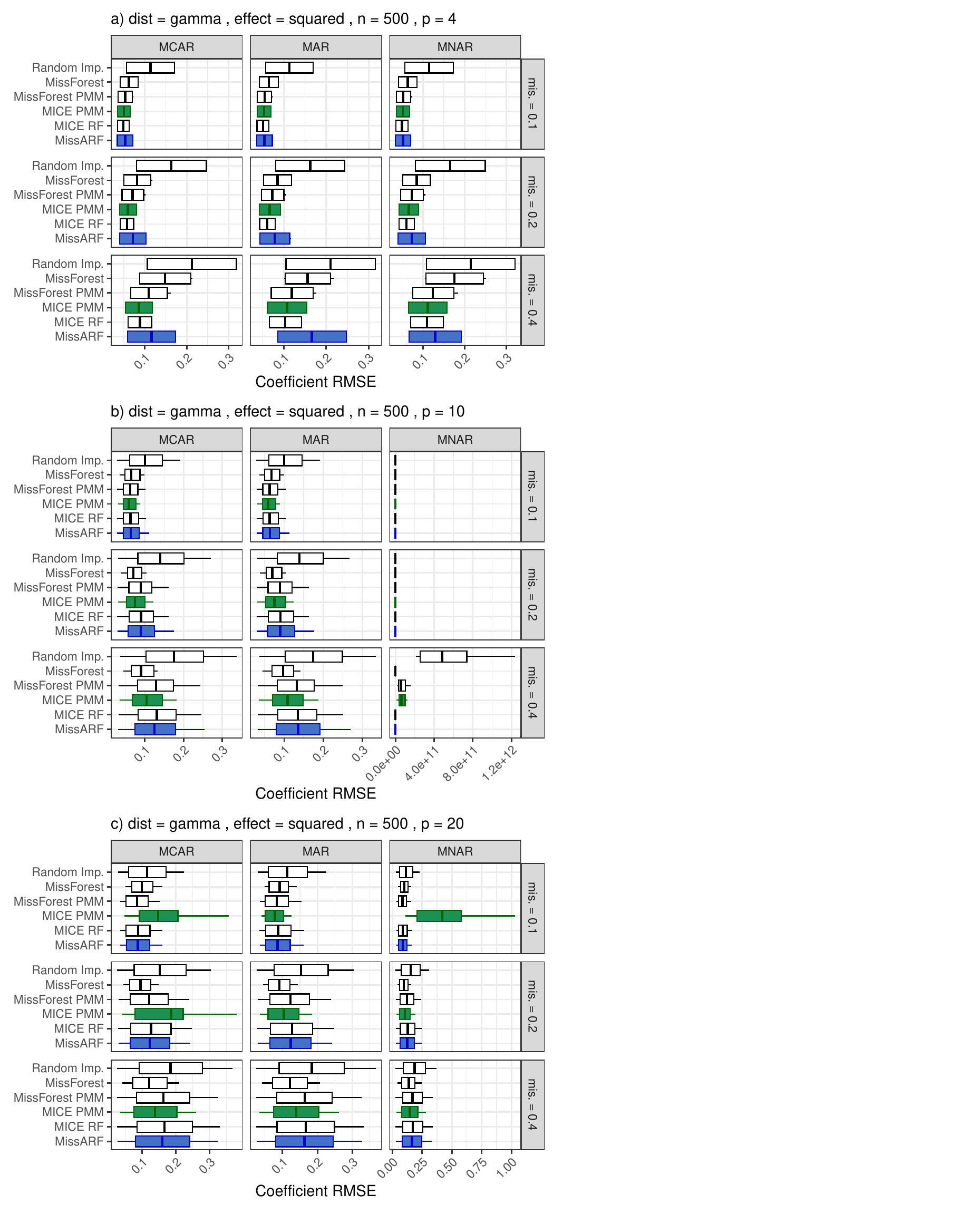}
\caption{ \textbf{RMSE of the regression coefficients} of the gamma distribution setting with a squared effect over different missingness patterns, dimensionality ($p$) and missingness rates (mis.) with $n=500$. The boxplots are plotted over the features, with MissARF (blue) and MICE PMM (green) highlighted.} \label{fig: logreg_rmse_squared_gamma_500}
\end{figure}

\begin{figure}[p]
\centering
\includegraphics[width=0.92\linewidth]{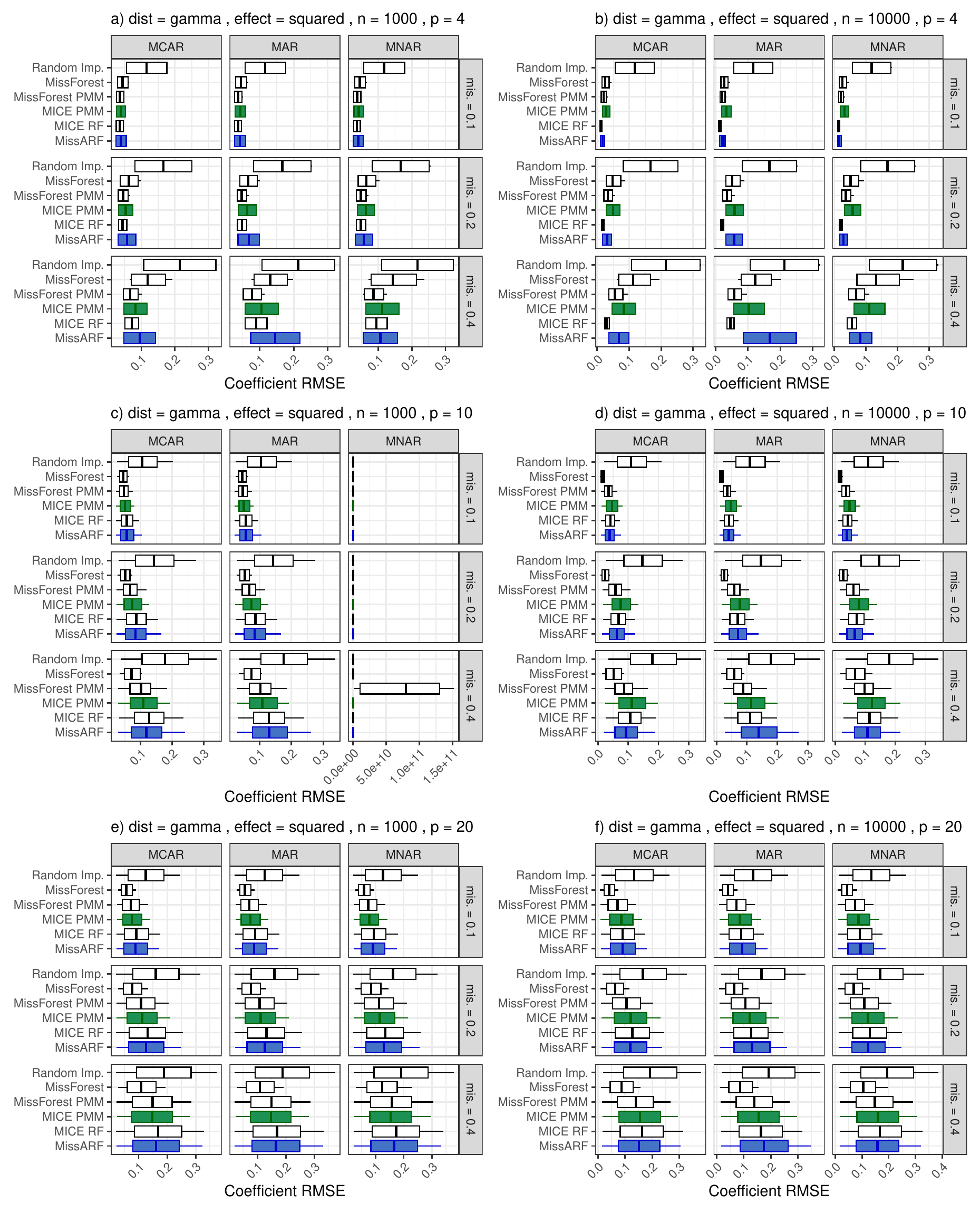}
\caption{ \textbf{RMSE of the regression coefficients} of the gamma distribution setting with a squared effect over different missingness patterns, dimensionality ($p$) and missingness rates (mis.) with $n=1000$ (left) and $n= 10,000$ (right). The boxplots are plotted over the features, with MissARF (blue) and MICE PMM (green) highlighted.} \label{fig: logreg_rmse_squared_gamma}
\end{figure}

\begin{figure}[p]
\centering
\includegraphics[width=0.92\linewidth]{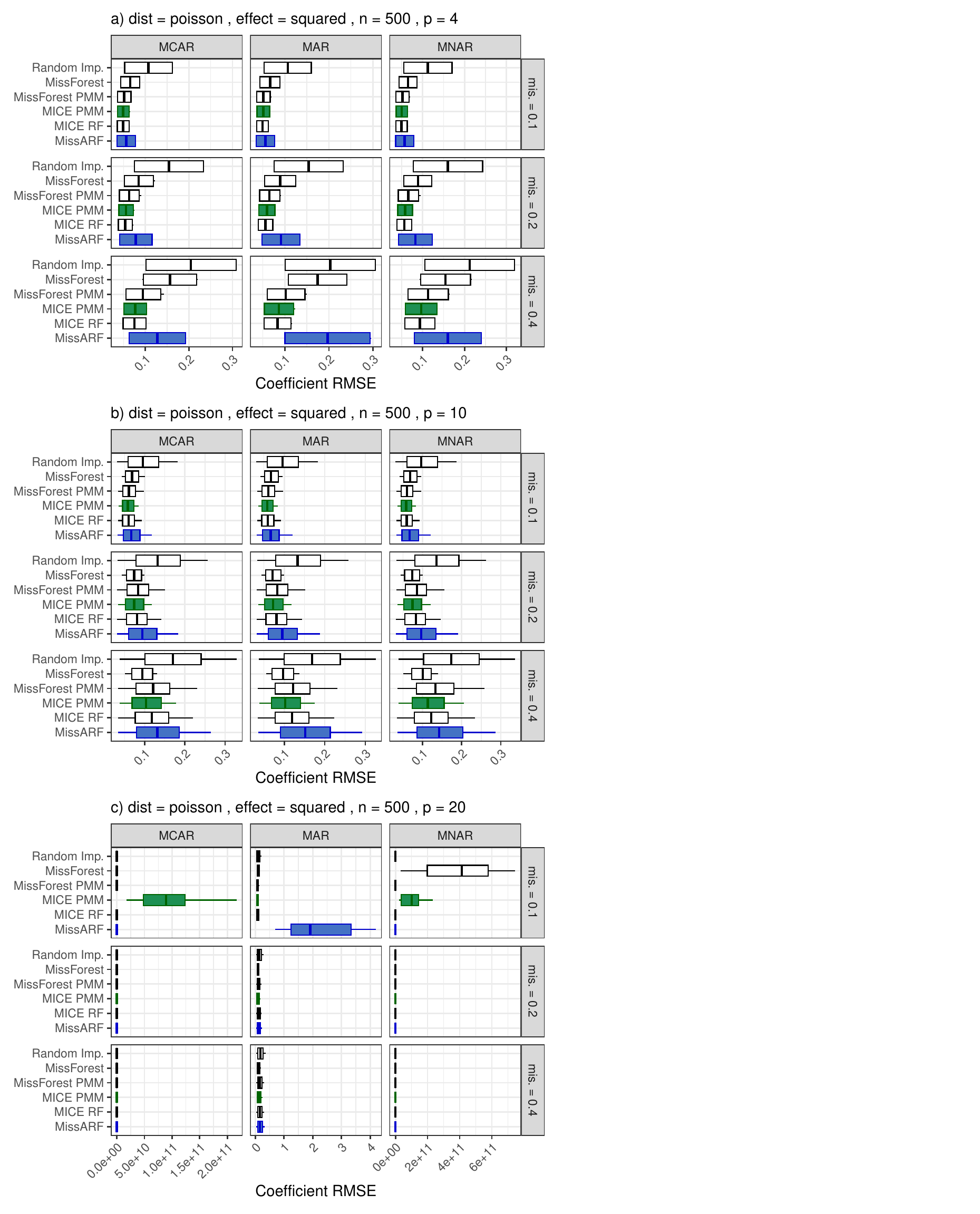}
\caption{ \textbf{RMSE of the regression coefficients} of the Poisson distribution setting with a squared effect over different missingness patterns, dimensionality ($p$) and missingness rates (mis.) with $n=500$. The boxplots are plotted over the features, with MissARF (blue) and MICE PMM (green) highlighted.} \label{fig: logreg_rmse_squared_poisson_500}
\end{figure}

\begin{figure}[p]
\centering
\includegraphics[width=0.92\linewidth]{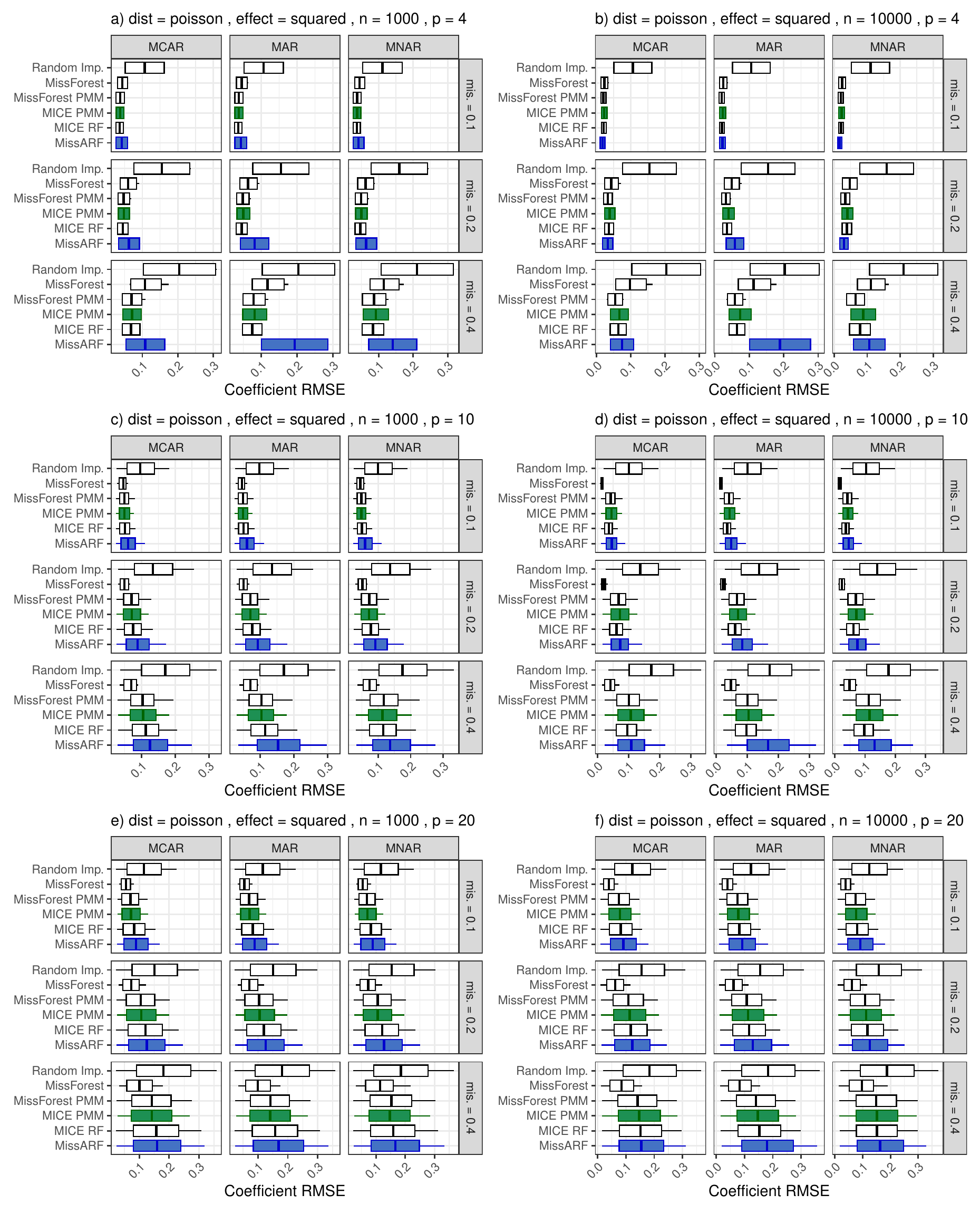}
\caption{ \textbf{RMSE of the regression coefficients} of the Poisson distribution setting with a squared effect over different missingness patterns, dimensionality ($p$) and missingness rates (mis.) with $n=1000$ (left) and $n= 10,000$ (right). The boxplots are plotted over the features, with MissARF (blue) and MICE PMM (green) highlighted.} \label{fig: logreg_rmse_squared_poisson}
\end{figure}

\clearpage
\section{Real Data}
\begin{figure}[ht]
    \centering
    \includegraphics[width=0.92\linewidth]{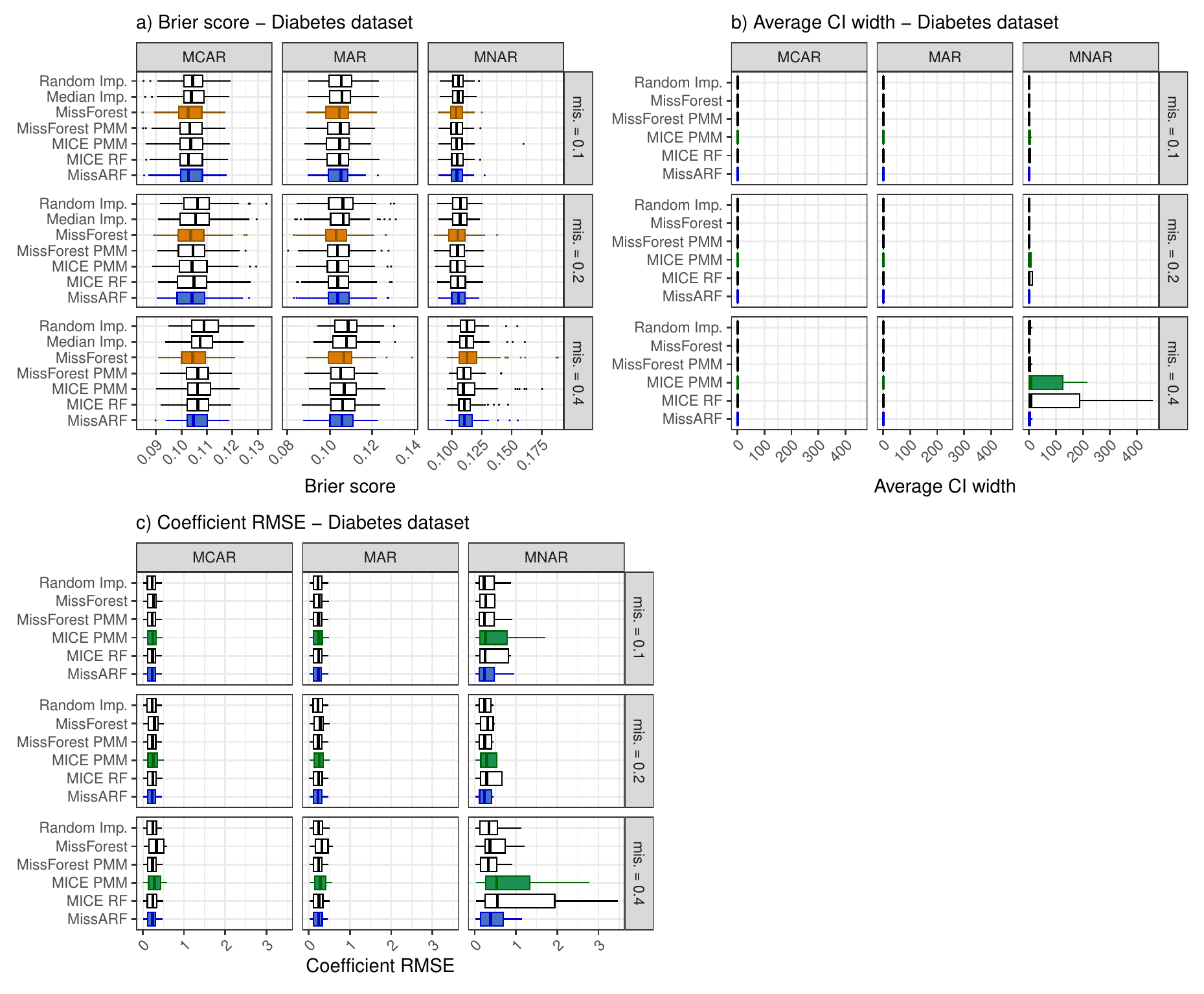}
    \caption{\textbf{Real data example results} over the different
missingness proportions and patterns for a) Brier score, b) average width of the confidence intervals and c) the RMSE of the regression coefficients. The boxplots are plotted over the replicates in a) and over the features in b)-c) with MissARF (blue), MissForest (orange) and MICE PMM (green) highlighted.}
    \label{fig: real data rest}
\end{figure}

\clearpage
\section{Runtime comparisons}
\label{sec:appx_runtime}
\begin{figure}[ht]
    \centering
    \includegraphics[width=0.92\linewidth]{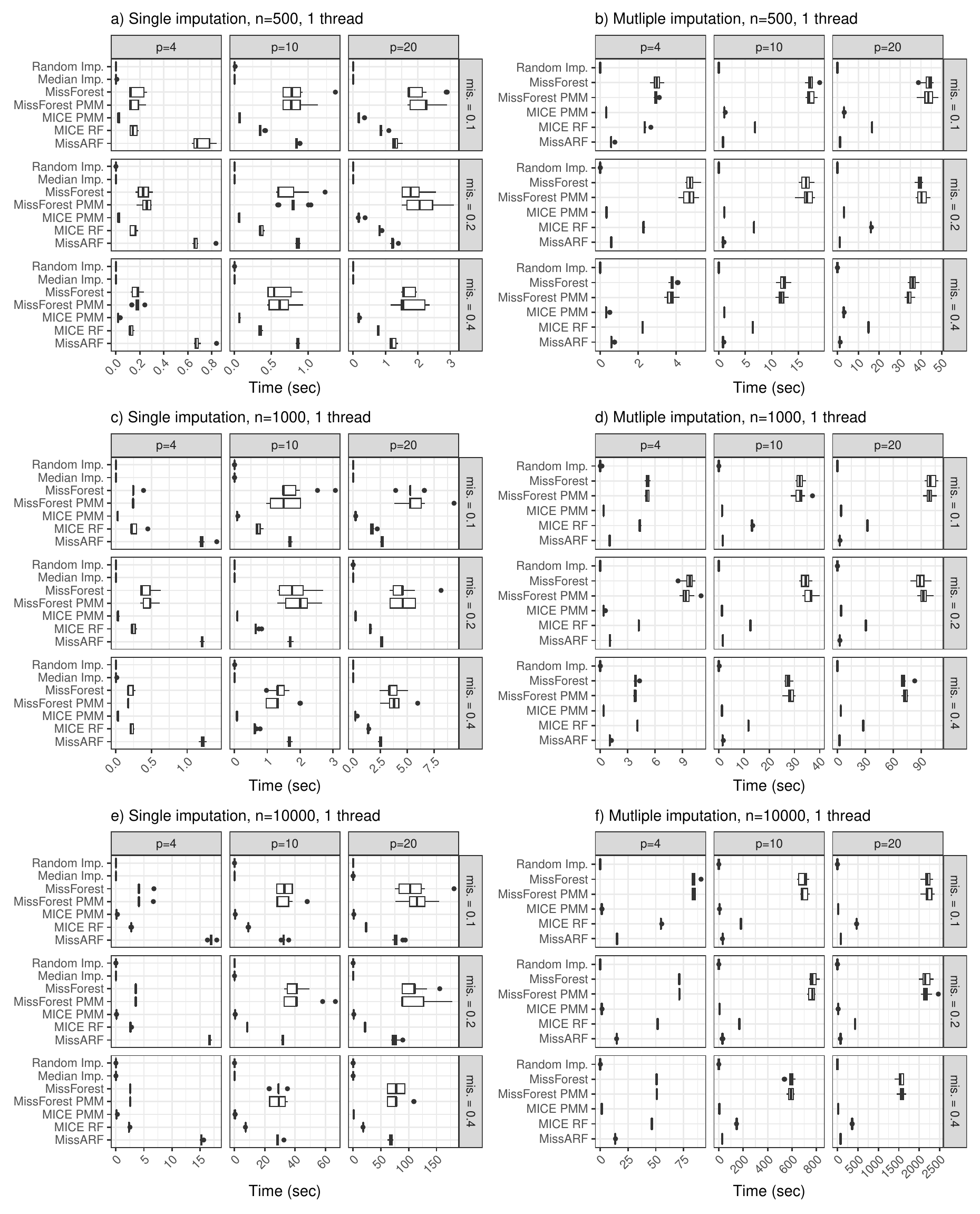}
    \caption{\textbf{Runtime comparison with one thread} of the different imputation methods across the dimensionality ($p$) and missingness proportions (mis.) with different $n$. The boxplots for single imputation (left) and multiple imputation (right) are plotted over the 10 replicates. The runtime is shown in seconds (sec).}
    \label{fig: runtime k1}
\end{figure}

\begin{figure}[p]
    \centering
    \includegraphics[width=0.92\linewidth]{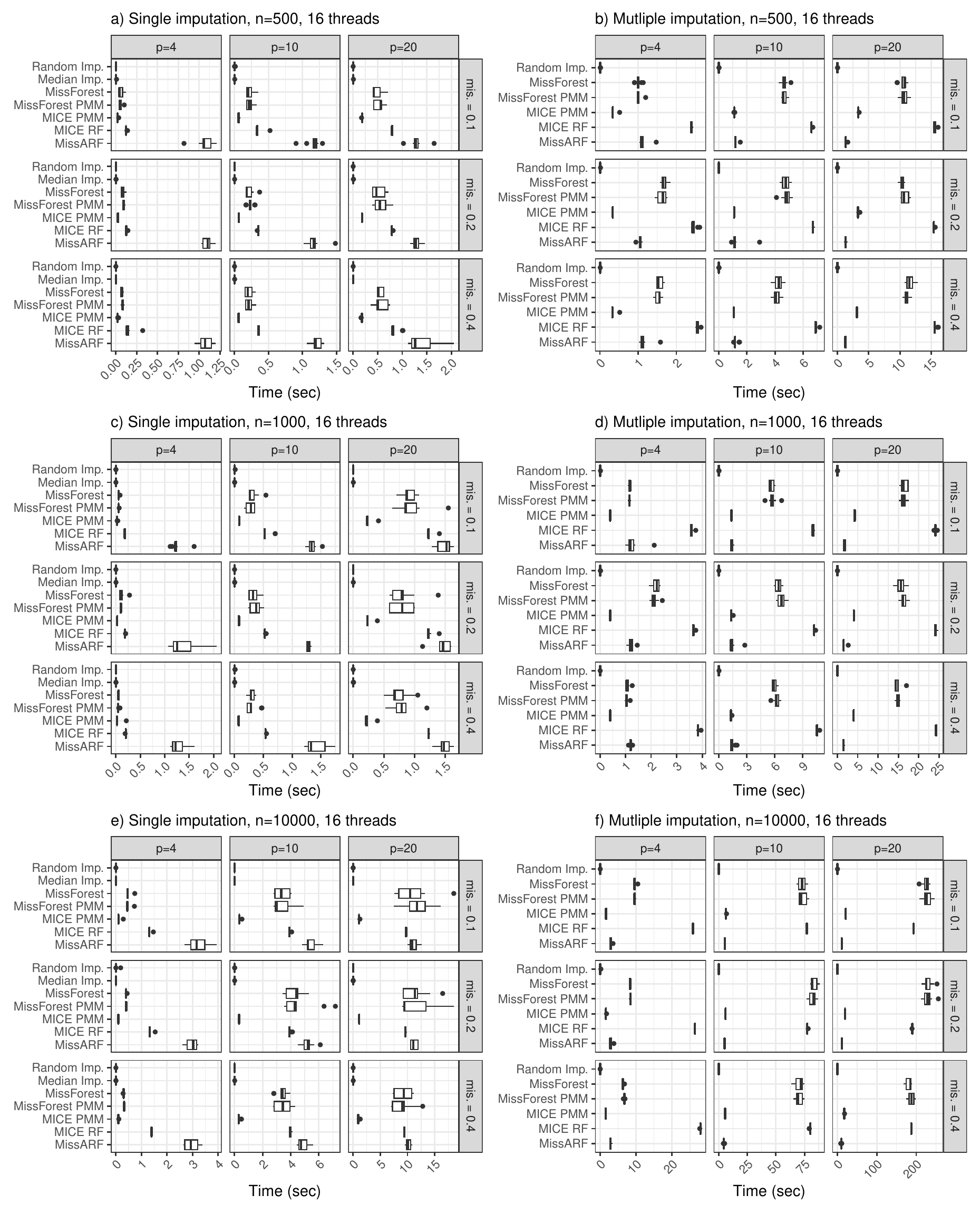}
    \caption{\textbf{Runtime comparison with 16 threads} of the different imputation methods across the dimensionality ($p$) and missingness proportions (mis.) with different $n$. The boxplots for single imputation (left) and multiple imputation (right) are plotted over the 10 replicates. The runtime is shown in seconds (sec).}
    \label{fig: runtime k16}
\end{figure}

Tables~\ref{tab:runtime k1} and \ref{tab:runtime k16} show more detailed runtime comparisons over different $n$. In Figures~\ref{fig: runtime k1} and ~\ref{fig: runtime k16}, we investigate runtime comparisons in more detail over different dimensions and missingness proportions. The runtime comparisons were performed using the following package versions: \texttt{missRanger\_2.6.1}, \texttt{arf\_0.2.4}, \texttt{mice\_3.17.0}, \texttt{simstudy\_0.8.0}, \texttt{missMethods\_0.4.0}. 
%Note, that for the simulations we used an earlier version.
\begin{table}[ht]
    \centering
\begin{tabular}{l|rr|rr|rr}
 \hline
 Method & \multicolumn{2}{c}{$n=500$} & \multicolumn{2}{c}{$n=1000$} & \multicolumn{2}{c}{$n=10 000$} \\
 & Single & Multiple  & Single & Multiple & Single & Multiple \\
 \hline
Random Imp. & $<$0.01 & 0.02 & $<$0.01 & 0.02 & $<$0.01 & 0.07\\
Median Imp. & $<$0.01 & - & $<$0.01 & - & 0.01 & -\\
MissForest & 0.92 & 19.59 & 2.17 & 41.59 & 45.11 & 906.93\\
MissForest PMM & 0.98 & 19.53 & 2.20 & 42.13 & 46.41 & 913.82\\
MICE PMM & 0.09 & 1.50 & 0.11 & 1.86 & 0.48 & 8.46\\
MICE RF & 0.44 & 8.26 & 0.84 & 15.66 & 10.63 & 211.44\\
MissARF & 0.93 & 0.87 & 1.84 & 1.70 & 40.42 & 40.77\\
\hline
\end{tabular}
    \caption{Runtime comparison of the different methods, executed with one thread. The measured average times in seconds for single and multiple imputation for each of the different $n$ are compared.}
    \label{tab:runtime k1}
\end{table}

\begin{table}[ht]
    \centering
\begin{tabular}{l|rr|rr|rr}
 \hline
 Method & \multicolumn{2}{c}{$n=500$} & \multicolumn{2}{c}{$n=1000$} & \multicolumn{2}{c}{$n=10 000$} \\
 & Single & Multiple  & Single & Multiple & Single & Multiple \\
 \hline
Random Imp. & $<$0.01 & 0.01 & $<$0.01 & 0.02 & 0.01 & 0.06\\

Median Imp. & $<$0.01 & - & $<$0.01 & - & 0.01 & -\\

MissForest & 0.27 & 5.59 & 0.40 & 7.70 & 4.86 & 98.33\\

MissForest PMM & 0.28 & 5.58 & 0.41 & 7.81 & 5.01 & 99.43\\

MICE PMM & 0.09 & 1.57 & 0.12 & 1.95 & 0.50 & 8.98\\

MICE RF & 0.43 & 8.24 & 0.66 & 12.73 & 4.96 & 98.42\\

MissARF & 1.20 & 1.23 & 1.38 & 1.44 & 6.34 & 6.26\\
\hline
\end{tabular}
    \caption{Runtime comparison of the different methods, executed with 16 threads. The measured average times in seconds for single and multiple imputation for each of the different $n$ are compared. }%Finally, the total time is displayed in the last column.}
    \label{tab:runtime k16}
\end{table}

\clearpage
\newpage
\section{Further experiments}
\label{sec:appx_further}
\subsection{Possible hyperparameter tuning}%\subsection{Hyperparameter tuning}
In the discussion, we state that MissARF has difficulties with higher missingness proportions and that the multiple imputation would benefit from a larger diversity of imputed values. We examine if the following approaches would improve the performance: 1) Adjusting the minimum node size, 2) increasing the number of trees, and 3) increasing the number of multiple imputations. For that we selected an example in which the coverage of MissARF was below the nominal level ($\sim 0.9$) for higher missingness: normal distribution with a linear effect with $p=4$, $n=1000$, and $0.4$ missingness with a MAR pattern (See Figure~\ref{fig: logreg_coverage_linear_normal}a), \ref{fig: logreg_aw_linear_normal}a) , \ref{fig: logreg_rmse_linear_normal}a)).

\begin{figure}[ht]
    \centering
    \includegraphics[width=0.92\linewidth]{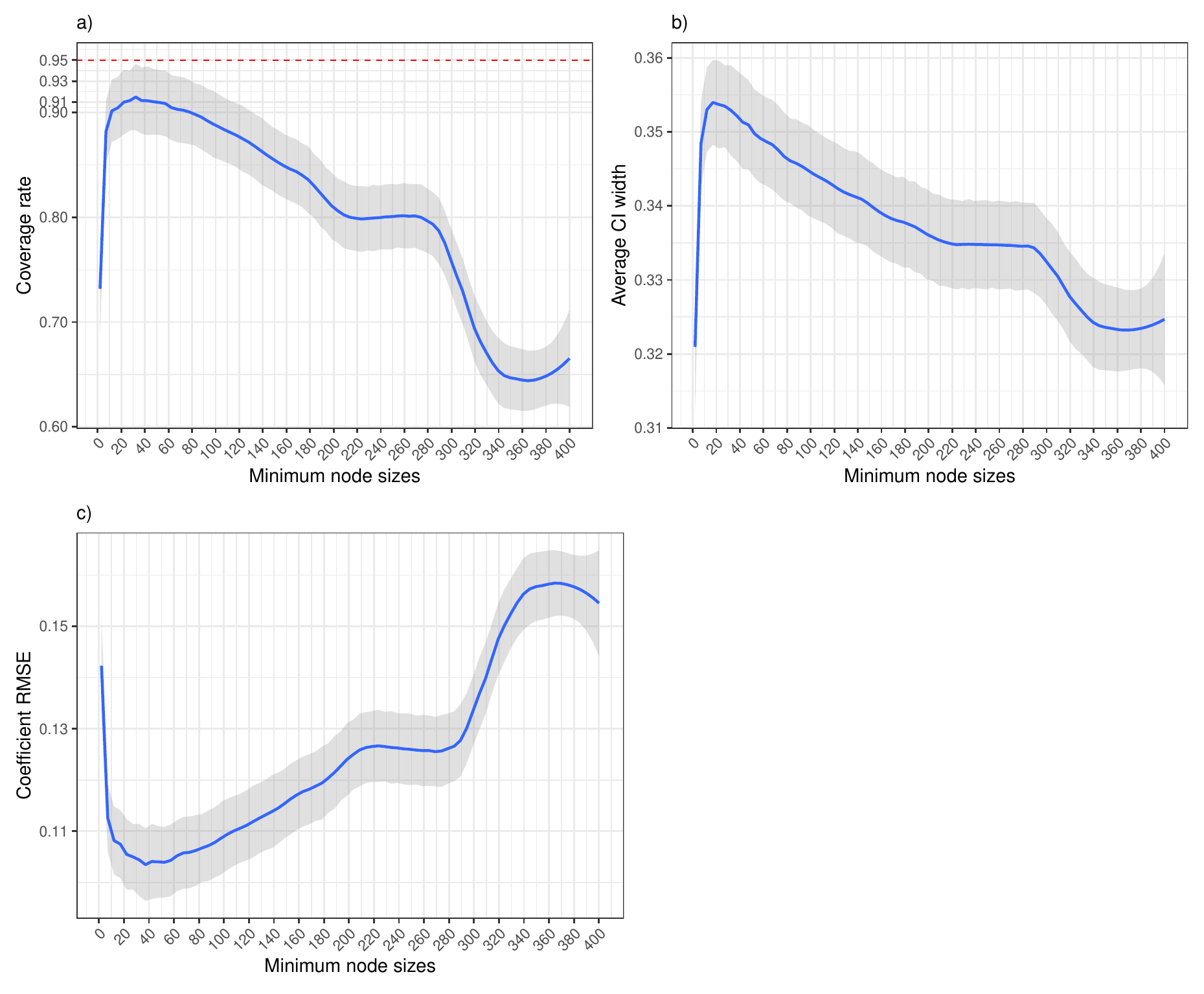}
    \caption{%Different min node sizes from 2 to 400 captured in a logscale with a 100 points.
    Mean values with the standard error for a) coverage rate, b) average confidence interval width and c) RMSE for minimum node sizes of 2-400 (100 points from a logarithmic scale) and 100 number of trees with MissARF for a normal distribution with a linear effect with $p=4$, $n=1000$, and $0.4$ missingness with a MAR pattern over 1000 replications. 
    }
    \label{fig: min node}
\end{figure}

\paragraph{Different number of node sizes:} The idea is that the larger the leaf node size, the more different variable values are considered for each imputed value, which could be helpful if the missingness rate is high. In Figure~\ref{fig: min node}, different minimum node sizes of 2-400 of a logarithmic scale grid are visualized. 
%For the coverage, we get a coverage above 0.9 for node sizes between 10 and 80 with a peak around 30. 
Very small node sizes lead to poor coverage and it increases drastically to a good coverage already by a node size of 10. We get a coverage above 0.9 for node sizes between 10 and 80 with a peak around 31 at 0.92. Then the coverage gets poorer and decreases with increasing node size. So choosing a node size of 31 instead of 10 improved the coverage by 0.02. The average confidence interval width starts with a small width for small node sizes, gets above 0.35 between 10 and 50 nodes with a peak around 20 and then decreases with larger node sizes. The RMSE starts with higher values for small node sizes and decreases with the number of node sizes. It is under 0.11 for node sizes between 10 and 110 with a minimum around 40 and increases after that with larger node sizes.

\begin{figure}[ht]
    \centering
    \includegraphics[width=0.92\linewidth]{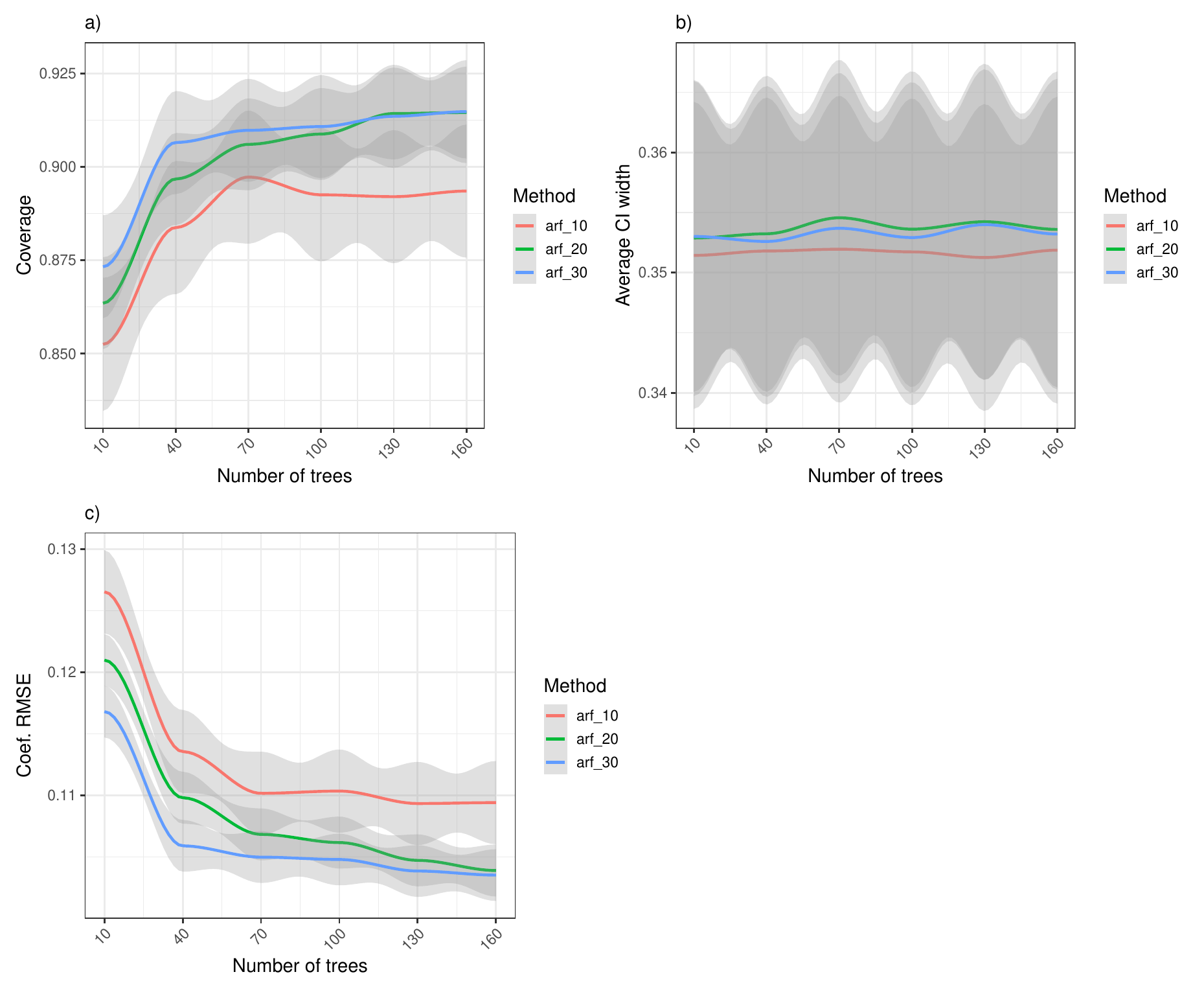}
    \caption{%Different min node sizes from 2 to 400 captured in a logscale with a 100 points.
    Mean values with the standard error for a) coverage rate, b) average confidence interval width and c) RMSE for minimum node sizes 10 (arf\_10), 20 (arf\_20) and 30 (arf\_30) and 10-160 number of trees with MissARF for a normal distribution with a linear effect with $p=4$, $n=1000$, and $0.4$ missingness with a MAR pattern over 1000 replications. 
    }
    \label{fig: min node num trees coverage}
\end{figure}

\begin{figure}[ht]
    \centering
    \includegraphics[width=0.5\linewidth]{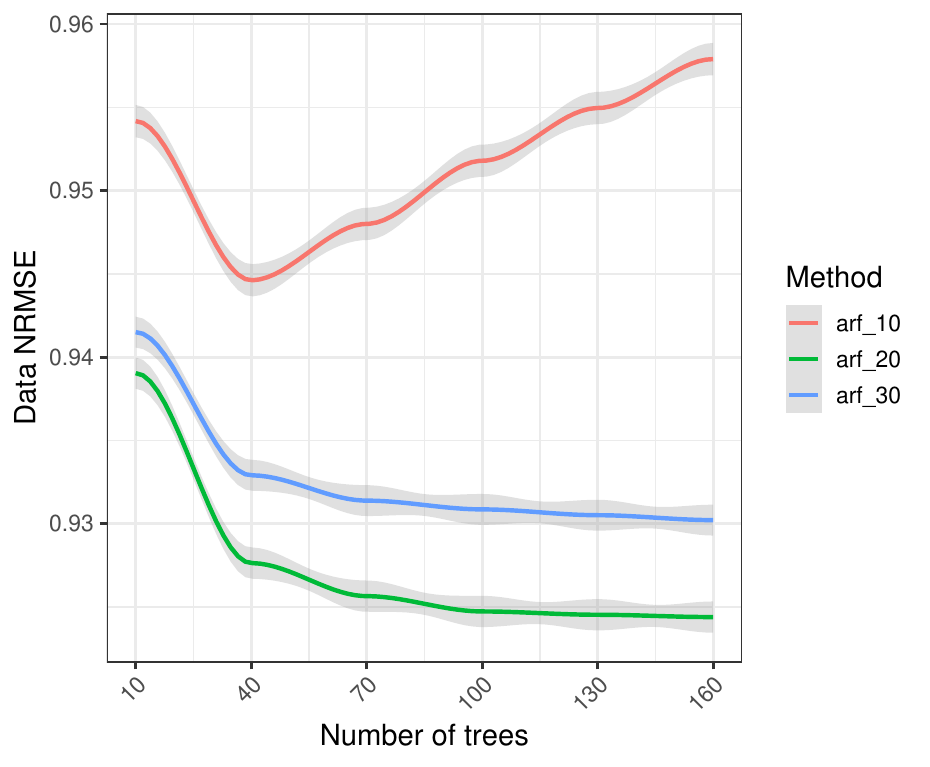}
    \caption{%Different min node sizes from 2 to 400 captured in a logscale with a 100 points.
    Mean values with the standard error for the NRMSE for minimum node sizes 10 (arf\_10), 20 (arf\_20) and 30 (arf\_30) and 10-160 number of trees with MissARF for a normal distribution with a linear effect with $p=20$, $n=1000$, and $0.4$ missingness with a MAR pattern over 1000 replications. 
    }
    \label{fig: min node num trees NRMSE}
\end{figure}

\paragraph{Number of trees:}
We examine the effect of different numbers of trees ranging from 10 to 160 for different minimum node sizes 10, 20 and 30 on coverage, average CI width and coefficient RMSE (Figure~\ref{fig: min node num trees coverage}) and on the NRMSE (Figure~\ref{fig: min node num trees NRMSE}). We chose $p=20$ for the NRMSE because MissARF performed worse than MissForest in the experiments and we aim to improve it. 

We observe that for a small number of node size 10, more trees harm the performance, whereas it does not have an effect with 20 or 30. For a minimum node size 10, the NRMSE has an optimum around 40, and then the performance weakens with growing number of trees. For the coverage we observe a similar effect with an optimum at 70. In both cases we observe worse performance for 10 as the minimum node size. For the NRMSE, we observe better performance for 20 as a minimum node size. For the coverage, and coefficient RMSE, the number of node size of 30 has a slight advantage to 20. The average width is slightly larger than for 10.

\begin{figure}[ht]
    \centering
    \includegraphics[width=0.92\linewidth]{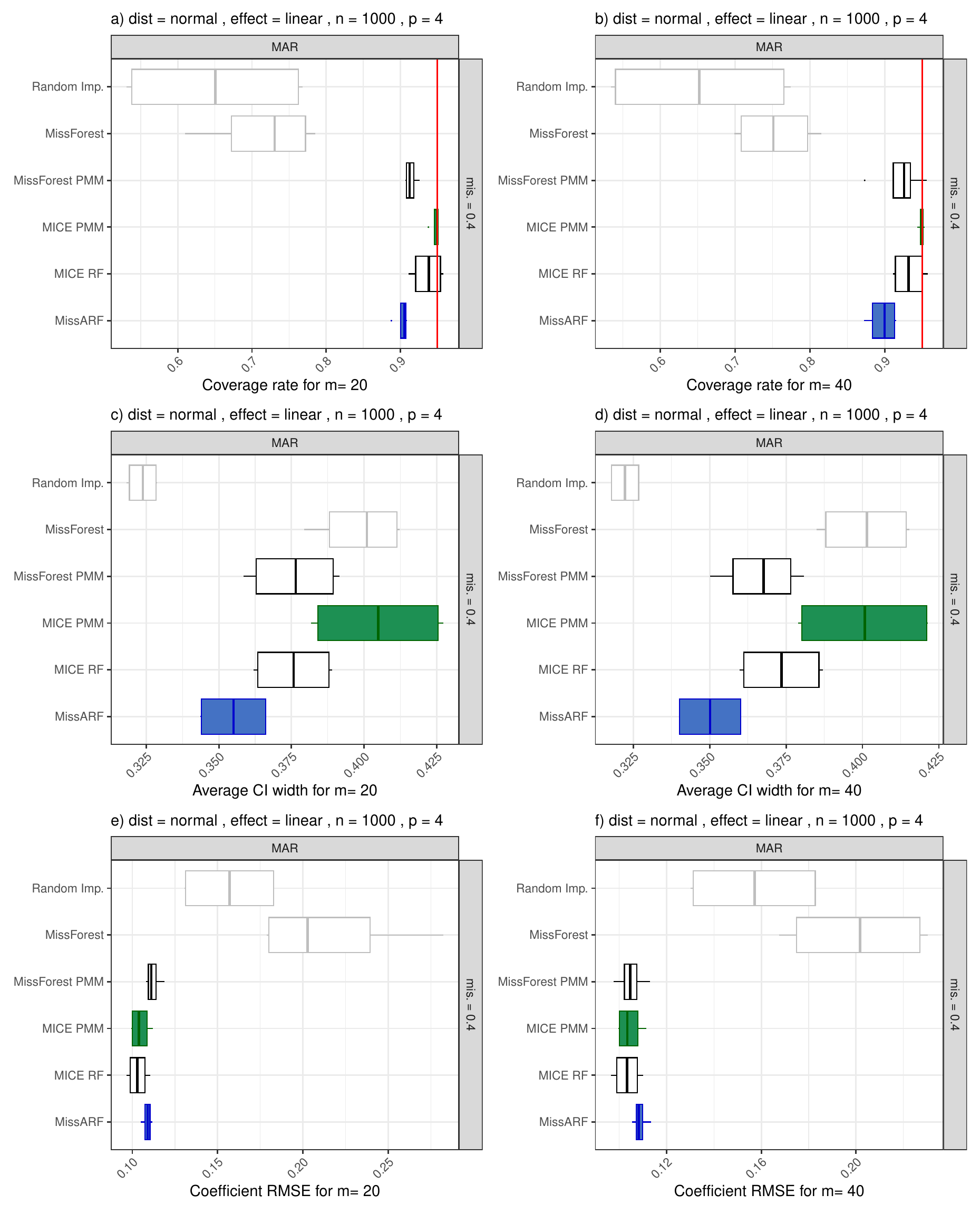}
    \caption{
The effect of the number of multiple imputations $m=20$ (left) and $m=40$ (right) for a-b) coverage rate, c-d) average confidence interval width and e-f) RMSE for a normal distribution with a linear effect with $p=4$, $n=1000$, and $0.4$ missingness with a MAR pattern. The boxplots are plotted over the features, with MissARF (blue) and MICE PMM (green) highlighted.
    }
    \label{fig: add. exp. multiple m40}
\end{figure}

\paragraph{More multiple imputations ($m=40$):}
%Thus, a possible solution is to increase the number of multiple imputations. 
In our experiments, we chose a fixed number of $m=20$ for simplicity and due to computational restrictions. However, White, Royston, and Wood\cite{white2011multiple} describe a rule of thumb for $m$ being ``at least equal to the percentage of incomplete cases''\cite{white2011multiple}\cite{molenberghs2014handbook}. When repeating the experiment with $m=40$ imputations as recommended by White, Royston, and Wood\cite{white2011multiple}, we would expect more stable results. However, in this example (see Figure~\ref{fig: add. exp. multiple m40}) we do not notice a big difference for coverage, average CI width and RMSE, and the boxplots are slightly wider than before. %and the variance of the boxplots are a bit brider than before. 

\paragraph{Conclusion:}All in all, we conclude that optimizing node size could improve the performance and therefore be considered as a hyperparameter to tune.

\subsection{Smaller sample sizes}
\label{sec:appx_further_small}
In practice, for example, in clinical research, we encounter small sample sizes. In the following, we examine the trend of the performance with smaller sample sizes for the normal distribution with a linear effect with $p=4$ and $0.2$ missingness with a MAR pattern for sample sizes between 50 and 500.

\begin{figure}[ht]
    \centering
    \includegraphics[width=0.5\linewidth]{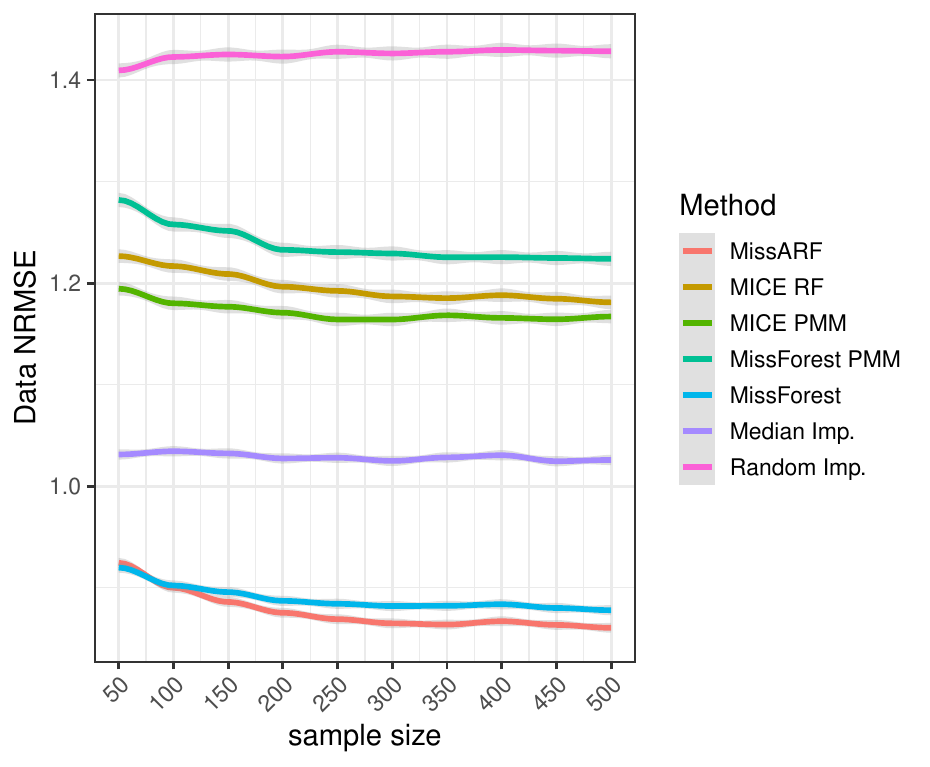}
    \caption{Mean values with the standard error for the NRMSE for sample sizes 50-500 for the different imputation methods. Results are for a normal distribution with a linear effect with $p=4$ and $0.2$ missingness with a MAR pattern over 1000 replications. 
    }
    \label{fig: sample sizes NRMSE}
\end{figure}

\begin{figure}[ht]
    \centering
    \includegraphics[width=0.5\linewidth]{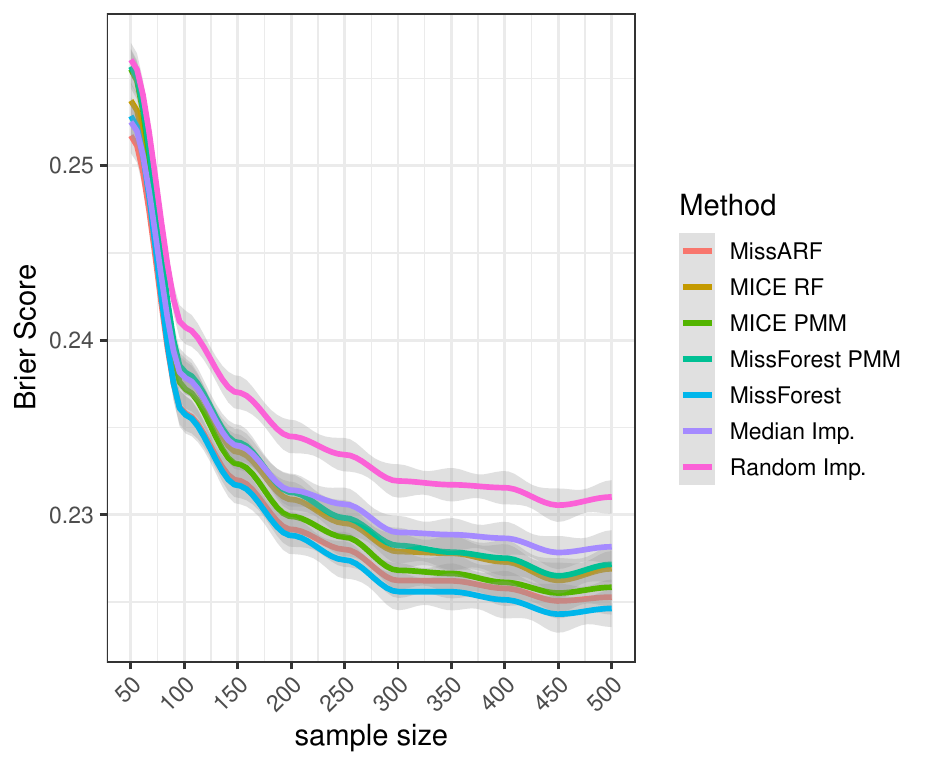}
    \caption{Mean values with the standard error for the Brier Score for sample sizes 50-500 for the different imputation methods. Results are for a normal distribution with a linear effect with $p=4$ and $0.2$ missingness with a MAR pattern over 1000 replications. 
    }
    \label{fig: sample sizes Brier Score}
\end{figure}

\begin{figure}[ht]
    \centering
    \includegraphics[width=0.92\linewidth]{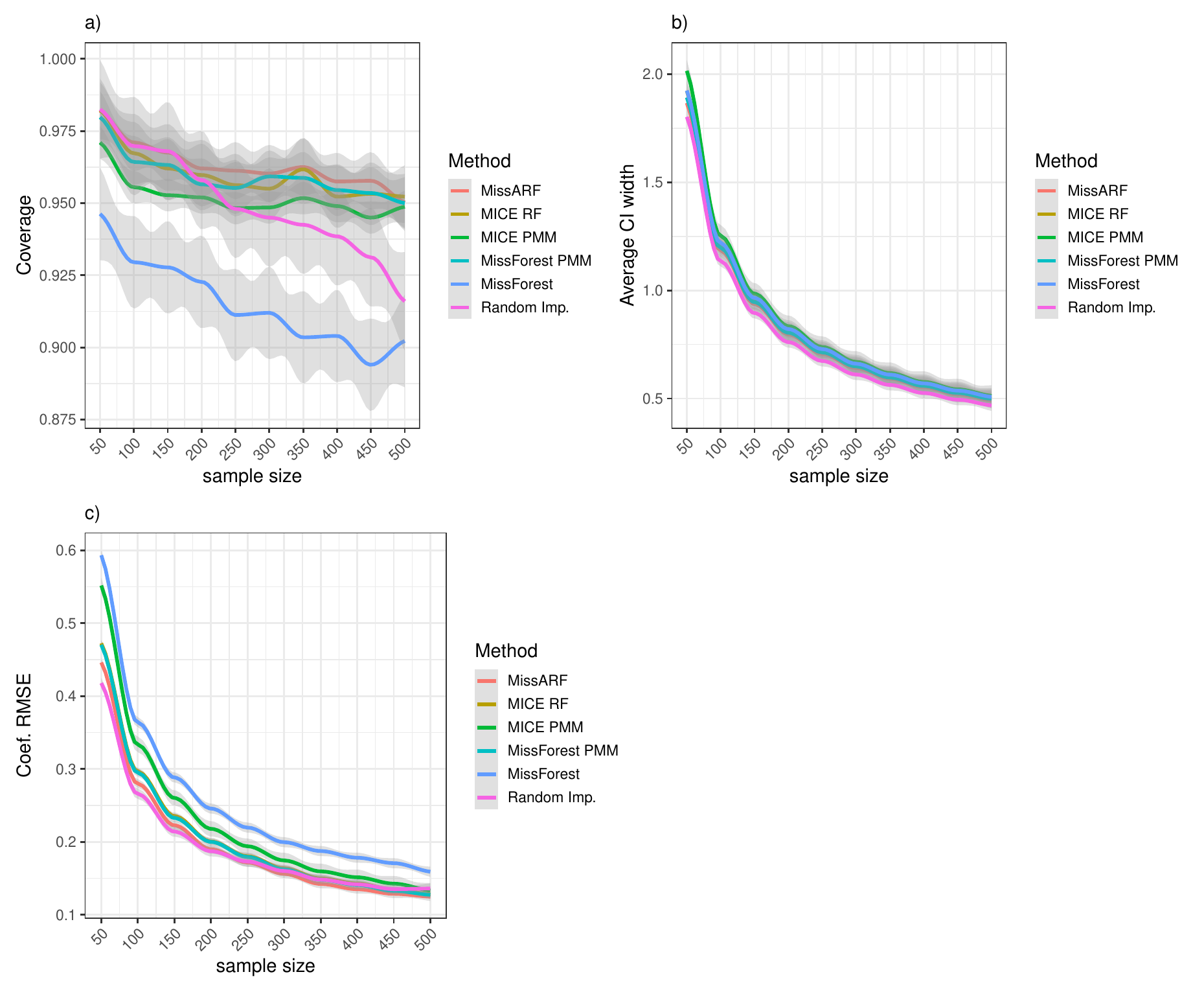}
    \caption{%Different min node sizes from 2 to 400 captured in a logscale with a 100 points.
    Mean values with the standard error for a) coverage rate, b) average confidence interval width and c) RMSE for sample sizes 50-500 for the different imputation methods. Results are for a normal distribution with a linear effect with $p=4$ and $0.2$ missingness with a MAR pattern over 1000 replications.
    }
    \label{fig: smaller sample coverage}
\end{figure}

In general, the performance of all metrics decreases for sample sizes smaller than 150. For the NRMSE, it only decreases slightly for all methods except for random and median imputation, which do not learn from data and are thus not affected much by sample size (Figure~\ref{fig: sample sizes NRMSE}). The Brier score (Figure~\ref{fig: sample sizes Brier Score}) declines more steeply for all methods compared to the NRMSE. For setting II (multiple imputation) (Figure~\ref{fig: smaller sample coverage}), all methods achieve higher coverage rates with wider average widths of the confidence intervals and higher RMSE values for a smaller sample size. All methods over-cover, except MissForest, which under-covers, and MICE PMM, which has constant good coverage close to 0.95. For $n=50$, all methods, except MissForest, over-cover and have high values for the RMSE of the regression coefficients. While for MissForest and random imputation the coverage decreases with growing sample size, for the other methods it remains similar after a sample size of 100. %, which indicate non-convergence of the GLM. 
%For setting II (multiple imputation) (Figure~\ref{fig: smaller sample coverage}), all methods achieve larger coverage rate values with wider average widths of the confidence intervals. All methods over-cover, except MissForest, which under-covers, and MICE PMM, which has constant good coverage close to 0.95. For $n=50$, all methods, except MissForest, over-cover and have high values for the RMSE of the regression coefficients.%, which indicate non-convergence of the GLM. 

\end{document}